\documentclass[twoside,11pt]{article}

\usepackage{jmlr2e}

\usepackage{amssymb}
\usepackage{graphicx}
\usepackage{stmaryrd,nicefrac}
\usepackage{latexsym,amsmath,amsfonts,amssymb}
\usepackage{url}

\usepackage{tikz}
\usetikzlibrary{arrows,shapes,positioning,shadows,trees}

\usepackage{afterpage}

\usepackage{longtable}
\usepackage{centernot}
\usepackage{makecell}

\usepackage{footnote}
\usepackage{longtable}
\makesavenoteenv{tabular}
\makesavenoteenv{table}

\newcommand{\reforder}{\tau}

\newcommand*{\defeq}{\mathrel{\vcenter{\baselineskip0.5ex \lineskiplimit0pt
			\hbox{\footnotesize.}\hbox{\footnotesize.}}}%
	=}

\makeatletter
\newcommand\footnoteref[1]{\protected@xdef\@thefnmark{\ref{#1}}\@footnotemark}
\makeatother

\newcommand{\cR}{{\cal R}}
\newcommand{\cA}{{\cal A}}
\newcommand{\cK}{{\cal K}}
\newcommand{\cM}{{\cal M}}

\newcommand{\cL}{{\cal L}}
\newcommand{\cS}{{\cal S}}
\newcommand{\cH}{{\cal H}}
\newcommand{\cG}{{\cal G}}
\newcommand{\cT}{{\cal T}}
\newcommand{\cP}{{\cal P}}

\newcommand{\br}{{\pi }}

\newcommand{\bQ}{{\bf Q}}
\newcommand{\bS}{{\bf S}}

\newcommand{\bN}{{\bf N}}
\newcommand{\bM}{{\bf M}}
\newcommand{\bSigma}{{\bf \Sigma}}
\newcommand{\bm}{{\bf m}}

\newcommand\R{\mathbb{R}}   % for the real numbers
\newcommand\N{\mathbb{N}}   % for the natural numbers

\newcommand{\prob}{\mathbf{P}} 
\newcommand{\exptd}{\mathbf{E}}

\newcommand{\IND}[1]{{{\mathbb I}\{{#1}\}}}
\newcommand{\Algo}[1]{\textsc{#1}}

\newcommand{\bigO}{{\cal O}}

\newcommand{\ro}{{\br^*}}
\newcommand{\roi}{\pi}

\newcommand{\KL}{\operatornamewithlimits{KL}}
\newcommand{\argmax}{\operatornamewithlimits{argmax}}
\newcommand{\argmin}{\operatornamewithlimits{argmin}}

\newcommand\BIN{\mathrm{CO}}
\newcommand\SE{\mathrm{SE}}
\newcommand\RW{\mathrm{RW}}
\newcommand{\CP}{\mathrm{CP}} % to denote the Copeland winner set

\usepackage{marginnote}
\usepackage{xcolor}

\firstpageno{1}

\usepackage{lastpage}
\jmlrheading{21}{2020}{1-\pageref{LastPage}}{8/18; Revised
12/19}{01/21}{18-546}{Viktor Bengs,  R{\'{o}}bert Busa-Fekete, Adil El Mesaoudi-Paul and Eyke H\"ullermeier}
\ShortHeadings{Preference-based Online Learning with Dueling Bandits: A Survey}{Bengs, Busa-Fekete, El Mesaoudi-Paul and H\"ullermeier}

\begin{document}
	
	\title{Preference-based Online Learning with \\ Dueling Bandits: A Survey}
	
	\author{\name Viktor Bengs \email viktor.bengs@upb.de \\
		\addr Heinz Nixdorf Institute and Department of Computer Science\\
		Paderborn University, Germany
				\AND
		\name R{\'{o}}bert Busa{-}Fekete \email busarobi@google.com \\
		\addr Google Research\\
		New York, NY, USA 
		\AND
		\name Adil El Mesaoudi-Paul \email adil.paul@upb.de \\
		\addr Heinz Nixdorf Institute and Department of Computer Science\\
		Paderborn University, Germany
				\AND
		\name Eyke H\"ullermeier \email eyke@upb.de \\
		\addr Heinz Nixdorf Institute and Department of Computer Science\\
		Paderborn University, Germany
	}
	
	\editor{Peter Auer}
	
	\maketitle
	
	\begin{abstract}%  <- trailing '%' for backward compatibility of .sty file
		In machine learning,  the notion of \emph{multi-armed bandits} refers to a class of online learning problems, in which an agent is supposed to simultaneously explore and exploit a given set of choice alternatives in the course of a sequential decision process. In the standard setting, the agent learns from stochastic feedback in the form of real-valued rewards. In many applications, however, numerical reward signals are not readily available---instead, only weaker information is provided, in particular relative preferences in the form of qualitative comparisons between pairs of alternatives. This observation has motivated the study of variants of the multi-armed bandit problem, in which more general representations are used both for the type of feedback to learn from and the target of prediction.  The aim of this paper is to provide a survey of the state of the art in this field, referred to as preference-based multi-armed bandits or dueling bandits. To this end, we provide an overview of problems that have been considered in the literature as well as methods for tackling them. Our taxonomy is mainly based on the assumptions made by these methods about the data-generating process and, related to this, the properties of the preference-based feedback.
	\end{abstract}

	\begin{keywords}
		Multi-armed bandits, online learning, preference learning, ranking, top-$k$ selection, exploration/exploitation, cumulative regret, sample complexity, PAC learning
	\end{keywords}
	
	%#################################
	\section{Introduction}
	Multi-armed bandit (MAB) algorithms have received considerable attention and have been studied quite intensely in machine learning in the recent past. The great interest in this topic is hardly surprising, given that the MAB setting is not only theoretically challenging but also practically useful, as can be seen from its use in a wide range of applications. For example, MAB algorithms turned out to offer effective solutions for problems in medical treatment design \citep{LaRo85,KuPr14}, online advertisement \citep{ChKuRaUp08}, and recommendation systems \citep{KoSaSt13}, just to mention a few.  
	
	The multi-armed bandit problem, or bandit problem for short, is one of the simplest instances of the sequential decision making problem, in which a \emph{learner} (also called decision maker or agent) needs to select \emph{options} from a given set of alternatives repeatedly in an online manner---referring to the metaphor of the eponymous gambling machine in casinos, these options are also associated with ``arms'' that can be ``pulled''. More specifically, the agent selects one option at a time and observes a numerical (and typically noisy) \emph{reward} signal providing information on the quality of that option. The goal of the learner is to optimize an evaluation criterion such as the \emph{error rate,} i.e., the expected percentage of suboptimal pulls, or the \emph{cumulative regret}, i.e., the expected difference between the sum of rewards that could have been obtained by playing the best arm (defined as the one generating the highest rewards on average) in each round and the sum of the rewards actually obtained. To achieve the desired goal, the online learner has to cope with the famous exploration/exploitation dilemma \citep{AuCeFi02,CeLu06,LaRo85}: It has to find a reasonable compromise between playing the arms that produced high rewards in the past (exploitation) and trying other, possibly even better arms the (expected) rewards of which are not precisely known so far (exploration).
	
	The assumption of a numerical reward signal is a potential limitation of the MAB setting. In fact, there are many practical applications in which it is hard or even impossible to quantify the quality of an option on a numerical scale. More generally, the lack of precise feedback or exact supervision has been observed in other branches of machine learning, too, and has led to the emergence of fields such as \emph{weakly supervised learning} and \emph{preference learning} \citep{FuHu11}. In the latter, feedback is typically represented in a purely qualitative way, namely in terms of pairwise comparisons or rankings. Feedback of this kind can be useful in online learning, too, as has been shown in online information retrieval \citep{Ho13,RaKuJo08}. %for example see the interleaving problem. 
	As another example, think of crowd-sourcing services like the Amazon Mechanical Turk, where simple questions such as pairwise comparisons between decision alternatives are asked to a group of annotators. The task is to approximate an underlying target ranking on the basis of these pairwise comparisons, which are possibly noisy and partially non-coherent \citep{ChBeCoHo13}. Another application worth mentioning is the ranking of XBox gamers based on their pairwise online duels; the ranking system of XBox is called TrueSkill$\texttrademark$ \citep{GuSaGrBu12}.
	
	%In some practical applications, the goal is to elicit the rank over a set of objects of interest based on the ranking provided by individuals. But to ask full rankings might be impossible or very time consuming where, for example, there are many objects to be ranked, or to provide full ranking by the individuals queried is too time-consuming from some reason. For example, think of crowd-sourcing services like the Amazon Mechanical Turk, where simple questions such as pairwise comparisons between decision alternatives are asked to a group of annotators. The task is to approximate an underlying target ranking on the basis of these pairwise comparisons, which are possibly noisy and partially inconsistent~\cite{ChBeCoHo13}. Another application worth mentioning is the ranking of XBox gamers based on their pairwise online duels; the ranking system of XBox is called TrueSkill$\texttrademark$\cite{GuSaGrBu12}. 
	
	Extending the multi-armed bandit setting to the case of preference-based feedback, i.e., the case in which the online learner is allowed to compare arms in a qualitative way, is therefore a promising idea. And indeed, extensions of that kind have received increasing attention in the recent years. The aim of this paper is to provide a survey of the state of the art in the field of preference-based multi-armed bandits (PB-MAB); it updates and significantly extends an earlier review by \cite{mpub296}. After recalling the basic setting of the problem in Section 2, we provide an overview of methods that have been proposed to tackle PB-MAB problems in Sections 3 and 4. Our taxonomy is mainly based on the assumptions made by these methods about the data-generating process or, more specifically, the properties of the pairwise comparisons between arms. Our survey is focused on the \emph{stochastic} MAB setup, in which feedback is generated according to an underlying (unknown but stationary) probabilistic process; we do not cover the case of \emph{adversarial} data-generating processes, which has recently received a lot of attention as well \citep{AiHaTa14,DuHoShSlZo15,GaUr15,ZiSe19}, except briefly in Section \ref{sec:extensions}, where also other extensions of the basic PB-MAB problem are discussed.
	Due to the surge of research interest in the multi-dueling variant of the basic PB-MAB problem in the recent past\footnote{This serves as an additional motivation for referring to both settings, i.e, dueling bandits and multi-dueling bandits, as the \emph{preference-based multi-armed bandits} such as implicitly proposed by this survey, since this notion unifies the latter two closely related fields in a quite intuitive way.}, we devote Section \ref{sec:multi_duel_extensions} to this extension.
	Finally, we highlight some interesting practical applications of the PB-MAB problem in Section \ref{sec_applications} prior to concluding with a discussion of open issues within the scope of the survey in Section \ref{sec_summary}.
	
	%#################################
	\section{The Basic Preference-based Multi-Armed Bandit Problem} \label{sec:ppe}
	The stochastic MAB problem with pairwise comparisons as actions has been studied under the notion of ``dueling bandits'' in several papers \citep{YuJo09,YuBrKlJo12}. Although this term has been introduced for a concrete setting with specific modeling assumptions \citep{SuZoHoYu18}, it is meanwhile used more broadly for variants of that setting, too. Throughout this paper, we shall use the terms ``dueling bandits'' and ``preference-based bandits'' synonymously, although the latter may indeed seem more convenient in light of the recent multi-dueling variant (cf.\ Section \ref{sec:multi_duel_extensions}).
	
	Consider a fixed set of arms (options) $\cA = \{a_{1}, \dots , a_{K} \}$. As actions, the learning algorithm (or simply the learner or agent) can perform a comparison between any pair of arms $a_i$ and $a_j$, i.e., the action space can be identified with the set of index pairs $(i,j)$ such that $1 \leq i \leq j \leq K$. We assume the feedback observable by the learner to be generated by an underlying (unknown) probabilistic process characterized by a \emph{preference relation} 
	$$
	\bQ = \left[ q_{i,j} \right]_{1\le i,j \le K} \in [0,1]^{K \times K} \enspace .
	$$ 
	More specifically, for each pair of actions $(a_{i},  a_{j})$, this relation specifies the pairwise preference probability 
	\begin{align}\label{eq:pairwisex}
	\prob \left( a_{i} \succ a_{j} \right) = q_{i,j}
	\end{align}
	of observing a preference for $a_{i}$ in a direct comparison with $a_{j}$. Thus, each $q_{i,j}$ specifies a Bernoulli distribution. These distributions are assumed to be stationary and independent, both across actions and iterations. Thus, whenever the learner takes action $(i,j)$, the outcome is distributed according to (\ref{eq:pairwisex}), regardless of the outcomes in previous iterations. 
	
	The relation $\bQ$ is reciprocal in the sense that $q_{i,j} = 1 - q_{j,i}$ for all $i , j \in [K] \defeq \{1, \ldots , K \}$. We note that, instead of only observing strict preferences, one may also allow a comparison to result in a \emph{tie} or an \emph{indifference}. In that case, the outcome is a trinomial instead of a binomial event. Since this generalization makes the problem technically more complicated, though without changing it conceptually, we shall not consider it further. \citet{BuSzWeChHu13,BuSzHu14} handle indifference by giving ``half a point'' to both arms, which, in expectation, is equivalent to deciding the winner by tossing a coin. Thus, the problem is essentially reduced to the case of binomial outcomes.  
	
	We say arm $a_{i}$ beats arm $a_{j}$ if $q_{i,j}>1/2$, i.e., if the probability of winning in a pairwise comparison is larger for $a_{i}$ than it is for $a_{j}$. Clearly, the closer $q_{i,j}$ is to $1/2$, the harder it becomes to distinguish the arms $a_{i}$ and $a_{j}$ based on a finite sample set from $\prob \left( a_{i} \succ a_{j} \right)$. In the worst case, when $q_{i,j} = 1/2$, one cannot decide which arm is better based on a finite number of pairwise comparisons. Therefore, 
	$$
	\Delta_{i,j} = q_{i,j} - \frac{1}{2}
	$$ 
	appears to be a reasonable quantity to characterize the hardness of a PB-MAB task (whatever goal the learner wants to achieve), which we call the \emph{calibrated pairwise preference probabilities}. Note that $\Delta_{i,j}$ can also be negative (opposed to the value-based setting, in which the usual quantity used for characterizing the complexity of a multi-armed bandit task is always positive and depends on the gap between the means of the best arm and the suboptimal arms). 
	
	%\subsection{Pairwise probability estimation}\label{sec:ppe}
	
	\subsection{Learning Protocol}
	
	The decision making process iterates in discrete steps, either through a finite time horizon $\mathbb{T} \defeq [T]=\{1, \ldots , T \}$ or an infinite horizon $\mathbb{T} \defeq \mathbb{N}$. 
	As mentioned above, the learner is allowed to compare two actions in each iteration $t \in \mathbb{T}$. Thus, in each iteration $t$, it selects an index pair $1 \leq i(t) \leq  j(t) \leq K$ and observes 
	%either $a_{i(t)} \succ a_{j(t)}$ (with probability $q_{i(t),j(t)}$) or $a_{j(t)} \succ a_{i(t)}$ (with probability $q_{j(t),i(t)}$). 
	$$
	\left\{
	\begin{array}{cl}
	a_{i(t)} \succ a_{j(t)}, & \text{ with probability } q_{i(t),j(t)} \\[2mm]
	a_{j(t)} \succ a_{i(t)}, & \text{ with probability } q_{j(t),i(t)}
	\end{array}
	\right. \enspace .
	$$
	%Considering the outcome as a binary event, we shall also use the notation $o^{t} = 1$ if the feedback was in favor of the first arm $a_{i(t)}$ and $o^{t} = 0$ if a preference for $a_{j(t)}$ was observed.  
	The pairwise probabilities $q_{i,j}$ can be estimated on the basis of finite sample sets. Consider the set of time steps among the first $t$ iterations, in which the learner decides to compare arms $a_{i}$ and $a_{j}$, and denote the size of this set by $n_{i,j}^{t}$. Moreover, denoting by $w_{i,j}^{t}$ and $w_{j,i}^{t}$ the frequency of ``wins'' of $a_i$ and $a_j$, respectively,  
	%$$
	%I^{t}_{i,j} = \left\{ \, \ell \in [t] \, \vert \, \left(i(\ell), j(\ell) \right) = (i,j) \, \right\} \enspace ,
	%$$ 
	%and the size of this set by $n_{i,j}^{t}=\#I^{t}_{i,j}$.\footnote{We omit the index $t$ if there is no risk of confusion.} 
	the proportion of wins of $a_i$ against $a_j$ up to iteration $t$ is then given by
	\[
	\widehat{q}^{\, t}_{i,j}= \frac{w_{i,j}^{t}}{n^{t}_{i,j}}  = \frac{w_{i,j}^{t}}{w_{i,j}^{t} + w_{j,i}^{t}}
	%\sum_{\ell \in I^{t}_{i,j}} \INDo^{\ell} 
	\enspace .
	\]
	Since our samples are assumed to be independent and identically distributed (i.i.d.), $\widehat{q}^{\, t}_{i,j}$ is a plausible estimate of the pairwise winning probability defined in (\ref{eq:pairwisex}). Yet, this estimate might be biased, since $n_{i,j}^{t}$ depends on the choice of the learner, which in turn depends on the data; therefore, $n_{i,j}^{t}$ itself is a random quantity. A high probability confidence interval $c_{i,j}^{t}$ for $q_{i,j}$ can be obtained based on the Hoeffding's bound \citep{Ho63}, which is commonly used in the bandit literature. Although the specific computation of the confidence intervals may differ from case to case, they are generally of the form $[\widehat{q}^{\, t}_{i,j} \pm c_{i,j}^{t}]$. Accordingly, if $\widehat{q}^{\, t}_{i,j} - c_{i,j}^{t} > 1/2$, arm $a_{i}$ beats arm $a_{j}$ with high probability; analogously, $a_{i}$ is beaten by arm $a_{j}$ with high probability, if $\widehat{q}^{\, t}_{i,j} + c_{i,j}^{t} < 1/2$. 
	
	\subsection{Learning Tasks} \label{subsec_targets}
	
	%\subsubsection{Evaluation criteria}
	%The goal of the online learner is usually stated as minimizing some kind of cumulative regret. Alternatively, in the ``pure exploration'' scenario, the goal is to identify the best arm (or the top-$k$ arms, or a ranking of all arms) both quickly and reliably. As an important difference between these two types of targets, note that the regret of a comparison of arms depends on the concrete arms being chosen, whereas the sample complexity penalizes each comparison equally. 
	
	%It is also worth mentioning that 
	The notion of optimality of an arm is far less obvious in the preference-based setting than it is in the value-based (numerical) setting. In the latter, the optimal arm is (usually) simply the one with the highest expected reward---more generally, the expected reward induces a natural total order on the set of arms $\cA$. In the preference-based case, the connection between the pairwise preferences $\bQ$ and the order induced by this relation on $\cA$ is less trivial; in particular, the latter may contain preferential cycles. In the following, we provide an overview of different notions of (sub-)optimality of arms and the related learning tasks.

	\subsubsection{Best Arm}
	In the preference-based setting, it appears natural to define the best arm as the one that is preferred to any other arm. Formally, $a_{i^*} \in \cA$ is the best arm if $\Delta_{i^{*},j}>0$ for all $j\in [K]\backslash\{i^{*}\}.$
	This definition corresponds to the definition of the so-called \emph{Condorcet winner.} In spite of being natural, it comes with the major drawback that a Condorcet winner is not guaranteed to exist for every preference relation $\bQ.$
	Due to this problem, various alternative notions for a best arm have been suggested, each with its own advantages and drawbacks.
	These different alternatives will be introduced in Section \ref{sec:noass}. For sake of simplicity, we subsequently assume the existence of a Condorcet winner and denote it by $a_{i^{*}}$ throughout the rest of this section.

	\subsubsection{Rankings of Arms} \label{subsec_target_rankings}
	Another target for the learner,  more ambitious  than merely finding an optimal arm, is to find an entire ranking over the arms. A ranking can be identified by a permutation $\pi:\{1,\ldots,K\} \to \{1,\ldots,K\}$, with $\pi(i)$ specifying the rank of arm $a_i\in \cA$ (and $\pi^{-1}(i)$ the index of the arm on the $i^{th}$ position). For a sensible definition of a target ranking of the arms, similar issues arise like in the specification of the best arm.
	
	Indeed, given a preference relation $\bQ$, the arguably most natural approach is to consider the ranking of the arms such that arm $a_i$ is ranked higher than another arm $a_j$ if and only if the former beats the latter ($\Delta_{i,j} > 0$). However, just like the Condorcet winner may fail to exist for a given preference relation $\bQ,$ such a ranking may not exist due to preferential cycles.  
	Again, to circumvent this problem, alternative definitions of target rankings have been proposed in the literature (cf.\ Section \ref{sec:noass}).
%	}

%	\editvik[
	\subsubsection{Top-positioned Arms}
	There are several applications in which the entire ranking over the arms is not of major relevance. Instead, only the ordering or merely the identity of the top-$k$ arms, i.e., the best $k$ elements of the underlying full ranking (assuming this ranking to exist), are of interest. This leads to the problems of \emph{top-$k$ ranking} and \emph{top-$k$ identification}
	
	In general, top-$k$ identification is an easier learning task, as it requires less information than the top-$k$ ranking (the former can be derived from the latter, but not the other way around). There are, however, examples of learning tasks for which both targets have the same complexity.
	
	\subsubsection{Estimation of the Preference Relation}
	
	The problem of estimating the preference relation $\bQ$, which has been tackled as an offline learning  task under the assumption of certain structural properties on $\bQ$ \citep{shah2016stochastically}, can also be considered in an online learning framework.
	%
	%\\
	A related target emerges by assuming that the preference probabilities in (\ref{eq:pairwisex}) are the marginals of a distribution over rankings (cf.\ Section \ref{sec:mall}), and one seeks to estimate the entire ranking distribution based on these marginals.
	%
	%This problem has also been studied in the offline learning setup \citep{sun2011estimating,sibony2015mra}.
%	]
	 
	\subsubsection{Near-optimal Targets} \label{subsec_near_opt_targets}
	There are various applications in which it suffices to produce a reasonably good approximation of the truly optimal target.
	Yet, compared to the classical MAB setting, approximation errors are less straightforward to define in the preference-based setup, again due to the lack of numerical rewards.
	In spite of this, for many of the just introduced targets, there have been attempts to define reasonable surrogates \citep{FaHaOrPiRa17,FaOrPiSu17,falahatgar2018limits}.
	
	% which will be considered in the following.

	For $\epsilon\in (0,1/2)$, an arm $a_i$ is called $\epsilon$-\emph{preferable} to an arm $a_j$ if $\Delta_{i,j} \geq -\epsilon$. Equipped with this notion, one can define an $\epsilon$-\emph{optimal arm} as an arm that is $\epsilon$-\emph{preferable} to the optimal arm $a_{i^{*}}$.
	%
	%Here, it is not crucial 
	%

	If the target is the natural ranking introduced in Section \ref{subsec_target_rankings}, which we shall identify by a permutation $\pi^*$ on $[K]$ such that $\pi^*(i)>\pi^*(j)$ implies $\Delta_{i,j}>0$, then an $\epsilon$-\emph{optimal} variant of $\pi^*$ is any permutation $\pi$ on $[K]$ such that  $\pi(i)>\pi(j)$ implies that $a_i$ is  $\epsilon$-preferable to arm $a_j.$
	For target rankings other than $\pi^*$, it is possible to define $\epsilon$-\emph{optimal} variants in a similar way, for example by specifying a sensible metric $d$ on the space of all permutations and calling $\epsilon$-optimal each ranking that is $\epsilon$-close to the target permutation with respect to $d.$ A more detailed discussion is deferred to Section \ref{sec:noass}.
%
%	}

	\subsection{Performance Measures} \label{subsec_preformance_measure}

	As usual in the realm of machine learning, there are different goals a learner may pursue, and correspondingly different ways to quantify the overall performance of a learner. The most prominent goals and performance measures will be discussed in the following. 
%}
	%The goal of the online learner is usually stated as minimizing some kind of cumulative regret. Alternatively, in the ``pure exploration'' scenario, the goal is to identify the best arm (or the top-$k$ arms, or a ranking of all arms) both quickly and reliably. As an important difference between these two types of targets, note that the regret of a comparison of arms depends on the concrete arms being chosen, whereas the sample complexity penalizes each comparison equally. 

	\subsubsection{Regret Bounds} \label{subsec_regret_bounds}
	In a preference-based setting, defining a reasonable notion of regret is not as straightforward as in the value-based setting, where the sub-optimality of an action can be expressed easily on a numerical scale. In particular, since the learner selects two arms to be compared in an iteration, the sub-optimality of both of these arms should be taken into account. A commonly used definition of regret is the following \citep{YuJo09,YuJo11,UrClFeNa13,ZoWhMuDe14}: Suppose the learner selects arms $a_{i(t)}$ and $a_{j(t)}$ in time step $t,$ then its regret per time is 
	$$r_{t,avg} \defeq  \frac{ \Delta_{i^{*},i(t)}+ \Delta_{i^{*},j(t)} }{2} \enspace ,
	$$
	i.e., the average quality (with respect to the target arm $a_{i^{*}}$) of the chosen arms in $t$, to which we refer as the \emph{average regret.}
	The \emph{cumulative regret} incurred by the learner $A$ up to time $T$ is then given by
	\begin{equation} \label{eq:regret}
	R^{T}_{A} \defeq \sum_{t=1}^{T} r_{t,avg} =  \sum_{t=1}^{T} \frac{ \Delta_{i^{*},i(t)}+ \Delta_{i^{*},j(t)} }{2}\enspace \, .
	\end{equation}
	For sake of brevity, we will suppress the dependency on $A$ in the notation of the cumulative regret and simply write $R^{T}$ if the learner is clear from the context.
	This regret takes into account the optimality of both arms, meaning that the learner has to select two nearly optimal arms to incur small regret. Note that this regret is zero if the optimal arm $a_{i^{*}}$ is compared to itself, i.e., if the learner effectively abstains from gathering further information and instead fully commits to the arm $a_{i^{*}}$.

	The regret defined in (\ref{eq:regret}) reflects the average quality of the decision made by the learner. Obviously, one can define a more strict resp.\ less strict regret by taking the maximum or minimum of the calibrated pairwise probabilities, respectively, instead of their average. Formally, the \emph{strong and weak regret} in time step $t$ are defined, respectively, as 
	\begin{align*}
	r_{t,\max} & \defeq \max 
	\left\{ \Delta_{i^{*}, i(t)} , \Delta_{i^{*},j(t)} \right\} \enspace , \quad 
	r_{t,\min}  \defeq \min 
	\left\{ \Delta_{i^{*},i(t)}, \Delta_{i^{*},j(t)} \right\} \enspace .
	\end{align*}
	Obviously, it holds that $r_{t,\min} \leq r_{t,avg} \leq r_{t,\max},$ so that one has a hierarchy among these notions of regret. Furthermore, note that some works refer to the average regret $r_{t,avg}$ also as the strong regret, which is due to the (arti-)fact that the average regret is zero only if both $a_{i(t)}$ and $a_{j(t)}$ correspond to the target arm $a_{i^{*}}$.
	This peculiarity of the average regret is a motivation for allowing $a_{i(t)}=a_{j(t)}$, which can be interpreted as a \emph{full commitment} to a single arm usually adopted as a ``final'' action, once being sure enough that the target arm has been found.

	From a theoretical point of view, a distinction between these regret definitions is important, as shown by \citet{ChFr17} and further discussed in Section \ref{sec_winner_stays_algo}. 
	In the rest of the paper, when speaking about the regret of a learner, we will refer to the regret in (\ref{eq:regret}) with respect to $r_{t,avg}$ unless otherwise stated.
	
	Another notion of regret per time is considered in the literature if the pairwise probabilities are modeled by utility functions, which will be discussed in Section \ref{subsec_utility_dueling_bandits}. However, this additional notion can be expressed as a linear transformation of the average regret $r_{t,avg}$, and consequently only scales the cumulative regret.

	In a theoretical analysis of an MAB algorithm, one is typically interested in providing a bound on the (cumulative) regret produced by that algorithm.
	We are going to distinguish two types of regret bound. The first one is the \emph{expected regret bound}, which is of the form 
	\begin{equation}\label{eq:expreg}
	\exptd \left[ R^{T} \right] \le B( \bQ, K, T ) \enspace ,
	\end{equation} 
	where $\exptd \left[ \cdot \right]$ is the expected value operator, $R^{T}$ is the regret accumulated till time step $T$, and $B( \cdot )$ is a positive real-valued function with the following arguments: the pairwise probabilities $\bQ$, the number of arms $K$, and the iteration number $T$. This function may additionally depend on parameters of the learner, for example a tuning- or hyper-parameter.
%	
%	
	%$\Rightarrow$ Concerning point 7 of Ref.1.]
%	
	However, we neglect this dependence here. The expectation is taken with respect to the stochastic nature of the data-generating process and the (possible) internal randomization of the online learner. The regret bound (\ref{eq:expreg}) is technically akin to the expected regret bound of value-based multi-armed bandit algorithms like the one that is calculated for UCB \citep{AuCeFi02}, although the parameters used for characterizing the complexity of the learning task are different. 
	
	The bound in (\ref{eq:expreg}) does not inform about how the regret achieved by the learner is concentrated around its expectation. Therefore, one might prefer to consider a second type of a reasonable regret bound, namely one that holds with high probability. This bound can be written in the form 
	$$
	\prob \Big( \, R^{T} < B( \bQ, K, T, \delta ) \, \Big) \ge 1-\delta \enspace .
	$$
	For simplicity, we also say that the regret achieved by the online learner is $\bigO (B( \bQ, K, T, \delta ))$ with high probability.

	Similar to the classical problem of online learning, if the goal is to minimize the regret, one is typically interested in \emph{no-regret} algorithms, i.e., algorithms the regret bound function $B$ of which grows sublinearly in the time horizon $T,$ if the remaining components are fixed.
	In general, there are two types of regret bounds. First, the \emph{gap-dependent regret bounds} depending on the calibrated preference probabilities with respect to the best arm. These bounds typically grow logarithmically with the time horizon $T$. Second, the \emph{gap-independent regret bounds}, which are typically given as a specific root function of the time horizon $T,$ but opposed to the first variant do not depend on the calibrated preference probabilities, though still on the number of arms $K$. The latter type of regret bounds corresponds in general to the worst-case learning scenario.

	% that is, the smallest $i \in \N^{+}$ for which there is no $i' \in \N^{+}$ such that $i' >i $ and $\prob( i' ) > 0 $.
	\subsubsection{PAC Setting}
	%\subsubsection{PAC algorithms}
	In many applications, one is willing to gain efficiency at the cost of optimality: The algorithm is allowed to return a solution that is only approximately optimal, though it is supposed to do so more quickly. 
	The variable of interest is then the \emph{sample complexity of the learner,} that is the number of pairwise comparisons it queries prior to termination for returning a nearly optimal target (cf.\ Section \ref{subsec_targets}). Such settings are referred to as \emph{probably approximately correct} (PAC) settings \citep{EvMaMa02} and have been studied extensively for the classical MAB problem \citep{mannor2004sample,bubeck2012regret}.
	
	A preference-based MAB algorithm is called $(\epsilon, \delta)$-PAC preference-based MAB algorithm with a \emph{sample complexity} $B( \bQ, K, \epsilon, \delta )$, if it terminates and returns an $\epsilon$-optimal target\footnote{Definitions of $\epsilon$-optimal targets are less straightforward in the PB-MAB problem. Such definitions will be given in subsequent sections.} with probability at least $1-\delta$, and the number of comparisons taken by the algorithm is at most $B( \bQ, K, \epsilon, \delta )$. 
	%and the corresponding bound is denoted $B( \bQ, K, \varepsilon, \delta )$. 
	%
	%Here, $1-\delta$ specifies a lower bound on the probability that the learner terminates and returns an $\epsilon$-optimal \editvik[target].
	%correct solution\footnote{Here, we consider the pure exploration setup with fixed confidence. Alternatively, one can fix the horizon and control the error of the recommendation \citep{AuBuMu10,BuMuSt11,BuWaVi13}.}.
	Note that only the number of the pairwise comparisons is taken into account, which means that pairwise comparisons are equally penalized, independently of the suboptimality of the arms chosen; in this regard, the setting differs from the goal of regret minimization.

	\subsubsection{Exact Sample Complexity}
	The exact sample complexity analysis is the strict version of the PAC learning goal described above, where the learner is supposed to return the correct target instead of a nearly optimal target. This setting is sometimes also called the $\delta$-PAC setting \citep{kaufmann2016complexity}. Obviously, when the allowed approximation error $\epsilon$ of the ($\epsilon,\delta$)-PAC learning scenario tends to zero, these two settings coincide. However, while in the PAC learning scenario the bound on the sample complexity focuses on the approximation error $\epsilon$ coming from the extrinsic problem formulation, the bounds for the exact sample complexity analysis focuses on the intrinsic hardness of the problem represented by the smallest calibrated preference probability. Hence, an upper bound on the exact sample complexity of a learner also provides an upper bound on its sample complexity in the PAC setting.
		% (cf.\ \citep{mannor2004sample}) 
	%The sample complexity analysis is considered in a ``pure exploration'' setup where the learner, in each iteration, must either select a pair of arms to be compared or terminate and return its recommendation. 
%	}
	
	% and also  in a ``pure exploration'' setup where the learner, in each iteration, must either select a pair of arms to be compared or terminate and return its recommendation. 
	%The \emph{sample complexity of the learner} is then the number of pairwise comparisons it queries prior to termination, and the corresponding bound is denoted $B( \bQ, K, \delta )$. Here, $1-\delta$ specifies a lower bound on the probability that the learner terminates and returns the correct solution\footnote{Here, we consider the pure exploration setup with fixed confidence. Alternatively, one can fix the horizon and control the error of the recommendation \citep{AuBuMu10,BuMuSt11,BuWaVi13}.}.
	%Note that only the number of the pairwise comparisons is taken into account, which means that pairwise comparisons are equally penalized, independently of the suboptimality of the arms chosen, \editvik[as opposed to the goal of regret minimization.
	%]

	%This kind of sample complexity analysis was originally introduced by Even-Dar \etal~\cite{EvMaMa02} for value-based MABs. 

	\subsection{Algorithm Classes}
	Despite the recency of the field of PB-MAB problems, a striking variety of algorithms has been developed to tackle the different targets and goals described above. 
	Many of these algorithms are based on similar ideas and essentially invoke the same learning principles, such as efficient sorting or the derivation of representative statistics.
	In the following, we propose a categorization of existing algorithms into different algorithm classes, each of them characterized by specific properties.
	Note, however, that the algorithm classes are not mutually exclusive, since some learning algorithms are combining essential concepts from different classes at the same time.
	%
%	}

	%\disvik{In my opinion, it is unreasonable to spend such a long section on the explore-then-exploit algorithms, as they are outdated by now. I would suggest to align them into the Generalization of MAB algorithm below as just another principle in the MAB-related algorithms.}
	
	%\subsubsection{Explore-then-exploit algorithms}
	%\editvik[ Many of the first emerging PB-MAB algorithms 
	
	%$\Rightarrow$ Concerning point 11 of Ref.1.]] 

	%\disvik{To me, this issue of knowing the time horizon in advance is not a problem due to the Doubling trick. Maybe we can mention instead of the work by \citep{ZoWhMuDe14} the new advances for the doubling trick: Besson and Kaufmann. ''What doubling tricks can and can't do for multi-armed bandits''.}

	\subsubsection{MAB-related Algorithms} \label{subsec_mab_related_algorithms}
	
	As the PB-MAB problem is a variant of the classical MAB problem, it seems quite natural to exploit established algorithmic ideas for the latter in order to construct algorithms for the former---hoping, of course, to preserve corresponding benefits.
	This connection is established in the existing methods in two ways:

	\begin{enumerate}
		\item[--] \emph{Reduction to MAB problems:}  The dueling bandits problem can be interpreted as a symmetric zero-sum game between two players \citep{owen1982game}, in the sense that one player always pulls the ``left'' arm of the duel and the other always the ``right'' arm. The winner of the duel gains a reward of one and the other a loss of one. 
		Thus, using two classical MAB algorithms to determine the choice of a player, respectively, results in a conversion of the dueling bandits problem into some sort of a meta-MAB problem allowing the transfer of well-established theoretical guarantees.
		\item[--] \emph{Generalization of MAB algorithms:} At the heart of the PB-MAB problem is the estimation of the pairwise preference probabilities $q_{i,j},$ as these encode the quality of the arms in a similar manner as the rewards in the MAB problem.
		Thus, another natural approach, especially for the task of regret minimization, is to adapt the different high-level ideas of MAB algorithms revolving around an appropriate estimation of the rewards in order to strike a balance between exploration and exploitation for the pairwise preference probabilities.
		For this purpose, several well-established concepts are available: the optimism in face of uncertainty principle most prominently represented by UCB-type algorithms  \citep{AuCeFrSc02}, the probability matching heuristic underlying Thompson sampling \citep{Th33}, the value-based racing task \citep{MaMo94,MaMo97}, and the explore-then-exploit principle \citep{robbins1952some}.
		%	
		%	\item[--] \editvik[\emph{Usage of online optimization  algorithms:} Under specific assumptions on the data-generating process of the preference relation $\bQ$ as well as on the set of arms, it is possible to cast the problem into a classical online optimization problem for which powerful online learning algorithms are available \citep{shalev2012online,hazan2016introduction}, if the problem is for instance of a convex nature.]
		%	
	\end{enumerate}

	\subsubsection{Online Optimization-based Algorithms}
	Under specific assumptions on the data-generating process as determined by the preference relation $\bQ$ as well as on the set of arms, it is possible to cast the problem as a classical online optimization problem. For problems of that kind, powerful online learning algorithms are available \citep{shalev2012online,hazan2016introduction}, which typically exploit properties such as convexity of the target (function).
	
%	\editvik[
	
	\subsubsection{Noisy-sorting Algorithms}
	
	If the target of the learner is to provide a ranking over the arms, the most natural approach would be to sort the arms according to their optimality\footnote{As already mentioned in Section \ref{subsec_targets}, it is not obvious how to define optimality of an arm in general. We ignore this problem for the time being.}, giving rise to the class of noisy-sorting algorithms \citep{AiChNe05}.
	The active sampling strategies underlying such algorithms mimic the behavior of efficient sorting algorithms, such as merge sort or quicksort. 
	%\cite{hoare1962quicksort} 
	The main difference between deterministic sorting and these sampling strategies is that, due to the assumed stochasticity of the observed feedback, the order between two arms can only be determined with a certain probability and requires repeated comparisons. 
	Moreover, to guarantee representativeness of the observed comparisons for the target ranking, one has to require certain regularity assumptions, as will be thoroughly discussed in the subsequent sections.
%	]

%	\editvik[
	
	\subsubsection{Tournament Algorithms}
	
	Another approach for the task of finding an optimal arm or  to identify the top-$k$  arms is based on the concept of tournament systems as commonly considered in sports and gaming.
	Just like in the various sports disciplines, there are different types of tournament styles a tournament algorithm can employ. 
	Yet, from a high-level point of view, all these algorithms are proceeding as follows. 
	First, all arms are divided into groups, and duels are successively conducted among the arms within one group.
	This phase is usually called a ``round''. At the end of each round, the size of the group is diminished by discarding the inferior arms, and some of the groups are merged.
	The procedure of rounds is repeated until the size of the target is reached through successive elimination.
	The main differences for algorithms of this type lie in the decomposition of the arms into groups (the group sizes), the order in which duels within a group are played, the duration of a round, and the merging procedure for the groups after each round.
%	]
	
	%\editeyke[Shouldn't we have at least one reference?]
	%\editvik[I haven't found a good one yet ...]
	
%	\editvik[
	\subsubsection{Challenge Algorithms} \label{subsub_challenge_algos}
	
	Just like tournament algorithms are inspired by tournament systems in sports, challenge algorithms are inspired by the challenge system most prominently represented by the World Chess Championship.
	Transferring the idea to the dueling bandits problem, an algorithm of the challenge-type keeps a reference arm (the current champion) as well as a set of comparison arms (the challengers), and compares the reference arm systematically with different comparison arms. This is done until either a challenger is defeated by the reference arm and discarded from the set of comparison arms, or the reference arm is defeated by a comparison arm, whereupon the latter becomes the new reference arm.
	In contrast to the challenge system in sports, it is allowed that the challenger arm might vary in each round and is not fixed for a certain number of comparisons with the reference arm.
%	
%	In particular, the reference arm might be dueled with a comp
%	]
	
	%#################################
	\section{Learning from Coherent Pairwise Comparisons}\label{sec:assonp}
	
	As explained in Section \ref{sec:ppe}, learning in the PB-MAB setting essentially boils down to estimating  the pairwise preference matrix $\bQ$, i.e., the pairwise probabilities $q_{i,j}$. Usually, however, the target of the agent's prediction is not the relation $\bQ$ itself, but the best arm or, more generally, a ranking $\succ$ of all arms $\cA$. 
	As discussed in Section \ref{subsec_targets}, the target might not be well-defined for a given preference relation $\bQ$, so that information about the latter may not be indicative of the former.  
	Hence, the pairwise probabilities $q_{i,j}$ should be sufficiently coherent, so as to allow the learner to approximate and eventually identify the target (at least in the limit when the sample size grows to infinity).  
	%Consequently, the least assumption to be made is a connection between $\bQ$ and $\succ$, so that information about the former is indicative of the latter. 
	%Or, stated differently, the pairwise probabilities $q_{i,j}$ should be sufficiently coherent, so as to allow the learner to approximate and eventually identify the target (at least in the limit when the sample size grows to infinity). 
	%For example, if the target is a ranking $\succ$ on $\cA$, then the $q_{i,j}$ should be somehow coherent with that ranking, e.g., in the sense that $a_i \succ a_j$ implies $q_{i,j} > 1/2$. 
	
	%While this is only an example of a consistency property that might be required, different consistency or regularity assumptions on the pairwise probabilities $\bQ$ have been proposed in the literature---needless to say, these assumptions have a major impact on how PB-MAB problems are tackled algorithmically. 
	Different consistency or regularity assumptions on the pairwise probabilities $\bQ$ have been proposed in the literature. Needless to say, these assumptions have a major impact on how PB-MAB problems are tackled algorithmically.
	In this section and the two following ones, we provide an overview of approaches to such problems, categorized according to these assumptions (cf.\ Figure \ref{fig:stpbmab}). 
	
	%%%%%%%%%%%%%%%%%%%%%%%%%%%%%%%%%%%%%%%%%%%%%%%%%%%%%%%%%%%%%%%%%%%%%%%%%%%%%%%%%%%%%%%%%%%%%%
	%% Tiks picture
	%%%%%%%%%%%%%%%%%%%%%%%%%%%%%%%%%%%%%%%%%%%%%%%%%%%%%%%%%%%%%%%%%%%%%%%%%%%%%%%%%%%%%%%%%%%%%%
	
	\tikzset{
		basic/.style  = {draw, drop shadow, font=\sffamily, rectangle},
		root/.style   = {basic, text width=6cm, rounded corners=1.5pt, thin, align=center,
			fill=gray!30},
		level 2/.style = {basic, rounded corners=4pt, thin,align=center, fill=blue!30,
			text width=9em},                   
		level 22/.style = {basic, rounded corners=5pt, thin,align=center, fill=red!30,
			text width=11.5em},                   
		level 3/.style = {basic, thin, align=left, fill=white!30, text width=14em}
	}
		\begin{figure}
		\centering
		\resizebox {.83\textwidth} {!} {
			\begin{tikzpicture}[
				level 1/.style={sibling distance=85mm},
				edge from parent/.style={->,draw},
				>=latex]
				
				% root of the the initial tree, level 1
				\node[root] {\normalsize Preference-based (stochastic) MAB}
				% The first level, as children of the initial tree
				child {node[level 2] (c1) {\small Coherent $\bQ$\\ Section \ref{sec:assonp} }}
				child {node[level 2] (c2) {\small Arbitrary $\bQ$ \\ Section \ref{sec:noass} }};
				%child {node[level 2] (c3) {Reduction methods \\ Section \ref{sec:reductions} }};
				
				% The second level, relatively positioned nodes
				\begin{scope}[every node/.style={level 3}]
					% Axiomatic approaches
					\node [level 22, below of = c1, xshift=20pt,  yshift=-10pt] (c101) {\footnotesize Axiomatic approaches \\ Section \ref{subsec:regaxiom} };
%					\node [below of = c101, xshift=60pt] (c11) {\scriptsize Interleaved filtering \citep{YuBrKlJo12} };
%					\node [below of = c11] (c12) {\scriptsize Beat the mean \citep{YuJo11} };
%					\node [below of = c12] (c13) {\scriptsize Relative Upper Confidence Bound \citep{ZoWhMuDe14}};
%					\node [below of = c13, yshift=-5pt] (c14) {\scriptsize Relative confidence sampling \citep{ZoWhDeMu14}};
%					\node [below of = c14] (c15) {\scriptsize MergeRUCB \citep{ZoWhDe15}};
%					\node [below of = c15, yshift=-5pt] (c16) {\scriptsize Relative minimum empirical divergence \citep{KoHoKaNa15}};
%					\node [below of = c16] (c17) {\scriptsize Verification based solution \citep{Ka16}};
%					\node [below of = c17] (c18) {\scriptsize Winner stays algorithms \citep{ChFr17}};
%					\node [below of = c18, yshift=-5pt] (c19) {\scriptsize Knockout tournaments \citep{FaOrPiSu17}};
%					\node [below of = c19, yshift=-5pt] (c110) {\scriptsize Binary-Search-Ranking \citep{FaOrPiSu17,falahatgar2018limits}};
%					\node [below of = c110, yshift=-5pt] (c111) {\scriptsize Single elimination tournament \citep{MoSuEl17}};
%					\node [below of = c111, yshift=-5pt] (c112) {\scriptsize Sequential elimination \citep{FaHaOrPiRa17}};
%					\node [below of = c112] (c113) {\scriptsize Opt-Max \citep{falahatgar2018limits}};
%					\node [below of = c113] (c114) {\scriptsize Approx-Pro \citep{falahatgar2018limits}};
%					\node [below of = c114] (c115) {\scriptsize Merge-DTS \citep{li2020mergedts}};
%					\node [below of = c115] (c116) {\scriptsize Beat-The-Winner  \citep{pekoez_ross_zhang_2020}};
					
					% Utility functions
					\node [level 22, below of = c101,  yshift=-10pt] (b101) {\footnotesize Utility functions \\ Section \ref{subsec_utility_dueling_bandits}};
%					\node [below of = b101, xshift=60pt] (b11)  {\scriptsize Dueling Bandit Gradient Descent \citep{YuJo09} };
%					\node [below of = b11, yshift=-5pt] (b12)  {\scriptsize Multi-Point Deterministic Gradient Descent \citep{ZhKi16}};
%					\node [below of = b12, yshift=-10pt] (b13) {\scriptsize 1. Doubler \citep{AiKaJo14} 2.MultiSBM \citep{AiKaJo14} };
%					\node [below of = b13, yshift=-10pt] (b14)               {\scriptsize Multisort \citep{MaGr17} };
%					\node [below of = b14, yshift=-5pt] (b15) {\scriptsize Winner stays \citep{ChFr17}};
%					\node [below of = b15, yshift=-5pt] (b16) {\scriptsize Dueling Bandits Temporary Elimination Algorithm \citep{ZiSe18}};
					
					% Statistical models
					\node [level 22, below of = b101,  yshift=-10pt] (d101) {\footnotesize Statistical models  \\ Section \ref{sec:mall}  };
				\end{scope}
				
				% lines from each level 1 node to every one of its "children"
%				\foreach \value in {1,...,16}
%				\draw[->] (c101.195) |- (c1\value.west);
%				
%				\foreach \value in {1,...,6}
%				\draw[->] (b101.195) |- (b1\value.west);
%				
%				\foreach \value in {1,2}
%				\draw[->] (d101.195) |- (d1\value.west);
%				
%				\foreach \value in {1,...,12}
%				\draw[->] (c2.195) |- (c2\value.west);
				
				%\foreach \value in {1,...,3}
				%  \draw[->] (c3.195) |- (c3\value.west);
				
				\draw[->] (c1.195) |- (c101.west);  
				\draw[->] (c1.195) |- (b101.west);  
				\draw[->] (c1.195) |- (d101.west);  
				
			\end{tikzpicture}
		}
		\caption{A taxonomy of (stochastic) PB-MAB algorithms.}
		\label{fig:stpbmab}
	\end{figure}

	\subsection{Axiomatic Approaches}\label{subsec:regaxiom}
	We begin this section by collecting various assumptions on pairwise preferences that can be found in the literature. As will be seen later on, by exploiting the (preference) structure imposed by these assumptions, the development of efficient algorithms will become possible.

	\begin{itemize}
		\item \emph{Low noise model}: $\Delta_{i,j}\neq 0$ for all $i\neq j$, and if $\Delta_{i,j}>0,$ then $\sum_{k=1}^K \Delta_{i,k} > \sum_{k=1}^K \Delta_{j,k}.$
		%	There is a total order $\succ$ on $\cA$, and there exists a universal constant $\alpha\in(0,\frac12)$ such that for any pair of arms with $a_{i} \succ a_{j}$, the pairwise probabilities satisfy $\Delta_{i,j} > \alpha $. 
		%	
		\item \emph{Total order over arms}: There is a total order $\succ$ on $\cA$, such that $a_{i} \succ a_{j}$ implies $\Delta_{i,j}>0$.  The existence of a total order over arms is closely related to different regularity assumptions for triplet of arms, including a stochastic version of the triangle inequality or relaxed notions of transitivity \citep{haddenhorst}:
		\begin{itemize}
			\item[--]  \emph{Strong stochastic transitivity $(\mathrm{SST})$}: The inequality $\Delta_{i,k} \ge \max \left\{ \Delta_{i,j} , \Delta_{j,k} \right\} $ holds for all pairwise distinct $i,j,k \in [K]$ such that $\Delta_{i,j}\geq 0$ and $\Delta_{j,k}\geq 0.$
			\item[--]  \emph{$\gamma$-relaxed stochastic transitivity $(\gamma-\mathrm{RST})$}: For $\gamma \in (0,1)$ and all pairwise distinct $i,j,k \in [K]$, such that $\Delta_{i,j}\geq 0$ and $\Delta_{j,k}\geq 0,$ the inequality $  \Delta_{i,k} \ge \gamma  \, \max \left\{ \Delta_{i,j}, \Delta_{j,k} \right\} $ holds.
			\item[--] \emph{Moderate stochastic transitivity $(\mathrm{MST})$}: The calibrated pairwise probabilities satisfy $  \Delta_{i,k} \ge \min \left\{ \Delta_{i,j}, \Delta_{j,k} \right\} $  for all pairwise distinct $i,j,k \in [K]$ such that $\Delta_{i,j}\geq 0$ and $\Delta_{j,k}\geq 0.$
			\item[--]  \emph{Weak stochastic transitivity $(\mathrm{WST})$}: For any triplet of arms $a_i,a_j,a_k\in \cA$, $\Delta_{i,j} \ge 0$ and $\Delta_{j,k} \ge 0$ implies $\Delta_{i,k} \ge 0 $.
			
			\item[--] \emph{Stochastic triangle inequality $(\mathrm{STI})$}: Given a total order over arms, then for any triplet of arms such that $a_{i} \succ a_{j} \succ a_{k}$, it holds that $\Delta_{i,k} \le \Delta_{i,j} + \Delta_{j,k} $.
		\end{itemize}
		%
		
		%	\item[--] \emph{Strong stochastic transitivity (SST)}: For any triplet of arms such that $a_{i} \succ a_{j} \succ a_{k}$, the  pairwise probabilities satisfy $\Delta_{i,k} \ge \max \left( \Delta_{i,j} , \Delta_{j,k} \right) $.
		%%	
		%	\item[--] \emph{Relaxed stochastic transitivity (RST)}: There is a $\gamma \in (0,1]$ such that, for any triplet of arms such that $a_{i} \succ a_{j} \succ a_{k}$, the  pairwise probabilities satisfy $  \Delta_{i,k} \ge \gamma  \, \max \left\{ \Delta_{i,j}, \Delta_{j,k} \right\} $.
		%	\item[--] \editvik[\emph{Moderate stochastic transitivity (MST)}: For any triplet of arms such that $a_{i} \succ a_{j} \succ a_{k}$, the  pairwise probabilities satisfy $  \Delta_{i,k} \ge \min \left\{ \Delta_{i,j}, \Delta_{j,k} \right\} $.]
		%	\item[--] \editvik[\emph{Weak stochastic transitivity (WST)}: For any triplet of arms such that $a_{i} \succ a_{j} \succ a_{k}$ it holds that $ (\Delta_{i,j} \ge 0 \wedge  \Delta_{j,k} \ge 0) \Rightarrow \Delta_{i,k} \ge 0 $.]
		
		\item \emph{General identifiability assumption}: There exists an arm $i^* \in [K]$ such that  for any $j \in [K] \backslash \{ i^* \}$ it holds that $  \min_{k \in  [K]  } \Delta_{i^*,k} - \Delta_{j,k} >0. $ 
		
		\item \emph{No ties}: The preference relation $\bQ$ is said to have no ties, if $\Delta_{i,j} \neq 0$ for all pairs of distinct arms $(a_i,a_j)$.
		
		\item \emph{Existence of a Condorcet winner}: An arm $a_{i}$ is considered a Condorcet winner if $\Delta_{i,j} > 0 $ for all $j\in [K]\backslash\{i\}$, i.e., if it beats all other arms in a pairwise comparison.
		%	\item[--] \editvik[\emph{Weak stochastic transitivity}:] For any triplet of arms such that $a_{i} \succ a_{j} \succ a_{k}$, the  pairwise probabilities satisfy that if $\Delta_{i,j} > 0$ and $  \Delta_{j,k} > 0$ then also $\Delta_{j,k} > 0 $.	
		
		%	\item[--] \emph{Sorted-by-probabilities model}: For any pair of arms such that $a_{i} \succ a_{j}$, it holds that $\Delta_{i,j} > 0 $.
		
		%	\item[--] \emph{Specific structural constraints on the preference matrix:} We will see an example of such constraint in subsection \ref{subsubsec:se}.
	\end{itemize}

\begin{center}
	\begin{figure}[h]
		\makebox[\textwidth]{
			\begin{tikzpicture}[scale=0.9]
				\draw[rounded corners=15pt,draw=black] (4,11) rectangle ++(9,-9); % box around total order
				\draw[rounded corners=15pt,draw=black,fill=white] (0,10) rectangle ++(8.5,-7.25); % box around the stochastic transitivities
				\draw[draw=black,fill=white] (7.25,8) rectangle ++(4.5,-5.5); % box around STI
				%\draw[rounded corners=15pt,draw=black] (15,6.5) rectangle ++(2,-2); % box around WST 
				\draw[draw=black] (0.5,7.5) rectangle ++(2,1); % box around SST
				\draw[draw=black] (0.5,5.5) rectangle ++(2,1); % box around MST
				\draw[draw=black] (0.5,3.5) rectangle ++(2,1); % box around WST
				\draw[draw=black] (4.25,5.5) rectangle ++(2,1); % box around RST
				\draw[rounded corners=12pt,draw=black] (4,13.5) rectangle ++(9,-1); % box around Uniform noise model
				\draw[rounded corners=12pt,draw=black] (4,1) rectangle ++(9,-1); % box around Condorcet 			
				\draw[rounded corners=12pt,draw=black] (1,1) rectangle ++(2,-1); % box around No ties 
				\draw[rounded corners=15pt,draw=black,fill=white] (6.5,6.5) rectangle ++(8.5,-2.5); % box around general identifiability
				\draw[thick,<-] (1.5,6.5)--(1.5,7.5); % arrow from SST to MSTs
				%		\draw[thick,<-] (7,3.5)--(7,5.5);
				\draw[thick,->] (2.5,8)--(5.25,6.5);% arrow from SST to \gamma RST
				\draw[thick,->] (5.25,5.5)--(2.5,4);% arrow from  \gamma RST to WST
				\draw[thick,->] (1.5,5.5)--(1.5,4.5);% arrow from MST to  WST
				%		\draw[thick,<-] (8,3)--(11,6.5);
				\draw[thick,<-] (8.5,11)--(8.5,12.5); % arrow from uniform noise to total order over arms
				%			\draw[thick,-] (13,13)--(14,13); % arrow from   total order over arms to  WST
				%			\draw[thick,->] (14,13)--(14,6.5); % arrow from   total order over arms to  WST
				\draw[thick,-] (14,4)--(14,0.5); % line from  general identifiability to Condorcet
				\draw[thick,->] (14,0.5)--(13,0.5); % arrow from  general identifiability to Condorcet
				\draw[thick,<-] (8.5,1)--(8.5,2); % arrow from   total order over arms to  Condorcet
				\draw[thick,<-] (2.8,1)--(6.5,2); % arrow from   total order over arms to  No ties
				
				%			\draw[thick,<-] (8.5,-1)--(8.5,0); % arrow from    general identifiability to Condorcet
				\draw[fill=black] (1.25,8) node {$\mathrm{SST}$};
				\draw[fill=black] (1.25,6) node {$\mathrm{MST}$};
				\draw[fill=black] (1.25,4) node {$\mathrm{WST}$};
				%\draw[fill=black] (16,5.5) node {$\mathrm{WST}$};
				\draw[fill=black] (5.25,6) node {$\gamma-\mathrm{RST}$};
				\draw[fill=black] (9.5,7) node {$\mathrm{STI}$};
				\draw[fill=black] (8.5,10.5) node {Total order over arms};
				\draw[fill=black] (4,9.5) node {Stochastic Transitivity};
				\draw[fill=black] (8.5,13) node {Low noise model}; 
				\draw[fill=black] (9,0.5) node {Condorcet winner};
				\draw[fill=black] (2,0.5) node {No Ties};
				\draw[fill=black] (11,5) node {General identifiability assumption};
				%\draw[fill=black] (9,0.5) node {General identifiability assumption};
				%\draw[fill=black] (8.5,-1.5) node {Condorcet winner};
			\end{tikzpicture} 
		}
		\caption{Relationship between the different axiomatic approaches.}
		\label{fig_axiom_relations_overview}
	\end{figure}
\end{center}
	%\editvik[Note that these properties only make sense if $|K|>2.$]
	Note that $\mathrm{WST}$ can also be formulated (as done by some authors) as follows: There is a ranking $\succ$ on $\cA$, such that $\Delta_{i,j}\geq 0$ whenever $a_i \succ a_j$.  Quite naturally, $\mathrm{WST}$ is a necessary and sufficient condition for the existence of a complete ranking of all arms, which is consistent with all pairwise preferences in the sense just mentioned.
		%
%	\editvik[Another regularity assumption, which is neither implied ]
		
	In Figure \ref{fig_axiom_relations_overview}, we give an overview of the relationships between the different assumptions, where an arrow represents that one condition implies another one.
	The implications are quite straightforward to prove, so that in the following we merely provide counterexamples showing why the reversed implications do not hold.
	However, it is worth noting that the low noise model assumption does not imply any of the triplet conditions within the total order assumption, such as $\mathrm{SST}$ or $\mathrm{STI}.$
	This is illustrated in Figure \ref{fig_axiom_relations_overview} by the angular rectangle around the triplet conditions.
	We start with the assumptions in the rounded rectangles in Figure \ref{fig_axiom_relations_overview}:
	\begin{itemize}
		\item Total order over arms $\centernot\implies  $ low noise model: Let 
		\begin{align*}
		\bQ =	\left(\begin{array} {cccc}
		0.5 & 1 & 1 & 1 \\
		0 & 0.5 & 0.6 & 0.6\\
		0 & 0.4 & 0.5 & 1 \\
		0 & 0.4 & 0 & 0.5
		\end{array}\right).
		\end{align*}
		Then, $a_1 \succ a_2 \succ a_3 \succ a_4$ is the total order for this preference relation matrix. 
		However, it holds that $\Delta_{2,3}>0$ but $ \sum_{k=1}^4 \Delta_{2,k} = -0.3 < -0.1 = \sum_{k=1}^{4} \Delta_{3,k}.$
		\item General identifiability assumption $\centernot\implies  $ total order over arms $\vee$  $\mathrm{SST}$: Let 
		\begin{align*}
		\bQ =	\left(\begin{array} {cccc}
		0.5 & 1 & 1 & 1 \\
		0 & 0.5 & 0.9 & 0.1\\
		0 & 0.1 & 0.5 & 0.9 \\
		0 & 0.9 & 0.1 & 0.5
		\end{array}\right).
		\end{align*}
		Then, for $i^*=1$, we have $\max_{j \in [K]\backslash\{1\}} \min_{k \in [K]} \Delta_{i^*,k} - \Delta_{j,k} = 0.1 >0,$ but since $\Delta_{2,3}>0$, $\Delta_{3,4}>0$, $\Delta_{2,4}<0$, there exists no total order for $\bQ$. 
		In addition,  $\mathrm{SST}$ does not hold because of $\Delta_{2,4} = -0.4 < 0.4 = \max\{\Delta_{2,3}, \Delta_{3,4} \}.$
		\item $\mathrm{STI}$ $\centernot\implies  $ general identifiability assumption: Let 
		\begin{align*}
		\bQ =	\left(\begin{array} {ccc}
		0.5 & 0.6 & 0.6  \\
		0.4 & 0.5 & 1 \\
		0.4 & 0  & 0.5 
		\end{array}\right).
		\end{align*}
		Then, there exists a total order $a_1 \succ a_2 \succ a_3$ and due to $\Delta_{1,3} = 0.1 \leq 0.6 = \Delta_{1,2} + \Delta_{2,3} $ the stochastic triangle inequality is fulfilled.
		However, it holds that  
		\begin{align*}
		& \min_{k=1,2,3} \Delta_{1,k} - \Delta_{2,k} = -0.4 < 0, \quad  
		  \min_{k=1,2,3} \Delta_{2,k} - \Delta_{1,k} = -0.1 <0   \\
		&  \min_{k=1,2,3} \Delta_{3,k} - \Delta_{1,k} = -0.6 <0,
		\end{align*}
		so that the general identifiability assumption does not hold.
		%	
		%	\item General identifiability assumption $\centernot\implies  $ WST: Let 
		%	
		\item  Total order over arms $\centernot\implies  $  $\mathrm{SST}$ : Consider the latter preference relation $\bQ.$ Then, there exists a total order of arms  but $\mathrm{SST}$  does not hold, due to  $\Delta_{1,3} = 0.1 < 0.5 = \max\{ \Delta_{1,2}, \Delta_{2,3} \}.$
		\item $\mathrm{WST}$ $\centernot\implies  $ total order over arms $\vee$ general identifiability assumption: Consider the preference relation $\bQ$ with all entries set to $0.5,$ then neither a total order of arms exists nor the general identifiability assumption holds, although $\mathrm{WST}$ is fulfilled.
		\item No ties $\centernot\implies  $ existence of a Condorcet winner: Let 
		\begin{align*}
			\bQ =	\left(\begin{array} {ccc}
				0.5 & 0.6 & 0.4  \\
				0.4 & 0.5 & 0.6 \\
				0.6 & 0.4  & 0.5 
			\end{array}\right).
		\end{align*} 
		Then $\bQ$ has no ties, but there exists no Condorcet winner, as each arm is beaten by one other arm.
	\end{itemize}
	We proceed with the non-obvious transitivity assumptions within the stochastic transitivity properties, namely the relation between MST and RST:
	\begin{itemize}
		\item  $\mathrm{MST}$ $\centernot\implies  $ $\gamma-\mathrm{RST}$: Let 	
		\begin{align*}
		\bQ =	\left(\begin{array} {ccc}
		0.5 & 0.5 + z  & 0.5 + x    \\
		0.5 - z  & 0.5 & 0.5 + y  \\
		0.5 - x  & 0.5 - y   & 0.5 
		\end{array}\right),
		\end{align*}
		where $x,y,z\in (0,1/2)$ are such that $z\leq \min\{x,y\}$ and $x/y <\gamma.$
		With this, it holds that $a_1 \succ a_2 \succ a_3$ is a total order for $\bQ$. Moreover, $\mathrm{MST}$ holds because of  $\Delta_{1,3}=x \geq z  = \min\{y,z\} = \min\{\Delta_{1,2},\Delta_{2,3}\}.$
		Yet, $\Delta_{1,3}=x < \gamma  y = \gamma \max\{\Delta_{1,2},\Delta_{2,3}\},$ which is equivalent to $x/y <\gamma.$
		\item  $\gamma-\mathrm{RST}$ $\centernot\implies  $ $\mathrm{MST}$: For any $\gamma \in (0,1)$, there exists some $\delta>1$ such that $\gamma \leq 1/\delta.$
		Fix one such $\delta,$ then for any $x\in(0,1/(2\delta))$, the preference relation given by 	
		\begin{align*}
		\bQ =	\left(\begin{array} {ccc}
		0.5 & 0.5 + \delta x  & 0.5 + x    \\
		0.5 - \delta x  & 0.5 & 0.5 + \delta x  \\
		0.5 - x  & 0.5 - \delta x   & 0.5 
		\end{array}\right),
		\end{align*}
		implies a total order of arms $a_1 \succ a_2 \succ a_3$ and  $\gamma-\mathrm{RST},$ since $\Delta_{1,3} = x \geq \gamma \delta x = \gamma \max\{\Delta_{1,2},\Delta_{2,3}\}.$
		On the other hand, $\mathrm{MST}$ is not fulfilled due to $\Delta_{1,3} = x < \delta x = \min\{\Delta_{1,2},\Delta_{2,3}\}.$
	\end{itemize}
	Finally, the stochastic triangle inequality is not implied by any transitivity assumptions stronger than $\mathrm{WST}$ or vice versa:
	\begin{itemize}
		\item  $\mathrm{SST}$ $\centernot\implies  $ $\mathrm{STI}$: For some $\varepsilon\in (0,1/6)$ let
		\begin{align*}
		\bQ =	\left(\begin{array} {ccc}
		0.5 & 0.5 + \varepsilon & 0.5 + 3 \varepsilon   \\
		0.5 - \varepsilon  & 0.5 & 0.5 + \varepsilon  \\
		0.5 - 3\varepsilon & 0.5 - \varepsilon  & 0.5 
		\end{array}\right).
		\end{align*}
		A total order for $\bQ$ exists, namely $a_1 \succ a_2 \succ a_3$, and $\mathrm{SST}$ is fulfilled, as $\Delta_{1,3} = 3\varepsilon \geq  \varepsilon = \max\{\Delta_{1,2},\Delta_{2,3}\}$. 
		However, $\Delta_{1,3} = 3\varepsilon > 2\varepsilon = \Delta_{1,2} +  \Delta_{2,3},$ so that $\mathrm{STI}$ is violated.
		\item  $\mathrm{STI}$ $\centernot\implies  $ $\mathrm{MST}$: For some $\varepsilon\in (0,1/2)$ let
		\begin{align*}
		\bQ =	\left(\begin{array} {ccc}
		0.5 & 1 & 0.5 +  \varepsilon   \\
		0  & 0.5 & 1  \\
		0.5 - \varepsilon & 0  & 0.5 
		\end{array}\right).
		\end{align*}
		Then, $a_1 \succ a_2 \succ a_3$ is a total order for $\bQ$, and $\mathrm{STI}$ is satisfied, since 
		$\Delta_{1,3} = \varepsilon \leq 1 = \Delta_{1,2} +  \Delta_{2,3}.$ 
		$\mathrm{MST}$ is not present, because of $\Delta_{1,3} = \varepsilon < 0.5= \min\{\Delta_{1,2},\Delta_{2,3}\}$
		\item  $\mathrm{STI}$ $\centernot\implies  $ $\gamma-\mathrm{RST}$: Consider the preference relation as just defined and set $\varepsilon=\gamma/4.$
		Then, $\gamma-\mathrm{RST}$ does not hold due to $\Delta_{1,3} = \gamma/4 < \gamma/2 = \gamma \max\{\Delta_{1,2},\Delta_{2,3}\}.$
	\end{itemize}
	%
	%Note that the assumption of a total order with arms separated by positive margins ensures the existence of a unique best arm, which in this case coincides with the Condorcet winner. 
	%Also note that strong stochastic transitivity is recovered from the relaxed stochastic transitivity for $\gamma = 1$.
	%
%	At last, note that these concepts are in fact not restricted to the case of finitely many arms but can be extended to the case of an infinite number of arms. 
%	%
%	Indeed, let $\cS$ be some space of arms, which is not necessarily finite\footnote{This space corresponds to our set of arms $\cA$. However, as we assume $\cA$ to be finite, we use another notation here. }, but such that $(\cS,\mathbb{S},\mu)$ is a measure space.
%	% 
%	Then, an extension of the low  noise assumption can be defined as follows: For all $a,\tilde a \in \cS$ with $a \neq \tilde a$, it holds that $\Delta_{a,\tilde a}\neq 0$, and if $\Delta_{a,\tilde a}>0$, then 
%	%
%	$$	\int_{\cS} \Delta_{a,a'}  \mathrm{d}\mu(a') > 	\int_{\cS} \Delta_{\tilde a,a'}  \, \mathrm{d}\mu(a').
%	$$
%	%
%%	for any weighting function $w:\cS \to \R_+$ for which the latter integrals are well defined.
%	%
%	It is straightforward to extend the other axiomatic properties to the case of a possibly infinite number of arms and deduce a relationship between these as in Figure \ref{fig_axiom_relations_overview}. 
	
	In the following, we review all known methods of the preference-based multi-armed bandit literature for the axiomatic approaches.
	%in a chronological order. 
	%
	In particular, the underlying assumptions, the considered goals, the theoretical guarantees (if known), as well as the accompanying novelties of the corresponding methods are concisely described and discussed. 
	Note that the order in which the algorithms are discussed is primarily according to the underlying target and secondarily according to their algorithmic ideas. 
	Finally, Tables \ref{tab:regaxiom}, \ref{tab:regaxiom_second} and \ref{tab:regaxiom_third} provide a concise overview of the existing methods for the tasks of regret minimization, $(\epsilon,\delta)$-PAC learning as well as minimization of the exact sample complexity.
	
	\begin{table}
		\scriptsize
		\begin{center}		
%						   {|p{2.5cm}|p{2cm}|p{2.25cm}|p{2.5cm}|p{5cm}|}
			\begin{tabular}{|p{1.5cm}|p{1.75cm}|p{2.25cm}|p{2.5cm}|p{5cm}|}
				\hline
				\textbf{Algorithm} & \textbf{Algorithm class(es)} & \textbf{Assumption(s)} & \textbf{Target(s) and goal(s) of learner} & \textbf{Theoretical guarantee(s)} \\
				\hline
				\hline
				Interleaved Filtering \newline (Section \ref{subsec_IF_algorithm})
%				\citep{YuBrKlJo12} 
				&  Generalization (Explore-then-exploit), Challenge 
				&			A priori known time horizon $T$, Total order over arms, SST and STI 
				%			Total order over arms, strong stochastic transitivity and stochastic triangle inequality &
				& Expected regret minimization
				& $\hphantom{text}$ \newline $\bigO \left( \frac{K \log T}{\min_{j \neq i^{*}} \Delta_{i^{*},j}  }  \right)$ \\
				\hline
				Beat the Mean \newline(Section \ref{subsec_btm_algorithm})
%				\citep{YuJo11} 
				&   Generalization (Explore-then-exploit), Challenge
				&	A priori known time horizon $T$, Total order over arms, RST and STI
				&  High probability regret  minimization
				& $\hphantom{text}$ \newline $\bigO \left( \frac{ K  \log T }{\gamma^{7} \min_{j \neq i^{*}} \Delta_{i^{*},j}  }\right) $ 				
				\\
				\hline
				Relative Upper Confidence Bound \newline (Section \ref{subsec_RUCB})
				%			(RUCB)
%				\citep{ZoWhMuDe14} 
				& Generalization (UCB) 
				& Existence of a Condorcet winner 
				& Expected and high probability regret minimization 
				& $\hphantom{text}$ \newline $\bigO \left(K^2 + \sum_{i\neq i^{*}} \frac{\log T}{\Delta^2_{i^{*},i}} \right)$ \\
				\hline
				MergeRUCB \newline (Section \ref{subsec_mergeRUCB})
				%				 \citep{ZoWhDe15} 
				& Generalization (UCB), tournament 
				& No ties\footnote{This assumption is simplified here for sake of convenience. Refer to Section \ref{subsec_mergeRUCB} for the exact formulation of the assumption.} and existence of a Condorcet winner 
				& High probability regret minimization  
				& $\hphantom{text}$ \newline	$\mathcal{O}\left(\frac{K\log(T)}{\min_{i,j:\, q_{i,j\neq 1/2}} \Delta_{i,j}^2  }  \right)$ \\
				\hline
				MergeDTS  \newline (Section \ref{sec_merge_dts})
				%				\citep{li2020mergedts} 
				& Generalization (ThompsonSampling), tournament 
				& A priori known time horizon $T$ and same assumptions as for MergeRUCB
				& High probability regret minimization 
				& $\hphantom{text}$ \newline	$\mathcal{O}\left(\frac{K\log(T)}{ \min_{i,j:\, q_{i,j\neq 1/2}} \Delta_{i,j}^2  }  \right)$ \\
				\hline
				Relative Confidence Sampling  \newline (Section \ref{subsec_RCS})
%				\citep{ZoWhDeMu14} 
				%			(RCS)
				& Generalization (UCB and Thompson sampling), tournament  
				& Existence of a Condorcet winner 
				& Expected and high probability regret minimization   
				& No theoretical guarantees \\
				\hline

				Relative Minimum Empirical Divergence  \newline (Section \ref{subsubsec:rmed})
%				\citep{KoHoKaNa15} 
				& Generalization (DMED)
				& Existence of a Condorcet winner 
				& Expected regret minimization
				&$\hphantom{text}$ \newline $\mathcal{O}\left( \sum_{i \neq i^* } \frac{ \Delta_{i^*,i} \, \log T   }{\KL(q_{i,i^{*}},1/2)}  + K^{2+\varepsilon}\right)$ \\
				\hline
				%			CondorcetSAVAGE \citep{UrClFeNa13} & 
				%			%			Total order over arms, relaxed stochastic transitivity and stochastic triangle inequality both relative to the best arm 
				%			Total order over arms, RST and STI
				%			& 1.Best arm \ 2.Best arm (PAC) & High probability regret and sample complexity minimization in the PAC setting  &
				%			1. $\bigO \left( \frac{ K  \log T }{\gamma^{7} \min_{j \neq i^{*}} \Delta_{i^{*},j}  }\right) $ \
				%			2. $\bigO \left( \frac{K }{ \gamma^6 \epsilon^2 } \log \frac{K}{\epsilon \delta} \right) $
				%			\\
				%			\hline
				%			
				%
				Winner Stays \newline (Section \ref{sec_winner_stays_algo})
%				\citep{ChFr17} 
				& Challenge 
				& No ties and either \newline 
				1. Existence of a Condorcet winner, or \newline 
				2. Total order over  arms 
				&  Expected regret minimization with 
				\newline (a) weak regret 
				\newline (b) strong regret
				& 1.(a): $ \mathcal{O}\left( \frac{K^2}{\min_{i,j} |\Delta_{i,j}|^3 } \right)$ \newline
				1.(b):   $ \mathcal{O}\left( \frac{K^2 }{\min_{i,j} \Delta_{i,j}^2 } + \frac{K \log T}{\min_{i,j} |\Delta_{i,j}| }\right)$ \newline
				2.(a): $ \mathcal{O}\left( \frac{K \log K}{\min_{i,j} \Delta_{i,j}^6 } \right) $ \newline
				2.(b): $ \mathcal{O}\left( \frac{K \log K}{\min_{i,j} \Delta_{i,j}^6 } + \frac{K \log T}{\min_{i,j} |\Delta_{i,j}| }\right) $
				\\
				\hline
%				\color{blue}
				Beat the Winner \newline (Section \ref{sec_beat_the_winner})                                                              
				%				\citep{pekoez_ross_zhang_2020} 
				& Challenge 	
				& Existence of a Condorcet winner 
				&  Expected regret minimization with weak regret
				& $\hphantom{text}$ \newline $ \mathcal{O}\left(  K^2 + \frac{K}{(1-\exp(-2\min_{j\neq i^*} |\Delta_{j}|^2))^2}  \right) $ 
				\\ 
				\hline	
			\end{tabular}
		\end{center}
				\caption{ Algorithms for the regret minimization task under axiomatic approaches. The index $i^{*}$ is representing the best arm and $\KL(p,q)$ is the Kullback-Leibler divergence of Bernoulli random variables with parameters $p$ and $q.$ }
				\label{tab:regaxiom}
%		}
	\end{table}
	\normalsize

	\begin{table}	
		\scriptsize
		\begin{center}		
			\begin{tabular}{|p{2.5cm}|p{1.75cm}|p{2.25cm}|p{2.5cm}|p{4cm}|}
%				{|p{2.5cm}|p{2cm}|p{2.25cm}|p{2.5cm}|p{5cm}|}
				\hline
				\textbf{Algorithm} & \textbf{Algorithm class(es)} & \textbf{Assumption(s)} & \textbf{Target(s) and goal(s) of learner} & \textbf{Theoretical guarantee(s)} \\
				\hline
				\hline	
				Beat the Mean \newline(Section \ref{subsec_btm_algorithm})
				%				\citep{YuJo11} 
				& Challenge
				& Total order over arms, $\gamma$-RST and STI
				& $(\epsilon,\delta)$-PAC for best arm   
				& $\hphantom{text}$ \newline $\bigO \left( \frac{K }{ \gamma^6 \epsilon^2 } \log \frac{K}{\epsilon \delta} \right) $
				\\
				\hline
				Knockout Tournaments \newline (Section \ref{subsec_Knockout_tournaments})
				%				\citep{FaOrPiSu17} 
				& Tournament 
				&			Total order over arms, SST, STI, $\gamma$-RST
				%			Strong stochastic transitivity and stochastic triangle inequality &
				&$(\epsilon,\delta)$-PAC for best arm  
				&
%					SST: \quad $ \bigO \left( \frac{K}{\epsilon^2} \log \frac{1}{\delta} \right)$ \newline
%				%			
%				$\gamma$-RST: 
				 $\hphantom{text}$ \newline $ \bigO \left( \frac{K }{\gamma^4 \epsilon^2} \log \frac{1}{\delta} \right)$ 
				\\
				\hline
				Sequential Elimination \newline (Section \ref{subsec_seq_elimination})
%				\citep{FaHaOrPiRa17} 
				& Challenge 
				& Total order over arms, SST 
				%			Strong stochastic transitivity & 
				
				&			$(\epsilon,\delta)$-PAC for best arm 
				%			Best arm and best ranking in the PAC setting
				&	$\hphantom{text}$ \newline	$ \bigO \left( \frac{K}{\epsilon^2} \log \frac{1}{\delta} \right)$  \\
				\hline
				Opt-Max \newline (Section \ref{opt-max})
%				\citep{falahatgar2018limits}
				& Tournament 	
				& Total order over arms, MST  
				& $(\epsilon,\delta)$-PAC for best arm  
				&	If 	$\delta\geq \min(1/K,\exp(-K^{1/4})):$	\newline $\bigO \left( \frac{K }{ \epsilon^2 } \log \frac{1}{ \delta} \right) $  
				\\ 
				\hline
				Binary-Search-Ranking \newline (Section \ref{sec:bsr})
				%				\citep{FaOrPiSu17,falahatgar2018limits} 
				& Tournament 
				&Total order over arms, SST, STI 
				&$(\epsilon,\delta)$-PAC for ranking 
				& 
%				If $\delta\leq 1/K:$ $\bigO \left( \frac{K (\log(K))^3 }{ \epsilon^2 } \log \frac{K}{ \delta} \right) $ \newline
				If $\delta\geq 1/K:$ $\bigO \left( \frac{K \log K}{ \epsilon^2 }  \right) $
				\\
				\hline	
				Noisy Quick Select \newline (Section \ref{sec_Quick-Select-based})
%				: \newline
%				1. Epsilon Quick Select \newline
%				2. Tournament $k$-Selection \newline (Section \ref{sec_Quick-Select-based}) \newline
				%				
				& Noisy-sorting, Tournament	
				& Total order over arms, SST, STI
				& ($\epsilon$,$\delta$)-PAC for Top-$k$ identification (for $1\leq k\leq K/2$)
				& $\hphantom{text}$ \newline 
				$\bigO \left( \frac{K }{ \epsilon^2 } \log \frac{k}{ \delta} \right) $ 
				\\ 
				\hline	
				Approx-Prob \newline (Section \ref{subsec_approx_prob})
%				\citep{falahatgar2018limits} 
				& Tournament 	
				& Total order over arms, SST and STI  
				& $(\epsilon,\delta)$-PAC for estimation of  $\bQ$ 
				& If $\delta\geq 1/K:$ \newline $\bigO \left( \frac{K  \min(K,1/\epsilon) \log(K) }{ \epsilon^2 } \right) $  
%				 \newline for $\delta\geq 1/K$
				%						
				\\ 
				\hline		
			\end{tabular}				
		\end{center}			
	\caption{ Algorithms for $(\epsilon,\delta)$-PAC tasks under axiomatic approaches.}
	\label{tab:regaxiom_second}
%		}
	\end{table}
	\normalsize

	\begin{table}
		\scriptsize
		\begin{center}		
			\begin{tabular}{|p{1.5cm}|p{1.75cm}|p{2.25cm}|p{2.5cm}|p{5cm}|}
%				{|p{2.5cm}|p{2cm}|p{2.25cm}|p{2.5cm}|p{5cm}|}
				\hline
				\textbf{Algorithm} & \textbf{Algorithm class(es)} & \textbf{Assumption(s)} & \textbf{Target(s) and goal(s) of learner} & \textbf{Theoretical guarantee(s)} \\
				\hline
				\hline	
				Robust Query Selection  \newline (Section \ref{subsec_robust_query_selection})
				%				\citep{ren2019sample} 
				& Generalization
				& Total order over arms and embedding of the arms into a $d$-dimensional Euclidean space
				&  Exact sample complexity for ranking
				&  $\hphantom{text}$ \newline  $\bigO\left( \frac{d}{\min_{1\leq i<j \leq K}  \Delta_{i,j}^2} \log^2\left(\frac{K}{\delta}\right) \right)$
				\\ 
				\hline
				Verification-based Solution \newline (Section \ref{subsubsec:verification_based})
				%				 \citep{Ka16}
				& Generalization (exploration and verification) 
				& Existence of a Condorcet winner 
				&   Exact sample complexity for best arm 
				&	 {\tiny$\mathcal{O} \Big( \sum_{i \neq i^*} \min_{j: q_{i,j}<1/2} \frac{\log \big(K/(\delta \Delta_{i,j}^2)\big)}{\Delta_{i,j}^2} \Big)   $ \newline 	$\hphantom{text}$ $  + \tilde{\mathcal{O}} \Big( \sum_{i \neq i^* } \Big( \Delta_{i^*,i}^{-2}  +  \sum_{j \neq i }  \Delta_{i,j}^{-2}  \Big) \Big)$} \\
				\hline
				Parallel Selection and Partition  \newline (Section \ref{sec_parallel_select})
				%				\citep{ren2019sample} 
				&  Noisy-sorting
				&  Total order over arms
				&  Exact sample/round complexity for top-$k$ identification
				& $\hphantom{text}$ \newline  $\bigO\left( \frac{K}{\min_{1\leq i<j \leq K} \Delta_{i,j}^2} \log(K) \right)$
				\\ 
				\hline
				Single Elimination Tournament \newline (Section \ref{subsec_single_eliminiation_tournament})
				%				\citep{MoSuEl17} 
				& Noisy-sorting, tournament
				& Total order over arms
				&  Exact sample complexity for \newline 1. Best arm  \newline 2. Top-$k$ ranking \newline 3. Top-$k$ identification  
				&  1.	$\bigO \left( \frac{K  \log \log(K) }{ \min_{j \neq i^{*}} \Delta_{i^{*},j}^2 } \right) $  \newline
				2. $ \bigO \left( \frac{ (K+k\log k) \max\{ \log k, \log \log K \} }{\min_{i \in [k]} \min_{j:j\geq i} \Delta_{i,j}^2} \right) $\newline 3. $ \bigO \left( \frac{ (K+k\log k) \max\{ \log k, \log \log K \} }{ \Delta_{(k),(k+1)}^2} \right) $ 
				\\
				\hline
				Sequential Elimination Exact Selection \newline (Section \ref{sec_Sequential-Elimination-Exact-Selection})
				%				\citep{ren20a} 
				%				\newline 
				%				\Algo{Sequential Elimination Exact $k$ Selection} \citep{ren20a} 
				& Noisy-sorting, challenge
				& Total order over arms, SST, STI
				&  Exact sample complexity for \newline 
				1. Best arm \newline
				2. Top-$k$ identification 
				& 1.  $\bigO  \left(  \sum_{i \in [K]} \frac{\log 1/\delta + \log \log 1/\Delta_{i,(k)}}{\Delta_{i,(k)}^2}  \right)$ \newline
				2. $\bigO  \left(  \sum_{i \in [K]} \frac{\log K/\delta + \log \log 1/\Delta_{i,(k)}}{\Delta_{i,(k)}^2}  \right)$
				\\ 
				\hline
				Iterative-Insertion-Ranking \newline (Section \ref{sec_iterative_insertion_ranking})
				%				\citep{ren2019sample} 
				& Noisy-sorting 	
				& Total order over arms, SST
				&  Exact sample complexity for ranking
				& $\hphantom{text}$ \newline {\tiny $\bigO \left( \sum_{i\in [K]} \frac{ (\log\log(\min_{j\neq i} \Delta_{i,j}^{-1}) + \log(K/\delta) )}{\min_{j\neq i} \Delta_{i,j}^{2}} \right)$}
				%		ren20a	
				\\ 
				\hline
			\end{tabular}
		\end{center}
	\caption{ Algorithms for exact sample complexity tasks (corresponding to the $(0,\delta)$-PAC learning scenario) under axiomatic approaches. The definitions of $\Delta_{i,(k)}$ and $\Delta_{(k),(k+1)}$ can be found in the respective sections.}
\label{tab:regaxiom_third}
		%		}
	\end{table}
	\normalsize

	\subsubsection{Interleaved Filtering} \label{subsec_IF_algorithm}
	Assuming a total order over arms, strong stochastic transitivity, and the stochastic triangle inequality, \citet{YuBrKlJo12} propose an explore-then-exploit algorithm. The exploration step consists of a simple sequential elimination strategy, called \Algo{Interleaved Filtering} (\Algo{IF}), which identifies the best arm with probability at least $1-\delta$. The \Algo{IF} algorithm successively selects an arm which is compared to other arms in a one-versus-all manner. More specifically, the currently selected arm $a_{i}$ is compared to the rest of the active (not yet eliminated) arms. If an arm $a_{j}$ beats $a_{i}$, that is, $\widehat{q}_{i,j} + c_{i,j} < 1/2$, then $a_{i}$ is eliminated, and $a_{j}$ is compared to the rest of the (active) arms, again in a one-versus-all manner. In addition, a simple pruning technique can be applied: if $\widehat{q}_{i,j} - c_{i,j} > 1/2$ for an arm $a_{j}$ at any time, then $a_{j}$ can be eliminated, as it cannot be the best arm anymore (with high probability) due to the underlying transitivity assumption. After the exploration step, the exploitation step simply takes the best arm $a_{\widehat{i}^{*}}$ found by $\Algo{IF}$ and repeatedly compares $a_{\widehat{i}^{*}}$ to itself. 
	
	The authors analyze the expected regret achieved by \Algo{IF}. Assuming the horizon $T$ to be finite and known in advance, they show that \Algo{IF} incurs an expected regret of order $\mathcal{O} \left( \frac{K}{\min_{j \neq i^{*}} \Delta_{i^{*},j}  } \log T \right),$ which is shown to be the lower bound in this case as well.
	%\[ 
	%\exptd \left[ R^{T}_{\Algo{IF}} \right] = \bigO \left( \frac{K}{\min_{j \neq i^{*}} \Delta_{i^{*},j}  } \log T \right) \enspace .
	%\]
	
	\subsubsection{Beat the Mean} \label{subsec_btm_algorithm}
	In a subsequent work, \citet{YuJo11} relax the strong stochastic transitivity property and only require relaxed stochastic transitivity for the pairwise probabilities. Further, both the relaxed stochastic transitivity and the stochastic triangle inequality are required to hold only relative to the best arm, i.e., only for triplets where $ i $ is the index of the best arm $a_{i^{*}}$.  
	
	With these relaxed properties, \citet{YuJo11} propose a preference-based online learning algorithm called \Algo{Beat-The-Mean} (\Algo{BTM}), which is an elimination strategy resembling \Algo{IF}. However, while \Algo{IF} compares a single arm to the rest of the (active) arms in a one-versus-all manner, \Algo{BTM} selects an arm with the fewest comparisons so far and pairs it with a randomly chosen arm from the set of active arms (using the uniform distribution). Based on the outcomes of the pairwise comparisons, a score $b_{i}$ is assigned to each active arm $a_{i}$, which is an empirical estimate of the probability that $a_{i}$ is winning in a pairwise comparison (not taking into account which arm it was compared to). The idea is that comparing an arm $a_{i}$ to the ``mean'' arm, which beats half of the arms, is equivalent to comparing $a_{i}$ to an arm randomly selected from the active set. One can deduce a confidence interval for the scores $b_{i}$, which allows for deciding whether the scores for two arms are significantly different. An arm is then eliminated as soon as there is another arm with a significantly higher score.

	In the regret analysis of \Algo{BTM}, a high probability bound is provided for a finite time horizon. More precisely, the regret accumulated by \Algo{BTM} is 
	$\bigO \left( \frac{ K}{\gamma^{7} \min_{j \neq i^{*}} \Delta_{i^{*},j}  } \log T \right)
	$
	with high probability. This result is stronger than the one proven for \Algo{IF}, in which only the expected regret is upper-bounded. Moreover, this high probability regret bound matches with the expected regret bound in the case $\gamma = 1$ (strong stochastic transitivity). 
	
	The authors also analyze the \Algo{BTM} algorithm in a PAC setting for finding the best arm, i.e., for any given $ \epsilon,\delta > 0 $, the algorithm should find an $\epsilon$-optimal arm (cf.\ Section \ref{subsec_near_opt_targets}) with probability at least $1-\delta,$ while keeping the number of overall duels as low as possible.
	It is shown that \Algo{BTM} is an $(\epsilon, \delta)$-PAC preference-based learner (by setting its input parameters appropriately) with a sample complexity of $\bigO ( \frac {K}{\gamma^{6} \epsilon^{2}} \log \frac{KN}{\delta} )$ if $N$ is large enough, that is, $N$ is the smallest positive integer for which $N=\left\lceil \frac{36}{ \gamma^{6} \epsilon^{2} } \log  \frac{K^{3} N }{\delta} \right\rceil $ holds. One may simplify this bound by noting that $N<N'=\left\lceil \frac{864 }{\gamma^{6}\epsilon^{2}} \log \frac{K}{\delta}\right\rceil$. Therefore, the sample complexity of BTM is 
	$\bigO \left( \frac{ K}{\gamma^{6} \epsilon^{2}} \log \frac{K \log (K / \delta)}{\delta \, \gamma \, \epsilon} \right).
	$
	
	%\subsubsection{Preference-based UCB}
	\subsubsection{Relative Upper Confidence Bound} \label{subsec_RUCB}
%	problem of minimizing the cumulative regret when the number of arms is large under the assumption that a Condorcet winner exists.
	In a work by \citet{ZoWhMuDe14}, the well-known \Algo{UCB} algorithm \citep{AuCeFi02} is adapted from the value-based to the preference-based MAP setup in order derive an algorithm minimizing the cumulative regret (see \eqref{eq:regret}). One of the main advantages of the proposed algorithm, called \Algo{RUCB} (for Relative UCB), is that only the existence of a Condorcet winner is required. The \Algo{RUCB} algorithm is based on the ``optimism in the face of uncertainty'' principle, which means that the arms to be compared next are selected based on the optimistic estimates of the pairwise probabilities, that is, based on the upper bounds $\widehat{q}_{i,j}+c_{i,j}$ of the confidence intervals. In an iteration step, \Algo{RUCB} selects the set of potential Condorcet winners for which all $\widehat{q}_{i,j}+c_{i,j}$ values are above $1/2$, and then selects an arm $a_{i}$ from this set uniformly at random. Finally, $a_{i}$ is compared to the arm $a_{j}$, where $j = \argmax_{\ell \neq i} \widehat{q}_{i,\ell}+c_{i,\ell}$, that may lead to the smallest regret, taking into account the optimistic estimates.
	
	In the analysis of the \Algo{RUCB} algorithm, horizonless bounds are provided, both for the expected and high probability regret. Thus, unlike the bounds for \Algo{IF} and \Algo{BTM}, these bounds are valid for each time step. Both the expected regret bound and high probability bound of \Algo{RUCB} are 
	%$\bigO (K^2 + K\log T).$
	$$\bigO \Big(K^2 + \sum_{i\neq i^{*}} \frac{\log T}{\Delta^2_{i^{*},i}} \Big).$$
	However, while the regret bounds of \Algo{IF} and \Algo{BTM} only depend on $\min_{j\neq i^{*}} \Delta_{i^{*},j}$, the constants are now of different nature, despite being still calculated based on the $\Delta_{i,j}$ values. Therefore, the regret bounds for \Algo{RUCB} are not directly comparable with those given for \Algo{IF} and \Algo{BTM}. Moreover, the regret bound for \Algo{IF} and \Algo{BTM} is derived based on the explore-and-exploit technique, which requires the knowledge of the horizon in advance, whereas regret bounds for \Algo{RUCB}, both high probability and expectation, are finite time bounds that hold for any time step $T$.

	\subsubsection{MergeRUCB} \label{subsec_mergeRUCB}
	\citet{ZoWhDe15}  consider the same problem as in the previous section, but with a special focus on learning scenarios in which the number of available arms is large.
	In order to keep the number of comparisons small, they propose the \Algo{MergeRUCB} algorithm which, using a similar divide-and-conquer strategy as the merge sort algorithm, proceeds by first grouping the arms in batches of a predefined size and then processing them separately before merging them together.
	In particular, only arms within the same batch can be compared with each other, but not arms in different batches.
	Due to the stochastic nature of the feedback, the local comparisons within each batch between two arms are run multiple times before eliminating inferior arms based on upper confidence bounds of the preference probabilities. 
	More precisely, an arm is eliminated within one batch if its upper confidence bound on the pairwise winning probability with respect to some arm in the same batch is below $1/2.$
%	
%	Due to the usage of upper confidence bounds 
%	 
	In each time step, \Algo{MergeRUCB} chooses one batch in a round-robin manner, while the choice for the arms compared within one batch is made by choosing one arm uniformly at random and comparing it with its worst competitor, i.e., the arm having the highest chance of leading to an elimination of the first chosen arm.
	In light of this, it is ensured that the batch sizes are reduced quickly, and in turn a merge step can be performed early, which happens as soon as the sum of the batch sizes are below a stage-wise geometrically decreasing threshold, while a stage corresponds to the overall number of conducted merge steps.
	Within one merge step, batches of smaller sizes are grouped together with batches of larger sizes. 
	This entire iterative process will eventually end with one single batch left, consisting of only one arm, which is then guaranteed to be the Condorcet winner with high probability.    

	For the theoretical analysis of \Algo{MergeRUCB}, it is assumed that for any pair of arms $(a_i,a_j)$ it holds that either their pairwise winning probability is different from $1/2$ (i.e., $\Delta_{i,j}\neq 0$), or they are inferior to any other arm $a_k$ in the sense that $\max\{\Delta_{i,k},\Delta_{j,k}\}<0$ holds.
	If the latter property holds for a pair of arms, then this pair is called \emph{uninformative}. 
	The authors assume that at most a third of the arms are uninformative, which guarantees that after a specific number of merge steps at least one arm will be present in the batch in order to eliminate all other arms of that batch.
	Moreover, these assumptions	allow one to derive a high probability bound on the cumulative regret of \Algo{MergeRUCB}, which is of the order 
	%$ \mathcal{O}(K \log T) $ 
	$\mathcal{O}\Big(\frac{K\log(T)}{ \min_{j \neq i: q_{i,j}\neq 1/2} \Delta_{i,j}^2  }  \Big)$,
	thereby eliminating the additive $K^2$ term in the regret bound of \Algo{RUCB}, as pairwise comparisons are only carried out within the local batches but not ``globally'' as in \Algo{RUCB}.

	\subsubsection{MergeDTS} \label{sec_merge_dts}
	Under the same assumptions as \Algo{MergeRUCB}, \citet{li2020mergedts} propose the Merge Double Thompson Sampling (\Algo{MergeDTS}) algorithm, which improves upon the former by using the Double Thompson Sampling algorithm (cf.\  Section \ref{subsec_dts}) in order to choose a pair of arms for the comparison within one batch.
	It is shown that \Algo{MergeDTS} enjoys the same order as \Algo{MergeRUCB} on its upper bound of the regret with high probability, but in contrast to the latter, needs to know the time horizon beforehand.
	However, in an extensive experimental study, the authors show that  \Algo{MergeDTS} is superior to \Algo{MergeRUCB} and other state of the art dueling bandits algorithms.
%	
%	The Double Thompson Sampling algorithm is a regret minimizing algorithm for a more general problem scenario than the and will be discussed 

	\subsubsection{Relative Confidence Sampling} \label{subsec_RCS}

	Again merely assuming the existence of a Condorcet winner and focusing on the minimization of the cumulative regret,  \citet{ZoWhDeMu14} introduce the relative confidence sampling (\Algo{RCS}) algorithm, which, in addition to the upper confidence bounds of the entries of the preference relation $\bQ$ (such as \Algo{RUCB}), maintains a Beta posterior distribution over the entries, respectively.
	The idea is that both upper confidence bounds and the Beta posterior distributions are used to suggest one arm each for the duel at one iteration step.
	To this end, a preference relation $\tilde \bQ \in [0,1]^{K\times K}$ is sampled\footnote{Strictly speaking, the preference relation $\tilde \bQ$ depends on the iteration step $t,$ as the Beta posterior distributions do. For sake of convenience, we suppress this dependency here in the notation.} according to the current Beta posterior distributions.
	If $\tilde \bQ $ has a Condorcet winner, then this arm is chosen as the first arm for the duel, otherwise the arm with the fewest picks according to the choice mechanism of the first case is used.
	As the second arm of the duel, \Algo{RCS} chooses the toughest  competitor of the first arm, namely the arm that has the highest (optimistic) chance to confute the Condorcet winner property of the latter (similar as in \Algo{MergeRUCB}) according to the current upper confidence bounds.
	The authors present experimental results on learning-to-rank data sets, revealing a satisfactory empirical performance of \Algo{RCS} regarding cumulative regret, but do not provide theoretical guarantees for \Algo{RCS} in terms of an upper bound on its cumulative regret.

	\subsubsection{Relative Minimum Empirical Divergence}\label{subsubsec:rmed}
	\citet{KoHoKaNa15} assume that the underlying preference matrix has a Condorcet winner, and propose three variants of the relative minimum empirical divergence (\Algo{RMED}) algorithm, which can be interpreted as a dueling bandits variant of the deterministic minimum empirical divergence (\Algo{DMED}) algorithm \citep{HoTa10} for the value-based MAB problem. 
%	\Algo{RMED} is based on the empirical Kullback-Leibler (KL) divergence between Bernoulli distributions with parameters corresponding to the probability that one arm being preferred to another one and draws arms that are likely to be the Condorcet winner with high probability. 
	More specifically, the algorithm revolves around \emph{the empirical divergence of an arm} $a_i$ at time $t$ defined by
	$$	I_{a_i}(t) = \sum_{ \{ a_j : \widehat{q}^{\, t}_{i,j} \leq 1/2  \}  } n_{i,j}^t \, \KL(\widehat{q}^{\, t}_{i,j},1/2),	$$
	where  $\KL(p,q)$ denotes the Kullback-Leibler divergence of Bernoulli random variables with success probabilities $p$ and $q.$
	As the exponential negative empirical divergence of an arm $a_i$ can be interpreted as the likelihood of being the Condorcet
	winner, one can define the empirical best arm by means of $i^*_t = \argmin_{i\in[K]} I_{a_i}(t).$
	Further, the algorithm maintains a set of potentially good arms defined by $C_t = \{ i \in [K] \, | \, I_{a_i}(t) - I_{i^*_t}(t) \leq  \log(t) + f(K) \}$  for some non-negative function $f$ that does not depend on $t.$
 	After a variant-specific exploration phase, all three variants of the \Algo{RMED} algorithm are in each time step essentially comparing one specifically chosen arm in $C_t$ (based on some ordering) with either the empirically best arm or, depending on the \Algo{RMED} variant, one specific arm based on the first chosen arm. 
	In particular, the first variant, called \Algo{RMED1}, chooses in case the empirically best arm is empirically not preferred over the first chosen arm, the arm which is empirically preferred the most over the latter.
	Here, ``empirically'' means that the choices are based on the current empirical pairwise estimates $(\widehat{q}^{\, t}_{i,j})_{1\leq i,j \leq K}.$
	For this variant, a bound on its expected regret of order $ \mathcal{O}\left( \sum_{i \neq i^* } \frac{\log T \,  \Delta_{i^*,i}}{\KL(q_{i,i^{*}},1/2)}  \right)  + \mathcal{O}(K^{2+\varepsilon})$ is shown, where $\varepsilon>0$ is a parameter of the algorithm specifying the used function $f$ for the set of potentially good arms.
	Further, using similar proof techniques as \cite{LaRo85} for showing asymptotic lower bounds in the standard MAB problem, they derive an asymptotic lower bound\footnote{This asymptotic lower bound $\Omega(C \cdot f(T))$ is to be understood as $\liminf_{T\to\infty}\frac{\exptd[R^T]}{f(T)} \geq C.$} of order $\Omega\left(  \sum_{i \neq i^* } \min_{j: q_{i,j}<1/2} \frac{ (\Delta_{i^*,i} + \Delta_{i^*,j} )\, \log T}{\KL(q_{i,j},1/2)}  \right)$ for the regret of any consistent dueling bandits algorithm for preference relations having a Condorcet winner or admitting a total order of the arms. 
	This result reveals that there is a gap between the upper bound of \Algo{RMED1} and the asymptotic lower bound regarding the constant factors.

	In order to obtain an algorithm having a regret bound with the constant factor matching the asymptotic lower bound (i.e., which is asymptotically optimal), they suggest the second variant of the \Algo{RMED} algorithm, called \Algo{RMED2},  which adapts the mechanism of  \Algo{RMED1} for choosing the second arm in order to obtain an estimate of the gap between the constant factor of \Algo{RMED1} and the asymptotic lower bound of the first chosen arm.
	Because the theoretical analysis of \Algo{RMED2} is cumbersome, the third variant of \Algo{RMED}, called \Algo{RMED2} Fixed Horizon (\Algo{RMED2FH}), is introduced.
	In contrast to the first two variants, \Algo{RMED2FH} needs to know the time horizon $T$ beforehand, because this is used to derive a ``non-exploring'' estimate for the gap between the constant factor of \Algo{RMED1} and the asymptotic lower bound of the first chosen arm, facilitating the theoretical analysis considerably.
	Under this assumption and the existence of a Condorcet winner, it is shown that \Algo{RMED2FH} enjoys an asymptotically optimal regret upper bound.
%	which is a static version of RMED2, and show that it is asymptotically optimal under the Condorcet assumption.

	\subsubsection{Winner Stays} \label{sec_winner_stays_algo}
	\citet{ChFr17} study the dueling bandits problem in the Condorcet winner setting, and consider both extreme cases for the regret, namely the strong regret specified by the instantaneous regret $r_{t,max}$ and the weak regret  $r_{t,min}$, which is $ 0 $ if either arm pulled is the Condorcet winner. 
	They propose the Winner Stays (\Algo{WS}) algorithm with variations for both kinds of regret. 
	\Algo{WS} for weak regret (\Algo{WS-W}) is a round-based challenge algorithm, where arms are dueled with each other in a batch of duels in each round, and each batch corresponds to a sequence of duels of the same pair consisting of the current champion and a ``worthy" challenger.
	The current champion is the arm that has currently the largest number of overall duels won, while the challenger is the arm that has not yet been considered in the current round and has currently the second largest number of overall duels won (possibly breaking ties).
	The batch of duels continues until either (i) the challenger manages to win against the current champion so many times that the champion's overall number of duels won is the third highest, or (ii) the champion wins against the challenger so many times that the challenger's overall number of duels won is the third highest.
	In case of the first event, the challenger becomes the current champion, while in the second event, the champion (winner) stays.
	In both cases, the defeated arm is not considered anymore in the current round and the next round starts after each available arm has been considered at least once in a batch of duels.

	If the underlying instantaneous regret is the strong regret, the authors propose the \Algo{WS} for strong regret (\Algo{WS-S}), which considers separate exploration and exploitation phases in epochs.
	Within each epoch $e \in \N$, first the $e$th round of the \Algo{WS-W} algorithm is conducted for the exploration phase, resulting in a current champion that is then dueled against itself (cf.\ ``fully commitment'' in Section \ref{subsec_regret_bounds}) in the exploitation phase for the purpose of keeping the cumulative strong regret low.
	In light of this, the length of an exploitation phase is exponentially increasing with the number of epochs passed.  

	Assuming that there are no ties in the underlying preference relation $\bQ,$ it is proven that unlike all regret bounds for cumulative average regret, the \Algo{WS-W} algorithm has expected cumulative weak regret that is constant in time.
	In particular, it is shown that \Algo{WS-W} has an  expected cumulative weak regret bound of order $ \mathcal{O}\left( \frac{K^2}{\min_{i,j} |\Delta_{i,j}|^3 } \right)$ under the assumption of an existing Condorcet winner, and of order $ \mathcal{O}\left( \frac{K \log K}{\min_{i,j} \Delta_{i,j}^6 } \right) $ under the assumption of an existing total order of arms.
	Further, it is proved that \Algo{WS-S} enjoys an expected cumulative strong regret bound of order $ \mathcal{O}\left( \frac{K^2 }{\min_{i,j} \Delta_{i,j}^2 } + \frac{K \log T}{\min_{i,j} |\Delta_{i,j}| }\right)$ in the Condorcet winner setting, and of order $ \mathcal{O}\left( \frac{K \log K}{\min_{i,j} \Delta_{i,j}^6 } + \frac{K \log T}{\min_{i,j} |\Delta_{i,j}| }\right) $ under the assumption of an existing total order of arms. 
	Both bounds are optimal regarding their dependence on the time horizon $T,$ but are not optimal regarding the dependence on the calibrated preference probabilities.
	The proof for \Algo{WS-W} is revised by \citet{pekoez_ross_zhang_2020}, who show that \Algo{WS-W} incurs in fact an expected regret upper bound of $ \mathcal{O}\left( \frac{K^2}{\min_{j\neq i^*} |\Delta_{i^*,j}|^2 } \right)$ under the assumption of an existing Condorcet winner, but without assuming that there are no ties in the underlying preference relation $\bQ.$

	It is worth noting that the analysis of the \Algo{WS} algorithms is in some sense unique, as the Gambler's ruin problem is used to upper bound the number of pulls of sub-optimal arms, whereas all regret minimizing algorithms reviewed so far make use of the Chernoff bound in some way.
	
	% Further, they also consider utility-based extensions of weak and strong regret, and show that their bounds also apply here, with a small modification. It is worth to mention that even if the regret bound of these algorithms are not optimal for the , they are unique in a sense that the Gambler's ruin problem is used to upper bound the number of pull of sub-optimal arms, whereas all regret optimization algorithm which we review in this study, make use of the Chernoff bound in some way.

	\subsubsection{Beat the Winner} \label{sec_beat_the_winner}
	Having the same target as \Algo{WS-W}, \citet{pekoez_ross_zhang_2020} propose the \Algo{Beat the Winner} (\Algo{BTW}) algorithm, which is a round-based challenge algorithm based on a queue structure of the arms. 
	Here, the first arm in the queue is the current champion while the second is its challenger.
	Both arms are compared in round $l\in \N$ as long as one of the two arms has won exactly $l$ many times, whereupon the defeated arm is set to the end of the queue and the winner is queued to the front (if necessary).
	It is shown that \Algo{BTW} enjoys an upper bound on its expected weak regret of $ \mathcal{O}\left(  K^2 + \frac{K}{(1-\exp(-2\min_{j\neq i^*} |\Delta_{j}|^2))^2}  \right),$ where the additive linear (in $K$) term might dominate the quadratic term for a small total number of arms $K.$

	As \Algo{BTW} does not consider the history of the past duels in an explicit way, the authors suggest the \Algo{Modified Beat the Winner} (\Algo{MBTW}) algorithm, which improves upon \Algo{BTW} by assigning scores to arms based on their history of wins and chooses the challenger in each round in a random manner based on their relative scores.
	The initial scores of the arms are set to 1 and increased/decreased by one for the winner/loser of a respective round, which, however, cannot fall below a score of one.
	Although \Algo{MBTW} is not theoretically analyzed, it is shown to have a satisfactory empirical performance compared to \Algo{BTW} as well as \Algo{WS-W} for the task of expected weak regret minimization.

	\subsubsection{Knockout Tournaments}  \label{subsec_Knockout_tournaments}
	Assuming a total order over the arms, $\gamma$-relaxed stochastic transitivity  as well as  stochastic triangle inequality, \citet{FaOrPiSu17}  consider the goals of finding the best arm as well as the best ranking (cf.\ Section \ref{sec:bsr}) in the $(\epsilon,\delta)$-PAC setting. 
	More specifically, for any given $ \epsilon,\delta > 0 $, the algorithm for finding the best arm must output an $\epsilon$-optimal arm (cf.\ Section \ref{subsec_near_opt_targets}) with probability at least $1-\delta.$
%	$ i $ such that, with probability at least $ 1-\delta $, for all $ j \neq i, \Delta_{i,j} \geq -\epsilon $, and the algorithm for the best ranking must output, with probability at least $ 1-\delta $, a ranking $\br$ such that $ \Delta_{i,j} \geq -\epsilon $ whenever $ r_i > r_j $.      
	
	For this purpose, they propose the \Algo{Knockout} algorithm, which proceeds in rounds, each of which corresponds to a knockout tournament with the goal of successively eliminating half of all currently remaining arms
%	
%	Certainly, the algorithm terminates as soon as 
	until only one single arm is left, which is then the suggested candidate for the best arm, i.e., an $\epsilon$-optimal arm.
	At the beginning of each round, all not yet eliminated arms are divided into pairs in a random manner.
	Then, all these pairs of arms are successively dueled with each other until either one is confident enough which arm is superior resp.\ inferior, or a certain (round-dependent) number of duels has been reached, whereupon the arm with the larger number of duels won proceeds to the next round, while the other one is eliminated.
	Thanks to the regularity assumptions made on $\bQ$, it is ensured by choosing a suitable maximal limit on the (round-dependent) number of duels that the highest ranked arm at the beginning of a round and the highest ranked arm at the end of a round are close in terms of their calibrated preference probabilities, which can be upper bounded by a term that is linear in the used approximation quality of a round.
%	
%	Due to the elimination strategy based on confidence levels possible approximation error propagation
	To maintain the overall confidence level $\delta$ and the approximation quality $\epsilon$ of the entire procedure, and at the same time prevent a larger sample complexity through a rough Bonferroni correction, both the round-wise confidence level and the round-wise approximation quality are geometrically progressing. 
%	Using Chernoff's bound to derive a suitable maximal limit for a round-dependent number of duels to maintain a specific  that the highest ranked arm at the beginning of a round 
%	
%	Thus, by choosing the confidence levels as well as  in an appropriate way by taking possible approximation error propagation and the regularity assumptions on $\bQ$ into account, 
	In this way, it can be shown that \Algo{Knockout} is an  $(\epsilon,\delta)$-PAC algorithm for finding the best arm with a sample complexity of $ \bigO \left( \frac{K}{\gamma^4 \epsilon^2} \log \frac{1}{\delta} \right)$.
	This improves upon the sample complexity shown for \Algo{BTM} for the same setting (cf.\ Section \ref{subsec_btm_algorithm}) and matches the lower bound for $\gamma=1$ (i.e., strong stochastic transitivity), which can be derived by  \cite{feige1994computing}.
%	
%	Computing with noisy information

	\subsubsection{Sequential Elimination} \label{subsec_seq_elimination}
	Seeking the same goals as \citet{FaOrPiSu17}, but this time only requiring a total order over the arms and strong stochastic transitivity (no stochastic triangle inequality), \citet{FaHaOrPiRa17} present the \Algo{Seq-Eliminate} algorithm in an ($\epsilon,\delta$)-PAC setting for finding the best arm, which uses $ \bigO \left( \frac{K}{\epsilon^2} \log \frac{1}{\delta} \right)$ comparisons.
	The \Algo{Seq-Eliminate} algorithm is a challenge algorithm (cf.\ Section \ref{subsub_challenge_algos}), which does not use an arm for dueling if this arm has been defeated once.
	In particular, \Algo{Seq-Eliminate}  starts by selecting a current ``champion" at random, and keeps dueling it with another random  arm (challenger) until the more preferred arm of the two is determined with a certain confidence. 
	It then proceeds to the next competition stage, after setting the winner from the last stage as the new champion and eliminating the loser. The algorithm stops as soon as only a single arm remains. 
	Unlike \Algo{Knockout}, the confidence levels within each competition stage are designed in a more adaptive way, which ensures that with this elimination procedure the overall confidence level $\delta$ as well as the approximation quality $\epsilon$ are maintained.

	Using a random choice to determine the first champion, \Algo{Seq-Eliminate} has in fact a sample complexity of  $ \bigO \left( \frac{K}{\epsilon^2} \log \frac{K}{\delta} \right),$ which is order optimal in the high confidence regime, i.e., if $\delta \leq 1/K,$ but not if $\delta>1/K.$
	In order to deal with confidence levels of the latter kind, the authors propose to find first a good initial champion, say $\tilde a$, by using \Algo{Seq-Eliminate} on a randomly sampled smaller subset $\cA$ of a specific size. 
%	
%	The basic idea is that, by applying \Algo{Seq-Eliminate} to a random subset of arms, it is possible to find a good initial champion with high probability, provided the size of the subset is suitably chosen.
%	
	Next, $\tilde a$ is used to obtain an auxiliary and potentially smaller subset of all arms, say $\tilde A,$ by dueling it with all arms multiple times in a competition stage manner (with specifically chosen confidence levels and approximation quality) and include an arm in $\tilde A$ only if it has managed to withstand $\tilde a$ in all stages.
	Finally, each arm in $\tilde A$ is dueled once again in a competition stage manner against $\tilde a$ until either $\tilde a$ is winning against each arm in $\tilde A$, making $\tilde a$ the final output, or $\tilde a$ is inferior to one arm in $\tilde A,$ whereupon \Algo{Seq-Eliminate} is used on $\tilde A$ to obtain the final output.
	With this modification of \Algo{Seq-Eliminate}, the authors show that one obtains an $(\epsilon,\delta)$-PAC algorithm for finding the best arm, which uses $ \bigO \left( \frac{K}{\epsilon^2} \log \frac{1}{\delta} \right)$ comparisons for $\delta>1/K.$

	\subsubsection{Opt-Max}\label{opt-max}
	Replacing the strong stochastic transitivity by the moderate stochastic transitivity assumption, \cite{falahatgar2018limits} study the problem of best arm identification in an $(\epsilon,\delta)$-PAC setting. 
	They present the \Algo{Opt-Max} algorithm, which makes heavily use of a subroutine, called \Algo{Soft-Seq-Elim}.
	The latter essentially operates as \Algo{Seq-Eliminate}, but can also refrain from eliminating one arm after a sequence of duels in a stage, if no clear winner based on a specific confidence level can be declared.
	Thus, \Algo{Soft-Seq-Elim} terminates if the current champion has not changed after dueling it with all active arms. 

The guarantees and the sample complexity of \Algo{Soft-Seq-Elim} critically depend on the number of changes of the champion: Although the worst-case sample complexity of the algorithm is quadratic, it runs fast (close to linear) and tends to yield correct answers when the number of required changes is small. In other words, \Algo{Soft-Seq-Elim} strongly benefits from the choice of a ``good'' initial champion (ideally the best arm) in the beginning.  
The \Algo{Opt-Max} algorithm consists of three variants of \Algo{Soft-Seq-Elim}, each of them tailored to a certain range of the confidence level $\delta$ by adapting it essentially in a similar manner as \Algo{Seq-Eliminate} for the low confidence regime (cf.\ Section \ref{subsec_seq_elimination}).
%
%Using a similar approach as for the low confidence regime for \Algo{Seq-Eliminate} tailored to three different ranges of confidence levels eventually defines the \Algo{Opt-Max} algorithm.

%On the basis of \Algo{Soft-Seq-Elim}, the authors propose three different maxing algorithms, which target different confidence levels $1-\delta$ (low, median, and high), and which are finally combined into \Algo{Opt-Max}. All these algorithms call \Algo{Soft-Seq-Elim} repeatedly with random subsets of the full set of arms (and eventually of course with all arms).
%
It is shown that \Algo{Opt-Max}  is an $(\epsilon,\delta)$-PAC algorithm for finding the best arm having a sample complexity of order $\bigO \left( \frac{K }{ \epsilon^2 } \log \frac{1}{ \delta} \right) $, at least if  $\delta\geq \min(1/K,\exp(-K^{1/4})).$
In light of the findings by \cite{feige1994computing}, the order of \Algo{Opt-Max}' sample complexity is optimal (cf.\ also Section \ref{subsec_Knockout_tournaments}).
Finally, the authors show that any algorithm for finding the best arm in the PAC-setting requires in the worst-case scenario a number of comparisons that scales with $K^2$ if merely weak stochastic transitivity is assumed.

\subsubsection{Binary-Search-Ranking}\label{sec:bsr}

For the ranking problem in an $(\epsilon,\delta)$-PAC setting, the authors of the \Algo{Knockout} algorithm (\citet{FaOrPiSu17}) also propose the \Algo{Binary-Search-Ranking} algorithm, assuming that the underlying preference relation $\bQ$ admits a total order over the arms and satisfies strong stochastic transitivity as well as the stochastic triangle inequality. 
%subsec_near_opt_targets

This algorithm consists of three major steps. In the first step, it randomly selects a set of arms of size $ \frac{K}{(\log K)^x} $, called anchors, and ranks them using a procedure called \Algo{Rank-x}---an $ (\epsilon,\delta) $-PAC ranking algorithm, which for any $ x>1 $, uses $ \bigO \left( \frac{K}{\epsilon^2}(\log K)^x\log \frac{K}{\delta} \right) $ comparisons, while at the same time creating $ \frac{K}{(\log K)^x}-1 $ bins between each two successive anchors. Then, in the second step, a random walk on a binary search tree is used to assign each arm to a bin. Finally, in the last step, the output ranking is produced. To this end, the arms that are close to an anchor are ranked close to it, while arms that are distant from two successive anchors are ranked using \Algo{Rank-x}.

In a subsequent work, \citet{falahatgar2018limits} propose an improvement of the latter, which gets rid of superfluous logarithmic terms in the sample complexity.
This improvement is achieved by modifying the components of the algorithm as follows. 
Each component is called a first time with the (high probability) guarantee of a correct output of $1-1/\log(K)$ instead of $1-1/K^5$, resulting in a smaller number of comparisons.
Then, the output of the respective component is checked, for which a small complexity can be shown. Finally, if the output is incorrect, the corresponding component is run again, but this time with a $1-1/K^5$  guarantee for its correctness.

It is shown that the (enhanced) \Algo{Binary-Search-Ranking} algorithm is an $(\epsilon,\delta)$-PAC algorithm for the ranking problem with sample complexity $ \bigO \left( \frac{K\log K }{\epsilon^2} \right) $ if $ \delta $ is set to $\frac{1}{K} $. 
Thus, the leading factor of the sample complexity of finding a nearly best arm differs from finding a nearly best ranking by a logarithmic factor.
This was to be expected, and simply reflects the difference in the worst-case complexity for finding the largest element in an array and sorting an array using an efficient sorting strategy.

\citet{FaOrPiSu17} derive a lower bound on the sample complexity of $ \Omega \left( \frac{K}{\epsilon^2} \log\frac{K}{\delta} \right) $ for any $(\epsilon,\delta)$-PAC algorithm for the ranking problem under the assumptions made by \Algo{Binary-Search-Ranking} on the preference relation $\bQ.$
For the best ranking problem in the PAC-setting, \cite{FaHaOrPiRa17} show that any algorithm needs $ \Omega(K^2) $ comparisons under the strong stochastic transitivity property. To this end, they consider a preference relation for which they reduce the problem of finding a $ 1/4 $-ranking to a problem of finding a coin with bias $ 1 $ among $ \frac{K(K-1)}{2}-1 $ other fair coins, showing that any algorithm requires at least a number of comparisons that scales quadratically with $K$.  
Finally, \cite{falahatgar2018limits} verify the same lower bound by assuming moderate stochastic transitivity together with the stochastic triangle inequality. 
This in particular shows that the stochastic triangle inequality facilitates the learning problem.

	\subsubsection{Top-$k$ identification via Quick Select}  \label{sec_Quick-Select-based}
	\citet{ren20a} consider the $(\epsilon,\delta)$-PAC learning scenario for the top-$k$ arms if the underlying preference relation has a total order over the arms, satisfies strong stochastic transitivity as well as the stochastic triangle inequality.
	Note that an $\epsilon$-approximation of the top-$k$ arms is any $k$-sized set of arms such that any arm within the set is $\epsilon$-preferable to any other arm, which is not in the subset (cf.\ Section \ref{subsec_near_opt_targets}).
	Moreover, it is throughout assumed that $k$ is at most $K/2.$ 

	The suggested algorithm, called Tournament-$k$-Selection (\Algo{T-$k$-S}), is a tournament algorithm proceeding in rounds, each of which consists of dividing the currently selectable arms into subsets of sizes at most $2k$, and then using a subroutine called Epsilon-Quick-Select (\Algo{EQS}) on each subset. This is done to eliminate arms that are (with a specific confidence) not belonging to the (nearly) top-$k$ arms of that subset.
	The entire process stops as soon as the number of noneliminated arms is $k$. The final output of \Algo{T-$k$-S} is then given by the remaining arms.

	The subroutine \Algo{EQS} is inspired by the well-known \Algo{QuickSelect} algorithm \citep{hoare1961algorithm} to find the $k$ largest items of an array.
	First, \Algo{EQS} chooses one arm randomly to be the pivot arm and then duels it with any other arm a specific number of times (similar as for each stage of \Algo{Seq-Eliminate}) leading to a partition of the set of arms into three subsets:
	One subset consisting of all arms one is confident enough that each of its elements are preferred resp.\ not preferred over the pivot arm, and one subset consisting of all arms one is not confident enough about the preference relation merged with the pivot arm itself.
	If the subset of surely preferred arms consists of more than $k$ elements, the \Algo{EQS} algorithm is applied on the latter subset.
	Otherwise, if the union of surely preferred arms and unsurely preferred arms consists of more than $k$ elements, then a $k$-sized subset is formed by merging the surely preferred arms and a randomly chosen subset (of the right size) of the unsurely preferred arms. 
	Finally, if the latter union has strictly less than $k$ elements, say $k',$ then this union is returned together with the subset returned by \Algo{EQS} for finding the top-$(k-k')$ arms on the subset consisting of the surely not preferred arms.
%	 \Algo{T-$k$-S} algorithm then uses the \Algo{EQS} to call 

	By choosing the round-wise confidence levels and approximation errors in a suitable way, \Algo{T-$k$-S} is shown to be an $(\epsilon,\delta)$-PAC algorithm for finding the top-$k$ arms with sample complexity $ \bigO\left( \frac{K}{\epsilon^2} \log\frac{k}{\delta} \right).$ 
	This sample complexity is optimal, as the authors show also a lower bound on the sample complexity of $ \Omega \left( \frac{K}{\epsilon^2} \log\frac{k}{\delta} \right) $ for any $(\epsilon,\delta)$-PAC algorithm for finding the top-$k$ arms under the assumptions made by \Algo{T-$k$-S}.
%	 a total order over the arms, strong stochastic transitivity as well as the stochastic triangle inequality.
%	by connecting the $\delta$-PAC scenario in the value-based multi-armed bandit problem and the 

	\subsubsection{Approx-Prob}  \label{subsec_approx_prob}
	
	\cite{falahatgar2018limits} consider the problem of approximating all pairwise probabilities to an accuracy of $\epsilon$ and present the \Algo{Approx-Prob} algorithm for this purpose with an optimal sample complexity of the form $\bigO \left( \frac{K  \min(K,1/\epsilon) \log(K) }{ \epsilon^2 } \right) $ for $\delta\geq 1/K$. \Algo{Approx-Prob} takes a pre-sorted list of the items in the form of an $\epsilon/8$-ranking as an input---this requires solving a ranking problem first, for which the \Algo{Binary-Search-Ranking} algorithm as discussed in Section \ref{sec:bsr} can be used. Given this input, the algorithm reduces the number of pairwise comparisons by exploiting the assumptions of strong stochastic transitivity and the stochastic triangle inequality for the pairwise probabilities. The key idea is that, under these regularity assumptions, the pairwise probabilities should obey constraints of the form $\Delta_{i-1,j} \leq  \Delta_{i,j} \leq \Delta_{i,j-1}$ for $i < j$. These constraints are used to correct empirical estimates $\hat{\Delta}_{i,j}$, which are obtained as relative winning frequencies from repeated pairwise comparisons of arms $a_i$ and $a_j$, or even to avoid such an estimation altogether. For example, if $\hat{\Delta}_{i-1,j} = \hat{\Delta}_{i,j-1}$, then by virtue of the above constraint, inequalities turn into equalities $\Delta_{i-1,j} =  \Delta_{i,j} = \Delta_{i,j-1}$, and $\hat{\Delta}_{i,j}$ is simply set to $\hat{\Delta}_{i-1,j}$ instead of being estimated. Such equalities are fostered by providing estimates on a grid, i.e., estimates of pairwise probabilities are always rounded off to the closest multiple of $\epsilon$. Moreover, to take the greatest advantage of the triplet-constraints, the items are compared in a specific order: $a_i$ is compared to $a_j$ in an outer loop for $i = 1, \ldots , K-1$ and an inner loop for $j = i+1, \ldots , K$.
	
	% under the same assumptions as in Section \ref{opt-max},
	%idea of the algorithm is to first find an $\epsilon/8$-ranking of the items and then further approximate the pairwise probabilities of only a subset of cleverly selected pairs.}
	
	%\editeyke[What does orderwise mean? This is not detailled enough] 

		\subsubsection{Robust Query Selection} \label{subsec_robust_query_selection}
		\citet{JaNo11} assume that there is a total order over the arms and each arm can be embedded into $\R^d$ by means of suitable location points.
		Further, they assume that there exists some (reference) point in $\R^d,$ such that the Euclidean distance of this point to these locations is coherent with the underlying total order in the sense that the closer an arm's location point is to the reference point, the higher its ranking in the underlying total order.
		While the locations are assumed to be known, the reference point is unknown.
		By first assuming that the outcome of a duel between two arms is deterministic, i.e., $\bQ \in \{0,1\}^{K\times K},$ the authors introduce the notion of ambiguity of a duel: If it is not possible to infer from the past duels of other pairs of arms which arm will be preferred over the other for a specific pair of arms, then the latter pair is called \emph{ambiguous}, otherwise it is called \emph{unambiguous}.

		In order to characterize the property of ambiguity in the presence of a (latent) reference point and the embedding of the arms, they relate the problem of identifying the underlying ranking of the arms to the problem of determining the label of $(d+1)$-dimensional points via linear separators (in an active way).
		With this, they introduce the \Algo{Query Selection} algorithm, which essentially samples a pair of arms uniformly at random from the set of pairs not considered so far, checks whether the pair is unambiguous in order to skip superfluous duels, and carries out the duel in case the pair is ambiguous.
		Once all pairs have been considered, the algorithm terminates and provably returns the correct ranking. 

		For the scenario in which the outcome of duels between two arms is not necessarily deterministic, the latter algorithm is modified to the \Algo{Robust Query Selection} algorithm.
		This algorithm essentially corresponds to \Algo{Query Selection}. However, due to the noisy outcomes, it conducts a pre-specified number of duels for an ambiguous pair of arms in order to decide which arm is preferred over the other, namely the arm which has won the majority of the noisy duels. 
		It is shown that if the number of duels carried out per ambiguous pair is set to $\nicefrac{\log\left(   \nicefrac{2K\log(K)}{\delta} \right)}{2 h^2},$ where $\delta \in (0,1)$ and $h\in (0,1/2)$ is such that $\min_{1\leq i<j \leq K}|\Delta_{i,j}|\geq h,$ then the \Algo{Robust Query Selection} algorithm returns the true underlying ranking and has a sample complexity of order $\bigO\left( \frac{d}{h^2} \log^2\left(\frac{K}{\delta}\right) \right).$

	\subsubsection{Verification-based Solution} \label{subsubsec:verification_based}
	
	\citet{Ka16} considers the problem of finding the best arm in structured MAB problems, which refers to a general bandit framework covering a variety of different bandit problems such as the classical MAB problem, linear bandits, combinatorial bandits and other bandit problems (see \cite{lattimore2020bandit} for an overview). 
	To this end, a general algorithmic framework is introduced, consisting of two subroutines which need to be specified for each underlying bandit problem.
	The first subroutine, referred to as \Algo{FindBestArm}, needs to be designed such that it identifies the best arm with a certain high probability, while using as few samples within the underlying bandit problem as possible.
	The second subroutine, referred to as \Algo{VerifyBestArm}, serves the purpose of verifying the optimality of the arm suggested by the first subroutine as the best arm with a certain degree of confidence. 
	In particular, this subroutine can either confirm or refuse the optimality of the suggested arm and can also receive additional information about the underlying bandit problem from the first subroutine.
	Both subroutines are run one after the other during iterative stages, where in each stage the confidence level for the second subroutine is geometrically decreased, while the error probability of the first subroutine is throughout a specific constant. 
	The iteration process transitions into the next stage only if the second subroutine has refused the optimality of the suggested arm, and the entire process stops as soon as the second subroutine has confirmed the optimality of the suggested arm.
	The rationale behind this procedure is that it seems to be easier to verify or refute the optimality of a candidate arm than to explicitly search for the best arm.
%	
%	\citet{Ka16} considers the problem of finding the best arm in stochastic MABs in the pure exploration setting with the goal of minimizing the sample complexity, focusing on the scenario where the failure probability is very small, and presents the Explore-Verify framework for improving the performance of the task in multiple generalizations of the MAB setting, including dueling bandits with the Condorcet assumption. The framework is based on the fact that in MAB problems with structure, the task of verifying the optimality of a candidate is easier than discovering the best arm, which leads to a design in which first the arms are explored and a candidate best arm is obtained with probability $ 1-\kappa $ for some constant $ \kappa $, and then it is verified whether the found arm is indeed the best with confidence $ 1-\delta $. If the exploration procedure was correct, the sample complexity will be the sum of the one of the exploration algorithm, which is independent of $ \delta $, and the one of the easier verification task, which depends on $ \delta $. Thus, for small values of $ \delta $, the savings are large, regardless of whether the sample complexity is dominated by that of the verification task, or by that of the original task with a constant failure probability. 
	
	In concrete terms for the setting of dueling bandits under the Condorcet assumption, a suggestion for both subroutines is made.
	The suggestion for the first subroutine initializes an active set consisting of all ordered pairs of arms, and duels all active pairs, say $(a_i,a_j)$,  until one of three specific conditions on the lower resp.\ upper confidence estimates of the underlying pairwise preference probabilities, i.e., $\widehat{q}^{\, t}_{i,j}-c_{i,j}^t$ resp.\ $\widehat{q}^{\, t}_{i,j}+c_{i,j}^t$, holds, whereupon the pair is removed from the active set.
	Once the active set is empty, the subroutine stops and returns as its candidate for the Condorcet winner the arm that has a lower confidence estimate greater than $1/2$ against any other arm.
	The three conditions are designed such that this condition is fulfilled for exactly one arm, which is likely to be the actual Condorcet winner.
	Here, $c_{i,j}^t$ is a confidence interval based on Hoeffding's inequality for maintaining the predefined error probability of the first subroutine.
%
%	within the until it finds, for each suboptimal arm $ a_i $, an arm $ a_j $ with $ q_{i,j} < 1/2 $; the exploration algorithm provides as output the identity of the optimal arm, together with the identity of an arm $ a_j(a_i) $ that beats $ a_i $ by the largest gap for each sub-optimal arm $ a_i $. 
	The second subroutine proceeds from the information available after the first subroutine, i.e., the candidate for the Condorcet winner as well as for each allegedly non-Condorcet winner arm its toughest competitor (similarly as for \Algo{RCS}), and the verification task consists of verifying that each non-Condorcet winner arm is indeed inferior to its toughest competitor based on confidence intervals  maintaining the confidence level for the second subroutine.
	It is shown that these two subroutines used in the general algorithmic framework lead to a sample complexity of order 
	\begin{align} \label{bound_verif}
			\mathcal{O} \left( \sum\nolimits_{i \neq i^*} \min_{j: q_{i,j}<1/2} \frac{\log \big(K/(\delta \Delta_{i,j}^2)\big)}{\Delta_{i,j}^2} \right)    + \tilde{\mathcal{O}} \left( \sum\nolimits_{i \neq i^* } \Big( \Delta_{i^*,i}^{-2}  +  \sum\nolimits_{j \neq i }  \Delta_{i,j}^{-2}  \Big) \right),
	\end{align}
	where $\tilde{\mathcal{O}}$ is hiding $\log$-factors.
	By considering the RMED algorithm of \cite{KoHoKaNa15} as an algorithm for sample complexity minimization and transforming its expected regret upper bound to a sample complexity bound, the method suggested by \cite{Ka16} achieves an improvement over the latter by a multiplicative factor $ K^{\epsilon}$, provided $ \delta $ is sufficiently large.

	\subsubsection{Parallel Selection and Partition} \label{sec_parallel_select}
	
	Under the assumption of a total order over the arms, the task of finding the $k$th best arm or a partition of $\cA$ into the set of best $k$ arms and its complement is considered by \citet{braverman2016parallel}.
	The authors are not only interested in the sample complexity of algorithms for the latter task, but also in their round complexity, which can be interpreted as the degree of an algorithm's adaptivity.
	To be more precise, an algorithm making all decisions on the order of the duels entirely ahead of time has the smallest possible round complexity (non-adaptive), while an algorithm deciding in each time step which duel to conduct has the largest possible round complexity  (fully adaptive).

	The authors provide algorithms that correctly solve the above tasks with high probability, having a low round complexity, while the sample complexity is of order  $$\bigO\left( \frac{K}{\min_{1\leq i<j \leq K} \Delta_{i,j}^2} \log(K) \right),$$ assuming the knowledge of $\min_{1\leq i<j \leq K} \Delta_{i,j}^2>0.$
	Further, it is shown that any algorithm, which can correctly return the $k$th best arm  with high probability, has a sample complexity of the same order as above, revealing the optimality of the suggested algorithmic solutions.
	
	Just as for the \Algo{Robust Query Selection} algorithm in Section \ref{subsec_robust_query_selection}, the authors first consider the case of deterministic outcomes of duels in order to design suitable algorithmic solutions and transfer them to the probabilistic scenario by repeating a duel between a specific pair of arms a certain number of times.  
	The major algorithmic procedure underlying the deterministic case first chooses a sufficiently large subset of $\cA$ in a random manner, say $\cA^S,$ and then reduces $\cA^S$ successively by choosing random anchor arms in $\cA^S,$ comparing these with all arms in $\cA^S$ and keeping all arms that are in some sort of interquartile range (with respect to the number of duels won) of the anchor arms.
	The rationale behind the latter iterative process is that the subset $\cA^S$ will reduce quickly to the $k$th best arm (or a small subset containing it) of the initially first chosen subset, which in turn is likely to be close\footnote{Closeness is here to be understood in terms of their ranks with respect to the ground truth ranking.} to the $k$th best arm in $\cA.$
	Hence, one (randomly chosen) arm within the reduced subset can be used as a kind of pivot in order to partition $\cA$ into the set of best $k$ arms and its complement by dueling the former with all arms in $\cA:$ All arms preferred over the pivot are in the top set, while the inferior arms are in its complement.

	By running the major procedure twice with suitable choices for the subset sizes as well as number of iterations, and then combining and filtering the results of both procedures, the authors show that the true partition (and also the $k$th best arm) can be identified with high probability.

	\subsubsection{Single Elimination Tournament} \label{subsec_single_eliminiation_tournament}
	Assuming the existence of a total order over the arms, \citet{MoSuEl17} study the top-$k$ identification problem as also considered by \citet{braverman2016parallel}, and additionally the  top-$k$ ranking problem, both with the goal to minimize the exact sample complexity while maintaining a predefined confidence $\delta.$ 
%	They first characterize an upper bound on the sample size required for both problems, and demonstrate how sample complexity can be reduced through active instead of passive ranking. 
%	
	To this end, they first present the \Algo{Select} algorithm for identifying the best arm, which can be seen as a customized single-elimination tournament consisting of multiple layers, where in each layer, pairs of arms are randomly built first, and on the basis of pairwise comparisons, one arm is retained and the other one is eliminated. This process is repeated until only one arm is left, which is then the suggestion for the best arm. Note that the \Algo{Select} algorithm is conceptually similar to \Algo{Knockout} (cf.\ Section \ref{subsec_Knockout_tournaments}).
	The authors subsequently show that the algorithm finds the best arm with probability at least $1-\delta$ and has a sample complexity of order $ \bigO (\frac{K \log (1/\delta)}{\Delta_{1,2}^2}) $. 
	
	\Algo{Select} is then generalized to the \Algo{Top} algorithm, which works for both top-$k$ ranking and identification, by first splitting the arms into $k$ sub-groups, then identifying the best arm in each sub-group using \Algo{Select}, and finally forming a short list that includes all winners from the sub-groups. For this list, they build a heap data structure, from which  the  top-$k$ arms are extracted one after another, while whenever a best arm is extracted from its list, the second best arm from that list is identified and reinserted into the short list. 
	Unfortunately, it is not shown by the authors how the sample complexity of \Algo{Top} scales in terms of $\delta,$. We conjecture it is of the order $ \bigO \left( \frac{ (K+k\log k) \, \log(1/\delta) \, \max\{ \log k, \log \log K \} }{ \log \log K \, \Delta_k} \right) $, where $ \Delta_k = \min_{i \in [k]} \min_{j:j\geq i} \Delta_{i,j}^2 $ in the case of top-$k$ ranking and  $ \Delta_k = \Delta_{(k),(k+1)}^2 $ in the case of top-$k$ identification, with $\Delta_{(k),(k+1)}$ denoting the calibrated pairwise preference probability of the arm with rank $k$ and the arm with rank $k+1$ according to the total order of the arms.
	In Table \ref{tab:regaxiom_third}, we report the sample complexity as stated by the authors.
	
	Similarly as in the $(\epsilon,\delta)$-PAC setting, the leading factor of the sample complexity of \Algo{Top}  differs from the one of \Algo{SELECT} by a logarithmic factor.
	\subsubsection{Sequential-Elimination-Exact-Selection}  \label{sec_Sequential-Elimination-Exact-Selection}
	Having the same goal as \cite{MoSuEl17} but imposing stricter assumptions on $\bQ$ (strong stochastic transitivity and stochastic triangle inequality), the authors of the \Algo{T-$k$-S} algorithm also propose the Sequential-Elimination-Exact-Best-Selection (\Algo{SEEBS}) algorithm for finding the best arm and the Sequential-Elimination-Exact-$k$-Selection (\Algo{SEEKS}) for identifying the top-$k$ arms.

	The \Algo{SEEBS} algorithm is a challenge algorithm proceeding in rounds.
	In each round the \Algo{T-$k$-S} algorithm is used to obtain a nearly best arm by setting $k$ to 1, which is used as the current ``champion''.
	The latter arm is then used in a similar manner as the pivot arm in the \Algo{QuickSelect} inspired subroutine \Algo{EQS} of \Algo{T-$k$-S} to partition the set of arms into three subsets (surely preferred, surely not preferred and undecided arms).
	All arms (``challengers'') assigned to the set of surely not preferred arms over the champion are eliminated.
	Adapting the confidence for the latter assignment as well as for the choice of the round-wise champion in an appropriate way over the rounds will ensure that eventually only one single arm is left and \Algo{SEEBS} maintains the overall confidence level $\delta.$
	Moreover, \Algo{SEEBS} reveals an expected sample complexity of $\bigO \left(  \sum_{i \in [K]} \frac{\log 1/\delta + \log \log 1/\Delta_{i,(k)}}{\Delta_{i,(k)}^2}  \right),$ which is shown to be nearly optimal by providing a lower bound of $\Omega \left(  \sum_{i \in [K]} \frac{\log 1/\delta }{\Delta_{i,(k)}^2} + \log \log 1/\Delta_{i,(k)} \right)$ for any algorithm able to identify the top-$k$ arms with probability at least $1-\delta$ under the above assumptions on $\bQ.$
	Here, the gaps $\Delta_{i,(k)}$ are defined for an arm $a_i$ by the calibrated pairwise preference probability of $a_i$ and the $(k+1)$st best arm if $a_i$ belongs to the top-$k$ arms, and otherwise by the calibrated pairwise preference probability of $a_i$ and the $k$th best arm.

	\Algo{SEEKS}  is proceeding in rounds, just like \Algo{SEEBS}, but uses two versions of \Algo{T-$k$-S} to successively build a pool of candidate arms for the top-$k$ set: The first variant is used to extract at the beginning of each round a smaller subset of all currently selectable arms with the size equal to the number of remaining arms for the final top-$k$ set, while the second variant is used to determine the nearly worst arm of the latter subset\footnote{Here, nearly worst is once again to be understood in terms of the rank in the underlying total order of the arms.}.
	The nearly worst arm is then used in a similar manner as the round-wise ``champion'' of the \Algo{SEEBS} algorithm to partition the initially extracted subset of arms into three subsets (surely preferred, surely not preferred and undecided arms) and include all surely preferred arms into the pool of candidates, and eliminate all surely not preferred \emph{and} the surely preferred arms from the list of selectable arms for the next round.   
	The procedure stops as soon as either the pool of candidates consists of at least $k$ arms, or the cardinality of the candidate pool and still selectable arms is at most $k.$
	In the first termination event, a randomly chosen $k$-sized subset of the candidate pool is the final top-$k$ set, while in the second termination event the candidate pool augmented by a random subset of the selectable arms in order to form a $k$-sized subset is used for the final top-$k$ set.
	Once again, adapting the round-wise confidence levels in an appropriate way over the rounds leads to a sample complexity of $\bigO  \left(  \sum_{i \in [K]} \frac{\log K/\delta + \log \log 1/\Delta_{i,(k)}}{\Delta_{i,(k)}^2}  \right)$ for  \Algo{SEEKS}.

	\subsubsection{Iterative-Insertion-Ranking}  \label{sec_iterative_insertion_ranking}
	Assuming that a total order over the arms exists and shall be found, \cite{ren2019sample}  leverage the idea of the \Algo{BinarySearch} algorithm \citep{feige1994computing} in order to deal with the case of an unknown lower bound for the calibrated pairwise preference probabilities $|\Delta_{i,j}|.$
	To this end, they introduce the \Algo{Iterative-Insertion-Ranking} (\Algo{IIR}) algorithm, which  essentially is a binary insertion sort to rank the arms, where the underlying tree structure is a \emph{preference interval tree} (PIT) as also used in \Algo{BinarySearch}. 

	A PIT is a binary tree where each node $n$ maintains a triple of elements $n[1],n[2],n[3]$, each of which is an element of $\cA \cup \{-\infty,\varnothing,+\infty\},$ where  $-\infty,\varnothing,+\infty$ are to be understood as symbols representing artificial arms.
	For each inner node $n$, it holds that $n[2] \neq \varnothing,$ while for all leaf nodes $l$, the second element $l[2]$ is always set to $\varnothing.$
	Further, each inner node $n$ has a pointer to its children $n.left$ and $n.right,$ respectively, and it holds that 
	$n[1]=n.left[1],$ $n[3]=n.right[3]$ as well as $n[2]=n.left[3]=n.right[1].$
	The root node $n_0$ of the PIT is such that $n_0[1]=-\infty$ and $n_0[3]=+\infty.$
	Given these conditions, the idea is that the sequence of leaf nodes from left to right represent the underlying total order of the arms: If $l_1,\ldots,l_{K+1}$ are the leaf nodes from left to right, then the total order of the arms is given by $l_{2}[1]\prec l_{3}[1] \prec \ldots \prec l_{K+1}[1]$ or equivalently $l_{1}[3]\prec l_{2}[3] \prec \ldots \prec l_{K}[3].$
	
	The \Algo{IIR} algorithm successively builds a PIT by inserting each arm one after the other, once being confident enough about its right position, by creating leaf nodes and updating the inner nodes accordingly.
	To this end, the algorithm is repeatedly trying to insert an arm $a_i$ into the PIT by moving through the PIT starting from the root and successively dueling $a_i$ with the (non-artificial) arms of the current node in order to determine the direction of the next move, or to insert the arm into the PIT\,---\,both based on certain confidence levels.
	Once a leaf node is created for the current arm, the insertion mechanism is started for a not yet included arm until all arms are included in the PIT.
	The used confidence levels are based on a guess on $\min_{j\neq i} \Delta_{i,j}$, and it is shown that the insertion of an arm is at the right position of the PIT with high probability once the guess is appropriate, while the guess is updated in case the arm cannot be inserted with certainty into the PIT. 
	In the latter case, the insertion mechanism is started from scratch with the same arm.

	A sample complexity of $ \bigO \left( \sum_{i\in [K]} \frac{ (\log\log(\min_{j\neq i} \Delta_{i,j}^{-1}) + \log(K/\delta) )}{\min_{j\neq i} \Delta_{i,j}^{2}} \right) $
	is derived for \Algo{IIR} in order to guarantee that the true underlying ranking is represented by the final PIT with probability at least $1-\delta.$
	This sample complexity is nearly optimal if strong stochastic transitivity holds, as the authors derive a lower bound on the sample complexity of any learner to return the true underlying ranking maintaining the confidence level $\delta$ of 
	$$  \Omega \left( \sum\nolimits_{i\in [K]} \frac{ (\log\log(  \tilde \Delta_i^{-1}) + \log(1/\delta) )}{\min_{j\neq i} \tilde \Delta_i^{2}} + \inf\left\{	\sum\nolimits_{i \in [K]} \tilde \Delta_i^{-2} 	\log(1/p_i)   \, \big| \, \sum\nolimits_{i \in [K]} p_i \leq 1	\right\} \right),$$ 
	where $\tilde \Delta_i$ denotes the smaller of the two calibrated pairwise preference probabilities of an arm $a_i$ to its (two)  adjoining arms according to the underlying true ranking.
	In particular, $\tilde \Delta_i$ and $\min_{j\neq i} \Delta_{i,j}$ are in general not the same, but coincide if $\bQ$ satisfies strong stochastic transitivity.
	
%
%	Each move is due to the stochastic feedback of the duels based on 
%	
	
%

	\subsection{Regularity Through Latent Utility Functions} \label{subsec_utility_dueling_bandits}
	The representation of preferences in terms of utility functions has a long history in decision theory \citep{Fi70}. The idea is that the absolute preference for each arm can be reflected by a real-valued utility degree.

	This idea applied to the dueling bandits setting  is also known as \emph{utility-based dueling bandits,} where one assumes the existence of a latent utility function $u:\cA \to \R$ with $u(a_i)$ representing the utility of an arm $a_i\in \cA.$
	Further, the probability of the outcome of a pairwise comparison between two arms $a_i,a_j$ is determined by the difference of their utilities.
	However, as the difference is not necessarily a value inside  the unit interval, one makes use of a link function $\sigma:\R \to [0,1]$ in order to map these differences of utilities to actual probabilities.
	Formally,
	\begin{align} \label{eq_utility_based_pairwise}
	\prob ( a_i \succ a_j ) =  \sigma\big(		u(a_i) - u(a_j)	\big),
	\end{align}
	so that $\Delta_{i,j}= \sigma\big(		u(a_i) - u(a_j)	\big) - 1/2$ in this setting.
	The minimal assumptions on the link function are as follows:
	\begin{itemize}
		\item [1.] strict monotonicity on $\sigma^{-1}(0,1)$, 
		%	on $\sigma^{-1}(0,1)$, 
		so that an arm with a higher utility than another arm will also have a higher probability to be chosen than the latter;
		\item [2.] $\sigma(0)=1/2$, which implies that two arms having the same utility have also the same chance to beat the other one, respectively.
	\end{itemize}
	In principle, any cumulative distribution function of a symmetric continuous random variable is in line with these conditions.
	The two most common link functions, which are both satisfying the conditions, are the \emph{logistic link function} $\sigma(x) = 1/(1+\exp(-x))$ and the \emph{linear link function}  $\sigma(x) =  \max\{ 0, \min\{1, 1/2 \cdot (1+x) \}\}.$ 

	Under the above assumptions on the link function and the additional assumption that the utility function is injective, the low noise model assumption holds for $\bQ.$
	Indeed, injectivity of $u$ implies for distinct arms $a_i,a_j\in \cA$ that 	
	$$ \Delta_{i,j}= \sigma\big(		u(a_i) - u(a_j)	\big) - 1/2 \neq 0 \, .
	$$
	%\\
	Without loss of generality, let $u(a_i)>u(a_j)$. Then, by the strict monotonicity of $\sigma$, we obtain
	$$	\Delta_{i,k}= \sigma\big(		u(a_i) - u(a_k)	\big) - 1/2 > \Delta_{j,k}= \sigma\big(		u(a_j) - u(a_k)	\big) - 1/2 	
	$$
	for all $k\in[K] \setminus \{ i,j \}$.	
	Thus, summing over all $k$ shows that the low noise model assumption is fulfilled.
	Note that in case $u$ is not injective, the implication does not hold in general and also a total order over the arms does not exist. However, one can still infer that $\mathrm{SST}$ is satisfied for the corresponding preference relation $\bQ.$

	Another appealing property of the approach based on utility functions is the possibility to model the dueling bandits problem for the case of infinitely many arms. 
	Let $\cS$ be some space of arms, which is not necessarily finite\footnote{This space corresponds to our set of arms $\cA$. However, as we assume $\cA$ to be finite, we use another notation here.}. 
	Then, one assumes the existence of $u:\cS \to \R$ with $u(a)$ representing the utility of an arm $a\in \cS.$
	With this, the pairwise comparison probabilities are modeled as in (\ref{eq_utility_based_pairwise}) by means of some suitable link function $\sigma.$
	Assume $u$ to be injective and $\sigma$ to satisfy the requirements above. Then, similar to the case of finitely many arms, one can infer that the low noise model assumption for the infinite arm scenario holds, while in case of a non-injective $u$,  $\mathrm{SST}$ still holds \footnote{Here, a suitable substitute for the sum occurring in the low noise model assumption is necessary.	One possible approach is available in the case, where $(\cS,\mathbb{S},\mu)$ is a measure space.
		Then, an extension of the low  noise assumption can be defined as follows: For all $a,\tilde a \in \cS$ with $a \neq \tilde a$, it holds that $\Delta_{a,\tilde a}\neq 0$, and if $\Delta_{a,\tilde a}>0$, then 
		$$	\int_{\cS} \Delta_{a,a'}  \mathrm{d}\mu(a') > 	\int_{\cS} \Delta_{\tilde a,a'}  \, \mathrm{d}\mu(a').
		$$}.

	In summary, the assumption of a latent utility function for the arms is a stronger assumption than the regularity properties in Section \ref{subsec:regaxiom}, provided the utility function is injective.
	However, due to the increased structural complexity of a space of infinitely many arms, the utility assumption seems to be necessary, as it provides a surrogate for the quantitative rewards making the learning possible based on merely qualitative feedback.
	
	The regret definition in the utility-based dueling bandits scenario is similar to the one in (\ref{eq:regret}) and given by
	% and can be written as
	\begin{align}\label{eq:regret_utility}
	R^{T} 
	= \sum_{t=1}^{T} \Delta_{a_{*}, a_{t}} + \Delta_{a_{*}, a^{\prime}_{t}}
	= \sum_{t=1}^{T} \sigma ( u(a_{*}) - u(a_{t}) ) + \sigma ( u(a_{*}) - u( a^{\prime}_{t} ) ) - 1 \enspace,
	\end{align}
	where $(a_{t},a^{\prime}_{t})$ is the pair of arms chosen in time step $t.$
	%%%% denoting the arm selected by the learner by i^{t} is not a good idea here since the space is continous
	Here, however, two variants are considered for the reference arm $a_{*}:$ 
	%there are 
	%
	\begin{itemize}
		\item [(i)]  $a_{*}$ is the best one known only in hindsight, which is inspired by the classical regret definition in the realm of online learning with full information.
		%	
		%	The  definition of $a_{*}$ is depen
		%	In other words, $a_{*}$ is the best arm among those evaluated during the search process.
		%	
		\item [(ii)] $a_{*}$ is an arm with highest utility, i.e., $a_{*} \in \begin{cases}
		\argmax_{ a \in \cA } u(a), & \mbox{finite arm case,} \\ \argmax_{ a \in \cS } u(a), & \mbox{infinite arm case.} 
		\end{cases}$
		
		%	\in  or $a_{*} \in \argmax_{ a \in \cS }$ (finite arm case) is the best one known only in hindsight,
		%	
	\end{itemize}
	The problem of finding the best arm in hindsight can be viewed as a noisy online optimization task \citep{FiBeMe11}, where the underlying search space is $\cS$, and the function values cannot be observed directly; instead, only noisy pairwise comparisons of function values (utilities) are available. In this framework, it is hard to have a reasonable estimate for the gradient, so that classical online optimization algorithms are not directly applicable.
	
	In the following, we investigate existing methods for PB-MAB problems that are posing a latent utility structure on the available arms. For reasons of clarity, we list all these approaches in Table \ref{tab:utility_based_algo} in the spirit of the previous section.
	
	%Obviously, such degrees immediately impose a total order on the set of alternatives. Typically, however, the utility degrees are assumed to be latent and not directly observable.  

	\begin{table}
		\scriptsize
		\begin{center}		
			\begin{tabular}{|p{2cm}|p{1.5cm}|p{3.25cm}|p{2.5cm}|p{3.5cm}|}
%				{|p{2.5cm}|p{2.5cm}|p{3.5cm}|p{2.5cm}|p{4cm}|}
				\hline
				\textbf{Algorithm} & \textbf{Algorithm class(es)} & \textbf{Assumptions} & \textbf{Target(s) and goal(s) of learner} & \textbf{Theoretical \newline guarantee(s)} \\
				\hline
				\hline
				Dueling Bandit Gradient Descent (Section \ref{subsec_DBGD})
%				\citep{YuJo09} 
				& Online optimization based
				& Strictly concave, Lipschitz-continuous  utility function and rotation-symmetric, second order Lipschitz-continuous  link function
				& Expected regret minimization with best arm in hindsight  
				& $\hphantom{text}$ \newline $\bigO \left( T^{3/4} \sqrt{d} \enspace  \right)$ \\
%				\hline
%				Multi-Point Deterministic Gradient Descent  (Section \ref{subsec_MPGD})
%%				 \citep{ZhKi16} 
%				&  Online optimization based
%				& No explicit assumptions
%				&   Satisfactory experimental performance for total ranking
%				&  No guarantees
%				\\
				\hline
				Noisy Comparison-based Stochastic Mirror Descent  (Section \ref{subsec_stochastic_mirror_descent})
%				\citep{Ku17} 
				& Online optimization based
				& Smoothness and concavity assumptions for the utility functions;  differentiability assumption \& shape-constraints on link function
				%			\& $\alpha$-strong concavity of utility functions, differentiability assumption \& shape-constraints on link function
				& Expected regret minimization with best arm in hindsight 
				& $\hphantom{text}$ \newline $\bigO \left( d  \sqrt{T \, \log T}  \right)$ 
				\\
				\hline
				Doubler \newline (Section \ref{subsec_dobler_multisbm})
%				\citep{AiKaJo14} 
				%			3. Sparring \citep{AiKaJo14}
				& Reduction-based 
				%			Total order over arms, relaxed stochastic transitivity and stochastic triangle inequality both relative to the best arm 
				& Linear link function
				& Expected regret minimization  for best arm   
				&  Finite arms: \newline {\tiny $\bigO \left( \frac{K  \log^2 T }{\min_{j \neq i^{*}} \Delta_{i^{*},j}  }\right) $ } \newline
				Infinite arms: \newline {\tiny $\bigO \left( \frac{d^2  \log^4 T }{\min_{a \neq a_{*}} \Delta_{a_{*},a}  }\right) $} \ \ $\quad$ resp.\
				{\tiny $\bigO \left(  \sqrt{d \, T \, \log^3(T)} \right) $ }
%				\newline 
%				2.\ Finite arms: \newline  $\bigO \left( \sum_{i \neq i^* }\frac{ \log T + K \log(K) }{ \Delta_{i^{*},j}  }\right) $
				%			3. No guarantees
				\\
				\hline		
				MultiSBM \newline (Section \ref{subsec_dobler_multisbm})
%				\citep{AiKaJo14} \
				%			3. Sparring \citep{AiKaJo14}
				& Reduction-based 
				%			Total order over arms, relaxed stochastic transitivity and stochastic triangle inequality both relative to the best arm 
				& Linear link function
				& Expected regret minimization  for best arm   
				& Finite arms: \newline  $\bigO \left( \sum_{i \neq i^* }\frac{ \log T + K \log(K) }{ \Delta_{i^{*},j}  }\right) $
				%			3. No guarantees
				\\
				\hline
				Winner stays (Section \ref{subsec_winner_stay_utility})
%				\citep{ChFr17} 
				&Tournament
				& Linear link function 
				& Expected weak and strong regret minimization for best arm   
				& Weak regret: \newline $ \mathcal{O}\left( \frac{K \log K}{\min_{i,j} \Delta_{i,j}^5 } \right) $ \newline
				Strong regret:	\newline  $ \mathcal{O}\left(  K \log K + K \log T  \right) $  
				\\
				\hline	
				Dueling Bandits Temporary Elimination Algorithm \newline (Section \ref{subsec_DBTEA})
%				\citep{ZiSe18} 
				&  Generic tournament
				%			Total order over arms, relaxed stochastic transitivity and stochastic triangle inequality both relative to the best arm 
				& Linear link function
				& Expected regret minimization  for best arm    
				& $\hphantom{text}$ \newline $\bigO \left( \sum_{i \neq i^* }\frac{ \log T }{ \Delta_{i^{*},j}  }\right) + \bigO \left(K\right)$ 
				\\
				\hline			
				Round-Efficient Dueling Bandits \newline (Section \ref{subsec_round_efficient_dueling})
				%				\citep{MaGr17} 
				& Tournament
				%			Total order over arms, relaxed stochastic transitivity and stochastic triangle inequality both relative to the best arm 
				& Linear link function
				& 1.\ Sample and round complexity minimization  for best arm \newline
				\newline
				\newline
				\newline
				 $2. \, (\epsilon,\delta)$-PAC variant of 1.
				& 
				1.\ Round complexity: $\bigO \left(  \frac{\log\big(  \frac{K}{\delta \, \min_{j\neq i^*} \Delta_{i^*,j}} \big)}{\min_{j\neq i^*} \Delta_{i^*,j}^2}  \right) $ 
				Sample complexity: $\bigO \left(  \sum_{j\neq i^*} \frac{\log\big(  \frac{K}{\delta \, \Delta_{i^*,j}} \big)}{ \Delta_{i^*,j}^2}  \right) $
				2.\ Round complexity: $\bigO \left(  \frac{\log\big(  \frac{K}{\delta \, \epsilon } \big)}{\epsilon^2}  \right) $ \newline
				Sample complexity: See \eqref{sample_complex_peer_grading}
				\\
				\hline
				Multisort \newline (Section \ref{subsec_multisort})
				%				\citep{MaGr17} 
				& Noisy-sorting
				%			Total order over arms, relaxed stochastic transitivity and stochastic triangle inequality both relative to the best arm 
				& Logistic link function and well-separated utilities
				& Sample complexity minimization  for total ranking  
				& $\hphantom{text}$ \newline $\bigO \left( K \log^6 K \right) $ 
				\\
				\hline
				
			\end{tabular}
		\end{center}
%	}
			\caption{Utility-based approaches for the dueling bandits problem. In the case of infinitely many arms, the dimension of the space of arms is denoted by $d$, while $i^{*}$ is the index representing the best arm in case of finite arms and $a^*$ for infinite arms.}
			\label{tab:utility_based_algo}
	\end{table}
	
	\subsubsection{Dueling Bandits Gradient Descent} \label{subsec_DBGD}
	The dueling bandits setting has originally been introduced by \cite{YuJo09}.
	The authors consider the case of a possibly infinite number of arms, which are all elements of a compact, convex subset of $\R^d$ containing the origin. 
	%and
	Further, they assume the existence of a strictly concave, Lipschitz-continuous utility function $u$ of the arms, as well as a rotation-symmetric, second-order Lipschitz-continuous link function $\sigma.$
	The goal of the learner is to minimize the (expected) regret defined as in (\ref{eq:regret_utility}), where the reference arm is the best arm in hindsight.
	As mentioned above, this problem can be seen as an online noisy optimization task, where, however, a reasonable estimate for the gradient is lacking.
	Therefore, the authors opt for applying an online convex optimization method \citep{FlKaMc05}, which does not require the gradient to be calculated explicitly, and instead optimizes the parameter by estimating an unbiased gradient approximation. 
	The suggested algorithm, called Dueling Bandit Gradient Descent (\Algo{DBGD}), is an iterative search method traversing the space $\cS$ over the time horizon $\{1,2,\ldots,T\}.$
	%that proceeds in discrete time step $1,,t,...$. 
	%
	For a time step $t$, let $a_{t}\in \cS$ be the current point. To determine the dueling arm for $a_{t},$ a random direction $u_{t}$ is drawn uniformly from the unit ball and $a^{\prime}_{t}$ is set to be $\Pi_{\cS} (a_{t} + \delta u_{t} ),$ where $ \Pi_{\cS}(\cdot) $ denotes the projection into $ \cS $, and $ \delta $ is some step parameter steering the degree of exploration.
	This point $a^{\prime}_{t}$ is compared with the current point $a_{t},$ with two possible consequences for the update of the algorithm: if $a_{t}$ wins the comparison, the subsequent point $a_{t+1}$ is set to be $a_{t},$ while in the case of a defeat, the subsequent point is $\Pi_{\cS} (a_{t} + \gamma u_{t} )$ with $\gamma>0.$ 
	%is used for the subsequent reference point.
	%
	Here, $\gamma$ is an exploitation parameter of the method, as it determines the step size with which an update is taken into the winner direction.
	%

	%In the theoretical analysis of the proposed method, called Dueling Bandit Gradient Descent (\Algo{DBGD}), 
	%Under a strong convexity assumption on $\delta$, an expected regret bound for the proposed algorithm is derived. 
	%
	Assuming that the space of arms $\cS$ is a $d$-dimensional ball of radius $R$, the expected regret of the proposed method
	%, called Dueling Bandit Gradient Descent (\Algo{DBGD}),
	is shown to be bounded by
	$\exptd[ R^{T} ] \le  2T^{3/4} \sqrt{10RdL}$
	for an appropriate choice of $\delta$ and $\gamma$,  where $ L $ is the product of the Lipschitz constants of the link function $\sigma$ and the latent utility function $u.$
		
%	]
	%
	%the regret definition is similar to the one in (\ref{eq:regret}), and can be written as
	%\[
	%R^{T} = \sum_{t=1}^{T} \delta ( a_{*}, a_{t} ) + \delta (a_{*}, a^{\prime}_{t} ) \enspace .
	%\]
	%%%%% denoting the arm selected by the learner by i^{t} is not a good idea here since the space is continous
	%Here, however, the reference arm $a_{*}$ is the best one known only in hindsight. In other words, $a_{*}$ is the best arm among those evaluated during the search process.
	
	%Under a strong convexity assumption on $\delta$, an expected regret bound for the proposed algorithm is derived. More specifically, assuming the search space $\cS$ to be given by the $d$-dimensional ball of radius $R$, the expected regret is
	%\[
	%\exptd[ R^{T} ] \le  2T^{3/4} \sqrt{10RdL} \enspace ,
	%\] 
	%where $ L $ is the Lipschitz constant of $ \delta $.
	
%	\subsubsection{Multiple-point Gradient Descent} \label{subsec_MPGD}
%	

	\subsubsection{Noisy Comparison-based Stochastic Mirror Descent} \label{subsec_stochastic_mirror_descent}
	%\subsubsection{Stochastic Mirror Descent}
	\citet{Ku17} studies the utility-based dueling bandits problem imposing convexity and smoothness assumptions for the utility function, which are stronger than those by \citet{YuJo09}, and which guarantee the existence of a unique minimizer of the utility function. Other assumptions concern the link function, which are weaker than those by \citet{AiKaJo14} (cf.\ Section \ref{subsec_dobler_multisbm}) and satisfied by common functions, including the logistic function used in the experiments by \citet{YuJo09}, the linear function used by \citet{AiKaJo14}, and the Gaussian distribution function. 
	
	Motivated by the fact that \cite{YuJo09} use a stochastic gradient descent algorithm for the utility-based dueling bandits problem, the authors propose to use a stochastic mirror descent algorithm, which is known to ensure near optimality in convex optimization problems. 
	To this end, the expected regret minimization problem of a utility-based dueling bandits problem is reduced to a locally-convex optimization problem, for which the proposed algorithm, called Noisy Comparison-based Stochastic Mirror Descent (\Algo{NC-SMD}) can be analyzed in the spirit of bandit convex optimization problems \citep{hazan2016introduction,kumagai2018introduction}.
	With this, it is shown that \Algo{NC-SMD} achieves a regret bound of $ \bigO \left( d \sqrt{T \log T} \right) $ in expectation  if the set of arms $\cS$ is a compact convex subset of $\R^d$ with non-empty interior and the learning rate of \Algo{NC-SMD} is suitably tuned (i.e., depending on the time horizon $T$).
	The latter regret bound is optimal except for a logarithmic factor, as they derive a lower bound of order $\Omega(d\sqrt{T})$ by relating their problem to a convex optimization problem.

	%\subsubsection{Doubler}
	\subsubsection{Reduction to Value-based MAB}  \label{subsec_dobler_multisbm}
	\citet{AiKaJo14} propose various methodologies to reduce the utility-based PB-MAB problem to the standard value-based MAB problem for the case of finitely many arms (denoted by $\cA$) as well as for infinitely many arms (denoted by $\cS$).
	In their setup, the utility of an arm is assumed to be in $[0,1]$. Formally, $u : \cS \rightarrow [0,1]$, and the link function is the linear link function 
	%$\sigma_{lin} ( x ) =\frac{1}{2} x$. 
	$\sigma(x)=\max\{ 0, \min\{1, 1/2 \cdot (1+x) \}\}.$
	Therefore, the probability of an arm $a\in \cS$ beating another arm $a' \in \cS$ is
	\[
	\prob ( a \succ a' ) = \frac{1 + u(a) - u(a')}{2} \enspace ,
	\]
	which is again in $[0,1]$. The regret considered is the one defined in (\ref{eq:regret_utility}), where the reference arm $a_{*}$ is the globally best arm with maximal utility. 
	
	\citet{AiKaJo14} propose two reduction techniques, \Algo{Doubler} and \Algo{MultiSBM}, for a finite and an infinite set of arms. In both techniques, value-based MAB algorithms such as \Algo{UCB} \citep{AuCeFi02} are used as a black box for driving the search in the space of arms. For a finite number of arms, value-based bandit instances are assigned to each arm, and these bandit algorithms are run in parallel. More specifically, assume that an arm $i(t)$ is selected in iteration $t$ (to be explained in more detail shortly). Then, the bandit instance that belongs to arm $i(t)$ suggests another arm $j(t)$. These two arms are then compared in iteration $t$, and the reward, which is 0 or 1, is assigned to the bandit algorithm that belongs to $i(t)$. In iteration $t+1$, the arm $j(t)$ suggested by the bandit algorithm is compared, that is, $i(t+1) = j(t)$. What is nice about this reduction technique is that, under some mild conditions on the performance of the bandit algorithm, the preference-based expected regret defined in (\ref{eq:regret}) is asymptotically identical to the one achieved by the value-based algorithm for the standard value-based MAB task. 
	
	For infinitely many arms, the reduction technique can be viewed as a two player game. A run is divided into epochs: the $\ell$th epoch starts in round $t = 2^\ell$ and ends in round $t = 2^{\ell+1}-1$, and in each epoch the players start a new game. During the $\ell$\/th epoch, the second player acts adaptively according to a strategy provided by the value-based bandit instance, which is able to handle infinitely many arms, such as the ConfidenceBall algorithm by \citet{DaHaKa08}. The first player obeys some stochastic strategy, which is based on the strategy of the second player from the previous epoch. That is, the first player always draws a random arm from the multi-set of arms that contains the arms selected by the second player in the previous epoch. This reduction technique incurs an extra $\log T$ factor to the expected regret of the value-based bandit instance.

%	\editvik[
	\subsubsection{Winner Stays} \label{subsec_winner_stay_utility}
	The WS-W resp.\ WS-S algorithm \citep{ChFr17}, which was already discussed in Section \ref{sec_winner_stays_algo}, has also been analyzed by the authors for the case of a utility-based dueling bandits scenario with a linear link function and finitely many arms.
	In particular, the weak regret resp.\ strong regret version of (\ref{eq:regret_utility}) are considered (cf.\ discussion following (\ref{eq:regret}))  and a bound of order $ \mathcal{O}\left( \frac{K \log K}{\min_{i,j} \Delta_{i,j}^5 } \right) $ is derived for WS-W for minimizing weak cumulative regret, while for minimizing strong  cumulative regret a bound of order $ \mathcal{O}\left(  K \log K + K \log T  \right) $  in expectation is inferred for WS-S.
%	Although the regret bounds of these algorithms are not optimal for this setting, they are unique in the sense that the Gambler's ruin problem is used to upper bound the number of pulls of sub-optimal arms, whereas all other regret optimization algorithms reviewed in this study make use of the Chernoff bound in one way or the other.
	%
	%More specifically, they considered 
	%\begin{align*}
	%%\label{eq:regret_utility}
	%%	
	%R^{T} 
	%%	
	%= \sum_{t=1}^{T} \Delta_{a_{*}, a_{t}} + \Delta_{a_{*}, a^{\prime}_{t}}
	%%	
	%= \sum_{t=1}^{T} \sigma ( u(a_{*}) - u(a_{t}) ) + \sigma ( u(a_{*}) - u( a^{\prime}_{t} ) ) - 1 \enspace,
	%%	
	%\end{align*}
	%%
	%where $(a_{t},a^{\prime}_{t})$ is the pair of arms chosen in time step $t.$
	
	%The regret definition in the utility-based dueling bandits scenario is similar to the one in (\ref{eq:regret}):
	%% and can be written as
	%\begin{align}\label{eq:regret_utility}
	%%	
	%R^{T} 
	%%	
	%= \sum_{t=1}^{T} \Delta_{a_{*}, a_{t}} + \Delta_{a_{*}, a^{\prime}_{t}}
	%%	
	%= \sum_{t=1}^{T} \sigma ( u(a_{*}) - u(a_{t}) ) + \sigma ( u(a_{*}) - u( a^{\prime}_{t} ) ) - 1 \enspace,
	%%	
	%\end{align}
	%%
	%where $(a_{t},a^{\prime}_{t})$ is the pair of arms chosen in time step $t.$
	
%	]
	
	\subsubsection{Dueling Bandits Temporary Elimination Algorithm} \label{subsec_DBTEA}
	\cite{ZiSe18} introduce the general identifiability assumption (cf.\ Section \ref{subsec:regaxiom}) and show that any link function as used by \citet{YuJo09} fulfills this assumption.
	Like \citet{AiKaJo14}, they study the dueling bandits problem for a linear link function with the goal of expected regret minimization. 
	For this purpose, they consider the sub-optimality gaps $\tilde \Delta_{a} \defeq \min_{a' \neq a^*}  \Delta_{a^*,a'} - \Delta_{a,a'},$ where $a^*$ is an identifiable arm emerging in the general identifiability assumption, which allows them to rewrite the expected regret in (\ref{eq:regret_utility}) as 
	\begin{align*}
	\exptd [R_A^{T}] 
	= \sum_{t=1}^{T} \exptd [\delta ( a_{*}, a_{t} ) + \delta (a_{*}, a^{\prime}_{t} )] 
%	
%	= \sum_{t=1}^{T} \exptd \left[ \frac{ 2u(a^*) - u(a_t)  - u(a^{\prime}_{t}) }{2} \right]  
%	
	& =   \sum_{a \neq a^* }^{T} \exptd [N_T(a)] \,  \tilde \Delta_a  \enspace ,
	\end{align*}
	where $N_T(a)$ is the total number of plays of an arm $a.$ 

	Based on this relationship between the regret and the sub-optimality gaps, they suggest the \Algo{Dueling Bandits Temporary Elimination Algorithm} (DBTEA), which is a phase-based anytime algorithm maintaining a set of active arms consisting of arms having a non-positive lower confidence bound on the sub-optimality gap. For these arms, one is not yet certain enough that the sub-optimality gap is indeed positive (it is only zero for optimal arms). Each phase consists of dueling arms from two random orders of the active set, such that each arm in the active set duels twice in a phase.
	
	Exploiting  the relationship of the regret per time and the sum of the sub-optimality gaps is at the heart of their approach, so that a different proof technique is required to obtain confidence bounds for the latter sum.
	For this purpose, they provide a novel anytime concentration inequality for the sum of sub-Gaussian random variables.
	
	The authors' theoretical analysis of the DBTEA leads to an upper bound on its expected regret of  $\mathcal{O}(K) + \mathcal{O}( \sum_{i\neq i^*} \frac{\log(T)}{\tilde \Delta_i} ).$  
	%improving on the approaches before with respect to the additive horizon independent term, that is, reducing $\mathcal{O}(K^2)$ to $\mathcal{O}(K).$
	%
	In numerical experiments, they illustrate that the additive term appearing in some of the theoretical upper bounds of state of the art algorithms (cf.\ Table \ref{tab:regaxiom}) might dominate the regret for problem instances with a large number of arms and moderate time horizons, so that the DBTEA is superior to some state of the art algorithms in these cases.
	
	The setting in \cite{ZiSe18} is in fact more general, as they introduce the \emph{factorized bandit problem}, in which each action consists of a Cartesian product of elementary actions.
	This is a generalization of the so-called rank-1 bandit problem \citep{katariya2017stochastic}, and allows to model the dueling bandits setting by considering $\cA  \times \cA $ as the action space, i.e., the action is to ``generate'' a duel between two arms.

	\subsubsection{Round-Efficient Dueling Bandits} \label{subsec_round_efficient_dueling}
	
	Inspired by real world scenarios such as sport tournaments, where participants simultaneously compete against each other at a time, \cite{lin2018efficient} consider the dueling bandits scenario where the learner maintains in each iteration a set of active arms, builds a set of disjoint pairs of these arms and obtains all (noisy) pairwise preferences of these pairs.
	As a matter of fact, this problem scenario can be seen as a special case of the multi-dueling bandit problem, which will be discussed in Section \ref{sec:multi_duel_extensions}, since the learner obtains preference feedback in the form of a sequence of pairwise preferences, or in other words a partial preference feedback (cf.\ Section \ref{subsec_feedback_partial_preference}).
	However, as the authors assume a latent utility for each arm and a linear link function underlying the probability in \eqref{eq_utility_based_pairwise}, their problem scenario fits quite well into the scope of this section.

	The goal is to find the best arm while keeping the total number of rounds as well as the total number of pairwise comparisons as small as possible.
	Here, the number of rounds is simply the number of iterations, i.e., how many times the learner carries out its action in the form of a pairwise comparison of the pairs of active arms.
	Due to the underlying latent utilities, the best arm is the one with the highest utility.
	The authors derive unbiased estimates and corresponding confidence intervals for the latent utilities of the active arms, and based on these design an algorithm which successively eliminates arms in the active set as soon as their estimated utility is below the lower confidence utility of the empirically best arm.
	At the beginning, apparently all arms are active and the algorithm simply partitions all available arms randomly into disjoint pairs.
	If the size of active arms is an odd number, i.e., there is one arm without a dueling partner, this arm is not compared with another arm until one of the arms is eliminated, whereupon the former builds a pair with the arm that competed with the eliminated one.
%	, but is composed to a dueling pair with the next available arm, which happens as soon 
	
	It is shown that this algorithm returns the arm with the highest utility with probability at least $1-\delta$ and has a round complexity of order 
 $$ \bigO \left(  \frac{1}{\min_{j\neq i^*} \Delta_{i^*,j}^2} \log\big(  \frac{K}{\delta \, \min_{j\neq i^*} \Delta_{i^*,j}} \big)  \right) $$
 and a (pairwise) sample complexity of order $\bigO \left(  \sum_{j\neq i^*} \frac{1}{ \Delta_{i^*,j}^2} \log\big(  \frac{K}{\delta \, \Delta_{i^*,j}} \big)  \right). $ 
	Further, it is shown that the algorithm is via a slight modification of its stopping criterion also an $(\epsilon,\delta)$-PAC learner with round complexity 
	$	\bigO \left(  \frac{1}{\epsilon^2} \log\big(  \frac{K}{\delta \, \epsilon } \big) \right)	$
	and sample complexity 
	\begin{align} \label{sample_complex_peer_grading}
		\bigO \left(  \sum_{j \notin \mathrm{CW}_\epsilon} \frac{\log\big(  \frac{K}{\delta \, \Delta_{i^*,j}} \big)}{ \Delta_{i^*,j}^2} + \frac{|\mathrm{CW}_\epsilon|}{\epsilon^2} \log \frac{K}{\epsilon \delta} \right), 
	\end{align}
	where $\mathrm{CW}_\epsilon= \{ i \in [K] \, | \, u_i \geq u_{i^*} - \epsilon  \}$ is the set of $\epsilon$-best arms, which is slightly different from the definition in Section \ref{subsec_near_opt_targets}.

	\subsubsection{Multisort} \label{subsec_multisort}
	\citet{MaGr17} address the ranking problem when comparison outcomes are generated from the Bradley-Terry (BT) \citep{BrTe52} probabilistic model with parameters $ u = (u_{1}, \ldots , u_{K})^\top \in \mathbb{R}_+^K$, which represent the utilities of the arms. Using the BT model, the probability that an arm $a_i$ is preferred to $a_j$ is given by
	\begin{align} \label{eq_bradley_terry}
		\prob ( a_i \succ a_j ) = \frac{1}{ 1 + \exp(-(u_{i}-u_{j}))} \enspace .
	\end{align}
	Thus, they end up with a utility-based dueling bandits problem with the logistic link function.
	The authors propose the \Algo{Multisort} algorithm for this setting, which essentially calls the \Algo{QuickSort} algorithm \citep{Ho62} multiple times, stores the obtained (noisy) rankings and aggregates them to a final ranking by using Copeland's method \citep{Co51}, in which the arms are increasingly sorted by their Copeland scores given by the total number of pairwise wins (cf.\ Section \ref{sec:noass}).
	In order to analyze the theoretical properties of \Algo{Multisort}, the authors assume that the utilities are well-separated, so that in particular a total order over the arms induced by the utilities exists.
	More specifically, it is assumed that the utility parameter $u$ is generated by a Poisson point process.
	Then, it is shown that one call of \Algo{QuickSort}  results in a ranking that is appropriately close to the underlying ranking of the arms with respect to Spearman's footrule distance given by $ F(\sigma,\tau) = \sum_{i=1}^{n} |\sigma(i)-\tau(i)| $, where $ \sigma(i) $ and $ \tau(i) $ are the ranks of $ i $ according to the rankings $ \sigma $ and $ \tau $, respectively. 
%	Under the assumption that the distance between adjacent parameters is stochastically uniform across the ranking, they first show that the output of a single call of the \Algo{QuickSort} algorithm \citep{Ho62} is a good approximation of the ground-truth ranking
%	in terms of the quality of a ranking estimate by its displacement with respect to the ground truth measured by the Spearman's footrule distance given by $ F(\sigma,\tau) = \sum_{i=1}^{n} |\sigma(i)-\tau(i)| $, where $ \sigma(i) $ and $ \tau(i) $ are the ranks of $ i $ according to the rankings $ \sigma $ and $ \tau $, respectively. 
	%
	With this, they can infer that using $ \bigO(\log^5 K) $ calls of \Algo{QuickSort} within the \Algo{Multisort} algorithm results in a ranking that is close to the total order over the arms with respect to Spearman's footrule distance  with high probability.
	As each call of \Algo{QuickSort} involves $ \bigO( K \log K) $ many pairwise comparisons, the total number of comparisons made by \Algo{Multisort} is of the order $\bigO \left( K \log^6 K \right),$ while the distance  of the final ranking to  the underlying ranking of the arms is $o(K)$ in terms of $F.$ 
	
%	are aggregated by using Copeland's method \citep{Co51}, in which the arms are increasingly sorted by their scores given by the total number of pairwise wins (cf.\ Section \ref{sec:noass}), in order to return the final ranking of their Multisort method.
%	In summary, they show that with a number of comparisons of order $\bigO \left( K \log^6 K \right) $, their method returns a ranking, which, with high probability, is close to the total order over the arms with respect to Spearman's footrule distance.
	%More precisely, they show that Multisort can recover the ground truth everywhere, except at a vanishing fraction of the items, i.e., all but a vanishing fraction of the arms are correctly ranked, based on which they propose an active-learning strategy that consists of repeatedly sorting the items. 
	%	
	%Then they show that the aggregation of $ O(\log ^5 K) $ independent runs of \Algo{QuickSort} using Copeland's method \citep{Co51}, in which the arms are increasingly sorted by their scores given by the total number of pairwise wins, can recover the ground truth everywhere, except at a vanishing fraction of the items, i.e., all but a vanishing fraction of the arms are correctly ranked, based on which they propose an active-learning strategy that consists of repeatedly sorting the items. 
	%

	For practical purposes, the authors suggest to run  \Algo{QuickSort} repeatedly for a budget of $ c $ pairwise comparisons, until the budget is exhausted, resulting in a set of $ c $ comparison pairs and their outcomes,  while ignoring the produced rankings themselves.
	Eventually, the final ranking is built by sorting the ML estimates of the BT model parameters based on the set of all $ c $ pairwise comparison outcomes.
	%More specifically, for a budget of $ c $ pairwise comparisons, they run \Algo{QuickSort} repeatedly until the budget is exhausted to get a set of $ c $ comparison pairs and their outcomes while ignoring the produced rankings themselves, and then they induce the final ranking estimate from the ML estimate over the set of all the $ c $ pairwise comparison outcomes.

	\subsection{Regularity Through Statistical Models}\label{sec:mall}
	Since one of the most general tasks in the realm of preference-based bandits is to elicit a ranking of the complete set of arms based on noisy (probabilistic) feedback, it is quite natural to establish a connection to statistical models of rank data \citep{Ma95}. 
	
	The idea of relating preference-based bandits to rank data models has been put forward by \citet{BuHuSz14}, who assume the underlying data-generating process to be given in the form of a probability distribution $\prob:\, \mathbb{S}_K \rightarrow [0,1]$. Here, $\mathbb{S}_K$ is the set of all permutations of $[K]$ (the symmetric group of order $K$) or, via a natural bijection, the set of all rankings (total orders) of the $K$ arms.    
	
	The probabilities for pairwise comparisons are then obtained as marginals of $\prob$. More specifically, with $\prob(\br)$ the probability of observing the ranking $\br$, the probability $q_{i,j}$ that $a_i$ is preferred to $a_j$ is obtained by summing over all rankings $\br$ in which $a_i$ precedes $a_j$:  
	\begin{align} 
	\label{eq:pairwisexy}
	q_{i,j} \defeq  \prob(a_i \succ a_j) = \sum_{\br \in \cL(r_{j} > r_{i})} \prob (\br) \enspace ,
	\end{align}
	where $\cL(r_{j} > r_{i}) \defeq \left\{ \br \in \mathbb{S}_K \,\vert\, \pi(j) > \pi(i) \right\}$ denotes the subset of permutations for which the rank  of $a_j$ is higher than the rank  of $a_i$  (smaller ranks indicate higher preference).
	In this setting, the learning problem essentially comes down to making inference about $\prob$ based on samples in the form of pairwise comparisons. 

	So far, only two ranking models have been considered in the realm of PB-MAB problems, namely the Mallows model \citep{Ma57} and the Plackett-Luce model \citep{Pl75,Lu59}.
	PB-MAB methods based on both models will be reviewed in the following subsections. Note that both models impose assumptions on the preference relation $\bQ$ that are stronger than those for the utility-based dueling bandits (and thus also stronger than those in Section \ref{subsec:regaxiom}), as the pairwise probabilities in (\ref{eq:pairwisexy}) of both models can be expressed in the form as stipulated by (\ref{eq_utility_based_pairwise}) for a suitable link function $\sigma$ and utility function $u:\cA \to \R.$
	
	Under the assumption of a statistical model, the two targets of finding an optimal arm resp.\ ranking (cf.\ Section \ref{subsec_targets}) can be formulated as follows. 

	\begin{enumerate} 
		\item[--] \emph{Optimal arm}: 
		%	The MPI problem consists of finding the most preferred item $i^{*}$, namely the 
		The optimal arm $i^{*}$ is the one having highest probability of being top-ranked: 
		\begin{align}
		i^{*} & \defeq \argmax_{1\le i \le K } \, \exptd_{\br \sim \prob}  
		\, \IND{ r_{i} = 1 } \notag  =\argmax_{1\le i \le K } \, \sum_{\br\in \cL(r_{i}=1)} \prob ( \br  ), \notag 
		\end{align} 
		where $\cL(r_{i} = 1 ) \defeq \left\{ \br \in \mathbb{S}_K \,\vert\,  \pi(i) = 1 \right\}$ denotes the subset of permutations for which the rank of $a_i$ is 1. 
		\item[--] \emph{Ranking}: The target ranking $\br^{*}$ is the mode of the distribution:
		\[
		\br^{*} \defeq \argmax_{\br \in \mathbb{S}_K}  \,  \prob (\br ).
		\]
		%	\item[--] The KLD problem calls for producing a good estimate $\widehat{\prob}$ of the distribution $\prob$, that is, an estimate with small KL divergence: 
		%	$$
		%	\KL \left( \prob, \widehat{\prob} \right) = \sum_{\br \in \mathbb{S}_K} \prob (\br ) \log \frac{\prob (\br)}{ \widehat{\prob} (\br) } < \epsilon 
		%	$$
	\end{enumerate}
%	}

	\subsubsection{Mallows} \label{subsec_mallows}
	\citet{BuHuSz14} assume the underlying probability distribution $\prob$ to be a Mallows model \citep{Ma57}, one of the most well-known and widely used statistical models of rank data \citep{Ma95}. The Mallows model or, more specifically, Mallows $\phi$-distribution is a parameterized, distance-based probability distribution that belongs to the family of exponential distributions:
	\begin{align}
	\label{eq:mallows}
	\prob ( \br \, \vert \, \phi, \ro ) = \frac{1}{ Z(\phi) } \, \phi^{d ( \br, \ro )} \enspace ,
	\end{align}
	where $\phi$ and $\ro$ are the parameters of the model: $\ro = (\pi^*_1, \ldots , \pi^*_K) \in \mathbb{S}_K$ is the location parameter (center ranking) specifying the position $\roi_i = \ro(i)$ of each arm $a_i$ in the ranking, and $\phi \in (0,1]$ is the spread parameter. Moreover, $d$ is the Kendall distance on rankings, that is, the number of discordant pairs:
	\[
	d(\br,\ro) \defeq \sum_{1 \leq i < j \leq K} \IND{ \, (\pi_{i} - \pi_{j})(\pi^*_{i}-\pi^*_{j})<0 \, }  \enspace ,
	\] 
	where $\IND{\cdot}$ denotes the indicator function.
	The normalization factor in (\ref{eq:mallows}) can be written as
	$$
	Z(\phi) = \sum_{\br\in \mathbb{S}_K} \prob ( \br \, \vert \, \phi, \ro )
	= \prod_{i=1}^{K-1}\sum_{j=0}^{i} \phi^{j}
	$$
	and thus only depends on the spread \citep{FlVe86}. Note that, since $d(\br,\ro)=0$ is equivalent to $\br = \ro$, the center ranking $\ro$ is the mode of $\prob( \cdot \, \vert \, \phi, \ro )$, that is, the most probable ranking according to the Mallows model.
	
	%In \citep{BuHuSz14}, three different goals of the learner, which are all meant to be achieved with probability at least $1-\delta$,  are considered, depending on whether the application calls for the prediction of a single arm, a full ranking of all arms, or the entire probability distribution: 
	
	%\begin{enumerate} 
	%	\item[--] The MPI problem consists of finding the most preferred item $i^{*}$, namely the item whose probability of being top-ranked is maximal: 
	%	\begin{align}
	%	i^{*} & =\argmax_{1\le i \le K } \, \exptd_{\br \sim \prob}  
	%	\, \IND{ r_{i} = 1 } \notag  =\argmax_{1\le i \le K } \, \sum_{\br\in \cL(r_{i}=1)} \prob ( \br  ) \notag 
	%	\end{align} 
	%	\item[--] The MPR problem consists of finding the most probable ranking $\br^{*}$:
	%	\[
	%	\br^{*} = \argmax_{\br \in \mathbb{S}_K}  \,  \prob (\br )
	%	\]
	%	\item[--] The KLD problem calls for producing a good estimate $\widehat{\prob}$ of the distribution $\prob$, that is, an estimate with small KL divergence: 
	%	$$
	%	\KL \left( \prob, \widehat{\prob} \right) = \sum_{\br \in \mathbb{S}_K} \prob (\br ) \log \frac{\prob (\br)}{ \widehat{\prob} (\br) } < \epsilon 
	%	$$
	%\end{enumerate}
	
	In the case of Mallows, it is easy to see that $\roi_{i}<\roi_{j}$ implies $q_{i,j}>1/2$ for any pair of items $a_{i}$ and $a_{j}$. That is, the center ranking defines a total order on the set of arms.
	% If an arm $a_{i}$ precedes another arm $a_{j}$ in the (center) ranking, then $a_{i}$ beats $a_{j}$ in a pairwise comparison\footnote{Recall that this property is an axiomatic assumption underlying the \Algo{IF} and \Algo{BTM} algorithms. Interestingly, the stochastic triangle inequality, which is also assumed by \citet{YuBrKlJo12}, is not satisfied for Mallows $\phi$-model \citep{Ma57}.}.  
	Moreover, as shown by \citet{Ma57}, the pairwise probabilities can be calculated analytically as functions of the model parameters $\phi$ and $\ro$ as follows:
	%\begin{Thm}[Mallows, 1957] %\cite{Ma57}
	%\label{thm:mar57}
	Assume the Mallows model with parameters $\phi$ and $\ro$. Then, for any pair of items $i$ and $j$ such that $\roi_{i}<\roi_{j}$, the pairwise probability is given by $q_{i,j}=g(\roi_{i},\roi_{j},\phi)$, where
	\[
	g(i,j,\phi)= h(j-i+1,\phi) - h(j-i,\phi) 
	\]
	with $h(k,\phi) = k/(1-\phi^{k})$.
	%\end{Thm}
	As shown by this result, the Mallows model even induces a structural property over the arms like for the utility-based dueling bandits, so that the Mallows model is an even stricter assumption than the total order assumption\footnote{Interestingly, the stochastic triangle inequality is not satisfied for the Mallows $\phi$-model \citep{Ma57}.}. 
	Further, one can show that the ``margin'' around $1/2$, i.e.,
	%Based on this result, one can show that the ``margin'' 
	$
	\min_{i \neq j} |1/2 - q_{i,j}|,
	$
	is relatively wide; more specifically, there is no pair $(i,j)$ such that $q_{i,j} \in ( \frac{\phi}{1+\phi},  \frac{1}{1+\phi} )$. Moreover, the result also implies that $q_{i,j}-q_{i,k} = \bigO(\ell \phi^\ell)$ for arms $a_i, a_j, a_k$ satisfying $\roi_{i}=\roi_{j}-\ell=\roi_k-\ell-1$ with $1<\ell$, and $q_{i,k}-q_{i,j} = \bigO(\ell \phi^\ell)$ for arms $a_i, a_j, a_k$ satisfying $\roi_{i}=\roi_{j}+\ell=\roi_k+\ell+1$ with $1<\ell$. Therefore, deciding whether an arm $a_j$ has higher or lower rank than $a_i$ (with respect to $\ro$) is easier than selecting the preferred arm from two candidates $a_j$ and $a_k$ for which $j,k\neq i$.
	
	%Based on these observations, one can devise an efficient algorithm for identifying the most preferred arm when the underlying distribution is Mallows. The algorithm proposed in \citep{BuHuSz14} for the MPI problem, called \Algo{MallowsMPI}, is similar to the one used for finding the largest element in an array. However, since a stochastic environment is assumed in which the outcomes of pairwise comparisons are random variables, a single comparison of two arms $a_{i}$ and $a_{j}$ is not enough; instead, they are compared until 
	For the problem of finding the optimal arm with high probability and a minimal sample complexity, \citet{BuHuSz14}  propose the
	\Algo{MallowsMPI} algorithm, which is inspired by the algorithm for finding the largest element in an array. 
	% Due to the stochastic environment is assumed in which the outcomes of pairwise comparisons are random variables, a single comparison of 
	Any two arms $a_{i}$ and $a_{j}$ are compared until
	\begin{equation}\label{eq:ovl}
	1/2 \notin \big[ \, \widehat{q}_{i,j}-c_{i,j},\widehat{q}_{i,j}+c_{i,j} \, \big] \enspace 
	\end{equation}
	holds.
	This simple strategy finds the most preferred arm with probability at least $1-\delta$ for a sample complexity that is of the form $\bigO \left( \frac{K}{\rho^2} \log \frac{K}{\delta \rho }\right)$, where $\rho = \frac{1-\phi}{1+\phi}$. 
	
	%For the MPR problem, a sampling strategy called \Algo{MallowsMerge} is proposed, which is based on the merge sort algorithm for selecting the arms to be compared. However, as in the case of MPI, two arms $a_{i}$ and $a_{j}$ are not only compared once but until condition (\ref{eq:ovl}) holds. The \Algo{MallowsMerge} algorithm finds the most probable ranking, which coincides with the center ranking of the Mallows model, with a sample complexity of 
	
	For the problem of finding the most probable ranking with high probability and a minimal sample complexity, a sampling strategy called \Algo{MallowsMerge} is proposed, which is based on the merge sort algorithm for selecting the arms to be compared. However, as in the case of finding the optimal arm, two arms $a_{i}$ and $a_{j}$ are not only compared once but until condition (\ref{eq:ovl}) holds. The \Algo{MallowsMerge} algorithm finds the most probable ranking, which coincides with the center ranking of the Mallows model, with a sample complexity of 
	$$
	\bigO  \left( \frac{K\log_2 K}{\rho^{2}} \log \frac{K\log_2 K}{\delta \rho}\right) \enspace ,
	$$
	where $\rho = \frac{1-\phi}{1+\phi}$. The leading factor of the sample complexity of \Algo{MallowsMerge} differs from the one of \Algo{MallowsMPI} by a logarithmic factor, which is in line with the results in Section \ref{subsec:regaxiom} for the exact sample complexity of best arm and top-$K$ ranking.
%	This was to be expected, and simply reflects the difference in the worst case complexity for finding the largest element in an array and sorting an array using the merge sort strategy. 
%	 similar as 
	
	Finally, the authors consider the KLD problem, which calls for producing a good estimate $\widehat{\prob}$ of the distribution $\prob$, that is, an estimate with small KL divergence: 
	$$
	\KL \left( \prob, \widehat{\prob} \right) = \sum_{\br \in \mathbb{S}_K} \prob (\br ) \log \frac{\prob (\br)}{ \widehat{\prob} (\br) } < \epsilon.
	$$
	The KLD problem turns out to be very hard for the case of Mallows, and even for small $K$, the sample complexity required for a good approximation of the underlying Mallows model is extremely high with respect to $\epsilon$. In \citep{BuHuSz14}, the existence of a polynomial algorithm for this problem (under the assumption of the Mallows model) was left as an open question.
	
	\subsubsection{Plackett-Luce} \label{subsec_pl_model}
	\citet{SzBuPaHu15} assume that the underlying probability distribution is a Plackett-Luce (PL) model \citep{Pl75,Lu59}. The PL model is parametrized by a vector $\theta = (\theta_1, \theta_2, \ldots , \theta_K) \in \mathbb{R}_+^K$, where each $\theta_i$ can be interpreted as the weight or ``strength'' of the arm $a_i$. The probability assigned by the PL model to a ranking represented by a permutation $\br \in \mathbb{S}_K$ is given by
	\begin{equation}\label{eq:pl}
	\mathbb{P}_\theta(\br) =  \prod_{i=1}^K \frac{\theta_{\br^{-1}(i)}}{\theta_{\br^{-1}(i)} + \theta_{\br^{-1}(i+1)} + \ldots + \theta_{\br^{-1}(K)}} \enspace ,
	\end{equation}
	where $\br^{-1}(i)$ is the index of the item on position $i$.
	The product on the right-hand side of (\ref{eq:pl}) is the probability of producing the ranking $\br$ in a \emph{stagewise} process: First, the item on the first position is selected, then the item on the second position, and so forth. In each step, the probability of an item to be chosen next is proportional to its weight. Consequently, items with a higher weight tend to occupy higher positions. In particular, the most probable ranking (i.e., the mode of the PL distribution) is simply obtained by sorting the items in decreasing order of their weight:
	\begin{equation*}
	%\label{eq:mpr}
	\reforder = \operatorname*{argmax}_{\br \in \mathbb{S}_K} \mathbb{P}_\theta (\br ) = \operatorname*{argsort}_{k \in [K]} \{ \theta_1, \ldots , \theta_K \} \enspace.
	\end{equation*} 
	It is worth noting that the pairwise probabilities (\ref{eq:pairwisexy}) for the PL model coincide with (\ref{eq_bradley_terry}) by setting the parameters $u_i$ of the Bradley-Terry model to $\log(\theta_i)$ for all $i \in \{1,\ldots ,K\}.$
	
	%Obviously, the PL model is invariant toward multiplication of $\theta$ with a constant $c >0$, i.e., $\pl_\theta(\pi) = \pl_{c \theta}(\pi)$ for all $\pi \in \S_K$ and $c > 0$. Consequently, $\theta$ can be normalized without loss of generality (and the number of degrees of freedom is only $K-1$ instead of $K$). Note that the most probable ranking, i.e., the mode of the PL distribution, is simply obtained by sorting the items in decreasing order of their weight:
	%\begin{equation}\label{eq:mpr}
	%\reforder = \argmax_{\pi \in \S_K} \pl_\theta (\pi ) = \argsort_{k \in [K]} \{ \theta_1, \ldots , \theta_K \} \, .
	%\end{equation}
	
	%The authors consider two different goals of the learner, which are both meant to be achieved with high probability. In the first problem, called PACI (for PAC-item), the goal is to find an item that is almost as good as the Condorcet winner, i.e., an item $ j $ such that $ |\Delta_{i^{\star},j}| < \epsilon $, where $ i^{\star} $ is the Condorcet winner, for which they devise the \Algo{PLPAC} algorithm with a sample complexity of $ \bigO(K \log K) $. 
	The authors consider two different targets of the learner, which are both meant to be achieved within an $(\epsilon,\delta)$-PAC learning setting. 
	In the first problem, the goal is to find an $\epsilon$-optimal arm, for which they devise the \Algo{PLPAC} algorithm with a sample complexity of $ \bigO\left( \frac{K}{\epsilon^2} \log\frac{ K}{\delta}\right) $. 
	%The second goal, called AMPR (approximate most probable ranking), is to find the approximately most probable ranking $ \br $, i.e., there is no pair of items $ 1\leq i,j \leq K, $ such that $ r_i^{\star} < r_j^{\star}, \, r_i > r_j $ and $ \Delta_{i,j} > \epsilon $, where $ \br^{\star} = \operatorname*{argmax}_{\br \in \mathbb{S}_K} \mathbb{P} (\br) $, for which they propose the \Algo{PLPAC-AMPR} algorithm, whose sample complexity is of order $ \bigO(K \log^2 K) $
	The second goal is to find an $\epsilon$-optimal ranking,  for which they propose the \Algo{PLPAC-AMPR} algorithm, whose sample complexity is of order $ \bigO\left( \frac{K}{\epsilon^2} \log^2 \frac{ K}{\delta}\right) $.
	
	Both algorithms are based on a budgeted version of the \Algo{QuickSort} algorithm \citep{Ho62}, which reduces its quadratic worst-case complexity to the order $ \bigO(K \log K) $, and in which the pairwise stability property is provably preserved (the pairwise marginals obtained from the distribution defined by the \Algo{QuickSort} algorithm coincide with the marginals of the PL distribution).
	
	%#################################
	\section{Learning from Non-coherent Pairwise Comparisons}\label{sec:noass}
	
	The methods presented in the previous section essentially proceed from a given target, for example a ranking $\succ$ of all arms, which is considered as a ``ground truth''. The preference feedback in the form of (stochastic) pairwise comparisons provide information about this target and, consequently, should obey certain consistency or regularity properties. This is perhaps most explicitly expressed in Section \ref{sec:mall}, in which the $q_{i,j}$ are derived as marginals of a probability distribution on the set of all rankings, which can be seen as modeling a noisy observation of the ground truth given in the form of the center ranking. 
	
	Another way to look at the problem is to start from the pairwise preferences $\bQ$ themselves, that is to say, to consider the pairwise probabilities $q_{i,j}$ as the ground truth. In tournaments in sports, for example, the $q_{i,j}$ may express the probabilities of one team $a_i$ beating another one $a_j$. In this case, there is no underlying ground truth ranking from which these probabilities are derived. Instead, it is just the other way around: A ranking is derived from the pairwise comparisons. Moreover, there is no reason for why the $q_{i,j}$ should be coherent in the sense that a natural total ordering of the arms or a Condorcet winner exists. In particular, preferential cycles and violations of transitivity are commonly observed in practice, so that alternative concepts for the definition of a target, such as an optimal arm or a ranking, are needed.   
	In the following sections, we provide an overview of approaches that adopt this point of view, thereby providing the missing supplement to the taxonomy of PB-MAB algorithms (cf.\ Figure \ref{fig:stpbmab})
	% a specific sense. In particular, preferential cycles and violations of transitivity are commonly observed in many applications.  

	\subsection{Alternative Target Concepts} \label{subsec_alternative_targets}
	
	%This is exactly the challenge faced by \emph{ranking procedures}, which have been studied quite intensely in operations research and decision theory \citep{Mo88,ChEnLaMa07}. A ranking procedure $\cR$ turns $\bQ$ into a complete preorder relation $\succ^{\cR}$ of the alternatives under consideration. Thus, another way to pose the preference-based MAB problem is to instantiate $\succ$ with $\succ^{\cR}$ as the target for prediction---the connection between $\bQ$ and $\succ$ is then established by the ranking procedure $\cR$, which of course needs to be given as part of the problem specification.  
	The challenge caused by non-coherence such as cycles or violations of transitivity is faced by \emph{ranking procedures}, which have been studied quite intensely in operations research and decision theory \citep{Mo88,ChEnLaMa07}. A ranking procedure $\cR$ turns $\bQ$ into a complete preorder relation $\succ^{\cR}$ of the arms under consideration. Thus, another way to pose the preference-based MAB problem is to instantiate $\succ$ with $\succ^{\cR}$ as the target for prediction---the connection between $\bQ$ and $\succ$ is then established by the ranking procedure $\cR$, which of course needs to be given as part of the problem specification.  
	
	Formally, a ranking procedure $\cR$ is a map $[0,1]^{K \times K} \rightarrow \mathcal{C}_{K}$, where $\mathcal{C}_{K}$ denotes the set of complete preorders on the set of arms. We denote the complete preorder produced by the ranking procedure $\cR$ on the basis of $\bQ$ by $\succ^{\cR}_{\bQ}$, or simply by $\succ^{\cR}$ if $\bQ$ is clear from the context. Below we present some of the most common instantiations of the ranking procedure $\cR$.%~\cite{BuSzWeChHu13}
	\begin{enumerate}
		\item[--] Copeland's ranking ($\BIN$) is defined as follows \citep{Mo88}: $a_i
		\succ^{\BIN} a_j$ if and only if $d_i > d_j$, where $d_i \defeq \# \{k \in [K] \, \vert \, 1/2 < q_{i,k} \}$ is the Copeland score of $ a_i $. The interpretation of this relation is very simple: An arm $a_i$ is preferred to $a_j$ whenever $a_i$ ``beats'' more arms than $a_j$ does.
		
		\item[--] The sum of expectations ($\SE$) (or Borda)  ranking is a ``soft'' version of $\BIN$: $a_i \succ^{\SE} a_j$ if and only if
		\begin{equation}\label{eq:sscore}
		q_i = \frac{1}{K-1}\sum_{k \neq i} q_{i,k} > \frac{1}{K-1}\sum_{k \neq j} q_{j,k} = q_j \enspace .
		\end{equation}
		\item[--] The idea of the random walk (\Algo{RW}) ranking is to handle the
		matrix $\bQ$ as a transition matrix of a Markov chain and order the arms based on its stationary distribution. More precisely, \Algo{RW} first transforms $\bQ$ into the stochastic matrix $\bS = \left[s_{i,j}\right]_{K\times K }$,
		where $s_{i,j}= q_{i,j}/\sum_{\ell=1}^K q_{\ell,i}$. Then, it determines 
		the stationary distribution $(v_1, \dots, v_K)$ for this matrix (i.e., the normalized eigenvector corresponding
		to the largest eigenvalue 1). Finally, the arms are sorted according to these probabilities: $a_i \succ^{\RW} a_j$ iff $v_i > v_j$. The $\RW$ ranking is directly motivated by the PageRank algorithm \citep{BrPa98}, which has been well studied in social choice theory \citep{CoShSi99,BrFi07} and rank aggregation \citep{NeOhSh12}, and which is widely used in many application fields \citep{BrPa98,KoBuPo08}.
	\end{enumerate}
	
%	\paragraph{Relationships between the total order over arms.}
	%
	Unlike the total order over arms, which may not exist for a specific preference relation $\bQ,$ all the just mentioned notions of a target ranking are defined for any preference relation $\bQ.$
	However, in case the total ordering over the arms exists, neither the random walk ranking nor the Borda ranking will necessarily coincide with the former, whereas the Copeland ranking provably does.
	Indeed, consider the preference relation 
	\begin{align} \label{ex_Borda_RW}
	\bQ = \left(  \begin{array}{ccc}
	0.5&0.55&0.55 \\ 0.45&0.5&1 \\ 0.45&0&0.5
	\end{array} \right) .
	\end{align}
	Then, there is a total ordering given by $a_1\succ a_2 \succ a_3,$ but the Borda ranking as well as the random walk ranking are $a_2 \succ^{\SE} a_1 \succ^{\SE} a_3,$ as $q_1=0.55,$ $q_2=0.725$, and $q_3=0.275.$ Furthermore, the random walk ranking coincides with the Borda ranking, that is $a_2 \succ^{\RW} a_1 \succ^{\RW} a_3,$ since the stationary distribution of the induced stochastic matrix of $\bQ$ is $(v_1,v_2,v_3) \approx (0.4107, 0.4147, 0.1746).$

	\newcommand{\vonNeumann}{\mathcal P}

	\citet{groves2019top} provide a sufficient condition for the coherence of the Borda ranking and the total order over arms in the case $\bQ$ fulfills strong stochastic transitivity.
	In particular, if for any two distinct arms $a_i,a_j \in \cA,$ there exists some arm $a_k$ such that $\Delta_{i,k} \neq \Delta_{j,k},$ then the Borda ranking is the same as the total order.
	Moreover, if the low noise assumption holds, then the Borda ranking as well as the random walk ranking coincide with the total ordering.
	
%	\paragraph{Alternatives to the Condorcet winner.}
	%
	The alternative notions for a target ranking also allow one to define alternatives for the Condorcet winner, which, just like the total order of the arms, may not exist for some preference relations $\bQ.$
	In particular, the following definitions of optimal arms are considered in the literature.
	\begin{enumerate}
		\item[--] \emph{Copeland winner}:
		The Copeland set $\CP(\bQ)$ is defined as the set of arms in $ [K] $ with highest Copeland score, i.e., $ \CP(\bQ) = \{  i \in [K]  \, | \, i \in \argmax_{j \in [K]} d_j \}.$  
		In other words, the Copeland set consists of arms having the top position in the preorder defined by $\succ^{\BIN}.$
		Each arm in the Copeland set has the property that it beats the maximal number of other arms, and hence is called a Copeland winner. 
		In contrast to the Condorcet winner, a Copeland winner can be beaten by another arm. 
		The cardinality of such arms, i.e., the number of arms by which a Copeland winner is beaten, will be denoted by $L_{C}.$ 
		
		\item[--] \emph{Borda winner/random walk winner}:
		Similarly as for the Copeland winner, it is possible to define the Borda winner resp.\ the random walk winner as the arms having the top position according to the preorder on the alternatives induced by $\succ^{\SE}$ resp.\ $\succ^{\RW}$.
		Just like in the case of the Copeland winner, these alternative notions of optimality of an arm do not exclude the possibility of being beaten by another arm.
		\item[--]  \emph{von Neumann winner}: Inspired by the game-theoretical point of view on the dueling bandits problem (cf.\ Section \ref{subsec_mab_related_algorithms}), one can define a zero-sum game matrix $P=2\bQ-1$. By setting the outcome of the game to $ +1 $ if $a_i$ wins and $ -1 $ if $a_j$ wins, so that the entry of $P$ at position $ (i,j) $  is the expected outcome of a duel between the arms $a_i$ and $a_j$.
		This matrix $P$ is skew-symmetric and specifies a zero-sum game, so that von Neumann's minmax theorem \citep{neumann1928theorie,owen1982game} implies the existence of a probability vector $\vonNeumann \in \Delta^{(K-1)}$ which is a maxmin strategy for the game described by $ P $\footnote{$\Delta^{(K-1)}$ is the $K-1$ probability simplex.}.
		This probability vector $\vonNeumann$ is  called the von Neumann winner. In contrast to the alternative notions of optimality, it thus represents a random strategy instead of a pure strategy (a single arm or a set of arms) as an optimality concept.
		The von Neumann winner specifically satisfies $\sum_{i=1}^K \vonNeumann(i) \, q_{i,j} \geq \sum_{i=1}^K \vonNeumann(i) \, q_{j,i}$ for any $j \in [K],$ so that it guarantees a higher probability of winning by choosing an arm according to $\vonNeumann$ than loosing against any arm $a_j.$
	\end{enumerate}

%	\paragraph{Relationships between the optimality concepts of arms.}
	% 
	The relationship between the alternative optimality concepts for arms and the Condorcet winner resembles their ranking counterparts. If the Condorcet winner exists, the Copeland set reduces to a singleton set with the Condorcet winner as its only element. The Borda resp.\ random walk winner, on the other hand, may differ from the Condorcet winner (consider the preference relation in (\ref{ex_Borda_RW})).
	However, if the low noise model assumption is satisfied, the Borda and random walk winner are both unique and equal the Condorcet winner.
	Regarding the von Neumann winner $\vonNeumann$, it is well known that it coincides with the Condorcet winner $a^{*}$ if the latter exists, in the sense that $\vonNeumann$ is the Dirac measure on the singleton set $\{a^{*}\}.$
	In cases where a Condorcet winner does not exist, the von Neumann winner will generally differ from the Copeland, Borda, and random walk winner (cf.\ Appendix B in \cite{DuHoShSlZo15}).
	% 
%	}

%	\editvik[
	\subsection{Algorithms for Non-coherent Preference Relations}
	In this section, we survey existing methods for PB-MAB problems that do not proceed from specific structural assumptions on $\bQ$. Instead, the targets of the learner could be derived from a possibly non-coherent preference relation.
	In Tables \ref{tab:inconsist_regret}---\ref{tab:inconsist_exact}, we summarize the approaches according to their algorithm class, the  targets and goals considered, as well as their theoretical guarantees.
	%, prior to elaborating on the methods.
	%for the dueling bandits problem with non-coherent pairwise comparisons 

	%
	\begin{table}	
		\scriptsize
		\begin{center}
			\begin{tabular}{|p{3cm}|p{2.5cm}|p{3cm}|p{4.75cm}|}
				\hline
				\textbf{Algorithm} & \textbf{Algorithm class}  & \textbf{Target(s) and goal(s) of learner} & \textbf{Theoretical guarantee(s)} \\
				\hline
				\hline 
				1.\ Copeland confidence bound  \newline \newline
				2.\ Scalable Copeland bandits \newline (Section \ref{subsec_ccb})  
				& 1.\ Generalization-based (UCB), tournament \newline
				2.\ Generalization-based (Explore-then-exploit), Reduction-based
				& High probability and expected regret minimization for finding the Copeland winner 
				& 1. $\mathcal{O}\left(   \frac{K^2 + K( |\CP(\bQ)| + L_{C}  ) \log T}{\min_{i\neq j} \Delta_{i,j}^2 }\right)$ \newline
				2.  $ \mathcal{O}\left( \frac{K (\log(K) +L_{C})\log(T)}{\min_{i\neq j} \Delta_{i,j}^2 } \right) $
				\\
				\hline
				Copeland winners relative minimum empirical divergence \newline (Section \ref{subsec_copeland_rmed})
				& Generalization-based (Explore-then-exploit, DMED)
				& Expected regret minimization for finding the Copeland winner 
				& {\tiny$\mathcal{O}\left( \min_{k \in \CP(\bQ) }C_k^*(\bQ) \log(T  \right)$\ \newline $ + \, o(\log(T))$ \newline resp.\
				$\mathcal{O}\left(   \frac{  K \, L_C \,\log T}{\min_{i\neq j} \Delta_{i,j}^2 }\right)$  } 
				\\
				\hline
				Double Thompson sampling \newline (Section \ref{subsec_dts})
				& Generalization (Thompson sampling and upper/lower confidence bounds)
				& Expected regret minimization for finding the Copeland winner 
				& {\tiny$\mathcal{O}\left(   \frac{K^2 \log \log T + K( |\CP(\bQ)| + L_{C}  ) \log T}{\min_{i\neq j} \Delta_{i,j}^2  }\right)$ \newline $ + \,  \bigO(  K^3 )$ } 
				\\
				\hline
				Sparse sparring \newline (Section \ref{subsec_sparse_sparring})
				& Reduction-based, generalization (successive elimination)
				& High probability regret minimization for the von Neumann winner 
				& \hphantom{ $(\epsilon,\delta)$ PAC learning for finding}  \newline See \eqref{eq_regert_spar2}
%				$ \tilde{\mathcal{O}}( \sqrt{sT \log(s/\delta)} + C(P) \log(1/\delta)^2 ) $ for a constant $C(P)$ independent of $T$ and $s$ being the support of the von Neumann winner
				\\
				\hline
				General tournament solutions \newline (Section \ref{subsec_gen_tournaments})
				& Generalization (UCB)
				%			, tournament
				& High probability regret minimization for finding the Copeland set, the top cycle, uncovered set, and Banks set 
				&  $\mathcal{O}\left(   K^2 +  \frac{\log T}{\Delta^{(TS)}} \right),$ where $\Delta^{(TS)}$ is a target specific complexity term 
				\\
				\hline
			\end{tabular}
			
		\end{center}
		%	}
		\caption{Approaches for regret minimization tasks in the dueling bandits problem with non-coherent pairwise comparisons.}
		\label{tab:inconsist_regret}
	\end{table} 

	\begin{table}	

		\scriptsize
		\begin{center}
			\begin{tabular}{|p{3cm}|p{2.5cm}|p{3cm}|p{4.75cm}|}
%				{|p{3cm}|p{2.5cm}|p{5.25cm}|p{4.75cm}|}
				\hline
				\textbf{Algorithm} & \textbf{Algorithm class}  & \textbf{Target(s) and goal(s) of learner} & \textbf{Theoretical guarantee(s)} \\
				\hline
				\hline 
				PAC rank elicitation \newline (Section \ref{subsec_pac_rank_elic})
%				\citep{BuSzHu14} 
				& Generalization-based (racing/successive elimination)
				& $(\epsilon,\delta)$-PAC learning of Copeland and Borda rankings with the NDP and MRD measures for near optimality
				&  $ \mathcal{O} \left( R(\epsilon,\bQ)  \log \frac{R(\epsilon,\bQ) }{\delta}    \right)$ with a function $R$ depending on $\epsilon$ and $\bQ.$
				\\
				\hline
				Borda-ranking \newline (Section \ref{subsec_borda_ranking})
				&	Generalization (explore-then-exploit), reduction-based
				&  $(\epsilon,\delta)$-PAC learning for finding \newline
				1. Borda winner \newline
				2. Borda ranking 
				&   \hphantom{ $(\epsilon,\delta)$ PAC learning for finding}  \newline 
				1.	$ \bigO \left( \frac{K}{\epsilon^2} \log \frac{1}{\delta} \right)$  \newline
				2. 	$ \bigO \left( \frac{K}{\epsilon^2} \log \frac{K}{\delta} \right)$
				\\
				\hline
				Round-Efficient Dueling Bandits \newline (Section \ref{subsec_round_efficient_dueling_borda})
				%				\citep{MaGr17} 
				& Tournament
				&  $(\epsilon,\delta)$-PAC sample and round complexity minimization for Borda winner 
				& 
				Round complexity: $ \bigO \left( \frac{1}{\epsilon^2} \log \frac{K}{\epsilon \delta} \right)$ \newline
				Sample complexity: $ \bigO \left( \frac{K}{\epsilon^2} \log \frac{K}{\epsilon \delta} \right)$ 
				\\
				\hline
				Hamming-LUCB \newline (Section \ref{subsec_hamming_lucb})
				&  Generalization (lower-upper confidence bound)
				& $(\epsilon,\delta)$-PAC learning of the top-$k$ Borda arms (and their complementary set) w.r.t.\ the Hamming distance
				&  \hphantom{ $(\epsilon,\delta)$ PAC learning for finding}  \newline  See \eqref{eq_sample_hamming_lucb}
%				$
%				\tilde{\bigO} \big( \sum_{i=1}^{k-h} \Delta_{i,k+1+h}^{-2} $ \newline
%				$ + \sum_{i=k+1+h}^{K} \Delta_{k-h,i}^{-2} \!\!\! +  \Delta_{k-h,k+1+h}^{-2} \big) $
				\\
				\hline
			\end{tabular}
			
		\end{center}			\caption{Approaches for $(\epsilon,\delta)$-PAC settings of the dueling bandits problem with non-coherent pairwise comparisons.}
	\label{tab:inconsist_PAC}
%	}
	\end{table}

\begin{table}	

	\scriptsize
	\begin{center}
		\begin{tabular}{|p{3cm}|p{2.5cm}|p{3cm}|p{4.75cm}|}
			%				{|p{3cm}|p{2.5cm}|p{5.25cm}|p{4.75cm}|}
			\hline
			\textbf{Algorithm} & \textbf{Algorithm class}  & \textbf{Target(s) and goal(s) of learner} & \textbf{Theoretical guarantee(s)} \\
			\hline
			\hline 
			Preference-based racing \newline (Section \ref{subsec_pref_race})
			%				\citep{BuSzWeChHu13} 
			& Generalization-based (racing/successive elimination)
			& Exact sample complexity minimization for top-$k$ ranking w.r.t. Copeland, Borda, and Random walk ranking 
			&  $\hphantom{text}$ \newline $ \mathcal{O} \left( \sum_{1\leq i < j \leq K}  \frac{\log \frac{K}{\delta} }{ (\Delta_{i,j})^2 }  \right)$
			\\
			\hline	
			Voting bandits \newline (Section \ref{subsec_voting_bandits})
			%				\citep{UrClFeNa13} 
			& Generic zooming algorithm 
			& Exact sample complexity minimization for finding Copeland and Borda winners 
			&  $\hphantom{text}$ \newline $\mathcal{O}\left( \sum_{1\leq i < j \leq K} \frac{  \log \frac{ K}{\delta \Delta_{i,j}} }{\Delta_{i,j} ^2}\right)$
			\\
			\hline 
			Successive elimination \newline (Section \ref{subsubsec_successive_elim})
			%				\citep{JaKaDeNo15} 
			& Reduction-based 
			&  Exact sample complexity for Borda winner  
			& See \eqref{eq_sample_complex_SECS}
			\\
			\hline
			Pairwise Optimal Computing Budget Allocation \&
			Pairwise Knowledge Gradient		\newline (Section \ref{subsec_bayes_seq_sample})
			& Generalization (Bayesian Expected Improvement Sampling)
			& Asymptotic optimality for top-$k$ arms with respect to the Borda ranking 
			& Consistency of the algorithm, but no bound on the sample complexity
			\\
			\hline
			Active ranking \newline (Section \ref{subsec_active_rank})
			& Generalization (racing/successive elimination)
			& Exact sample complexity minimization for Borda coarse ranking 
			& \hphantom{ $(\epsilon,\delta)$ PAC learning for finding}  \newline See \eqref{eq_ar_sample_complex}
%			$\bigO \left( \sum_{i=1}^{K}\Delta_i^{-2}\,\log \left(\frac{K}{\delta\Delta_i} \right) \right)$
			\\
			\hline
		\end{tabular}
		\caption{Approaches for exact sample complexity tasks of the dueling bandits problem with non-coherent pairwise comparisons.}
		\label{tab:inconsist_exact}
	\end{center}
	%	}
\end{table}

	\subsubsection{Copeland Confidence Bound} \label{subsec_ccb}
	
	\citet{ZoKaWhDe15} introduce what is currently known in the literature as the Copeland dueling bandit problem, namely the task of regret minimization with the cumulative regret as in \eqref{eq:regret} and the instantaneous regret suffered by the learner when comparing $a_{i(t)}$ and $a_{j(t)}$ at time $t$ set to $ 2\text{cp}(a_C)-\text{cp}(a_{i(t)})-\text{cp}(a_{j(t)})$, where $ a_C \in \CP(\bQ)$ is a Copeland winner of the underlying preference relation $\bQ$ and $\text{cp}(a_i)=\frac{d_i}{K-1}$ is the normalized Copeland score of an arm $a_i\in \cA.$ 
	It is worth noting that in case $\bQ$ has a Condorcet winner, all theoretical guarantees in the form of regret bounds derived for algorithms on a Copeland dueling bandit problem can be transferred to regret bounds on the cumulative regret in \eqref{eq:regret} by a multiplicative constant.
	In particular, algorithms for Copeland dueling bandits can be used for the regret minimization task considered in Section \ref{subsec:regaxiom}  as well, since the former is a more general problem scenario than the latter.
%	
%	However, as the 
	
	For the Copeland dueling bandit problem, 	\citet{ZoKaWhDe15} suggest two algorithms: First, the Copeland confidence bound (\Algo{CCB}) algorithm, which is suited for learning problems with a small number of arms and borrows some ideas underlying the \Algo{RUCB} algorithm (cf.\ Section \ref{subsec_RUCB}).
	Second, the scalable Copeland bandits (\Algo{SCB}) algorithm, which can deal more efficiently with learning problems having a large number of available arms and makes use of the idea underlying the \Algo{KL-UCB} algorithm for value-based MAB problems \citep{garivier2011kl,cappe2013kullback} in order to find a Copeland winner arm.
	
	\Algo{CCB} is making use of the principle underlying UCB, i.e., the optimism in the face of uncertainty principle, and additionally its pessimistic counterpart by considering upper confidence and lower confidence estimates of the pairwise preference probabilities simultaneously.
	More specifically, in each time step the optimistic/pessimistic Copeland scores are computed based on the upper/lower confidence estimates. These scores are then used to (i) determine a set of optimistic Copeland winners from which, except for a probabilistic exploration, the first arm of the pair is sampled (almost) uniformly at random\footnote{The ``almost'' is due to a slight bias given to arms in the optimistic Copeland winners set, which have shown to be good candidates over the time.}, (ii) eliminate all likely non-Copeland winner arms, and (iii) create a set consisting of tough competitors for each arm.
	The key idea underlying the choice of the second arm is then to use an arm which has the highest potential to refute that the first chosen arm is a Copeland winner.
	More specifically, among all arms having a lower confidence estimate for the pairwise preference probability against the first arm at most 1/2, the one with the lower confidence estimate closest to $1/2$ is chosen (possibly breaking ties).
	Thus, if the lower confidence estimate for the pairwise preference probability against the first arm exceeds $1/2$, then the upper confidence estimate for the pairwise preference probability of the first arm being preferred over the second arm falls below $1/2$, which in turn lowers the first arm's optimistic Copeland score.
	In order to introduce additional ``exploration'' in this regard, the second arm with the above property is picked with probability $1/2$, respectively, from the set of all tough competitors of the first arm and among all arms.
	Assuming that there are no ties, i.e., $q_{i,j}\neq 1/2$ for all pairs $(a_i,a_j)$, $i \neq j$, the authors provide for \Algo{CCB} both high probability and expected (Copeland) regret bounds of the form $\mathcal{O}\left(   \frac{K^2 + K( |\CP(\bQ)| + L_{C}  ) \log T}{\min_{i\neq j} \Delta_{i,j}^2 }\right)$. 
%	\\
	In fact, the shown regret bound is stricter, because the gap term $\min_{i\neq j} \Delta_{i,j}^2$ can be replaced by a slightly larger gap term.
	
	\Algo{SCB} is an explore-then-exploit algorithm, which is made horizonless by using the doubling or squaring trick \citep{CeLu06}.
	During its exploration phase within one of the doubling periods, the algorithm uses a subroutine, which is a variation of the \Algo{KL-UCB} algorithm for a best arm identification problem of a value-based MAB setting with zero--one rewards.
	In order to specify a reasonable reward mechanism for each arm, the algorithm simulates, each time a value-based reward of the pulled arm $a_i$ is expected, the following process: First, another arm $a_j \in \cA \backslash \{a_i\}$ is chosen uniformly at random, then $a_i$ and $a_j$ are dueled with each other a specific period-dependent number of times, resulting in a reward of $1$ in case $a_i$ has won the majority of the duels and a reward of $0$ otherwise.
	It can be shown that the best arm(s) based on these induced reward distributions correspond(s) to the Copeland winner(s).
	After the \Algo{KL-UCB} variant has identified the best arm using this reward simulation process, \Algo{SCB} starts the exploitation phase of the doubling period by dueling the prior arm against itself (cf.\ ``full commitment" in Section \ref{subsec_regret_bounds}).
	If the available time budget of the corresponding doubling period is not sufficient to let the exploration phase come to an end, no exploitation phase is conducted in this period and instead the exploration phase of the next period is started.
	Once again assuming that the preference relation has no ties, \Algo{SCB} achieves an expected (Copeland) regret bound of $ \mathcal{O}\left( \frac{K (\log(K) +L_{C})\log(T)}{\min_{i\neq j} \Delta_{i,j}^2 } \right),$ which has a more favorable dependency on the number of arms $K$ than the regret bound  of \Algo{CCB} and does not depend on the number of Copeland winners $\CP(\bQ).$

	\subsubsection{Copeland Winners Relative Minimum Empirical Divergence} \label{subsec_copeland_rmed}

	\citet{koHoNa16} consider the Copeland dueling bandit problem as in \cite{ZoKaWhDe15}, but modify the instantaneous regret by a multiplicative constant factor.
	For this problem scenario, they propose the Copeland Winners Relative Minimum Empirical Divergence (\Algo{CW-RMED}) algorithm, which, just like the relative minimum empirical divergence algorithm (cf.\ Section \ref{subsubsec:rmed}), is inspired by the DMED algorithm \citep{HoTa10} for the value-based MAB problem. 
	Thus, the design idea underlying the algorithm is guided by the lower bound for the respective bandit problem.
	To this end, the authors  derive an asymptotic regret lower bound for any no-regret algorithm under the assumption that there are no ties in the preference relation $\bQ$, which is of the form $\Omega\left( \min_{k \in \CP(\bQ) }C_k^*(\bQ) \log(T) \right),$ where  $C_k^*(\bQ)$ is a complexity term coming from $\bQ$ and depending on the Kullback-Leibler divergence of Bernoulli random variables with success probability equal to the pairwise probabilities and $1/2$ in the same spirit as in Section \ref{subsubsec:rmed}.
	This complexity term $C_k^*(\bQ)$ has a natural interpretation coming from the proof of the lower bound, namely $C_k^*(\bQ) \log(T)$ represents the smallest possible cumulative regret to ensure that $k \in \CP(\bQ),$ i.e., arm $a_k$ is a Copeland winner.

	As a consequence, \Algo{CW-RMED} revolves around the natural estimate of the complexity term by using the current estimates of the entries of the preference relation. 
	Formally, $C_k^*(\hat \bQ_t)$ with  $\hat \bQ_t = [\widehat{q}^{\, t}_{i,j}]_{1\le i,j \le K}$ and  $i^*_t \in \argmin_{i\in[K]} C_i^*(\hat \bQ_t)$ are used, where the latter set specifies the candidates for the likeliest Copeland winners.
	After making sure that every pair of arms has been dueled a sufficient time-dependent number of times and all pairwise preference estimates are far enough away from 1/2, the \Algo{CW-RMED} checks whether the current these estimates provide enough evidence in favor of some $i^*_t$ being indeed a Copeland winner.
	If this is the case, the corresponding arm is dueled against itself (cf.\ ``full commitment" in Section \ref{subsec_regret_bounds}) in the upcoming time steps until the confidence of the belief is doubtful, in which case the pair of arms with the highest chance to regain confidence is dueled.

	Because the computation necessary to determine the latter pair is in fact cumbersome, the authors propose the Efficient \Algo{CW-RMED} (\Algo{ECW-RMED}) algorithm, which gets rid of the latter computational problem by slightly modifying the target complexity term $C_k^*$ to a reasonable surrogate $\tilde C_k^*.$
	As a consequence, the exploratory behavior of  \Algo{ECW-RMED} is different from \Algo{CW-RMED} and leads to a (Copeland) regret bound of the former of $\mathcal{O}\left( \min_{k \in \CP(\bQ) } \tilde C_k^*(\bQ) \log(T)  \right) + o(\log(T)),$ while the latter has provably an expected regret bound of $\mathcal{O}\left( \min_{k \in \CP(\bQ) }C_k^*(\bQ) \log(T)  \right) + o(\log(T)).$
	However, both complexity terms are the same if there is more than one Copeland winner, i.e., $|\CP(\bQ)|\geq 2.$

	Finally, the authors relate the resulting regret bound of \Algo{ECW-RMED} to \Algo{CCB}, showing improvements regarding the multiplicative constants. 
	More specifically, they show that, for the sake of comparison, the latter bounds can be further bounded by $\mathcal{O}\left(   \frac{  K \, L_C \,\log T}{\min_{i\neq j} \Delta_{i,j}^2 }\right)$, revealing that the dependency on $ \CP(\bQ)$ as a multiplicative factor in the (Copeland) regret can indeed be eliminated as it is the case for \Algo{SCB}.

	\subsubsection{Double Thompson Sampling} \label{subsec_dts}
	For the Copeland dueling bandit problem, 	\citet{WuLi16} suggest the Double Thompson Sampling (\Algo{D-TS}) algorithm as well as an enhanced version of it (\Algo{D-TS$ ^{+} $}). 
	Both algorithms rely on the well-known Thompson sampling \citep{Th33,AgGo13,KaKoMu12,ChLi11} algorithm for value-based MAB problems in the sense that Beta posterior distributions over the entries of the preference relation are maintained as in the \Algo{RCS} algorithm (cf.\ Section \ref{subsec_RCS}) and are consequently used to suggest arms for the duel at one iteration step.
	Additionally, both algorithms borrow some ideas underlying the design of the \Algo{CCB} algorithm  for choosing a maximum informative second arm to possibly refute that the first chosen arm is a Copeland winner.
	To be more precise, \Algo{D-TS} first samples a preference relation $\tilde \bQ \in [0,1]^{K\times K}$ in each iteration\footnote{Strictly speaking, the preference relation $\tilde \bQ$ depends on the iteration step $t,$ as the Beta posterior distributions do, but we suppress this dependency here in the notation for sake of convenience.} according to the current Beta posterior distributions and chooses for the first arm the one with the largest Copeland score on $\tilde \bQ $ among all arms currently having the largest optimistic Copeland score (based on upper confidence estimates), while possible ties are broken randomly.
	Then, for choosing the second arm, \Algo{D-TS} uses once again the current Beta posterior distributions to sample for any other arm the pairwise preference probability for being preferred over the first arm, and chooses the arm with the largest sampled pairwise preference probability among all arms having a lower confidence estimate for the pairwise preference probability against the first arm of at most $1/2$ (possibly breaking ties).
	Thus, the mechanism in choosing the second arm is similar to \Algo{CCB}, but differs due to the usage of the sampled pairwise preference probabilities instead of the upper confidence estimates.

	The enhanced version \Algo{D-TS$ ^{+} $} differs from \Algo{D-TS} in the possible tie breaking for choosing the first arm.
	More specifically, \Algo{D-TS$ ^{+} $}  additionally estimates the cumulative Copeland regret of an arm dueled against any other arm through the history of the sampled preference relations $\tilde \bQ$ and, in the case of ties, chooses among all optimistic Copeland winners the one with the lowest estimated regret.
%	

%	for Copeland dueling bandits, including Condorcet bandits as a special case.
%	to choose an optimal action that maximizes the expected reward according to the agent's current belief (randomly drawn according to the current prior). Moreover, to avoid engaging in suboptimal comparisons, it utilizes the idea of confidence bounds and eliminates arms that are unlikely to be the winner. More specifically, it maintains a posterior distribution for the preference matrix, and chooses the pair of arms for comparison according to two sets of samples, independently drawn from the posterior distributions, which are then updated according to the comparison results. % 
%	\Algo{D-TS} relies on Thompson sampling \citep{Th33,AgGo13,KaKoMu12,ChLi11} to choose an optimal action that maximizes the expected reward according to the agent's current belief (randomly drawn according to the current prior). Moreover, to avoid engaging in suboptimal comparisons, it utilizes the idea of confidence bounds and eliminates arms that are unlikely to be the winner. More specifically, it maintains a posterior distribution for the preference matrix, and chooses the pair of arms for comparison according to two sets of samples, independently drawn from the posterior distributions, which are then updated according to the comparison results. 
%	

	It is shown that \Algo{D-TS} achieves an expected cumulative (Copeland)  regret bound of $\bigO \left( \frac{K^2 \log T}{\min_{i\neq j} \Delta_{i,j}^2}  + K^2 \right),$ provided there are no ties in the preference relation.
%	
%	Using the fact that the distribution of the samples only depends on the historic comparison results and not on the time step $ t $, and referring to a back substitution argument, they further refine this bound to 
	Under an additional assumption on the pairwise probabilities of non-Copeland winner arms, the expected cumulative (Copeland)  regret bound of \Algo{D-TS}  is refined to
	$$ \bigO \left(   \frac{K^2 \log \log T + K( |\CP(\bQ)| + L_{C}  ) \log T}{\min_{i\neq j} \Delta_{i,j}^2 } + K^3 \right).
	$$ 
	The latter bounds are also shown to be valid for \Algo{D-TS$ ^{+} $}.
%	Since \Algo{D-TS} breaks ties randomly, and thus tends to explore all potential winners, its regret scales with the number of winners. The authors therefore propose an enhanced version of \Algo{D-TS}, referred to as  \Algo{D-TS$ ^{+} $}, which achieves the same regret bound, but performs better in practice, especially in the case of multiple winners. This is accomplished by a strategy for carefully breaking ties according to estimated regret.

	\subsubsection{Sparse Sparring} \label{subsec_sparse_sparring}
	
	\citet{BaKaScZo16} adopt the game-theoretic view of dueling bandits, as described in the definition of the von Neumann winner.
	The authors aim for algorithms the performance of which approach the performance of the von Neumann winner. To this end, they consider the cumulative regret up to time $ T $ as $ \max_{k\in K}\sum_{t=1}^{T} (P_{k,i(t)}+P_{k,j(t)})/2,$ where $P$ is the zero-sum game matrix of $\bQ$ (see the beginning of the section).
	This notion of regret was initially introduced by \citet{DuHoShSlZo15}, which will be discussed in another context in Section \ref{subsec_contextual_duels}.
	
	The construction idea of the proposed Sparse Sparring (\Algo{SPAR2}) algorithm is guided by the observation that the von Neumann winner is usually sparse in the sense that only $s \ll K$ many of its entries are non-zero, which can be interpreted as having only a small set of good arms, while most of the arms are far from being optimal. 

	\Algo{SPAR2}  makes use of the sparring idea suggested by \cite{AiKaJo14} (cf.\ Section \ref{subsec_adversarial}).
	In a nutshell, two algorithms for the value-based MAB problem are used to determine the pair of arms for the duel. 
	One of the algorithms suggests the first arm, the other the second arm, and after having conducted the duel, the algorithm which suggested the winning arm obtains a reward of 1, the other a reward of -1.
	The key idea of \Algo{SPAR2} is to use this sparring approach with two independent copies of the \Algo{Exp3} algorithm  \citep{AuCeFrSc02}  and simultaneously maintain confidence regions around the empirical estimate of $P$ in order to successively reduce the set of suggestible arms for the  \Algo{Exp3} algorithms until only the arms in the sparse support of the actual von Neumann winner arm remain.
	Roughly speaking, the latter reduction is realized by constantly checking the conformity of the supports of all von Neumann winners extractable from the confidence regions around the empirical estimate of $P$ and eliminating an arm as soon as it is the first time not within the support of all these von Neumann winners.
	Because this approach is computationally inefficient, the authors propose a more efficient workaround, for which results on the stability of the von Neumann support are derived to introduce more conservative yet practicable conditions for monitoring the conformity of the supports.
	
%	 
%	follows the action elimination principle of \cite{EvMaMa02}, in which actions that cannot belong to the support of the von Neumann winner are eliminated based on confidence bounds on probability estimates. For the remaining actions, the sparring idea of \cite{AiKaJo14} is applied (cf.\ Section \ref{subsec_adversarial}): Two independent copies of the \Algo{Exp3} \citep{AuCeFrSc02} algorithm are maintained, the estimate of $ P $ is successively improved, and actions are excluded appropriately.
%	
	
%	

%	Under the assumptions of uniqueness of the von Neumann winner and sparsity (there is only a small set of ``good'' actions, with the rest being strictly inferior), the authors present the algorithm.
	Under the assumption that the von Neumann winner is unique and has $s$ many non-zero probabilities, it is shown that \Algo{SPAR2} enjoys a cumulative regret bound of 
	\begin{align} \label{eq_regert_spar2}
		\min \left\{  \tilde{\mathcal{O}}( \sqrt{sT \log(s/\delta)} + C(P) \log(1/\delta)^2 )  \, , \, \bigO (  \sqrt{KT\log(K/\delta)} ) \right\}
	\end{align}
	with probability at least $1-\delta$, where $C(P)$ is some constant depending only on the zero-sum game matrix of $\bQ.$
	This improves upon the regret bound of $\tilde{\mathcal{O}}(\sqrt{KT} )$, which holds for the straightforward approach of simply using the two independent copies of the \Algo{Exp3} algorithm.

	\subsubsection{General Tournament Solutions} \label{subsec_gen_tournaments}
	\citet{RaRaAg16} consider general tournament solutions from social choice theory \citep{BrBrHaMo16} as sets of winning arms.
	To this end, they relate a preference relation $ Q $ with a tournament graph by means of $ \cT_Q = ([K],E_Q) $ with $[K]$ as the set of vertices and $ E_Q=\{(i,j):  q_{i,j} > 1/2 \}$ as the set of edges.
	Moreover, a sub-tournament $ \cT = (V,E) $ of $\cT_Q$ with $ V \subseteq [K] $ and $ E  \subseteq E_Q $  is called \emph{maximal acyclic}, if it is acyclic and no other sub-tournament of $\cT$ is acyclic. 
	With this, alternative general tournament solutions besides the Copeland set and the Condorcet winner can be defined as follows:
%	
	% which all always exist, and  which include:
	\begin{enumerate} 
		%\item[--] The Copeland set defined as the set of arms in $ [K] $ that beat the maximal number of arms, i.e., all arms $a_j$ with $ j \in \argmax_{i \in [K] } \sum_{j\neq i} \IND{a_j \succ a_i} $ , 
		\item[--] the \emph{top cycle} (or also known as the \emph{Smith-set}) defined as the smallest set $ W \subseteq [K] $ for which $  q_{i,j}>1/2 $ for all $i \in W$ and $j \notin W $ holds, 
		\item[--] the \emph{uncovered set} defined as the set of all arms that are not covered by any other arm, where an arm $ a_i $ is \emph{covered} by an arm $ a_j $ if $ q_{i,j}>1/2 $ and for any $k\in [K]\backslash\{i,j\}$ it holds that $ q_{j,k}>1/2 $ implies $ q_{i,k}>1/2  $, 
		\item[--] and the \emph{Banks set} defined as the set consisting of the maximal elements of all maximal acyclic sub-tournaments of $ \cT_Q$.
%		, where a tournament associated with a preference matrix $ P $ is $ \cT_P = ([K],E_P) $ with $ E_P=\{(i,j): a_i\succ_P a_j\} $, and a sub-tournament $ \cT = (V,E) $ with $ V \subseteq [K] $ and $ E = E|_{V \times V} $ is said to be maximal acyclic if it is acyclic and no other sub-tournament containing it is acyclic. 
	\end{enumerate}
	It is worth noting that in case a Condorcet winner exists, all these alternative tournament solutions sets coincide and are singleton sets consisting of the Condorcet winner.
	
	The authors measure the regret per time relative to the tournament solution of interest, i.e., a suitable notion of individual regret of an arm with respect to a tournament solution is defined and used to define the common pairwise regret.
	In order to minimize this notion of regret, a generic upper confidence bound (UCB) style dueling bandits algorithm is developed, \Algo{UCB-TS}, which can be instantiated for a specific tournament solution. 

	The generic algorithm maintains upper confidence bounds on the pairwise preference probabilities and updates these as well as all necessary statistics after observing the outcome of a duel.
	The composition of the duel is determined by a selection procedure especially designed for a specific tournament solution. 
	For each tournament solution, one suitable instantiation of the selection procedure is proposed, respectively.
	All selection procedures are conceptionally similar to \Algo{RUCB} (cf.\ Section \ref{subsec_RUCB}) or \Algo{CCB} (cf.\ Section \ref{subsec_ccb}): The first arm is chosen to be a candidate for the respective tournament solution based on the upper confidence bounds, while the second arm is chosen as the one having the highest potential to refute the validity that the first arm is indeed a (respective) tournament solution.
%	Each of the selection procedures adopts an ``optimism followed by pessimism approach'' to manage the exploration-exploitation tradeoff. More specifically, the first arm is selected as the potential winning arm based on the UCBs, and the second arm as the one that has the greatest chance of invalidating the first arm as a winning arm.
%	 which is suggested 
%	, which are updated in each trial, after the algorithm observes the feedback for a pair of arms it selects based on the current UCB matrix using a selection procedure designed for a specific tournament solution. Each of the selection procedures adopts an ``optimism followed by pessimism approach'' to manage the exploration-exploitation tradeoff. More specifically, the first arm is selected as the potential winning arm based on the UCBs, and the second arm as the one that has the greatest chance of invalidating the first arm as a winning arm.
	
	For each tournament solution, say TS, the specific instantiation of the generic algorithm is theoretically analyzed by showing high probability bounds on the regret of the form   $\mathcal{O}\left(   K^2 +  \frac{\log T}{\Delta^{(\text{TS})}} \right)$, where $\Delta^{(\text{TS})}$ is a complexity term representing the difficulty of a specific tournament solution comparable to the complexity term $\sum_{j \neq i^{*}} \Delta_{i^*,j}$ occurring in the regret bound of \Algo{RUCB} (cf.\ Section \ref{subsec_RUCB}), which assumes the existence of a Condorcet winner $a_{i^*}$. 
	In the worst-case, these bounds are of the order $\mathcal{O}\left(   K^2 \log T \right)$, and it remains an open question whether a lower bound of the same order can be shown.
%	, revealing a similar complexity as the alternative targets in \citep{UrClFeNa13,BuSzHu14}, see Section \ref{subsec_voting_bandits} and \ref{subsec_pac_rank_elic}.

	\subsubsection{PAC Rank Elicitation} \label{subsec_pac_rank_elic}
%	In a subsequent work by  an extended version of the top-k selection problem is considered. 
	\citet{BuSzHu14} were one of the first authors considering the goal of finding a ranking that is close to the ranking produced by some ranking procedure $\cR$ in an $(\epsilon,\delta)$-PAC learning scenario. 
	To formalize the notion of closeness of the (predicted) permutation $\tau$ with a (target) order $\succ$, two distance measures are considered.
First, the \emph{number of discordant pairs} (\Algo{NDP}), which is closely connected to Kendall's rank correlation \citep{ke55}, and can be expressed as  
\[
d_{\cK}(\tau, \succ) =   \sum_{i=1}^K \sum_{j \neq i } \IND{ \tau_j < \tau_i } \IND{ a_i \succ a_j }  \enspace,
\]
where $\tau_{i}$ denotes the rank of arm $a_i$ in the permutation $\tau$.
%where $\prec$ denotes the strict part of $\preceq$. 
The second is the \emph{maximum rank difference} (\Algo{MRD}) defined as the maximum difference between the rank of an arm $a_i$ according to $\tau$ and $\succ$, respectively.  More specifically, since $\succ$ is a partial but not necessarily total order, $\tau$ is compared to the set $\mathcal{L}^\succ$ of its linear extensions\footnote{$\tau \in \mathcal{L}^\succ$ iff $\forall\, i, j \in [K]: \, (a_{i} \succ a_{j}) \Rightarrow (\tau_{i} < \tau_{j}$)}: 
\[
d_{\cM}(\tau, \succ) = \min_{ \tau' \in \mathcal{L}^\succ } \max_{1\le i\le K}  \vert \tau_{i} -\tau'_{i} \vert  \enspace.
\]
The authors point out the fact that, regarding the induced order relation $\succ^{\BIN}$, ranking procedures such as Copeland might be strongly influenced by a minimal change of a value $q_{i,j} \approx \frac{1}{2}$. Consequently, the number of samples needed to assure (with high probability) a certain approximation quality $\epsilon$ may become arbitrarily large. A similar problem arises for $\succ^{\SE}$ as a target order if some of the individual Borda scores $q_i$ are very close or equal to each other.

As a practical (yet meaningful) solution to this problem, the authors propose to impose stronger requirements on the order of relations such as $\succ^{\BIN}$ and $\succ^{\SE}$ in order to make these more ``partial''. 
To this end, let
$d^{*}_{i} \defeq \#\left\{ k  \, \vert \, 1/2 + \gamma < q_{i,k}, i\neq k
\right\}$ denote the number of arms that are preferred over $a_i$ with a margin $\gamma > 0$, and let $s^{*}_{i} \defeq \#\left\{ k : \vert 1/2 - q_{i,k} \vert \le \gamma, \, i \neq k \right\}$.
Then, the $\gamma$-insensitive Copeland relation is defined as follows: $a_i \succ^{\BIN_\gamma} a_j$ if and only if $d^{*}_{i} + s^{*}_i >  d^{*}_{j}$. Likewise, in the case of the Borda ranking $\succ^{\SE}$, small differences of the $q_{i}$ are neglected and the $\gamma$-insensitive sum of expectations relation is defined as follows: $a_i \succ^{\SE_\gamma} a_j$ if and only if $q_{i}+\gamma > q_{j}$.

These $\gamma$-insensitive extensions are interval (and hence partial) orders, that is, they are obtained by characterizing each arm $a_i$ by the interval $[d_i^*, d_i^* + s_i^*]$ and sorting intervals according to $[a,b] \succ [a',b']$ iff $b > a'$. It is readily shown that $\succ^{\BIN_\gamma} \,\subseteq\, \succ^{\BIN_{\gamma'}} \,\subseteq\, \succ^{\BIN}$ for $\gamma > \gamma'$, with equality $\succ^{\BIN_0} \,\equiv\, \succ^{\BIN}$ if $q_{i,j} \neq 1/2$ for all $i \neq j \in [K]$ (and similarly for the Borda ranking). The parameter $\gamma$ controls the strictness of the order relations, and thereby the difficulty of the rank elicitation task. %Note that here the ranking procedure produces strict orders as opposed to the previous setup~\cite{BuSzWeChHu13}.

%	As mentioned above, the task in PAC rank elicitation is to approximate $\succ^{\cR}$ without knowing the $q_{i,j}$. Instead, relevant information can only be obtained through sampling pairwise comparisons from the underlying distribution. Thus, the arms can be compared in a pairwise manner, and a single sample essentially informs about a pairwise preference between two arms $a_i$ and $a_j$. The goal is to devise a \emph{sampling strategy} that keeps the size of the sample (the sample complexity) as small as possible while producing an estimation $\succ$ that is ``good'' in a PAC sense: $\succ$ is supposed to be sufficiently ``close'' to $\succ^{\cR}$ with high probability. Actually, the algorithms by \citet{BuSzHu14} even produce a total order as a prediction, i.e., $\succ$ is a ranking that can be represented by a permutation $\tau$ of order $K$, where $\tau_{i}$ denotes the rank of arm $a_i$ in the order. 

The authors propose four different methods for the two $\gamma$-sensitive ranking procedures, along with the two distance measures described above. Each algorithm calculates a surrogate ranking based on the empirical estimate of the preference relation whose distance can be upper-bounded again based on some statistics of the empirical estimates of preference. The sampling is carried out in a greedy manner in every case, in the sense that those arms are compared which are supposed to result in a maximum decrease of the upper bound calculated for the surrogate ranking. 

An expected sample complexity bound is derived for the $\gamma$-sensitive Cope\-land ranking procedure along with the MRD distance in a similar way like \citet{Ka11} and \citet{KaTeAuSt12}. The bound is of the form $\bigO \left( R_1 \log \left( \frac{R_1}{\delta} \right)\right)$, where $R_{1}=R_{1}(\gamma,\epsilon,\bQ)$ is a task/problem dependent constant.  More specifically, $R_{1}$ depends on the (sorted) calibrated pairwise preference probabilities $\Delta_{i,j}$, and on the robustness of the ranking procedure to small changes in the preference matrix (i.e., on how much the ranking produced by the ranking procedure might be changed in terms of the MRD distance if the preference matrix is slightly altered) as well as on the desired approximation quality $\epsilon$. 
An expected sample complexity is also derived for the $\gamma$-insensitive sum of expectations ranking procedure along with the MRD distance with a similar flavor as for the $\gamma$-sensitive Copeland ranking procedure. The analysis of the NDP distance is more difficult, since small changes in the preference matrix may strongly change the ranking in terms of the NDP distance. The sample complexity analysis for this distance has therefore been left as an open question.

	\subsubsection{Borda-Ranking} \label{subsec_borda_ranking}
	
	\citet{FaHaOrPiRa17} consider the  $(\epsilon,\delta)$-PAC learning scenario with respect to the nearly optimal Borda winner resp.\ Borda ranking.
	% Borda-score metric without any assumptions. 
	%The Borda score of an arm $ a_i $ is $ s(a_i)=\frac{1}{K}\sum_{j} q_{i,j} $, which gives its probability of winning against a randomly selected arm from the rest of arms. An arm $ a_i $ such that $ s(i) = \max_{j}s(j) $ is called Borda maximal or winner. 
	Here, an arm $ a_i $ such that $ q_i \geq \max_{j}q_j - \epsilon $ is called an $ \epsilon $-Borda optimal arm (cf.\ Section \ref{subsec_near_opt_targets}). 
	A permutation $ \pi \in \mathbb{S}_K $ such that 
	%$ 1\leq j\leq K-1, s(a_j)\geq s(a_{j+1}) $ is called a Borda ranking. A permutation $ a_1,\ldots,a_K $ such that for all 
	$q_{\pi^{-1}(k)} \geq q_{\pi^{-1}(j)} - \epsilon $ holds for $1\leq j < k \leq K,$ is called an $ \epsilon $-Borda ranking.
	
	The authors show that the problem of finding an $ \epsilon $-Borda optimal arm can be solved using $\bigO \left( \frac{K}{\epsilon^2} \log \frac{1}{\delta} \right)$ many pairwise comparisons. This is done by showing that PAC-optimal algorithms for the standard MAB setting can be used to solve the Borda score setting using the so-called \emph{Borda reduction} of the dueling bandits to the standard MAB problem.
	This reduction is based on the fact that the pull of an arm $a_i$ with an expected reward equal to the Borda score $q_i$ can be simulated by conducting a duel of arm $ a_i $ with another randomly selected arm and returning a reward of $1$ in case $a_i$ has won and $0$ otherwise (i.e., a Bernoulli MAB problem).
	
	For the problem of finding an $ \epsilon $-Borda ranking, they present the \Algo{Borda-ranking} algorithm that requires	$ \bigO \left( \frac{K}{\epsilon^2} \log \frac{K}{\delta} \right)$ comparisons. The \Algo{Borda-ranking} algorithm first approximates the Borda score of each arm with an approximation error of $ \epsilon/2 $ with confidence at least $1-\delta/K$ by using the Hoeffding's inequality and the Borda reduction. 
	Then the arms are simply sorted according to the approximated scores, which, thanks to an underlying Bonferroni correction, results in an $ \epsilon $-Borda optimal ranking with confidence at least $1-\delta$ .

	\subsubsection{Round-Efficient Dueling Bandits} \label{subsec_round_efficient_dueling_borda}
	Within the scenario as presented in Section \ref{subsec_round_efficient_dueling}, \cite{lin2018efficient} also analyze the goal of finding a Borda winner in an $(\epsilon,\delta)$-PAC learning scenario.
	To this end, they modify their original algorithm  by letting all arms be active throughout the time and replacing their utility estimates by suitable Borda score estimates.
	The rationale behind keeping all arms active is that each active arm is dueled with a randomly chosen arm (cf.\ \emph{Borda reduction} in Section \ref{subsec_borda_ranking}), which allows to derive  unbiased estimates of the Borda scores.
	Although the authors do not mention the explicit stopping criterion for this modified algorithm, the proof of its theoretical guarantees suggests that the algorithm is just run for a specific number of times $T_{\epsilon,\delta}$, which needs to be chosen appropriately, and then returns the arm with the largest estimated Borda score.
	By using $T_{\epsilon,\delta} =  \bigO \left( \frac{1}{\epsilon^2} \log \frac{K}{\epsilon \delta} \right)$, it is shown that, with probability at least $1-\delta$, an $ \epsilon $-Borda optimal arm is returned. Consequently, the round complexity is of the same order as the chosen $T_{\epsilon,\delta},$ while the pairwise sampling complexity is $K$ times this order.
%	 of the order $\bigO \left( \frac{K}{\epsilon^2} \log \frac{K}{\epsilon \delta} \right).$
%	
	Regarding the results of \citet{FaHaOrPiRa17} in the previous section, it seems that the $\log(\epsilon^{-1})$ term occurring in these bounds could in fact be eliminated\footnote{Note that \citet{lin2018efficient} assume that the learner obtains $K$ many pairwise preference observations per iteration, while \citet{FaHaOrPiRa17} assume only one. Consequently, the \emph{pairwise} sampling complexity by \cite{lin2018efficient} can be compared with the sample complexity by \citet{FaHaOrPiRa17}.  }. 

	\subsubsection{Hamming-LUCB} \label{subsec_hamming_lucb}
	
	\cite{heckel2018approximate} consider the task of partitioning  $\cA$ into the top-$k$ arms, say $\cA_{k},$ and its complement $\cA_{k}^\complement$ with respect to the Borda scores in an $(\epsilon,\delta)$-PAC setting. 
	To this end, the approximation quality of a suggested partition $(\hat{\cA_{k}}, \hat{\cA_{k}^\complement})$ is assessed by means of the Hamming distance defined as $D_{H}(A, B) \defeq \left| (A \cup B) \setminus (A \cap B) \right| $ for any subsets $A,B \subseteq \cA.$
	In light of this, the approximation quality $\epsilon$ takes only values in the non-negative integers.
	
	The authors present the \Algo{Hamming-LUCB} algorithm inspired by the Lower-Upper Confidence Bound (\Algo{LUCB}) algorithm \citep{KaTeAuSt12}, which maintains a set consisting of the current $k-\epsilon$  best arms by considering the Borda score counterpart of the lower confidence bound of the $(k-\epsilon)$th best arm and a set of the current $n-k-\epsilon$ worst arms based on the Borda score upper confidence bound of the  $(n-k-\epsilon)$th worst arm.
	The guiding principle underlying the design of \Algo{Hamming-LUCB} is to be confident as quickly as possible that the former set contains arms with larger Borda scores than the latter set by dueling an arm, which potentially belongs, based on its current Borda score confidence intervals, to both the actual $k-\epsilon$  best arms and the actual $n-k-\epsilon$ worst arms, with a randomly chosen arm (cf.\ \emph{Borda reduction} in Section \ref{subsec_borda_ranking}). 
	Because there is potentially more than one arm with the latter property, the one with the widest confidence interval is chosen (possibly breaking ties).
	As soon as the lower confidence bound of the current $(k-\epsilon)$th best arm exceeds the upper confidence bound of the current  $(n-k-\epsilon)$th worst arm, the \Algo{Hamming-LUCB} algorithm returns the $k$ currently best arms as $\hat{\cA_{k}}$ and correspondingly  $\hat{\cA_{k}^\complement}$  as $\cA \backslash \hat{\cA_{k}}$.

	Assuming that all Borda scores are distinct, it is shown that, in order to return a partition that is $(2\epsilon)$-close to the true underlying partition in terms of the Hamming-distance with probability at least $1-\delta$, \Algo{Hamming-LUCB} requires a number of comparisons of order
%	sample complexity that is close to optimal up to logarithmic factors, and takes the form 
	\begin{align} \label{eq_sample_hamming_lucb}
	\begin{split}
%					%			
		\bigO \Big( \log\Big( \frac{K}{\delta}\Big) \Big( &\epsilon \cdot f_1(\Delta_{(k-\epsilon),(k+1+\epsilon)}^{\SE})  
		+  \sum_{i=1}^{k-\epsilon} f_1(\Delta_{(i),(k+1+\epsilon)}^{\SE})+ \sum_{i=k+1+\epsilon}^{K}  f_1(\Delta_{(k-\epsilon),(i)}^{\SE})  \Big) \Big) \, ,
	\end{split}	
	\end{align}
	where $\epsilon \in \N_{0},$ $f_1(x)=\frac{\log 2 \log(2/x)}{x^2}$ and for $i<j$ the difference between the Borda scores of the arms with the $i$th and $j$th best Borda score (i.e., $q_{(i)}, q_{(j)}$) is denoted by $\Delta_{(i),(j)}^{\SE} = q_{(i)} - q_{(j)}.$ 
	Further, they show a lower bound on the sample complexity for any algorithm, which is an $(\epsilon,\delta)$-PAC algorithm for the considered problem scenario for any preference relation $\bQ$ satisfying $q_{i,j}\geq 3/8$ for any distinct $i,j\in [K]$, of order
	\begin{align*} 
%		\label{eq_sample_hamming_lucb}
		\begin{split}
			%					%			
			\Omega \left( \log\Big( \frac{1}{\delta}\Big) \Big(   \sum\nolimits_{i=1}^{k-2\epsilon} f_0(\Delta_{(i),(k+1+2\epsilon)}^{\SE})  + \sum\nolimits_{i=k+1+2\epsilon}^{K}  f_0(\Delta_{(k-2\epsilon),(i)}^{\SE}) \Big) \right) \, , \quad f_0(x)=x^{-2}.
		\end{split}	
	\end{align*}
	In fact, an even more refined lower bound is derived that also involves the Borda score gaps of arms having a Borda score close to the top-$k$ arm with respect to the Borda scores.
	Nevertheless, the authors show that, up to logarithmic terms, \Algo{Hamming-LUCB} is  already optimal with respect to its sample complexity, if compared with the lower bound above.

	Finally, it is shown that imposing a parametric assumption on the underlying pairwise preference probabilities such as the Bradley-Terry model in \eqref{eq_bradley_terry} does not lead to qualitatively stricter lower bounds for the sample complexity.
%	 
%	
	
%	%	such that with high probability the items in the first set have a larger score than the items in the second set. 
%	The remaining items are then arbitrarily distributed to the two sets to get a Hamming-accurate ranking with high probability.
%	to find an $h$-Hamming-accurate ranking with probability at least $1-\delta$, where %a ranking with $\left|\hat{\cS}_l\right|=\left|\cS_l\right|$ is 
%	$h$-Hamming-accuracy means that $D_H(\cS, \hat{\cS}) \leq 2h$ and $D_H(\cS', \hat{\cS}') \leq 2h$. 
%	y
%	and the notation $\tilde{\bigO}$ absorbs factors logarithmic in $k$ and doubly logarithmic in the gaps. 
%	Note that an $h$-Hamming-accurate ranking is an $\epsilon$-optimal target with respect to the Hamming metric.      
	
%	The algorithm is based on actively identifying the sets $\hat{\cS}$ and $\hat{\cS}'$ consisting of $k-h$ and $n-k-h$ items, respectively, such that with high probability the items in the first set have a larger score than the items in the second set. The remaining items are then arbitrarily distributed to the two sets to get a Hamming-accurate ranking with high probability. 

	 \subsubsection{Preference-based Racing} \label{subsec_pref_race}
	 The learning problem considered by \cite{BuSzWeChHu13} is to find, for some $k < K$, the top-$k$ arms with respect to the Copeland, Borda as well as random walk ranking procedures with high probability, while simultaneously keeping the sampling complexity as low as possible. To this end, three different learning algorithms are proposed in the finite horizon case, with the horizon given in advance. In principle, these learning problems are very similar to the value-based racing task \citep{MaMo94,MaMo97}, where the goal is to select the $k$ arms with the highest means. However, in the preference-based case, the ranking over the arms is determined by the ranking procedure instead of the means. Accordingly, the algorithms proposed by \cite{BuSzWeChHu13} consist of a successive selection and rejection strategy. The sample complexity bounds of all algorithms are of the form $\bigO ( \sum_{1 \leq i < j \leq K} \Delta_{i,j}^{-2}  \log K/\delta )$. Thus, they are not as tight in the number of arms as those considered in Section \ref{sec:assonp}. This is mainly due to the lack of any assumptions on the structure of $\bQ$. Since there are no regularities, and hence no redundancies in $\bQ$ that could be exploited, a sufficiently good estimation of the entire relation is needed to guarantee a good approximation of the target ranking in the worst-case. 
	 %For example, using the Copeland ranking procedure, if most of the $d_{i}$ values are equal to each other, then almost every entry of $\bQ$ needs to be known whether they are significantly bigger than $1/2$ or smaller. Nevertheless, these algorithm can be applied to any task with arbitrary preference matrix.
	 
	 %These algorithms were directly motivated by the Evolution Strategy optimization where in each optimization iteration the top-K candidates need to be selected~\cite{HeIg08,HeIg09}. By replacing the value-based selection mechanism in an ES strategy with a preference-based one such as proposed in \cite{BuSzWeChHu13}, one can obtain a more general stochastic optimization procedure which does not require the function values themselves, but only stochastic pairwise comparisons of the candidate solutions. In~\cite{BuSzWeChHu14}, this preference-based optimization framework is applied to the policy search problem in Markov Decision Processes where it is only assumed that trajectories can be only compared in a pairwise manner like in~\cite{FuHuChPa12}.
	 
	 \subsubsection{Voting Bandits} \label{subsec_voting_bandits}
	  \citet{UrClFeNa13} consider a general bandit learning setting involving $N$ many (unknown) distributions $P_1,\ldots,P_N$ with respective means $\mu_1, \ldots , \mu_N \in [0,1]$, a decision set $\mathcal D$ as well as a known utility function $u:\mathcal D \times [0,1]^N \to \R_+,$ which gives rise to the optimal decision (set) by means of $\argmax_{d\in \mathcal D} u(d,\mu).$
	  The goal is to find, with high probability, an optimal decision in the typical bandit learning protocol, i.e., only one distribution can be queried at a time to obtain a sample, by using as few as possible sample queries to the distributions.

	  By choosing these features in a suitable way, this general learning setting can be customized to the problem of finding an optimal arm in the value-based MAB or the dueling bandits learning scenario, respectively.
	  Indeed, the dueling bandits learning scenario, for instance, can be recovered by setting the decision set as $[K]$, $N=\binom{K}{2}$ and identifying the distributions $P_1,\ldots,P_N$ with the Bernoulli distributions corresponding to the random outcomes of duels, so that $\mu_1,\ldots,\mu_N$ correspond to the entries in the upper triangle matrix of $\bQ.$
	  In light of this, one obtains the problem of finding a Copeland winner, by using the utility function $u(i,\bQ) = d_i,$ while the utility function $u(i,\bQ) = q_i$  leads to the problem of finding a Borda winner\footnote{For sake of convenience, we abuse here the notation by using $\bQ$ in the second argument of $u$ .}.
	  Assuming the existence of a Condorcet winner, the problem of finding the latter can be modeled by using the same utility function as in the case of Copeland winner.

	For the general bandit learning scenario, the authors propose the \emph{Sensitivity Analysis of Variables for Generic Exploration} (\Algo{SAVAGE}) algorithm, which maintains an $N$-dimensional confidence region for the mean vector $(\mu_1,\ldots,\mu_N)^\top$ as well as a set $A\subset [N]$ indicating which distribution is (still) relevant for the final optimal decision.
	In order to make the latter set meaningful, the \Algo{SAVAGE} algorithm relies on a subroutine \Algo{IndepTest}, which needs to be customized for the concrete task of the bandit problem at hand, by exploiting structural properties of the utility function in a suitable way.
	The key idea is that by constantly observing samples from the underlying distributions, the confidence region will concentrate around the singleton set consisting of the mean vector and, in the course of time, allows  one to infer which distribution is irrelevant for the final decision and, consequently, can be removed from the set under consideration.
	To this end, \Algo{SAVAGE} simply queries a sample from each distribution within the set $A$ in a round-robin manner, and reduces this set by means of the subroutine \Algo{IndepTest}.
	
	For the problems of finding a Copeland or Borda winner, the authors design explicit subroutines \Algo{IndepTest}, respectively, which roughly speaking exclude an arm as soon as there exists another arm whose pessimistic Copeland/Borda score is larger than the optimistic Copeland/Borda score of the former.
	Under the assumption of an existing Condorcet Winner, the authors propose to instantiate \Algo{IndepTest} such that it removes an arm simply by checking if this arm is even pessimistically preferred over another arm, which can be expressed as a condition on the utility function used in the Copeland case.

	The sample complexity of \Algo{SAVAGE} for finding the best arm (Copeland or Borda winner) with probability at least $1-\delta$ is shown to be of order 
	$$ \mathcal{O}\left( \sum\nolimits_{1\leq i < j \leq K} \frac{ 1 }{\Delta_{i,j} ^2}  \log \frac{ K}{\delta \Delta_{i,j}}  \right),$$
	by implicitly assuming that there are no ties in $\bQ,$ so that $\Delta_{i,j}\neq 0$ for all distinct $i,j.$ 
	
	Further, the authors suggest to use \Algo{SAVAGE} for the task of regret minimization if the time horizon is known in advance. To this end, they adopt an explore-then-exploit strategy in the spirit of \Algo{IF} or \Algo{BTM} (cf.\ Section \ref{subsec_IF_algorithm} and \ref{subsec_btm_algorithm}).
	However, for all winner concepts, the resulting regret bounds are of the form
	$$	\mathcal{O}\left( \sum\nolimits_{1\leq i < j \leq K} \frac{ 1 }{\Delta_{i,j} ^2} \log (K T)\right)	,	$$
	which are in general not strict for the Condorcet winner (cf.\ Section \ref{subsubsec:rmed}) or the Copeland winner (cf.\ Section \ref{subsec_copeland_rmed}) as the target.
%
%	 \citet{UrClFeNa13} consider a setup similar to the one of \cite{BuSzHu14}. Again, a ranking procedure is assumed that produces a ranking over the arms, and the goal of the learner is to find a best arm according to this ranking (instead of the top-k). Note that a ranking procedure only defines a complete preorder, which means there can be more than one ``best'' arm. The authors propose an algorithm called \Algo{SAVAGE} as a general solution to this problem, which can be adapted to various ranking procedures. Concretely, two procedures are used in their study: the Copeland procedure, in which case the best arm is the Copeland winner, and the sum of expectations (or Borda counts), where the best arm is the Borda winner. Moreover, they also devise a method to find the Condorcet winner, assuming it exists.
%	 
%%	 At high level, the algorithm compares pairs of arms in a round robin fashion and drop pairs of arms from consideration  as soon as it is safe to do so, according to the following rule.
%	 The sample complexity of the implementations by \citet{UrClFeNa13} are of the order $\mathcal{O}(   \frac{K^2   \log K/\delta)}{\min_{i\neq j} \Delta_{i,j} })$ in general. Just like for \citet{BuSzHu14}, this is the price they pay for a ``model-free'' learning procedure that does not make any assumptions on the structure of the preference matrix. 
%%	 The analysis of the authors is more general, because they also investigate the infinite horizon case, where a time limit is not given in advance. 
%	 

	 \subsubsection{Successive Elimination} \label{subsubsec_successive_elim}
	 Interested in finding the best arm according to the Borda-score with a small number of comparisons, \citet{JaKaDeNo15}  consider a specific type of structural constraint on the preference relation $\bQ$, which assumes a (small) set of arms, the top candidates, that are similar to each other with respect to their mutual pairwise preference probabilities, and a (large) set of arms that would be barely preferred over one randomly chosen arm among the top candidates, i.e., arms having a large Borda score gap to the top candidates.
	 This assumption imposes some kind of sparsity on the preference relation and is motivated by numerous practical problem scenarios.
	 
	 They first show that, under such a sparsity assumption, the Borda reduction (cf.\ Section \ref{subsec_borda_ranking}) in combination with a suitable best arm identification algorithm of the value-based MAB problem may result in a poor sample complexity.
%	 
%	 in which the number of samples required only depends on the Borda scores, but not on the individual entries of the preference matrix . 
%	 
	 Subsequently, they propose the Successive Elimination with Comparison Sparsity (\Algo{SECS}) algorithm, which essentially invokes the successive elimination (\Algo{SE}) algorithm of \citet{EvMaMa06} for the value-based MAB problem with the Borda reduction, but enhances upon the latter straightforward approach by explicitly making use of the possible sparsity of the underlying preference relation $\bQ$. To this end, partial Borda score gaps are considered, where the Borda score in \eqref{eq:sscore} is computed only on a subset of $\cA.$ 
	 More specifically, for a specific number of iterations $T_0$ (input of \Algo{SECS}), the \Algo{SECS} algorithm uses only the elimination criterion of \Algo{SE}, which removes arms having large (estimated) Borda score gaps to any another arm.
	 After exceeding $T_0$ iterations, the extra elimination criterion is also activated, which eliminates arms having a large partial Borda score gap estimated by using a set of potential top candidates, thereby exploiting a possible sparsity.
%arge partial Bordagaps
	 
%	 
	
	 Under the above sparsity assumption and uniqueness of the Borda winner, a high probability upper bound on the  sample complexity of \Algo{SECS} of order 
	\begin{align} \label{eq_sample_complex_SECS}
	\begin{split}
		\bigO\Big(\sum_{ i \neq i^* } \min \Big\{ 
		\max\Big\{ \frac{1}{ \Delta_{min}^{\SE} }  \log \frac{K}{ \Delta_{min}^{\SE} \delta} 
		, \ &\frac{1}{ K (\Delta_{i}^{\SE})^2 } \log \frac{K}{(\Delta_{i}^{\SE})^2 \delta}   \Big\} 
		, \\  &\frac{1}{ (\Delta_{i}^{\SE})^2 } \log \frac{K}{(\Delta_{i}^{\SE})^2  \delta}  \Big\}  \Big)
	\end{split}
	\end{align}
	 is shown, where $i^*$ denotes the index of the arm with the largest Borda score, $\Delta_{min}^{\SE}=\min_{i\neq i^*} \Delta_{i}^{\SE}$ and $\Delta_{i}^{\SE} = q_{i^{*}} - q_i.$
	 Here, the first term within the minimum is representing the improvement over the straightforward \Algo{SE} with Borda reduction, whose sample complexity would correspond to the second term within the minimum.
	 Further, a lower bound of order $\Omega\left(\sum_{ i \neq i^* } \frac{1}{ (q_{i^{*}} - q_i)^2 } \log \frac{1}{\delta}   \right)$ on the expected sample complexity of finding the Borda winner with confidence $1-\delta \in (0.85,1)$ is shown, which holds if $q_{i,j} \in [3/8,5/8]$ for any pair $i,j\in[K]$.
	 %	  determines which of two arms is better on the basis of their performance with respect to a sparse set of comparison arms, leading to significant sample complexity improvements compared to the Borda reduction scheme. Basically, \Algo{SECS} implements the together with the Borda reduction and an additional elimination criterion that exploits sparsity. More specifically, \Algo{SECS} maintains an active set of arms of potential Borda winners, and in each time step, chooses an arm uniformly at random and compares it with all the arms in the active set. The algorithm terminates as soon as only one arm remains.

	\subsubsection{Bayesian Sequential Sampling} \label{subsec_bayes_seq_sample}
	
	Seeking to find the top-$k$ Borda scored arms as in  \citep{heckel2018approximate}, the work by \citep{groves2019top} investigates adaptations of well-established Bayesian methods used in the Simulation Optimization community \citep{branke2007selecting} to obtain learning algorithms with a small sample complexity.
	For this purpose, they consider the \emph{probability of correct selection} given by
	$$	\prob(q_i > q_j \, \forall i \in \cS, \, j \notin \cS), \quad \cS \subset [K], |\cS|=k,	$$
	as the performance metric for the learner.
	This metric is estimated by the \emph{approximated expected probability of correct selection} (AEPCS) defined by
	$$ \prob\Big(	\big( \bigcap\nolimits_{i \in \hat{\cS}} \{ q_{i} >c \} \big)	\cap 	\big( \bigcap\nolimits_{j \notin \hat{\cS}} \{ q_{j} <c \} \big)		\Big), \quad c>0,  $$
	where $\hat{\cS}\subset [K]$ is the $k$-sized set of  arms with the current highest estimated Borda scores.
	The consideration of AEPCS is reasonable as it is a lower bound for the sample version of the probability of correct selection, i.e.,
	$	\prob(q_i > q_j \, \forall i \in \hat{\cS}, \, j \notin \hat{\cS}).$
	Based on a normal distribution assumption on the estimated Borda scores, the Pairwise Optimal Computing Budget Allocation (\Algo{POCBA}) algorithm is suggested, which samples in each round the pair of arms having the maximal expected increase on the AEPCS until the AEPCS criterion reaches a predefined threshold.
	
	The authors further suggest to approximate the expected value of information gain of a single duel by the probability that the outcome of the duel will change the current set consisting of the highest estimated Borda scores $\hat{\cS}.$
	Exploiting again the normal assumption on the estimated Borda scores, the authors propose the Pairwise Knowledge Gradient (\Algo{PKG}) algorithm, which is successively choosing the pair of arms maximizing the approximated information gain based on its current Borda score estimates.  
	The algorithm terminates similar as \Algo{POCBA} as soon as the AEPCS criterion exceeds some specific threshold.
	
	Both algorithms are shown to be asymptotically optimal, that is, if the algorithms are run with an infinite sampling budget the probability of a correct selection is tending to one.
	
	\subsubsection{Active Ranking} \label{subsec_active_rank}
	
	The coarse ranking problem is the task of sorting random variables according to a specific parameter of their associated distribution into clusters of predefined sizes \citep{katariya2018adaptive}\footnote{The minimal requirement on these parameters is that they admit an order relation.}.
	Concretely, for a given number of clusters $c \in \{2,3,\ldots,K\}$ and for given cluster boundaries $ 1 \leq k_1 < k_2 < \ldots < k_c = K $ it is intended to identify the $k_1$ random variables with the ``best'' parameters, and from the remaining $K-k_1$ ones the $k_2-k_1$ random variables  with the ``best'' parameters, and so forth.
	Transferring this idea to the realm of bandit problems, it is quite natural to use the expected value of an arm's reward distribution as the parameter of interest in the value-based MAB problem. In the dueling bandits setting, on the other side, the choice of a suitable parameter is again far from obvious, due to the absence of numerical rewards.
	\citet{HeShRaWa16} propose to use the Borda score of an arm as the parameter of interest and study the sample complexity of algorithms to identify a coarse ranking with high probability.
	It is worth noting that the coarse ranking task is a generalization of finding the best arm ($k_1=1,k_2=K$) and identifying the top-$k$ arms problem ($k_1=k,k_2=K$), and also includes the task of finding a total ranking over the arms ($k_i=i$ for $i\in[K]$). 	
	%	

	%	
%	\citet{HeShRaWa16} consider the coarse ranking problem in the dueling bandit setting with the Borda scores as the surrogates for the expected rewards. 
%	
%	for a given tolerance parameter $ \delta \in (0,1) $, they consider finding an estimation $\hat{\br}$ of the true ranking $\br $ such that $ \prob ( \hat{r}_{i}=r_{i} \text{ for all } i \in [K] ) \geq 1-\delta $, while minimizing the number of comparisons queried.
	
%	\citet{HeShRaWa16} consider the Borda ranking problem without requiring any other structural properties of the underlying preference relation $\bQ$. Concretely, for a given tolerance parameter $ \delta \in (0,1) $, they consider finding an estimation $\hat{\br}$ of the true ranking $\br $ such that $ \prob ( \hat{r}_{i}=r_{i} \text{ for all } i \in [K] ) \geq 1-\delta $, while minimizing the number of comparisons queried.
	
	The authors propose the Active Ranking (\Algo{AR}) algorithm, which uses the idea underlying racing or successive elimination strategies \citep{Pa64,MaMo94,EvMaMa06}. 
	More specifically, \Algo{AR}  maintains a set of active arms consisting of arms which have not been assigned to a cluster yet, from which one arm is dueled with another arm chosen uniformly at random from the set of all arms (without the first arm) in a round-robin manner.
	Once the upper and lower confidence bounds on the Borda scores of the active arms reveal that one of the active arms belongs to a certain cluster, the latter is assigned to this cluster and removed from the set of active arms.
%	
%	the algorithm maintains estimates of the Borda scores of the arms obtained based on comparisons with randomly chosen arms, and assigns arms to ranks once being confident enough. It terminates when all arms are ranked. 
	
	Assuming that all Borda scores are distinct,  the authors show that \Algo{AR} outputs the correct coarse ranking with probability at least $1-\delta \in [0.86,1)$ and has a sample complexity of the order 
	\begin{align}\label{eq_ar_sample_complex}
		\begin{split}
			\bigO \Big( \log(K/\delta) \Big(   \sum_{i=1}^{k_1}  & f_1(\Delta_{(i),(k_1+1)}^{\SE})
			+ \sum_{l=2}^{c-1} \sum_{i=k_{l-1}+1}^{k_l} \max \big\{ f_1(\Delta_{(k_{l-1} ),(i)}^{\SE}) \, , \,  f_1(\Delta_{(i ),(k_l + 1)}^{\SE}) \big\} 
			\\ &+ \sum_{i=k_{c-1}+1}^{K} f_1(\Delta_{(k_{c-1} ),(i)}^{\SE}) \Big) \Big)  \enspace , 
		\end{split}
	\end{align}
	where $f_1(x)=\frac{\log 2 \log(2/x)}{x^2}$ and $ \Delta_{(i),(j)}^{\SE} $ is as in Section \ref{subsec_hamming_lucb}. 
	Additionally, a lower bound for any algorithm that outputs the correct coarse ranking with probability at least $1-\delta \in [0.86,1)$ is shown:
	\begin{align*} 
		\begin{split}
			%					%			
			\Omega \Big( \log(1/\delta) \Big(   \sum_{i=1}^{k_1}  & f_0(\Delta_{(i),(k_1+1)}^{\SE})
			+ \sum_{l=2}^{c-1} \sum_{i=k_{l-1}+1}^{k_l} \max \big\{ f_0(\Delta_{(k_{l-1} ),(i)}^{\SE}) \, , \,  f_0(\Delta_{(i ),(k_l + 1)}^{\SE}) \big\} 
			\\ &+ \sum_{i=k_{c-1}+1}^{K} f_0(\Delta_{(k_{c-1} ),(i)}^{\SE}) \Big) \Big)  \enspace ,
		\end{split}	
	\end{align*}
	where $f_0(x) = x^{-2}.$
	This lower bound holds if the underlying preference relations $\bQ$ is such that $q_{i,j}\geq 3/8$ for any distinct $i,j\in [K]$ and shows that \Algo{AR} is nearly optimal.

	Last but not least, similar as in the PAC learning scenario in Section \ref{subsec_hamming_lucb}, it is shown that parametric assumptions on the underlying pairwise preference probabilities do not lead to qualitatively stricter lower bounds for the sample complexity for the considered coarse ranking task.
%revealing that \Algo{Hamming-LUCB} is up to logarithmic terms optimal with respect to its sample complexity.
	
%	They show that the algorithm is optimal up to logarithmic factors  and, moreover, that imposing parametric models such as Bradley-Terry-Luce can reduce the sample complexity by at most a logarithmic factor.

	%#################################
	\section{Further Extensions}
	\label{sec:extensions}
	
	In this section, we review different generalizations and extensions of the setting of preference-based (dueling) bandits as discussed in the previous sections.

	\subsection{Adversarial Dueling Bandits} \label{subsec_adversarial}
	
	\citet{AiKaJo14} consider the adversarial variant of the utility-based dueling bandits problem, in which no stochastic assumption on the utilities of the arms is required, i.e., the latent utility of each arm may change over iterations. The authors suggest to apply the reduction algorithm \Algo{SPARRING}, which has originally been designed for stochastic settings, with an adversarial bandit algorithm such as the Exponential-weight algorithm for Exploration and Exploitation (\Algo{EXP3}) \citep{AuCeFrSc02} as a black-box MAB. More specifically, the algorithm uses two separate MABs, one for each arm. On receiving a relative feedback about a duel, one instantiation of \Algo{EXP3} only updates its weight for one arm and the other instantiation only updates for the other arm. For this \Algo{SPARRING} reduction, it is shown that the $ \mathcal{O}(\sqrt{KT \ln K}) $ upper bound of \Algo{EXP3} in the adversarial MAB problem is preserved. 
	
	Adversarial utility-based dueling bandits are also studied by \citet{GaUrCl15}. They suggest the Relative Exponential-weight for Exploration and Exploitation (\Algo{REX3}) algorithm, which is an extension of the \Algo{EXP3} algorithm to the dueling bandits setting with feedback in the form of pairwise preferences. 
	The key property observed by the authors for the optimal arm in expectation in a particular time step is that, in addition to its property of maximizing the absolute reward, it is also the one maximizing the regret of any fixed opponent strategy (a role that might be played by the algorithms' strategy itself).
%	authors notice that the best arm in expectation at a specific time is not only the one that maximizes the absolute reward, but also the one that maximizes the regret of any fixed reference strategy against it (a role which might be played by the algorithms' strategy itself). 
	This observation provides a possibility to estimate the individual rewards of two arms involved in a comparison, despite having access only to a relative value, and thus allowing them to adapt the  \Algo{EXP3} algorithm to the dueling bandits setting. In addition to providing a general lower bound of order $ \Omega(\sqrt{KT}) $ on the regret of any algorithm, using the reduction to the classical MAB problem by \cite{AiKaJo14}, they prove an upper bound on the expected regret of order $ \mathcal{O}(\sqrt{KT \ln K}) $ for \Algo{REX3}, which is of the same order as the regret bound of \Algo{EXP3} for adversarial MABs, and the one by \cite{AiKaJo14}.
	
	Finally, \citet{ZiSe19} consider the adversarial utility-based dueling setting as well and feed the \Algo{SPARRING} algorithm with their proposed \Algo{TSALLIS-INF} algorithm for the adversarial MAB problem. This leads to an $ \mathcal{O}(\sqrt{KT \ln K}) $ upper bound of \Algo{SPARRING}.
	In addition, they consider the stochastically constrained adversarial setting, in which, contrary to the adversarial setting, a best arm is fixed throughout the time horizon.
	For this setup, they show an expected regret upper bound for \Algo{TSALLIS-INF} of order  $ \bigO( \sum_{j \neq i^*} \frac{\ln(T)}{\Delta_{i^*,j}}) + \bigO(K)$ resulting in an expected regret upper bound of the same order for \Algo{SPARRING} in the stochastically constrained adversarial version of the utility-based dueling bandits setting.

	\subsection{Contextual Dueling Bandits} \label{subsec_contextual_duels}
	
	\citet{DuHoShSlZo15} extend the dueling bandits framework to incorporate contextual information in the spirit of MAB problems \citep{auer2002using}. More precisely, the learner is supposed to optimize its choice of arms  in the course of an iterative learning process of the following kind: In each round, the learner observes a random context, chooses a pair of actions, conducts a duel between them, and observes an outcome in the form of a pairwise preference. The authors consider the solution concept of a von Neumann winner and present three algorithms for online learning and for approximating such a winner from batch-like data, while measuring performance of a learner by means of the regret as in Section \ref{subsec_sparse_sparring}.
%	\cite{BaKaScZo16}.  

	The authors first present an algorithm that shares similarities with the sparring approach of \citet{AiKaJo14} discussed in Section \ref{subsec_adversarial}: Two separate independent copies of the multi-armed bandit algorithm \Algo{Exp4.P} \citep{BeLaLiReSc11}, which is designed to work with a pre-specified space of multi-armed bandit algorithms in a contextual setting, are run against each other. 
	Both copies are using the same action space, context space and space of multi-armed bandit algorithms as for the original problem. In each round, the environment (or nature) specifies a context and a preference matrix, while only the context is revealed to both copies, which then select an action, respectively. 
	A duel is run between the two selected actions, and the (negated) outcome (cf.\ the zero-sum game matrix in Section \ref{subsec_alternative_targets}) is forwarded as feedback to the first (second) copy. 
	This approach, called \Algo{Sparring Exp4.P}, leads to a regret that is upper bounded by $ \bigO(\sqrt{KT\ln(|\Pi|/\delta)}),$ where $\Pi$ is the space of MAB algorithms, with probability at least $1-\delta$, and requires time and space linear in the term $|\Pi|$, which leads to computational issues in cases where $\Pi$ is large. 
	
	To deal with the latter issue, the authors furthermore propose a general approach for constructing an approximate von Neumann winner. 
	To this end, the problem of finding a von Neumann winner is substituted by a more convenient empirical version of the latter problem. 
%	 to be solved to a more tractable form on the basis of a collection of empirical exploration data
%
	Assuming the existence of a classification oracle on $\Pi$, which can find the minimum cost MAB algorithm in $\Pi$ if it is provided with the cost of each action on each sequence of contexts, the authors propose two algorithms: \Algo{SparringFPL}, which is sparring two copies of the Follow-the-Perturbed-Leader algorithm \citep{KaVe05}, and \Algo{ProjectedGD}, which builds on the projected gradient descent algorithm \citep{Zi03}, but is sparring essentially against its  worst-case opponent strategy. 
	The two algorithms are primarily solving a compact game based on the more convenient empirical problem version, which, however, is equivalent to computing an approximate von Neumann winner. While their regret bound is weaker than for $\Algo{Sparring Exp4.P}$, namely $ \bigO((KT)^{2/3} \ln^{1/3}(|\Pi|/\delta)),$  they require time and space that depend only logarithmically on  $|\Pi|$.
	Further, the running time as well as the number of oracle calls in order to return an approximate von Neumann winner is theoretically analyzed.
%	

	%%%%%%%%%%%
	Based on the Perceptron algorithm, \citet{CoCr14} develop the \Algo{SHAMPO} (SHared Annotator for Multiple PrOblems) algorithm for online multi-task learning with a shared annotator, in which learning is performed in rounds. In each round, each of $ K $ different learners receives an input and predicts its label. A shared stochastic mechanism then annotates one of the $ K $ inputs, and the learner receiving the feedback updates its prediction rule. The authors show that this algorithm can be used to solve the contextual dueling bandits problem when a decoupling of exploration and exploitation is allowed.  
	
	To pick a task to be labeled, \Algo{SHAMPO} performs an exploration-exploitation strategy in which tasks are randomly queried, with a bias towards tasks that involve a high uncertainty about the labeling. To perform an update on the parameter vector representing the model, the algorithm applies the Perceptron update rule to the true label revealed for the task chosen. 
	
	\subsection{Dueling Bandits on Posets}
	
	\citet{AuRa17} extend the dueling bandits problem to partially ordered sets (posets), allowing pairs of arms to be incomparable. They consider the problem of identifying, with a minimal number of pairwise comparisons in a pure exploration setting, the set of maximal arms or Pareto set among all available arms. The main challenge in this framework is the problem of indistinguishability: The learner may be unsure whether two arms are actually comparable and just very close to each other, or whether they are indeed incomparable, no matter how many times the arms are compared. Without any additional information, it might then be impossible to recover the exact Pareto set. 
	
	The authors first devise the \Algo{UnchainedBandits} algorithm to find a nearly optimal approximation of the Pareto front of any poset. The strategy implemented by the algorithm is based on a peeling approach that offers a way to control the number of comparison of arms that are in fact indistinguishable. The authors provide a high probability regret bound of $ \bigO \left( K \, \text{width}(S)\log \frac{K}{\delta} \sum_{i,i \notin \mathcal{P}} \frac{1}{\Delta_i} \right) $, where $ S $ is the poset, $ \text{width}(S) $ is its width defined as the maximum size of an antichain (a subset in which every pair is incomparable), $ \cP $ is the Pareto front, $ \Delta_i $ is the regret associated with arm $a_i$ defined as the maximum difference between arm $a_i$ and the best arm comparable to $a_i$, and the regret incurred by comparing two arms $a_i$ and $a_j$ is defined by $ \Delta_i+\Delta_j $.
	
	Further, by making use of the concept of decoys, the authors show that \Algo{UnchainedBandits} can recover the exact set $ \mathcal{P} $, incurring regret that is comparable to the former one---except for an extra term due to the regret incurred by the use of decoys---with a sample complexity that is upper bounded by $ \bigO ( K \text{width}(S) \log(NK^2/\delta)/\Delta^2 ) $, where $ N $ is a positive integer related to a weaker form of distinguishability and $ \Delta $ is a parameter of the decoys. The concept of decoys is an idea inspired by works from social sciences and psychology, intended to force a decision maker to choose a specific option (the target option) by presenting her/him a choice between the target option and another option dominated by the latter in specific aspects (the decoy option). 
	
	\subsection{Graphical Dueling Bandits}

	\citet{DiGeMa11} consider the bandit problem over a graph, the structure of which defines the set of possible comparisons. More specifically, they assume that there is an inherent and unknown value per node (arm), and that the graph describes the allowed comparisons: two nodes are connected by an edge if they can be dueled with each other. 
	Such a duel returns a random number in $[-1,1]$, the expected value of which is the difference between the values of the two nodes. Thus, unlike the traditional dueling bandits setup, the topology is not a complete graph, and non-adjacent nodes can only be compared indirectly by sampling all the edges along a path connecting them.  
	Further, the learner receives feedback in the form of a numerical value.
	
	The authors consider different topologies and focus on the sample complexity for finding the optimal arm (largest latent value) in the PAC setting. 
%	They provide algorithms that construct estimates of edge reward differences, and combine these estimates into a node selection procedure, together with their sample complexities, in the case when the edges are bounded. 
	For the linear topology, in which each node is comparable to at most two other nodes, they present an algorithm that samples all edges, computes the empirical mean of each edge, and based on these statistics, assigns a value to each node reflecting its alleged optimality. The sample complexity is $ \bigO(\frac{K^2}{\max \{ \epsilon, u\}^2} \log(\frac{1}{\delta})) $, where $ u $ is the difference between the node with the highest value and the node with the second highest value. 
	
	For the tree topology, that is, a topology in the form of a tree, the authors present an algorithm which essentially reduces the graphical bandit problem to graphical bandit problems with linear topologies by treating each path from the root to a leaf as a line graph and use one each of these graphs the algorithm above. This leads to a sample complexity of $ \bigO(\frac{KD}{\max \{ \epsilon, u\}^2} \log(\frac{|L|}{\delta})) $, where $ D $ is the diameter of the tree and $ L $ the set of leaves. 
	
	For the network topology, that is, general connected and undirected graphs, the authors present the Network Node Elimination (\Algo{NNE}) algorithm, which is inspired by the action elimination procedure of \citet{EvMaMa06}. This algorithm has a sample complexity upper bounded by $ \frac{KD}{(\max \{ \epsilon, u\}/\log K)^2} \log(\frac{K}{\delta/\log K}) $. Further, they consider the contextualized version of the problem in the spirit of Section \ref{subsec_contextual_duels}, and show that a version of the \Algo{NNE} algorithm achieves a sample complexity of the form $ \bigO(B \log^2 B) $, where $ B = \frac{KD}{(\epsilon/\log K)^2} \log \left( \frac{K}{\delta/\log K} \right)d^2 $, and $ d $ is the dimension of the feature vectors.

	\cite{Ka16} also applies the suggested verification algorithm to graphical bandits, which improves upon the latter approach by  logarithmic terms.

	\subsection{Dueling Bandits with Dependent Arms}

	Focusing once again on minimizing the weak regret, \citet{ChFr16} study the utility-based dueling bandits, where each utility $u(\theta,x_{i})$ of an arm $a_i$ is determined by a known utility function $u:\R^{d'} \times \R^d \to \R$ of a known (and fixed) $d$-dimensional arm-specific feature $x_i$ and some unknown $d'$-dimensional weight parameter $\theta.$
	Thus, dependency among arms can be modeled if $d'$ is smaller than $d,$ which in turn can be exploited by a learner.
	
	The authors introduce the Comparing The Best (\Algo{CTB}) algorithm, which maintains $2^K$ many cells, one for each possible pairwise preference sequence involving every pair of arms, and assigns scores to each cell based on the number of times a pairwise preference corresponding to the cell has been observed.
	In each iteration, the first arm of the duel is the one that is optimal with respect to the cell(s) with the largest score, while the second arm is the arm that is optimal with respect to the cell(s) having largest score, and where the optimal arm is different from the first one.
	
	It is shown that \Algo{CTB} enjoys a bound on its expected cumulative weak regret of order 
	$ \bigO\left(		\frac{K^2 M'}{\min_{i\neq j} \Delta_{i,j}^2}	(\max_{i,j} u(\theta,x_i) - u(\theta,x_j)) \right),$
	where $M'$ is a constant which is $\Theta(	2^K )$ in general, but $O(K^{2d'})$ for special cases.
%	
%	which is constant in $T$. 
%	The dependence on $K$ is of the form $2^K$ in the worst case and $K^{2d}$ when the utility function is linear in the dimension of the space of preferences and arm features $d$.
	
%	 on the idea of cells that correspond to possible orderings of the arms by utility. The algorithm uses optional prior information to initialize each cell with a score, which can be interpreted as a monotone transformation of the posterior probability that the unknown preference vector is in this cell. It updates these scores based on results from duels, where arms are chosen for a duel by selecting two cells that have different best arms, and are together most likely to contain the unknown preference vector. 
	
	Further, under specific conditions, the authors show that the algorithm can be implemented in a favorable way in order to cope with  high computational costs due to the large number of cells.
%	Apparently, the algorithm h
%		
%	As a consequence, the authors suggest a general implementation that is appropriate for a small number of arms, and a computationally efficient one for a larger number of arms. It can be used when prior information that can be expressed as an initial  score for each pair of arms is available. 

	\subsection{Partial Monitoring Games}

	The dueling bandits problem can be seen as a special case of the partial monitoring (PM) problem \citep{BaPaSz11,Ba13,BaFoPARASz14}---a generic model for sequential decision-making with incomplete feedback, which is defined by a quintuple $ (\bN,\bM,\bSigma,\cL,\cH) $, where $\bN$ is the set of actions, $\bM$ is the set of outcomes, and $\bSigma $ is the feedback alphabet; the loss function $\cL$ associates a real-valued loss $\cL(I,J)$ with each action $ I \in \bN $ and outcome $ J \in \bM $, and the feedback function $ \cH $ associates a feedback symbol $ \cH(I,J) \in \bSigma $.
%	\citet{GaUr15} study the dueling bandits problem as an instance of a partial monitoring (PM) problem \citep{BaPaSz11,Ba13,BaFoPARASz14}---a generic model for sequential decision-making with incomplete feedback, which is defined by a quintuple $ (\bN,\bM,\bSigma,\cL,\cH) $, where $\bN$ is the set of actions, $\bM$ is the set of outcomes, and $\bSigma $ is the feedback alphabet; the loss function $\cL$ associates a real-valued loss $\cL(I,J)$ with each action $ I \in \bN $ and outcome $ J \in \bM $, and the feedback function $ \cH $ associates a feedback symbol $ \cH(I,J) \in \bSigma $. 
	In each round of a PM game, first the opponent chooses an outcome $ J_t $ from $ \bM $, and the learner an action $ I_t $ from $ \bN $. Then, the learner suffers the loss $ \cL(I_t,J_t) $  and receives the feedback $ \cH(I_t,J_t) $. The performance of a learner is measured by means of the expected cumulative regret against the best single-action strategy 
	\begin{align} \label{def_regret_pm}
			R^T = \max_{i \in \bN} \sum_{t=1}^{T} \cL(I_t,J_t)-\cL(i,J_t).
	\end{align}
	Following \citet{GaUr15} the utility-based dueling bandits problem  with a linear link function (and ties) can be encoded as a PM problem with the set of actions given by the set of all pairs of arms $ \bN = \{ (i,j):1\leq i,j \leq K, i\leq j \} $, the alphabet $ \bSigma = \{0,1/2,1\} $, and the set of outcomes given as vectors $\bm =(m_1,\ldots,m_K) \in \bM = [0,1]^K$, where $ m_i $ is simply the utility of arm $a_i$, which can be interpreted as its instantaneous gain, so that we differ from the notation as in Section \ref{subsec_utility_dueling_bandits}. 
	After the environment selects an outcome $ \bm \in \bM $ and the learner a duel $ (i,j) \in \bN $, the instantaneous loss\footnote{
		Note that \citet{GaUr15} work with the instantaneous gain $\cG((i,j),\bm) =  \frac{\bm_i+\bm_j}{2}$.} is 
	\[ \cL((i,j),\bm) = 1 - \frac{\bm_i+\bm_j}{2} \enspace ,
	\] 
	and the feedback is    
	\begin{equation*}
	X=
	\begin{cases}
	0, & \text{if}\ \bm_i<\bm_j \\
	1/2, & \text{if}\ \bm_i=\bm_j \\
	1, & \text{if}\ \bm_i>\bm_j 
	\end{cases} \enspace .
	\end{equation*}
	Using the PM formalism, the authors prove that the dueling bandits problem is an easy instance according to the hierarchy of PM problems \citep{BaPaSz11,Ba13}, i.e., it is possible to achieve a regret bound of order $\tilde{\bigO}(\sqrt{T})$ and the difference of  the loss vectors of adjacent actions are locally observable, where regret is measured as in \eqref{def_regret_pm}.

	Quite recently, \citet{kirschner2020information} suggested the Information Directed Sampling (\Algo{IDS}) algorithm for linear partial monitoring games and also consider the utility-based dueling bandits setting above as a special case. 
	Their results lead to a regret bound of order $\tilde{\bigO}(\sqrt{KT}),$ which is the same as \Algo{REX3} (cf.\ Section \ref{subsec_adversarial}).
%	which is, as noted by  \citet{GaUr15}, the same bound as \Algo{REX} (see Section \ref{subsec_adversarial}).
%	
	
%	Further, they survey existing PM algorithms and their optimality to solve dueling bandits problems efficiently, with respect to time $ T $ and number of actions $ K $. Their study reveals that the $ REX3 $ algorithm \citep{GaUrCl15} for adversarial utility-based dueling bandits with a regret of $ \tilde{\bigO}(\sqrt{KT}) $ is the only optimal algorithm with respect to $T$ and $K$, and that the \Algo{SAVAGE} algorithm \citep{UrClFeNa13} for general stochastic dueling bandits with a regret of $ \bigO(K^2 \log T) $ is optimal in $T$ but not in $K$.

	\subsection{Dueling Bandits for Qualitative Feedback}
	
	\citet{Xu_Honda_Sugiyama_2019} consider the qualitative dueling bandits problem, which is a variant of the MAB problem with reward feedback of the pulled arm on an ordinal scale, i.e., qualitative feedback, instead on a numerical scale, i.e., quantitative feedback.
	This setting was introduced by \citet{szorenyi2015qualitative}, but analyzed there in a rather classical MAB setting. Instead, following ideas by \citet{BuSzWeChHu13}, the authors embed the problem into a dueling bandits setting.
	More precisely, for each pair of arms $(i,j)$, they define the pairwise winning probability in (\ref{eq:pairwisex}) as $q_{i,j} = \prob(X_i > X_j) + \nicefrac{1}{2} \prob(X_i=X_j),$ where $X_i,X_j$ are mutually independent random variables with values in the considered ordinal scale. The law of $X_i$ is $\nu_i$, representing the qualitative feedback mechanism of an arm $a_i.$

	With this definition of pairwise probabilities, one can orchestrate the dueling bandits mechanism for this framework by simultaneously pulling two arms and observing which one has a higher ordinal reward, breaking ties at random.
	The authors seek to minimize the expected cumulative regret, with the cumulative regret given as 
	\begin{equation*}
	\exptd [R^{T} ] = \sum_{t=1}^T 	\Delta_{a_{i^*},a_{i(t)}} \, ,
	\end{equation*}
	where $a_{i^*}$ is either the Condorcet winner or the Borda winner, and $i(t)$ the index of the arm pulled in time step $t$.
	%
	%Note that this is a 
	
	For the case of the Condorcet winner, they suggest the \Algo{Thompson Condorcet sampling} algorithm, which uses for each arm $a_i$ a Dirichlet distribution as the prior distribution for the probability vector specified by the law $\nu_i$. In each time step, the algorithm generates a random sample from the posterior distribution and pulls the Condorcet winner based on these random samples\footnote{If no Condorcet winner exists for the sample, a new random sample is drawn.}.
	
	For the case of the Borda winner, the \Algo{Thompson Borda sampling} algorithm and the \Algo{Borda-UCB} algorithm are introduced. The \Algo{Thompson Borda sampling} algorithm follows the same idea as \Algo{Thompson Condorcet sampling}, only replacing the determination of the Condorcet winner by the Borda winner of the posterior samples.
	The \Algo{Borda-UCB} algorithm is based on the UCB algorithm. It pulls the arm with the highest upper confidence bound on the estimated Borda score if this is also the arm with the highest number of pulls. Otherwise, all arms not having the highest number of pulls are pulled in order to enhance the estimation of the Borda scores.
	
	Furthermore, in the case of an existing Condorcet winner, the authors show an expected regret bound for \Algo{Thompson Condorcet sampling} of order $\mathcal{O}(K \log(T))$. They also show that the constant appearing in the bound can be arbitrary small compared to the constants in the lower bound for the dueling bandits problem applied to the qualitative dueling bandits problem.
	For the case of the Borda winner, they show that \Algo{Thompson Borda sampling} suffers an expected regret which is polynomial with respect to $T$, while \Algo{Borda-UCB} achieves expected regret bounds of order $\mathcal{O}(K\log(T)),$ which, as verified by the authors, matches the theoretical lower bound.

	\subsection{Combinatorial Pure Exploration for Dueling Bandits}
	The combinatorial pure exploration for dueling bandits problem is a variation of the combinatorial pure exploration MAB problem \citep{chen2014combinatorial} introduced by	\cite{chen2020combinatorial}.
	In this setting, there exists a known bipartite graph $G=(\cA,\cP,E)$ with partitions $\cA=\{a_1,\ldots,a_K\}$ and $\cP=\{p_1,\ldots,p_l\},$ where $l \leq K,$ and a set of edges $E \subset \cA \times \cP$ with the following meaning:
	The arms correspond to candidates, while elements in $\cP$ are specific positions, and each edge $e_{i,k}=
	(a_i,p_k) \in E$ represents that arm $a_i$ is available for position $p_k.$
	Moreover, the existence of an unknown preference matrix $\bQ_G \in [0,1]^{|E|\times|E|}$ associated with $G$ is assumed. The entries of this matrix are $(q_{e_{i,k},e_{i',k'}})_{e_{i,k},e_{i',k'} \in E}$, which are zero if $k\neq k'$ and otherwise indicate the stochastic preference of arm $a_i$ over $a_{i'}$ for position $p_k.$
%	
%	Note that $\bQ_G$ is not a reciprocal relation anymore by setting its entries corresponding to incomparable arms to zero, but, however, alleviates many   

%	
	The goal is to find an optimal maximum matching\footnote{A matching of a bipartite graph $G=(\cA,\cP,E)$ is a sub-graph $G'$ with a set of edges $E'\subset E$ such that there exist no two edges $e'=(a',p'),e''=(a'',p'')$ in $E'$ with $p'=p''.$ A maximum matching is a matching of $G$ with the maximal possible number of edges.} $M$ in $G$ based on $\bQ_G,$  where it is once again not obvious how to define the notion of an optimal maximum matching without real-valued rewards (cf.\ Section \ref{subsec_targets}).
	As a remedy, the authors introduce the pairwise preference of a maximum matching $M$ over a maximum matching $M'$ (both with respect to $G$) by means of
	$$	 q^G_{M,M'} = \frac{1}{l} \sum_{k=1}^l q_{ M(k),M'(k)  },		$$
	where $M(k)$ (or $M'(k)$) denotes the edge in $M$ (or $M'$) with $p_k$ as its endpoint. 
	With this, it is possible to leverage the winner concepts of the dueling bandits: Let $\cM$ be the set of all maximum matchings in $G,$ then
	\begin{itemize}
		\item [--] $M^*$ is a Condorcet matching winner  if $q^G_{M^*,M}> 1/2 $ holds for any $M\in \cM\backslash\{M^*\};$
		\item [--] $M^*_{\SE}$ is a Borda matching winner if it has the highest Borda matching score defined for any $M\in \cM$ by 
		$ q^G_{M}  = \frac{1}{|\cM|} \sum_{M' \in \cM} q^G_{M,M'}.$
	\end{itemize}
	Once again, the  Condorcet matching winner  might not exist for $\bQ_G,$ while a Borda matching winner always exists, but might not be unique and does not necessarily coincide with the former if it exists.

%\\
	For the Borda matching winner  as the underlying goal, one could use the Borda reduction technique (cf.\ Section \ref{subsec_borda_ranking}) and make use of the \Algo{CLUCB} for the combinatorial pure exploration MAB problem suggested by \cite{chen2014combinatorial}.
	However, the na\"ive usage of the Borda reduction would require an exact uniformly at random sampling from the set of all maximum matchings in $G,$ which in turn could lead to exponential costs.
	Therefore, the authors modify this approach leading to the \Algo{CLUCB-Borda-PAC} and the \Algo{CLUCB-Borda-Exact} algorithm, which replace the exact uniformly at random sampling procedure by means of a substitutional sampling procedure generating approximately uniform random samples of $\cM,$ but with polynomial costs.
	As the approximate sampling procedure introduces a bias into the natural estimates, both algorithms are  adjusted accordingly.

	\Algo{CLUCB-Borda-PAC} is shown to be an $(\epsilon,\delta)$-PAC algorithm for finding the Borda matching winner  with a sample complexity of order $\bigO	\left( \frac{1}{H_\epsilon^{\SE}} \log\left( \frac{H_\epsilon^{\SE}}{\delta}	\right) \right),$
	where $H_\epsilon^{\SE} = \sum_{e \in E} \min\{	\frac{C_G^2}{(\Delta_e^{\SE})^2}, \frac{1}{\epsilon^2}		\}$ with $\Delta_e^{\SE}$ being some specific gap terms and $C_G>0$ some constant depending on the structure of the bipartite graph $G.$
	For \Algo{CLUCB-Borda-PAC}, the authors verify its correctness to return the Borda matching winner with probability at least $1-\delta$ and derive a sample complexity bound of order 
	 $$\bigO \left( C_G^2 \cdot H^{\SE} \cdot \log (l/\Delta_{\min}^{\SE}) \cdot \log\left( \frac{C_G \, H^{\SE}}{\delta} \right) + \log\log(l/\Delta_{\min}^{\SE})  \right),$$
	where $ H^{\SE} = \sum_{e \in E}	\frac{1}{(\Delta_e^{\SE})^2}$ and $\Delta_{\min}^{\SE} = \min_{e \in E} \Delta_e^{\SE}.$
	Further, a lower bound  of order $\Omega\left(  H^{\SE} \log (1/\delta)  	 \right)$ for this learning scenario is shown on a specific subclass of problem instances, so that both algorithms are almost optimal regarding their sample complexity for such problem instances.
	 
	Assuming the existence of a Condorcet matching winner, the authors suggest two algorithms, \Algo{CAR-Cond} and  \Algo{CAR-Parallel}, for identifying it.
	\Algo{CAR-Parallel} uses $k$ many parallel variants of the \Algo{CAR-Cond} algorithm, which are adapting the verification idea of \citet{Ka16} (cf.\ Section \ref{subsubsec:verification_based}), with suitably chosen confidence levels in order to profit from the variant \Algo{CAR-Cond} having the most favorable sample complexity of all parallel variants.
	The underlying key idea of \Algo{CAR-Cond} is that the Condorcet matching winner can be expressed as the solution of a specific optimization problem, which in turn can be relaxed to a convex optimization problem.
	Having access to some oracle that can return an approximate solution under some specific constraints and a suitable guarantee, the \Algo{CAR-Cond} algorithm maintains a set of undecided edges and uses upper as well as lower confidence estimates of the entries in $\bQ_G$ to iteratively check whether an undecided edge $e$ is likely an element of the Condorcet matching winner or not.
%	
%	The check is conducted by computing the approximate solution quality in four variants: Whether $e$ is included or excluded in the matching solution based on the upper or the lower confidence estimates.
%%	
%	If the solution quality with $e$ pessimistically included is up to some margin larger than the solution quality with $e$ pessimistically excluded, $e$ is added to the solution matching, while if the solution quality with $e$ optimistically included is up to some margin smaller than the solution quality with $e$ pessimistically excluded, $e$ is added to the set of edges not belonging to the solution matching.
% 

	Both algorithms admit a polynomial running time and the authors derive respective bounds on the sample complexity:
	\Algo{CAR-Cond} has a bound of order 
	$$	\bigO\left( 	\sum\nolimits_{j=1}^l \sum\nolimits_{e,e'\in E : e\neq e' } \frac{1}{(\Delta_{e,e'})^2} \log\left(	\frac{K^G}{\delta \Delta_{e,e'} }	\right) \right),		$$
	where $K^G$ is the number of all possible duels within $G,$ and $\Delta_{e,e'}$ is some gap term, while the sample complexity bound of \Algo{CAR-Parallel} is qualitatively similar to \eqref{bound_verif} by replacing $K$ by $K^G,$  $\Delta_{i,j}$ by $\Delta_{e,e'},$ and the sums by sums as in the display above.

	%#################################
	\section{Multi-Dueling Bandits}
	\label{sec:multi_duel_extensions}
	
	\newcommand{\bA}{\mathbb{A}}
	\renewcommand{\S}{\mathcal{S}}

	Although there are various practical scenarios in which the underlying sequential decision process can be modeled by means of the dueling bandits setting, because each of the decisions corresponds to a qualitative comparison of two choice alternatives, this modeling approach is apparently not sufficient to capture more general sequential decision processes in which qualitative comparisons of more than two available choice alternatives can be carried out at once.
	Such more general variants occur in many fields of application such as recommender systems, where a set of items (videos, songs, etc.) are displayed to a user, whereupon the latter expresses her preference over the displayed items in the form of a discrete choice.
	Another relevant field is web search, where usually an ordered list of the allegedly most relevant websites related to a user's query is returned, resulting in observing a click (or no-click) of the user for the presented website collection.
	Last but not least, any application involving social choices fits into the more general variant as well, because the underlying data correspond to individual opinions oftentimes expressed in the form of a (partial) ranking over specific choice alternatives.
%	the most favorable item

	Motivated by the limited coverage of the dueling bandits modeling approach for such practically relevant problem scenarios, research interest in the so-called \emph{multi-dueling bandits} problem has recently increased. The latter is a generalization of the dueling bandits problem allowing the learner a great deal of latitude with regard to the available actions.
	This generalization is conceptually similar to how the \emph{Combinatorial Bandits} \citep{cesa2012combinatorial} generalize the classical value-based MAB problem by allowing the learner to select subsets of arms in each iteration as well, whereupon feedback either in the form of rewards of each single arm in the selected subset (semi-bandit feedback) or the total sum of the rewards (bandit feedback) is observed.
	However, just like the underlying basic variants, these two generalizations are still fundamentally different, since the feedback obtained by the learner in the multi-dueling bandits setting is of a qualitative nature, while the feedback in the combinatorial bandits is of a quantitative or numerical nature. 
	
	There are some active fields of research that are closely related to the multi-dueling bandits problem, in the sense that the learner also receives some kind of preference feedback related to the choice alternatives (arms). 
	These frameworks and their similarities as well as differences will be discussed at the end of this section.

	Due to the different and in particular more general action space compared to the basic dueling bandits problem, there are a couple of novelties emerging in the modeling of the learning scenario of multi-dueling bandit problems, which shall be described in the following.

	\subsection{Learning Protocol} \label{subsec_learning_protocol_mulit_duel}
	As already pointed out, one of the main differences between the dueling and the multi-dueling bandits   problem concerns the action space of the learner, which we shall denote by $\bA$ throughout this section. 
	In the setting of multi-dueling bandits, the action space $\bA$ corresponds to a family of subsets of $\cA=\{a_1,\ldots,a_K\},$ which does not necessarily have to be the family of all possible pairs of arms in $\cA$, while the decision making process still iterates in discrete steps, either through a finite time horizon $\mathbb{T} \defeq [T]=\{1, \ldots , T \}$ or an infinite horizon $\mathbb{T} \defeq \mathbb{N}.$  
	In each iteration $t \in \mathbb{T}$, the learner can perform as its action a comparison of the arms within the selected subset $A_t \in \bA$, resulting in a qualitative feedback (cf.\ Section \ref{subsec_feedback_multi}).
	Practically motivated forms of the underlying action space $\bA$ include the following:
	\begin{itemize}
		\item $\bA_{l},$ all subsets of $\cA$ of a fixed size $l\in \{ 2, \ldots , K \}.$
		\item $\bA_{l}^+,$ the union of $\bA_{l}$ and $\{a_1\},\ldots,\{a_K\}.$
		\item $\bA_{\leq l},$ all subsets of $\cA$ with cardinality at least two but at most $l$ for some fixed $l\in \{ 2, \ldots , K \}.$
		\item $\bA_{\leq l}^+,$ the union of $\bA_{\leq l}$ and $\{a_1\},\ldots,\{a_K\}.$
		\item $\bA_{\text{full}} \defeq \bA_{\leq K}^+,$ all non-empty subsets of $\cA.$ 
	\end{itemize}
	By characterizing the admissible ``full commitment" to one arm of the dueling bandits by means of the singleton sets $\{a_1\},\ldots,\{a_K\}$, the action space of the basic dueling bandits setting corresponds to the action space $\bA_{\leq 2}^+.$ 
	Needless to say, the complexity of most of the above action spaces is much higher than the complexity of the action space underlying the dueling bandits.
	
	\subsection{Feedback Mechanisms} \label{subsec_feedback_multi}
	
	The fundamental assumption of observing a qualitative feedback underlying the dueling bandits or basic preference-based multi-armed bandits can be manifested in various ways if the learner expects a qualitative comparison of the arms involved in the performed action.
	In the following, we therefore discuss all possible types of feedback, throughout assuming that $A_t \in \bA$ is the action of the learner in iteration $t$, and $I(A_t) \subseteq [K]$ is the corresponding index set of the arms in $A_t.$  
	\subsubsection{All Pairwise Preferences} \label{subsec_all_pairwise_feedback}
	One natural way to leverage the concepts of the dueling bandits to the multi-dueling bandits is by assuming that all (noisy) pairwise preferences among the involved arms in $A_t$ are observed, which in turn are still governed by an underlying (unknown) preference relation $\bQ$ specifying the pairwise preference probabilities.
	Formally, if $|A_t|\geq 2$, the feedback\footnote{In case $|A_t| = 1,$ i.e., a ``full commitment" to one arm (compare Section \ref{subsec_regret_bounds}), the learner observes obviously no preference information. } consists of a sequence of length $\binom{|A_t|}{2},$ each element of which is either $\{a_i \succ a_j \}$ or $\{a_i \prec a_j \}$ for some distinct $a_i,a_j\in A_t,$ where the probabilistic mechanism generating such a pairwise preference is given by \eqref{eq:pairwisex}: 
	$$ \prob \left(  a_i \succ a_j \right)  = q_{i,j}, \quad   \prob \left(  a_i \prec a_j \right)  = 1- q_{i,j} = q_{j,i}.$$

	\subsubsection{The Most Preferred Arm}
	From a high level point of view, by comparing two specific arms in the dueling bandits problem, the learner obtains as its feedback the most preferred arm among these two.
	Thus, one can easily generalize the concept underlying this type of feedback to cases with more than two arms involved in the comparison by assuming that the information received is still only the most preferred arm, but now among \emph{all arms} involved  in $A_t.$   
	Formally, the feedback is a partial ranking of the form $a_{i} \succ A_t \backslash \{a_{i} \}$ for exactly one $a_{i} \in A_t,$ which is observed with probability
	\begin{align} \label{defi_winner_feedback}
%		\prob(a_i \, | \, A_t)
		q_{i| I(A_t)} \defeq  \prob(a_{i} \succ A_t \backslash \{a_{i} \}).
	\end{align} 
	Such type of feedback is of great importance especially in the fields of economics or social sciences and studied there under the notion of \emph{discrete choice models} \citep{train2009discrete}.
	Discrete choice models specify the probability that an individual chooses one alternative among a given set of choice alternatives, which is essentially represented by the left-hand side of \eqref{defi_winner_feedback} by regarding choice alternatives and arms as the same.
	Formally, a discrete choice model assumes a latent family of categorical distributions $(q_{i|I(A)})_{A\in \bA, i\in I(A)}$ for every admissible set of choice alternatives (arms) $A \in \bA.$

	A popular class of such categorical distribution families is specified by a \emph{Random Utility Model} (RUM), where each arm (choice alternative) $a_i \in \cA$ is assumed to be equipped with a (latent) utility $\theta_i>0$ and the discrete choice probability is given by
	\begin{align}\label{defi_rums}
%		\prob(a_i \, | \, A_t)
		q_{i| I(A_t)} 	  = \prob \left( u_i  = \max_{a_j \in A_t} u_j  \right),
	\end{align}
	where $u_i = \theta_i + \zeta_{i}$ and $\zeta_1,\ldots,\zeta_K$ is an identically distributed sample of some probability distribution $\prob^*.$
	Apparently, the probability specified by \eqref{defi_rums} depends on the concrete values of the latent utilities as well as on the distribution  $\prob^*$ of the noise terms  $\zeta_1,\ldots,\zeta_K$. 
	Different parametric probability models are used for the random noise in order to assess the probability in \eqref{defi_rums}, whereas it is common to assume that the noise terms are i.i.d.
%	Here, different assumptions on  $\prob^*$ and 
%	
	
	In general, the right-hand side of \eqref{defi_rums} has no closed analytical form and even worse, might be computationally costly. 
	Nevertheless, RUMs still provide a convenient way to facilitate theoretical considerations, as there is a natural ordering of the arms due to the underlying latent utility structure on the one side (cf.\ Section \ref{subsec_utility_dueling_bandits}), and on the other side, RUMs offer a quite intuitive explanation of the feedback generation: The individual (or nature) first assigns each available choice alternative (arm) a noisy utility $(u_i)_{a_i \in A_t}$, then sorts the available choice alternatives according to their noisy utilities, and finally chooses the one with the highest noisy utility.

	Here, the use of noise terms to perturb the actual utilities is reasonable, because external effects such as information asymmetry or impreciseness in the choice mechanism might occur, which can lead to a biased perception of the actual utility of each choice alternative.
	In other words, the individual (or nature) may not act perfectly according to the latent utilities $\theta_1,\ldots,\theta_K$ of the choice alternatives.
%	, but instead deviates from the 

	Some popular special cases of a RUM include the following:
	\begin{itemize}
		\item The \emph{multinomial logit (MNL) model}, where $\zeta_1,\ldots,\zeta_K$ is an i.i.d.\ sample of a standard Gumbel distribution. Quite interestingly, the MNL model admits a closed analytical form for \eqref{defi_rums} given by 
%		eq_bradley_terry
		\begin{align} \label{eq_mnl_model}
		%	\prob(a_i \, | \, A_t)
				q_{i| I(A_t)} 	  = \frac{\exp(\theta_i)}{ \sum_{a_j \in A_t} \exp(\theta_j)  } \enspace .
		\end{align}
		\item The \emph{multinomial probit (MNP) model}, where $\zeta_1,\ldots,\zeta_K$ is an i.i.d.\ sample of a standard Gaussian distribution. In contrast to the MNL model, there is in general no closed analytical form for \eqref{defi_rums} for the MNP model.
	\end{itemize} 

	Another natural way to come up with a reasonable family of categorical distributions is by assuming a probability distribution $\prob:\, \mathbb{S}_K \rightarrow [0,1]$ such as in Section \ref{sec:mall}.
	The probability of the most preferred arm in \eqref{defi_winner_feedback} can be obtained in the same spirit as in \eqref{eq:pairwisexy} by summing over all rankings $\pi \in \mathbb{S}_K $ in which $a_i$ precedes all arms in $A_t \backslash \{a_i\}:$
	\begin{align} \label{eq:marginals}
%		\prob(a_{i} \succ A_t \backslash \{a_{i} \}) 
% \prob(a_i \, | \, A_t) 
		q_{i| I(A_t)}  = \sum_{ \pi \in \mathbb{S}_K \,:\, \pi(j) > \pi(i) \, ,  \, \forall j \in I(A_t) \backslash \{i\} } \prob (\pi) \enspace ,
	\end{align} 
	where $\pi(j)$ is the rank of $a_j \in A_t$  with respect to $\pi$ (smaller ranks indicate higher preference).
	It is worth noting that the Plackett-Luce (PL) model (cf.\ Section \ref{subsec_pl_model}) coincides with the MNL model in this regard, i.e., if the score parameter $\theta \in \R_+^K$ of the PL model is set to $(\exp(\theta_1),\ldots, \exp(\theta_K) )^\top ,$ then the marginals in \eqref{eq:marginals} of the PL model are equivalent to the right-hand side of \eqref{eq_mnl_model}.

	\subsubsection{The $l'$ Most Preferred Arms}
	Another practically relevant type of feedback is an ordered list of $l' \in \{ 1, \ldots , |A_t| \}$ arms involved in the action at time $t.$ 
	Here, it is reasonable to consider the action space $\bA_{l}$ with $l\geq l',$ or to allow that $l'$ can vary with each iteration if the action space is $\bA_{\leq l}$ for instance.
	Formally, a partial ranking $\pi \in \mathbb{S}_{K|I(A_t)}$ of the form 
	$$ a_{\pi^{-1}(1)} \succ a_{\pi^{-1}(2)} \succ \ldots \succ a_{\pi^{-1}(l')} \succ A_t\backslash\{    a_{\pi^{-1}(1)}, a_{\pi^{-1}(2)},\ldots, a_{\pi^{-1}(l')}   \}    $$ 
	is observed as the feedback, where $\mathbb{S}_{K|I(A_t)}$ is the set of all permutations restricted to the set $I(A_t)$\footnote{The index set $I(A_t)$ corresponding to the arms in $A_t$ has to be used, as $\mathbb{S}_{K}$ has been defined as the set of all permutations on $[K] \subset \N$.} and $\pi^{-1}(i)$ is the index of the arm having the $i$th best rank with respect to $\pi.$

	The two probabilistic approaches for modeling the feedback generation in the previous case can here be used in a similar way: 
	\begin{itemize}
		\item \emph{Probabilistic ranking model ---} 	Assuming an underlying probabilistic model for $\prob$ on $\mathbb{S}_{K},$ the probability of obtaining $\pi \in \mathbb{S}_{K|I(A_t)}$  is given by 
		\begin{align*} 
			%		\label{eq:marginals_partial}
			%	
%			\prob(  \pi | A_t) = 
			\sum_{ \tilde \pi \in \mathbb{S}_K(\pi,I(A_t))} \prob (\tilde \pi) \enspace ,
		\end{align*} 
		where 
		\begin{align*}
			\mathbb{S}_K(\pi,I(A_t)) = \Big\{  \tilde \pi \in \mathbb{S}_K  \, \big\vert \,   \forall i & \in\{2,\ldots,l'\}:  \, \tilde \pi (\pi^{-1}(i)) > \tilde \pi (\pi^{-1}(i-1)),  \, \\
			&  \forall j \in  I(A_t)  \backslash\{ \pi^{-1}(1),\ldots,\pi^{-1}(l') \}: \, \tilde \pi(j) > \tilde \pi (\pi^{-1}(l'))  \Big\}
		\end{align*}
		is the set of all rankings respecting the order of the arms in $A_t$ according to $\pi$ (smaller ranks indicate higher preference).
%		
%		Here, $ \tilde \pi_{|  \pi^{-1}([l']) }$ is the restriction of $\tilde \pi$ to the (index) set of the arms ranked by $\pi.$
		%
%		
		\item \emph{Discrete choice model ---} 	By making the assumption of an underlying discrete choice model $(q_{i|I(A)})_{A\in \bA, i\in I(A)}$, the probability of observing the partial ranking $\pi \in \mathbb{S}_{K|I(A_t)}$ is equal to
		\begin{align*}
			%		
%			\prob(  \pi  |A_t) = 
			\prod_{i=1}^{l'} 	q_{\pi^{-1}(i)| I(A_t)  \backslash\{ \pi^{-1}(1),\ldots,\pi^{-1}(i-1) \}   }  \enspace .
			%		\prob(  a_{\pi^{-1}(1)} | A_t ) \cdot \prob(  a_{\pi^{-1}(2)} | A_t \backslash \{ a_{\pi^{-1}(1)} \} ) \cdot \ldots \cdot \prob(  a_{\pi^{-1}(l')} | A_t \backslash\{ a_{\pi^{-1}(1)},\ldots,a_{\pi^{-1}(l'-1)} \} ).
			%		
		\end{align*}
		In words, the latter probability is equivalent to the process, in which iteratively the most preferred arm among all ``remaining'' arms is (stochastically) determined, where in each iteration, the set of remaining arms is reduced by removing the most preferred arm of the previous iteration.
		Note that this is similar to the stagewise generation process underlying the PL model (cf.\ Section \ref{subsec_pl_model}).

	\end{itemize}

	Apparently, if $l'= 1$, then the feedback about the $l'$ most preferred arms and the feedback about the most preferred arm coincide. Consequently, for $l'>1$, the former type of feedback  is more informative than the latter.
	However, in general it is not possible to compare the information content of the feedback about the $l'$ most preferred arms and the feedback in the form of all pairwise preferences even if $l'= |A_t|.$
	Although it is true that a partial ranking occurring in the latter case can be transformed into a sequence of all pairwise preferences via the technique of \emph{rank-breaking} \citep{azari2013generalized}, it is an open question  whether this technique  introduces a bias for the pairwise preference estimates\footnote{For the PL model it is known that no bias occurs (cf.\ Section \ref{subsec_toparm_PL_multiwise}) due to its independence from irrelevant alternatives (IIA) property \citep{AlvoLu14}.}.
	The other way around, a (noisy) sequence of all pairwise preferences might not be aggregated into a consensus ranking of all arms involved in $A_t.$ 
	\subsubsection{Partial Preferences} \label{subsec_feedback_partial_preference}
	Finally, the most general type of feedback one may think of is certainly that of a sequence of partial preferences of the arms involved in the learner's action at time $t,$ which opposed to the feedback scenario involving the $l'$ most preferred arms can now be present in any form, i.e., a sequence of $k$-wise preferences with $k\in \{ 1, \ldots , |A_t| \}$ or combinations of the latter for different values of $k$.
	Once again, an underlying probabilistic model for $\prob$ on $\mathbb{S}_{K}$ or a discrete choice model can be leveraged in order to specify the probability of observing such a partial ranking.

	\subsection{Learning Tasks}
	
	The learning tasks considered in multi-dueling bandits are mostly the same as specified in Section \ref{subsec_targets} for dueling bandits, but in contrast to the latter additionally give rise to the task of finding the \emph{best subset of arms}, which can be seen as a generalization of the task of finding the best arm or finding the top-$k$ arms. 
	Similarly as in the dueling bandits setting, it is oftentimes not obvious how to define a reasonable notion of a best arm, let alone  an optimal or best subset of arms in the multi-dueling bandits variant.
	This issue is once again mainly due to the different types of feedback in the multi-dueling bandits variant, which may suggest different notions of optimality.
	In the following, we will only discuss the notions of best arm, best subset of arms, and reasonable target rankings.

	\subsubsection{Best Arm and Ranking of Arms}
	Clearly, if the underlying feedback mechanism allows one to make inference about the underlying preference relation $\bQ$, then the most natural way to define the best arm/the target ranking is to consider the Condorcet winner/total order of the arms, or if its existence is doubtful to opt for alternative notions of a best arm/target ranking as specified in Section \ref{subsec_alternative_targets}.
	This is most obviously the case in the feedback scenario, in which all pairwise preferences are observed, but can also be the case for other types of feedback as proposed by \citet{saha2018battle} (cf.\ Section \ref{subsec_battling_bandits}).
	However, it might be the case that the underlying feedback mechanism is impractical to make inference about the preference relation $\bQ.$ 
	Then, a workaround to establish the notion of a best arm or a reasonable target ranking can be derived in a similar manner as in Section \ref{subsec_utility_dueling_bandits} under the assumption of an underlying discrete choice model, or as in Section \ref{sec:mall} under the assumption of a probabilistic model $\prob$ on $\mathbb{S}_{K}.$
	
	Another reasonable possibility to answer the question of optimality of an arm exists if the feedback mechanism allows inference about the most preferred arm in a subset. 
	In such cases, one can generalize the concepts of a Condorcet winner or (some of) the alternative notions of best arm in the spirit of \citet{agarwal2020choice}:
	\begin{itemize}
		\item \emph{Generalized Condorcet winner ---} 
%		$\prob(a_{i^*} \, | \, A) > \prob(a_j \, | \, A)  $ 
		There exists an arm $a_{i^*} \in \cA$ such that $ q_{i^*| I(A) } >  q_{j| I(A) }  $ for any $A\in \bA$ with $a_{i^*},a_{j}\in A.$ 
		Here, $I(A)\subseteq [K]$ corresponds to the index set of the arms in $A\in \bA.$
%		and $a_j \in A\backslash \{a_{i^*}\}.$
%		
		In words, for any subset it is contained in, the arm $a_{i^*}$ has the highest probability to be the most preferred arm compared to all other arms in that subset. 
		\item \emph{Generalized Copeland winner ---} Define the generalized (normalized) Copeland scores of an arm $a_i\in \cA$ by means of
		$$d_i^{\text{gen}} \defeq \frac{1}{|\bA(i)|} \sum_{A \in \bA(i) } \IND{ \forall a_j \in A\backslash\{a_i\} : \,  q_{i|I(A)} >    q_{j|I(A)} }\enspace ,$$
%		$$d_i^{\text{gen}} \defeq \frac{1}{|\bA(i)|} \sum_{A \in \bA(i) } \IND{\prob(a_i \, | \, A) = \max_{a \in A} \prob(a \, | \, A)  }\enspace ,$$
%		
		where $\bA(i)=\{A \in \bA: a_i \in A \} $ is the subfamily of sets in $\bA,$ which contain $a_i.$
		Each arm with highest generalized Copeland score is called a generalized Copeland winner.
		\item \emph{Generalized Borda winner ---} Define the generalized sum of expectations or generalized Borda scores by 
		\begin{align*}
%			\label{eq:sscore}
%			
%q_i^{\text{gen}} = \frac{1}{ |\bA(i)| }\sum_{A \in \bA(i) } \prob(a_i \, | \, A)  \enspace .
			q_i^{\text{gen}} = \frac{1}{ |\bA(i)| }\sum_{A \in \bA(i) } q_{i|I(A)} \enspace .
		\end{align*} 
		Each arm with highest generalized Borda score is called a generalized Borda winner.
	\end{itemize}
	It is easy to check that if the action space corresponds to the action space of the basic dueling bandits setting (i.e.,  $\bA_{\leq 2}^+$), all these generalized notions coincide with the corresponding dueling bandits variant, because $q_{i|I(A)} = q_{i,j}$ if $A=\{a_i,a_j\}.$
	However, the major drawback of these generalized concepts is the combinatorial challenge involved in checking the necessary conditions, respectively. 
	Moreover, the drawbacks of the underlying basic variants are still preserved in the sense that the generalized Condorcet winner may not exist and the generalized Copeland/Borda winner may not be unique, but if the generalized Condorcet winner exists, then there is only one generalized Copeland winner, namely the generalized Condorcet winner itself\footnote{This holds for all action spaces listed in Section \ref{subsec_learning_protocol_mulit_duel}. \label{footnote_Copeland}}.
	Nevertheless, the generalized Borda winner can still differ from the generalized Condorcet winner in case it exists, and may still not be unique.

	Finally, it is straightforward to define the notions of a generalized Copeland/Borda ranking using the generalized Copeland/Borda scores similarly as in Section \ref{subsec_alternative_targets}. 
	The natural ranking in the spirit of Section \ref{subsec_target_rankings} is to consider $a_i$ to be ranked lower than $a_j$ if $q_{i|I(A)} > q_{j|I(A)}  $ for any $A\in \bA$ with $a_{i},a_{j}\in A.$
	Once again, such a ranking may not exist in general, while the generalized Copeland/Borda ranking always exists and agrees/might disagree with the former if it exists\footnoteref{footnote_Copeland}.
	Assuming a RUM for the underlying feedback mechanism, one obtains a natural ranking by sorting the latent utilities, while under the assumption of an underlying probabilistic model for $\prob$ on $\mathbb{S}_K,$ the target ranking can be defined as in Section \ref{sec:mall} via the mode of the distribution.

	\subsubsection{Best Subset of Arms}
	In the case of $\bA=\bA_{l}$, it is of special interest to define a notion of a best subset of arms. In fact, in this scenario it is quite natural to recommend an optimal action $\bA_{l}$, which in turn is exactly a subset in $\cA$ of size $l.$
	For this purpose, the notions of reasonable rankings in the previous section can be used in a straightforward way to establish the notion of an optimal subset of arms by setting it equal to the top-$l$ arms according to the corresponding ranking.

	However, under the assumption of a RUM with latent utilities $\theta=(\theta_1,\ldots,\theta_K)^\top \in \R_+^K$ for the feedback mechanism,  it is also sensible to define the expected utility of an action $A \in \bA$ by
	\begin{align} \label{defi_expected_utility}
		U_\theta(A) = \sum_{a_i \in A} \theta_i \cdot  q_{i|I(A)} \, ,
	\end{align}
	where the occurring probabilities above depend on $\theta$ as well, see \eqref{defi_rums}.
	In this way, one may also define the best subset(s) of arms by $A^* \in \argmax_{ A \in \bA } U_\theta(A),$ i.e., the action(s) with the maximum achievable expected utility.
	It is worth noting that this best subset is not necessarily composed of the arms with highest utility (cf.\ Section \ref{subsec_preselection}).

	\subsection{Performance Measures} \label{subsec_performance_measure_multi}
	Quite unsurprisingly, the performance measures of interest are essentially the same as in Section \ref{subsec_preformance_measure}, i.e., sample complexity for ($\epsilon,\delta$)-PAC learning scenarios and cumulative regret for regret minimization tasks.
	However, the generality of the multi-dueling bandits allows one to define a reasonable regret per time in an even greater variety.
	By now, three variants  have been considered in the literature for defining the regret per time. These variants shall be reviewed in the following, where we again denote by $A_t\in \bA$ the action used in time $t.$
	\begin{itemize}
		\item \emph{Generalized averaged regret ---} If the feedback mechanism allows inference about the pairwise preferences, then this regret is the straightforward extension of the average regret used in (\ref{eq:regret}), i.e.,
		\begin{align*}
			r_{t,avg}^{\text{gen}} \defeq  \frac{ \sum_{a_j\in {A_t}} \Delta_{i^{*},j} }{|A_t|} \enspace,
		\end{align*}
		where $i^*$ is the target arm, e.g., the Condorcet winner assuming it exists.
		Under the assumption of an underlying RUM with latent utilities $\theta=(\theta_1,\ldots,\theta_K)^\top \in \R_+^K$ for the feedback mechanism, another variant of the generalized averaged regret can be defined quite naturally based on the latent utilities:
		\begin{align*}
			r_{t,avg(\theta)}^{\text{gen}} \defeq  \frac{ \sum_{a_j\in {A_t}} (\theta_{i^*} - \theta_j) }{|A_t|} \enspace,
		\end{align*}
		where $i^*$ is the index of the arm with the highest utility.
		The notions of weak and strong regret per time can be generalized in a straightforward way.
		Similarly as in the dueling bandits case, the generalized averaged regret is only zero if the action in time $t$ is the singleton set $\{a_{i^{*}}\}$ corresponding to a full commitment to this arm.
%		 $a_{i(t)}$ and $a_{j(t)}$  which is due to the (arti-)fact that the average regret  are exactly the target arm .
%		
		\item \emph{Regret by shortfall of preference probabilities  ---} 
		If the feedback mechanism allows inference about the probability that an arm is the most preferred among a subset of arms, then one can consider the cumulated shortfall of the probabilities of all arms in $A_t$ to be the most preferred one if the target arm is also present.
		Formally, 
		\begin{align*}
%			%	
%r_{t,sh} \defeq    \sum_{a_j\in {A_t}} \prob(a_{i^*}  |  A_t \cup \{  a_{i^*}\}  )    - \prob(a_{j}  |  A_t \cup \{  a_{i^*}\}  )  \enspace,
%%	
			%	
			r_{t,sh} \defeq    
			\sum_{a_j\in {A_t}} q_{i^*   |  A_t \cup \{  a_{i^*}\}  }   - q_{j   |  A_t \cup \{  a_{i^*}\}  }  \enspace,
		\end{align*}
		where $i^*$ is the target arm, e.g., the (generalized) Condorcet winner assuming it exists.
		\item \emph{Expected-utility regret ---}
		Recalling the expected utility of an action $A_t$ in \eqref{defi_expected_utility}, the expected-utility regret in time $t$ is then defined as
		$$r_{t,\theta} = \max_{A \in \bA} \, U_\theta(A)  - U_\theta(A_t). $$
		In words, the difference between the maximum achievable expected utility and the expected utility of the used action.
	\end{itemize}
	Finally, it is worth noting that although the feedback in the form of the most preferred arm in the multi-dueling bandits is seemingly more informative than the one in the dueling bandits, there are a couple of ``negative'' results in the sense that this seemingly larger amount of information does not necessarily lead to a theoretical improvement (in order sense) of the underlying performance measure for the same target.
	For instance, \citet{ren2019sample} show that if $\bA=\bA_{l}$ and the feedback is the most preferred arm for each action, then the lower bound on the sample complexity of every algorithm to return the actual total order of the arms (assuming its existence) with confidence $1-\delta$ is the same as for the dueling bandits problem (cf.\ Section \ref{sec_iterative_insertion_ranking}).
	Other results in this spirit will be discussed in the following section.
%	are shown by \citet{saha2018battle} (Section \ref{subsec_battling_bandits}), \citet{saha2019pac} (Section \ref{}) .
%	 
	However, if the feedback corresponds to the $l'$ most preferred arms, the theoretical guarantees are indeed improving with the value of $l'.$
	
	\subsection{Algorithmic Approaches}
%	
%	In spite of the recency of the even more restricted learning scenario underlying the basic PB-MAB problem, there have been quite  many works dealing with the general PB-MAB problem, in which the metaphorically \emph{duel} between two used choice alternatives (arms) is brought to another level.
%	
	In the following, we provide an overview of existing methods for the generalized PB-MAB problem. For reasons of clarity, we summarize all approaches in Tables \ref{tab:multi_dueling_algo} and \ref{tab:multi_dueling_algo2}, maintaining the conceptional structure used in the previous sections.

	\begin{table}
		
		\scriptsize
		\begin{center}		
			\begin{tabular}{|p{2cm}|p{1.75cm}|p{3cm}|p{2.5cm}|p{3.5cm}|}
				\hline
				\textbf{Algorithm} & \textbf{Algorithm class} & \textbf{Action space, feedback and further assumptions}  & \textbf{Target(s) and goal(s) of learner} & \textbf{Theoretical guarantee(s)} \\
				\hline
				\hline
				Multileaving \newline  (Section \ref{subsec_multileaving})
				& Online optimization-based			
				&  $\bA_{l};$ Partial rankings; No further explicit assumptions
				& Satisfactory experimental performance
				& No theoretical guarantees \\
				\hline
				Multi-Dueling Bandit \newline  (Section \ref{subsec_multidueling})
				%				\citep{YuJo09} 
				& Generalization (UCB)				
				&  $\bA_{\text{full}};$ All pairwise preferences; Existence of a Condorcet winner
				& Expected regret minimization with generalized average regret
				& No theoretical guarantees \\
				\hline
				%			
%				Multileave Gradient Descent \newline  (Section \ref{subsec_multileaving})
%				%				\citep{Ku17} 
%				& Online optimization based
%				& $\bA_{l};$ Partial rankings; No further explicit assumptions
%				%			\& $\alpha$-strong concavity of utility functions, differentiability assumption \& shape-constraints on link function
%				& Satisfactory experimental performance for NDCG
%				& No theoretical guarantees
%				%				
%				\\
%				\hline
%				Counterfactual Dueling Bandits \newline  (Section \ref{subsec_counterfactual})
%				%				\citep{Ku17} 
%				& Online optimization based
%				& $\bA_{l};$ Partial rankings; No further explicit assumptions
%				%			\& $\alpha$-strong concavity of utility functions, differentiability assumption \& shape-constraints on link function
%				& Satisfactory experimental performance for NDCG
%				& No theoretical guarantees
%				\\
%				\hline
				%	
				SelfSparring \newline  (Section \ref{subsec_selfsparring})
				%				\citep{Ku17} 
				& Reduction-based 
				& $\bA_{\leq l}$; Partial rankings; Total order of arms and  approximate linearity 
				& Expected regret minimization with generalized average regret
				& Asymptotically no-regret
				\\
				\hline
				Battling-Doubler \newline (Section \ref{subsec_battling_bandits})
				%				\citep{Ku17} 
				& Reduction-based 
				& $\bA_{\leq l};$ Most preferred arm; Linear pairwise subset choice model
				%			& 
				%			\& $\alpha$-strong concavity of utility functions, differentiability assumption \& shape-constraints on link function
				&  Expected regret minimization with generalized average regret
				& Finite arms: \newline { $\bigO \left( \frac{l K  \log^2 T }{\min_{j \neq i^{*}} \Delta_{i^{*},j}  }\right) $ }
				resp.\
				{ $\bigO \left(  l \sqrt{K \, T \, \log^3(T)} \right) $ }
				\newline
				Infinite arms: \newline { $\bigO \left( \frac{l d^2  \log^4 T }{\min_{a \neq a_{*}} \Delta_{a_{*},a}  }\right) $} \ \ $\quad$ 
				%				\newline 
				\\
				\hline
				Battling-MulitSBM \newline  (Section \ref{subsec_battling_bandits})
				%				\citep{Ku17} 
				& Reduction-based 
				& $\bA_{\leq l};$ Most preferred arm; Linear pairwise subset choice model
				%			\& $\alpha$-strong concavity of utility functions, differentiability assumption \& shape-constraints on link function
				& Expected regret minimization with generalized average regret
				&  $\hphantom{text}$ \newline  {  $\bigO \left( \frac{l K  \log T }{\min_{j \neq i^{*}} \Delta_{i^{*},j}  }\right) $ }
				\\
				\hline
				Battling-Duel \newline  (Section \ref{subsec_battling_bandits})
				%				\citep{Ku17} 
				& Reduction-based 
				& $\bA_{\leq 2}^+;$ Most preferred arm; Pairwise subset choice model
				%			\& $\alpha$-strong concavity of utility functions, differentiability assumption \& shape-constraints on link function
				& Expected regret minimization with generalized average regret
				& Same as the used black-box dueling bandit algorithm
				\\
				\hline
				MaxMin-UCB \newline  (Section \ref{subsec_maxminUCB})
				%				\citep{Ku17} 
				& Generalization (UCB)
				& $\bA_{\leq l} \, \& \, \bA_{l'};$ $l'$ most preferred arms; Underlying MNL model
				%			\& $\alpha$-strong concavity of utility functions, differentiability assumption \& shape-constraints on link function
				& Regret minimization with \newline
				1.\ Gen.\ avg.\ regret 
				2.\ Gen.\ average top-$l'$ regret
				& $\hphantom{text}$ \newline $\hphantom{text}$ \newline
				1.\ See \eqref{regret_bound_maxminucb} \newline
				2.\ See \eqref{regret_bound_recmaxmin}
				\\
				\hline		
				Thresholding-Random-
				Confidence-Bound \newline  (Section \ref{subsec_preselection})
				%				\citep{Ku17} 
				& Generalization (UCB)
				& $\bA_{l}$;  Most preferred arm; Underlying MNL model
				%			\& $\alpha$-strong concavity of utility functions, differentiability assumption \& shape-constraints on link function
				&  Expected regret minimization with expected-utility regret
				&  $\hphantom{text}$ \newline $\bigO(\sqrt{K T \log(T)})$
				\\
				\hline
				Confidence-Bound-Racing \newline  (Section \ref{subsec_preselection})
				%				\citep{Ku17} 
				& Generalization (Racing)
				& $\bA_{\text{full}};$ Most preferred arm; Underlying MNL model
				%			\& $\alpha$-strong concavity of utility functions, differentiability assumption \& shape-constraints on link function
				& Expected regret minimization with expected-utility regret 
				& $\bigO( \sum_{i \neq i^*} \frac{\log(T)}{(\theta_{i^*}-\theta_i)^2} )$
				\\
				\hline
				Winner Beats All \newline (Section \ref{subsec_choice_bandits})
				%				\citep{Ku17} 
				& Challenge
				& $\bA_{\leq l}^+;$ Most preferred arm; Existence of a generalized Condorcet winner
				%			\& $\alpha$-strong concavity of utility functions, differentiability assumption \& shape-constraints on link function
				& Expected regret minimization with regret by shortfall
				&  {\tiny $ \bigO\left(		\frac{K \log(T) }{\Delta_{\text{GCW}}^2}		\right)$ } \newline
				MNL Model: {\tiny $ \bigO\left(		\frac{K \log(T) }{\theta_{i^*}-\theta_i}		\right)$}
				\\
				\hline
			\end{tabular}
		\end{center}
				\caption{Approaches for the multi-dueling bandits problem for regret minimization tasks. }
				\label{tab:multi_dueling_algo}
		%	}
	\end{table}
\begin{table}
	\scriptsize
	\begin{center}		
		\begin{tabular}{|p{2.2cm}|p{1.75cm}|p{2.8cm}|p{2.5cm}|p{3.5cm}|}
			\hline
			\textbf{Algorithm} & \textbf{Algorithm class} & \textbf{Action space, feedback and further assumptions}  & \textbf{Target(s) and goal(s) of learner} & \textbf{Theoretical guarantee(s)} \\
			\hline
			\hline
			Trace-the-best \newline  (Section \ref{subsec_toparm_PL_multiwise})
			%				\citep{Ku17} 
			& Challenge
			& $\bA_{l};$ $l'$ most preferred arms;  MNL model
			%			\& $\alpha$-strong concavity of utility functions, differentiability assumption \& shape-constraints on link function
			&  ($\epsilon,\delta$)-PAC setting for finding best arm
			& $\hphantom{text}$  \newline  
			$\bigO\left( \frac{K}{l' \, \epsilon^2} \log \frac{K}{\delta}  \right)$
			\\
			\hline
			Divide-and-Battle \newline  (Section \ref{subsec_toparm_PL_multiwise})
			%				\citep{Ku17} 
			& Tournament
			& $\bA_{l};$  $l'$ most preferred arms; MNL model
			&  ($\epsilon,\delta$)-PAC setting for finding best arm
			& 
			$\hphantom{text}$  \newline  $\bigO\left( \frac{K}{l' \, \epsilon^2} \log \frac{l}{\delta}  \right)$ or \eqref{sample_complex_divide_and_battle}
			\\
			\hline
			Halving-Battle \newline  (Section \ref{subsec_toparm_PL_multiwise})
			%				\citep{Ku17} 
			& Tournament, Generalization (Successive-Elimination)
			& $\bA_{\leq l};$ Most preferred arm; MNL model
			%			\& $\alpha$-strong concavity of utility functions, differentiability assumption \& shape-constraints on link function
			&  ($\epsilon,\delta$)-PAC setting for finding best arm
			& $\hphantom{text}$  \newline $\bigO\left( \frac{K}{\epsilon^2} \log \frac{1}{\delta}  \right)$ 
			\\
			\hline
			PAC-Wrapper \newline  (Section \ref{subsec_toparm_PL_multiwise})
			%				\citep{Ku17} 
			& Tournament
			& $\bA_{l};$  $l'$ most preferred arms; MNL model
			&  ($\epsilon,\delta$)-PAC setting for finding best arm
			& $\hphantom{text}$  \newline  See \eqref{sample_complex_pac_wrapper}
			\\
			\hline
			Uniform-Allocation \newline  (Section \ref{subsec_toparm_PL_multiwise})
			%				\citep{Ku17} 
			& Tournament, Generalization (Successive-Elimination)
			& $\bA_{l};$  $l'$ most preferred arms; MNL model
			& Small error of probability for finding best arm with fixed budget $B$ for total number of actions
			& $\hphantom{text}$  \newline  $\bigO \left( \log n \exp \left(	 \frac{l' B \Delta_{min}(\theta) }{K+l\log(l)}	\right)	\right)$\newline\newline
			{\tiny $\Delta_{min}(\theta)=\min_{i\neq i^*} \theta_{i^*} - \theta_i$}
			\\
			\hline
			Sequential-Pairwise-Battle \newline  (Section \ref{subsec_toparm_RUMS_multiwise})
			& Tournament
			& $\bA_{l};$ $l'$ most preferred arms; IID-RUM
			%			\& $\alpha$-strong concavity of utility functions, differentiability assumption \& shape-constraints on link function
			&  ($\epsilon,\delta$)-PAC setting for finding ranking of arms
			&  $\hphantom{text}$  \newline 
			 $\bigO \left( \frac{K}{l' \, C_{\text{RUM}}^2 \, \epsilon^2} \log \frac{l}{\delta} \right)$
			%			2.\ $\bigO\left( \frac{K}{l' \, \epsilon^2} \log \frac{l}{\delta}  \right)$ Sequential-Pairwise-Battle
			\\
			\hline
			Beat-the-Pivot \newline  (Section \ref{subsec_ranking_PL_multiwise})
			%				\citep{Ku17} 
			& Challenge
			& $\bA_{l};$ $l'$ most preferred arms;  MNL model
			%			\& $\alpha$-strong concavity of utility functions, differentiability assumption \& shape-constraints on link function
			&  ($\epsilon,\delta$)-PAC setting for finding ranking of arms
			& $\hphantom{text}$  \newline  $\bigO\left( \frac{K}{l' \, \epsilon^2} \log \frac{K}{\delta}  \right)$
			\\
			\hline
			Score-and-Rank \newline  (Section \ref{subsec_ranking_PL_multiwise})
			%				\citep{Ku17} 
			& Challenge
			& $\bA_{l};$ Most preferred arm; MNL model
			%			\& $\alpha$-strong concavity of utility functions, differentiability assumption \& shape-constraints on link function
			&  ($\epsilon,\delta$)-PAC setting for finding ranking of arms
			&  $\hphantom{text}$  \newline $\bigO \left( \frac{K}{\epsilon^2} \log \frac{K}{\delta} \right)$
%			2.\ $\bigO\left( \frac{K}{l' \, \epsilon^2} \log \frac{l}{\delta}  \right)$ 
			\\
			\hline
			AlgPairwise \&  
			AlgMulti-wise \newline 
			(Section \ref{subsec_topk_PL_multiwise})
			%				\citep{Ku17} 
			& Tournament
			& $\bA_{\leq l};$ Most preferred arm; MNL model
			%			\& $\alpha$-strong concavity of utility functions, differentiability assumption \& shape-constraints on link function
			&  Exact sample complexity for finding top-$k$ ranking
			& $\hphantom{text}$  \newline   See \eqref{sample_complex_alg_pairwise}
			\\
			\hline
		\end{tabular}		
	\caption{Approaches for the multi-dueling bandits problem for $(\epsilon,\delta)$-PAC learning. }
	\label{tab:multi_dueling_algo2}
	\end{center}
	%	}
\end{table}

	\subsubsection{Multileaving} \label{subsec_multileaving}
	
	In the general scenario of the online learning-to-rank problem, it is assumed that there exists some set of objects such as web pages and a (possibly infinite) pool of possible ranking models (or rankers) producing a ranked list of the objects, which can be displayed to users.
	The goal is to find the best ranking model by learning from preference feedback produced by the users.
	The online learning-to-rank problem can be cast quite naturally as a PB-MAB problem by considering the ranking models as arms.
	The pairwise preference feedback of two arms as in the dueling bandits problem is then usually the result of an \emph{interleaving} method \citep{RaKuJo08}, which allows one to infer a preference between the two ranking models.
	Roughly speaking, this is achieved by interleaving the suggested lists of objects, presenting them to a user, and monitoring which object is chosen by a user.
	As a matter of fact, the first work on dueling bandits by \cite{YuJo09} is also strongly motivated by online learning-to-rank problems.

	The multi-dueling bandits counterparts of the interleaving methods are the \emph{multileaving} methods \citep{schuth2014multileaved,schuth2015probabilistic}, which extend interleaving methods by allowing to incorporate more than two ranking models at a time, and even more general feedback coming from the user. 
	Based on such multileaving methods, \citet{ScOoWhDe16} propose the Multileave Gradient Descent (\Algo{MGD}) algorithm, which extends \Algo{DBGD} (cf.\ Section \ref{subsec_DBGD}) by first using multiple random directions for the multiwise comparison with the current point, representing the current best ranker, and modifying the update of the current point accordingly to the more general user feedback. 
	To this end, two variants of \Algo{MGD} are suggested, which differ in the way the observed preferences are used for the update: \Algo{MGD} winner takes all (\Algo{MGD-W}) and \Algo{MGD} mean winner (\Algo{MGD-M}).
	In \Algo{MGD-W}, the current point is updated by sampling uniformly at random one of the directions that were preferred over the current point, and move into this direction with a certain step size.
	This is basically the straightforward extension of \Algo{DBGD}.
	The \Algo{MGD-M} algorithm updates the current point by moving into the ``mean'' direction, i.e., the mean of all directions that were preferred over the current point, again with a pre-specified step size.
	\citet{OoScDe16} consider the \Algo{MGD-M} algorithm as well, but replace the used multileaving method by \citet{ScOoWhDe16}, team draft multileaving, by an extension of the probabilistic interleave method \citep{HoWhDe11}.

	Another attempt to improve the practical performance of the \Algo{DBGD} algorithm by means of multileaving methods is suggested by \citet{ZhKi16}.
	They construct the Dual-Point Dueling Bandit Gradient Descent (\Algo{DP-DBGD}) algorithm and the Multi-Point Deterministic Gradient Descent (\Algo{MP-DGD}) algorithm, which, similar to \Algo{MGD}, both take multiple directions within one time step into account, but are focused on reducing variances in the underlying gradient approximation of \Algo{DBGD}.
	For this purpose, \Algo{DP-DBGD} uses two opposite stochastic directions from the current point for the exploration, instead of only one as in \Algo{DBGD}, while exploration in \Algo{MP-DGD} is carried out by using all points lying on the directions determined by the (non-random) standard unit basis vectors with a specific distance from the current point.
	The update for \Algo{DP-DBGD} corresponds to the winner takes all strategy as in \Algo{MGD-W}, while \Algo{MP-DGD} updates the current point successively into the directions based on the standard unit basis vectors (with a certain step size), which were preferred over the current point.
	Both algorithms are combined with the proposed multileaving method \emph{Contextual Interleaving}, which takes previous results from the conducted explorations into account to compose the final interleaved list of objects.

	As all these algorithms are highly motivated from a practical perspective, they are thoroughly analyzed in various experiments on data sets from the field of information retrieval.
	Needless to say, all approaches perform superior compared to \Algo{DBGD} in terms of common metrics such as Normalized Discounted Cumulative Gain (NDCG).
 	However, theoretical analyses like for \Algo{DBGD} are not provided for these approaches. 

	Last nut not least, \cite{pereira2019online} propose the Counterfactual Dueling Bandits (\Algo{CDB}) algorithm inspired by the specific task of music recommendation, where the objects of interest correspond to songs, which can be recommended to a user.
	However, in their setting the user obtains at each time step only one recommended song instead of a multileaved list of songs, which results in a binary feedback, i.e., whether the recommended song is played until the end or whether it is skipped.
	Moreover, the recommended song is solely determined by the current point (current ranking model) and a set of candidate songs generated by some retrieval function, but the conceptional idea for updating the current point before the recommendation is made reveals some similarities to the multi-dueling approach as in \Algo{MGD}.
	More specifically, \Algo{CDB} compares the current point with multiple random directions (exploratory ranking models) based on rewards, which are estimated in a counterfactual manner by considering the rank assigned by the corresponding ranking model to the previously recommended song, and updates the current point successively into the directions (with a specific step size) having a higher estimated reward than the reward of the current point, which is based on the observed binary feedback.

	\subsubsection{Multi-Dueling} \label{subsec_multidueling}
	Also inspired by online learning-to-rank problems, \citet{BrSeCoLi16} consider the multi-dueling bandits problem with feedback in the form of all pairwise preferences (cf.\ Section \ref{subsec_all_pairwise_feedback}) and the possibility to choose all possible (non-empty) subsets of arms as actions for the learner.
	Under the assumption that the underlying preference relation $\bQ$ has a Condorcet winner, the multi-dueling bandits algorithm (\Algo{MDB}) is proposed, which, based on two types of upper confidence estimates for each pairwise preference probability, maintains a set of optimistic Condorcet winners in the spirit of \Algo{RUCB}, respectively.
	Here, one of the types is more conservative than the other by means of a multiplicative factor.
	In case the set of Condorcet winners based on the non-conservative upper confidence estimates is a singleton set, the \Algo{MDB} makes a full commitment to the corresponding arm, and otherwise composes the subset for its action by all arms within the set of Condorcet winners based on the conservative upper confidence estimates.
	The key idea underlying the usage of the conservative upper confidence estimates is to boost exploration, which in turn should lead to a quick exploitation by means of the non-conservative upper confidence estimates.

	It is shown that the algorithm performs well in empirical experiments, if the performance measure is the cumulative regret with the generalized averaged regret for the regret per time.
	However, the authors do not provide any theoretical guarantees for \Algo{MDB}.
	%	
	
	%	\citet{BrSeCoLi16} consider the problem of online ranker evaluation and address the question of which rankers to compare in each iteration. To this end, they generalize the $ K $-armed dueling bandits to the \emph{multi-dueling} bandit framework, which is based on simultaneous comparisons of more than two rankers through \emph{multileaving}, as opposed to comparisons of only pairs of rankers through interleaving. They assume the existence of a Condorcet winner and aim at selecting subsets of arms, so that the cumulative regret is minimized, where the regret of a set of arms is the average of the regrets of the individual arms with respect to the Condorcet winner. The authors then propose the multi-dueling bandits algorithm (\Algo{MDB}), which plays arms that, according to optimistic estimates of the pairwise winning probabilities, are most likely to be the Condorcet winner. More specifically, when only a single candidate remains, \Algo{MDB} plays only that candidate, and when there are multiple candidates, \Algo{MDB} explores by comparing all of them, together with additional arms obtained using wider upper confidence bounds on the probabilities to increase parallel exploration.
	
%	\subsubsection{Counterfactual Dueling Bandits} \label{subsec_counterfactual}

	\subsubsection{SelfSparring} \label{subsec_selfsparring}

	Allowing the learner to choose a subset of size at most $l$ as its action and assuming feedback in the form of some (though not necessarily all) pairwise preferences of the arms involved in the comparison, \citet{SuZhBuYu17} investigate the multi-dueling bandits problem in two settings: A finite arm scenario with an existing total order of arms and another scenario based on a Gaussian process modeling of the preference probabilities allowing an infinite number of arms.
	Inspired by the idea of the sparring approach of \citet{AiKaJo14} (cf.\ Section \ref{subsec_adversarial}), the authors propose the \Algo{SelfSparring} framework, in which $l$ value-based MAB algorithms are used to control the choice of each of the $l$ arms to be drawn in each iteration.
	For both scenarios, i.e., the utility-based and the one based on Gaussian process, an instantiation using $l$ many Thompson Sampling instances of \Algo{SelfSparring} is suggested:  \Algo{IndSelfSparring} and \Algo{KernelSelfSparring}.

	In \Algo{IndSelfSparring}, the underlying Thompson sampling algorithms use independent Beta prior and posterior distributions for the arm choice mechanism.
	Under the assumption of approximate linearity\footnote{This is a generalization of the linear utility-based setting of \cite{AiKaJo14}, where for any triplet of arms such that $ a_i\succ a_j\succ a_k $ and some constant $ \gamma>0 $, it holds that $\Delta_{i,j}-\Delta_{j,k}\geq \gamma \Delta_{i,k}$.} they show that the algorithm converges to the optimal arm with asymptotically optimal no-regret rate of $ \bigO(K\ln(T)/\Delta) $, where the regret is the generalized average regret, up to a constant, and $ \Delta $ is the calibrated pairwise preference between the best two arms.

	The key idea underlying the design of \Algo{KernelSelfSparring} is that  $u(a) = \prob(a\succ a^*),$ where $a^*$ is the best arm, corresponds to a sample of a Gaussian process $GP(\mu(a),k(a,a'))$ with unknown mean function $\mu:\cS \to [0,1]$ and some appropriate unknown covariance function $k:\cS \times \cS \to \R,$ which needs to be positive semidefinite.
	Due to the covariance function, it is possible to allow dependencies among the arms in contrast to the previous scenario considered for \Algo{IndSelfSparring}.
	On account of the Gaussian process modeling, the underlying Thompson sampling algorithms of \Algo{KernelSelfSparring} are using Gaussian process priors and posteriors for the arm choice mechanism.

	\subsubsection{Battling Bandits} \label{subsec_battling_bandits}

	Picking up the dueling bandits metaphor, \citet{saha2018battle} refer to the scenario of receiving feedback in the form of the most preferred arm in a multi-dueling bandits setting as the \emph{battling bandits}.
%	
%	As receiving winner feedback for  comparison involves more than two arms while revealing only a single winner, it is more natural to speak of a ``battle'' of the arms rather than a duel, whence the \citet{saha2018battle} refer to this variant as \emph{battling bandits}.]
	%
	%\citet{saha2018battle} introduce another variant of the dueling bandits setting with multiple arm comparisons, where the feedback consists of the information which of the compared arms is the most preferred one.
	% among this subset is obtained as a feedback. 
	%
	%
	%
	They introduce the \emph{pairwise-subset choice model}, which is a probabilistic choice model based on the preference relation $\bQ$ for the considered feedback mechanism.
	In particular, they assume that the probability that an arm $a_{i}$ is the most preferred among a subset of arms $ A \subset \cA$ with $|A|=l$ and $a_{i} \in A$ is 
	\begin{align} \label{eq_discrete_choice_probability}
		\prob(a_i \, | \, A) = \sum_{j\neq i: a_j \in A}  \frac{2 \, q_{i,j}}{l(l-1)} \, .	
	\end{align}
	They first propose two algorithms, \Algo{Battling-Doubler} and \Algo{Battling-MultiSBM}, which are generalizations of \Algo{Doubler} and \Algo{MultiSBM} for the dueling bandits case (cf.\ Section \ref{subsec_dobler_multisbm}).
	%
	%Battling-Doubler uses again only one single standard MAB algorithm  
	Further, a third algorithm, \Algo{Battling-Duel}, is suggested, using some dueling bandits algorithm as a black-box in the following way.
	In each round, the black-box dueling bandits algorithm suggests a pair of arms $(a_{i(t)},a_{j(t)})$ for the duel, which is then used to define the set $A_t$ for the battle by replicating $a_{i(t)}$ resp.\ $a_{j(t)}$ for $\lfloor l/2 \rfloor$ resp.\ $\lceil l/2 \rceil$ or  $\lceil l/2 \rceil$ resp.\ $\lfloor l/2 \rfloor$ times, each with equal probability.
	With this, the received feedback is either the information that $a_{i(t)}$ is preferred over $a_{j(t)}$ or vice versa, so that this feedback can be propagated to the black-box dueling bandits algorithm.
	As a consequence, the underlying action space of  \Algo{Battling-Duel} is in fact the same as in the dueling bandits problem, while \Algo{Battling-Doubler} and \Algo{Battling-MultiSBM} consider the action space $\bA_{\leq l}.$
	
	For the theoretical analysis, the authors assume the existence of a Condorcet winner for $\bQ$ and consider the generalized average regret.
	Under these assumptions and any probability model (\ref{eq_discrete_choice_probability}), they show that the  \Algo{Battling-Duel} algorithm has the same bound (up to a multiplicative constant) on its expected regret as the underlying black-box dueling bandits algorithm for the cumulative regret in \eqref{eq:regret}.
	More interestingly, they show that the lower bound under these assumptions is of the same order as in the dueling bandits case and does not decrease with the size of compared arms. 
	This confirms that, from a theoretical point of view, there is no advantage in comparing subsets of more than two arms for such pairwise-subset choice models if feedback is given in the form of the most preferred arm.
	
	For \Algo{Battling-Doubler} and \Algo{Battling-MultiSBM}, respectively, gap-dependent as well as gap-independent upper bounds on the expected regret are proved under the additional assumption that the pairwise probabilities are given by (\ref{eq_utility_based_pairwise}) for the linear link function. 
	However, all these bounds exhibit a leading multiplicative term $l$ (see Table \ref{tab:multi_dueling_algo}) and, consequently, are not optimal with respect to their dependency on $l.$

	\subsubsection{MaxMin-UCB} \label{subsec_maxminUCB}
	
	\cite{saha2019combinatorial} assume a PL model with utility parameter $\theta$ (cf.\ Section \ref{subsec_pl_model}) for the generation of the feedback in the form of the $l'$ most preferred arms if the action space is $\bA_{\leq l},$ where $l'$ is some non-negative integer strictly smaller than $l.$
	Under this assumption, the method of \emph{rank-breaking} \citep{soufiani2014computing}, in which a ranking is decomposed into all possible pairwise preference relations coherent with the ranking, does not introduce any bias in the empirical estimation of the pairwise preference probabilities.
	This is due to the independence from irrelevant alternatives (IIA) property of the PL model (cf.\ \cite{AlvoLu14}).
	They suggest the \Algo{MaxMin-UCB} algorithm, which is essentially a generalization of the \Algo{RUCB} algorithm (cf.\ Section \ref{subsec_RUCB}) using only subsets of sizes $l'+1$ or $1$ (full commitment to one arm) for its actions.
	More specifically, \Algo{MaxMin-UCB} maintains also a set of optimistic Condorcet winners\footnote{The Condorcet winner exists due to the PL model.} based on the upper confidence estimates of the pairwise preference probabilities extracted by the rank-breaking method and  composes the subset of size $l'+1$ by first choosing one potential optimistic Condorcet winner and then iteratively adding an arm to the subset until the subset has a size of $l'+1.$
	Here, the added arm in one iteration corresponds to the overall worst competitor regarding all previously added arms based on the upper confidence estimates.
	In case the set of optimistic Condorcet winners is a singleton set, the algorithm makes a full commitment to this arm, i.e., chooses this singleton set as its action subset.
	It is shown that \Algo{MaxMin-UCB} has a regret bound both with high probability and in expectation of order
	\begin{align} \label{regret_bound_maxminucb}
			\bigO\left(	\frac{ \log T}{l'} \max_{i\neq i^*}\frac{1}{\Delta^2_{i}(\theta)}	\sum_{i\neq i^{*}} \Delta_i(\theta)   	\right)	,	
	\end{align}
	if the regret per time is measured by the generalized average regret with latent utilities. 
	Here, $\Delta_i(\theta) = \theta_{i^*} - \theta_i$ denotes the difference between the utilities of the arm $a_{i^*}$ with the largest utility and arm $a_i.$
	To complement the theoretical analysis, an asymptotic lower bound of order
	$$		\Omega\left(  \frac{K \, \log(T)}{l' \, \min_{i\neq i^*} \, \Delta_{i}(\theta)} \right)	$$
	is shown.
	
	Under the same assumptions as above, but with the action space $\bA_{l'},$ the authors then investigate  the task of minimizing the (cumulative) generalized top-$l'$ regret, where the regret per time for using action $A_t$ is measured by means of
	$$	r_{t,avg(\theta)}^{l'}  \defeq  \frac{ \sum_{i=1}^{l'} \theta_{(i)} - \sum_{a_j\in {A_t}}  \theta_j }{|A_t|} \enspace, 	$$
	with $\theta_{(1)}\geq \theta_{(2)} \geq \ldots \geq \theta_{(K)}.$
	For this purpose, they suggest the \Algo{Rec-MaxMin-UCB} algorithm, which differs from \Algo{MaxMin-UCB} as it now chooses a subset of size $l'$ for its action, but also makes use of the iterative subset construction based on the upper confidence estimates in order to compose the subset of the allegedly top-$l'$ arms. 
	This algorithm has a generalized top-$l'$ regret bound both with high probability and in expectation of order
	\begin{align}\label{regret_bound_recmaxmin}
					\bigO\left(	\frac{ \log T}{l'} 	\sum_{i \geq l'+1}^K  \frac{\Delta_{(l'),(i)}(\theta)}{\min_{ j \in [l'-1 ] }  (q_{(l'),(j)} - q_{(i),(j)})^2  }	\right)	,
	\end{align}
	where $q_{(i),(j)}$ corresponds to the $(i,j)$th entry of the permuted preference relation matrix $\bQ$ such that the $k$th row corresponds to the pairwise preference probabilities of the arm with the $k$th largest utility.
	The authors also provide an asymptotic lower bound of order 
	$$		\Omega\left(  \frac{(K-l') \, \log(T)}{l' \,  (\theta_{(l')} -\theta_{(l'+1)} ) } \right),$$
	and show that \Algo{Rec-MaxMin-UCB} is asymptotically optimal with respect to $l',K$ and $T.$ 

	Finally, the authors show that regret by shortfall of preference probabilities (cf.\ Section \ref{subsec_performance_measure_multi}) and the theoretically analyzed generalized average regret are equivalent up to constants if the components of the PL model parameters are elements of a compact interval.   
	
	\subsubsection{Preselection Bandits} \label{subsec_preselection}
	
	Interpreting the action within a multi-dueling bandits problem as a preselection for a user, who in turn chooses an arm from this subset, \cite{bengs20a} introduce the notion of \emph{preselection bandits}, along with the question what a good preselection of arms should look like.
	To this end, they assume feedback in the form of the most preferred arm, latent utilities $\theta_i>0$ for each arm $a_i$, and introduce the expected utility in \eqref{defi_expected_utility} to assess the value of a preselection.
	They further restrict the analysis to the PL model (cf.\ Section \ref{subsec_pl_model}) and model the utility parameters of the form $\theta_i = v_i^\gamma$ in order to incorporate the degree of preciseness of a user via the parameter $\gamma>0.$ 

	For the case $\bA=\bA_{l},$ it is shown that the optimal preselection (i.e., actions in $\bA$) does not necessarily consist of the arms with highest latent utilities. Instead, it is composed of top and worst arms, while the arm with highest utility is always in the optimal preselection.
	They propose the \Algo{Thresholded-Random-Confidence-Bound} (\Algo{TRCB}) algorithm, which maintains estimates of the relative utility parameters $\theta_i/\theta_j$ for distinct $i,j\in [K]$ as well as confidence bounds on the latter ratios, and uses the preselection (subset of arms) having the largest expected utility for randomly sampled values within the (trimmed) confidence intervals of the ratios $\theta_i/\theta_j$. 
%	
%	\Algo{TRCB} samples in each round values within the confidence intervals of the latter ratios for $j$ set to the 
%	
	It is shown that the algorithm has an expected-utility regret bound of $\bigO(\sqrt{K T \log(T)}),$ which is nearly optimal as they derive a lower bound of $\Omega(\sqrt{K T})$ if $l \leq K/4.$ 
	Interestingly, the cardinality $l$ of the preselections does not play a role in these bounds, although the action space has complexity $\Theta(K^l)$, suggesting that the computation of the optimal preselection with maximal expected utility involves a seemingly difficult combinatorial task.
	However, as shown by the authors, one can derive structural properties on the expected utility that considerably facilitate the latter optimization problem.
%	computation of the 
%	
%	This is due to a structural property 
	
	For the case $\bA=\bA_{\text{full}},$ where the optimal preselection (i.e., actions in $\bA$)  is obviously the singleton set consisting of the arm with highest utility, they suggest the \Algo{Confidence-Bound-Racing} (\Algo{CBR}) algorithm.
	\Algo{CBR} keeps track of the pairwise preference probability estimates, which are unbiased thanks to the appealing properties of the PL model, and composes the preselections in each iteration by a random process as follows.
	First, the arm with the highest empirical wins is selected as the current leader and an arm is randomly included into the preselection based on the share of its confidence interval on its pairwise preference probability of being preferred over the current leader above $1/2.$
	The larger the share, the higher the chance of being included, while an arm is excluded from consideration once the entire confidence interval is below $1/2.$
	This algorithm enjoys a bound on its expected-utility regret of $\bigO( \sum_{i \neq i^*} \frac{\log(T)}{(\theta_{i^*}-\theta_i)^2} ),$ where $i^*$ denotes the index of the arm with the highest utility.
	Both a worst-case lower bound of $\Omega(\sqrt{T})$ and an asymptotic problem-dependent lower bound of $\Omega(K \log(T))$ are shown for this problem scenario.

	\subsubsection{Choice Bandits} \label{subsec_choice_bandits}
	
	\cite{agarwal2020choice} consider the multi-dueling setting, where the learner is allowed to choose any non-empty subset of arms of size up to $l\leq K$ and is provided with the feedback of the most preferred arm among the chosen subset.
	The authors assume the existence of a generalized Condorcet winner and suggest the \Algo{Winner Beats All} (\Algo{WBA}) algorithm, which proceeds in rounds and revolves around finding a suitable champion arm, which ideally should be the generalized Condorcet winner, and quite naturally should be involved in any chosen action subset.
	To this end, \Algo{WBA} maintains a round-dependent set of possible challengers and chooses in each iteration within each round the champion arm as the one being empirically preferred over the maximum number of challenger arms, which have not been used in the current round.
	In particular, the champion arm can change several times within one round.
	The action subset in each iteration within one round is then composed of the current champion arm and an arbitrary subset of size up to $l-1$ of the possible challengers, which have not been used in the current round (possibly also adding already used challenger arms if the set is small). 
	Once all possible challengers have been used in one round, \Algo{WBA} computes the set of potential challengers for the next round by
	$$ \{ a_i \in \cA \, | \, \exists A \subset \cA:  I_{a_i }(t,A) \geq  \log(t) + f(K,|A|) \} \, , $$
	where $f$ is some appropriately chosen non-negative function, and the empirical divergence $I_{a_i }(t,A)$ of an arm $a_i$ at time $t$ for a subset $A\subset \cA$ (similarly as \Algo{RMED} in Section \ref{subsubsec:rmed}) is defined by
	$$	I_{a_i}(t,A) = \sum_{ \{ a_j \in A : \, \widehat{q}^{\, t}_{i,j|j} \leq 1/2  \}  } n_{i,j|j}^t \, \KL(\widehat{q}^{\, t}_{i,j|j},1/2) \, . 	$$
	Here, $\widehat{q}^{\, t}_{i,j|j}$ and $n_{i,j|j}^t$ differ from $\widehat{q}^{\, t}_{i,j}$ and $n_{i,j}^t$ by conditioning on the event that $j$ was the current champion while a preference of the form $a_i \succ A_t\backslash\{a_i\}$ with $a_j\in A_t$ was observed, in order to compute the respective quantities.

	The motivation of using the latter pairwise preference estimates is that unlike in the case of an underlying PL model (or MNL model) the pairwise estimates $\widehat{q}^{\, t}_{i,j}$ might be biased, but $\widehat{q}^{\, t}_{i^*,j|j}$ are concentrating around a suitable term which is larger than 1/2 if $i^*$ is the generalized Condorcet winner.
	As a consequence, the current champion chosen in the above way will quickly become the generalized Condorcet winner and the set of possible challengers will reduce quickly as well.
	Equipped with this concentration result, the authors show an expected regret bound for \Algo{WBA} of order
	$ \bigO\left(		\frac{K \log(T) }{\Delta_{\text{GCW}}^2}		\right) + \bigO\left( \frac{K^2 \log(K)}{\Delta_{\text{GCW}}^2} \right),		$ 
	where $\Delta_{\text{GCW}}$ is some complexity parameter for the most-preferred-arm probabilities in \eqref{defi_winner_feedback} under the assumption of an existing generalized Condorcet winner, and the regret is measured by the shortfall of preference probabilities regret.
	In the case of an MNL model with utility parameter $\theta$, the expected regret bound can be refined to 
	$ \bigO\left(		\frac{K \log(T) }{\Delta_{i}(\theta)}		\right) + \bigO\left( \frac{K^2 \log(K)}{ \min_{i\neq i^*} \Delta^2_{i}(\theta) } \right),		$ 
	where $\Delta_{i}(\theta)$ is as in Section \ref{subsec_maxminUCB}.
	For their learning scenario, the authors also show an asymptotic lower bound (for any consistent learner) of order $ \Omega\left(		\frac{K \log(T) }{\Delta_{\text{GCW}}}		\right),$ which corresponds to $ \Omega\left(		\frac{K \log(T) }{\Delta_{i}(\theta)}		\right)$ in the case of the MNL model, so that \Algo{WBA} is order optimal for the MNL model.

	\subsubsection{Nearly Best Arm of Plackett-Luce Model} \label{subsec_toparm_PL_multiwise}
	Again assuming a PL model and allowing the learner to choose actions in $\bA_{l}$, whereupon the most preferred arm is returned as the feedback, \citet{saha2019pac} analyze the $(\epsilon,\delta)$-PAC learning setting of finding the best arm.
	More specifically, if $\theta_i\in \R_+$ is the (latent) utility of arm $a_i$ according to the PL model, then one seeks to find an arm $a_i$ such that $\Delta_{i^{*},i} > - \varepsilon$ for some $\varepsilon\in(0,1/2)$ with probability at least $1-\delta$ , where $i^* \in \argmax_{i \in [K]} \theta_i.$
%	the feedback of the multiple comparison is assumed to be generated according to the marginals of a Plackett-Luce model (see Section \ref{subsec_pl_model}), i.e., 
%	%%
%	\begin{equation}\label{eq:pl_marginal}
%		%
%		\prob(a_i \, | \, a_S) = \frac{\theta_{i}}{ \sum_{j\neq i, a_j \in a_S}^K \theta_{j} } \enspace,
%		%
%	\end{equation}
%	%%
%	where $\theta_i\in \R_+$ is the (latent) utility of arm $a_i.$
	%
%	Then, one seeks to find an arm $a_i$ such that $\Delta_{i^{*},i} > - \varepsilon$ for some $\varepsilon\in(0,1/2).$
	%
	
	The authors propose two algorithms for this learning scenario.
	The first one, \Algo{Trace-the-best}, is a rather typical challenge algorithm, which generalizes the idea of \Algo{Interleaved Filtering} (cf.\ Section \ref{subsec_IF_algorithm}) by breaking the most preferred arm feedback into pairwise winner feedback, i.e., the overall most preferred arm is preferred in a direct comparison with each arm involved in the action subset.
	In each challenge round, the action subset is composed of the current champion and a randomly chosen $(l-1)$-sized subset of active arms (possibly adding non-active arms), and this action subset is chosen for a specific number of iterations depending on the approximation quality $\epsilon$ and the confidence level $\delta.$
	At the end of the round, the current champion is replaced if and only if there is an arm being empirically preferred  over the former by a margin $\epsilon/2.$ 
	All arms except the possibly replaced champion of the subset become non-active, and \Algo{Trace-the-best} proceeds with the above sampling procedure until only one arm remains active.

	The second one, \Algo{Divide-and-Battle}, is a tournament-based algorithm  that first divides the set of arms into subsets of size $l$, and each subset is successively used as the action for a specific number of times (depending on $\epsilon$ and $\delta$). 
	Then, all arms within one subset are eliminated except for the arm having the largest number of wins, and \Algo{Divide-and-Battle} proceeds in the same manner as for the first step on the remaining active arms until only a single active arm is left.

	\Algo{Trace-the-best} enjoys a sample complexity of order $ \bigO \left( \frac{K}{\epsilon^2} \log \frac{K}{\delta} \right)$ to return an $\epsilon$-best arm with probability at least $1-\delta,$ while \Algo{Divide-and-Battle} improves upon the logarithmic term revealing a sample complexity of order $ \bigO \left( \frac{K}{\epsilon^2} \log \frac{l}{\delta} \right).$
	It is shown that these bounds are optimal up to the logarithmic term, respectively, as the authors prove that in this learning scenario the lower bound on the sample complexity is the same as for the dueling bandits scenario (cf.\ Section \ref{subsec_seq_elimination} and Section \ref{opt-max}). 
%	Quite interestingly, the authors also show that this complexity corresponds to the lower bound for the battling bandit case as well.
	%
	%The authors verify that the latter bound is also the lower bound for the 
	%
	Thus, similarly as for the regret minimization scenarios considered before, the possibility of comparing more than two arms at the same time does not result in a better theoretical performance for feedback in the form of the most preferred arm.

	However, the authors also consider the more general feedback scenario of obtaining the $l'$ most preferred arms in each time step.
%	, where a preference relation of the compared arms in the form of a partial ranking over the $m \leq k$ most preferred arms is obtained in each time step.
	%
	For this kind of feedback, they provide a lower bound of order  $ \Omega \left( \frac{K}{l' \epsilon^2} \log \frac{1}{\delta} \right)$ for the sample complexity, showing that the learning task is facilitated thanks to the more informative feedback similarly as in the regret minimization scenario in Section \ref{subsec_maxminUCB}.
	To this end, they extend their algorithms \Algo{Trace-the-best} and  \Algo{Divide-and-Battle} for this kind of feedback by using the rank-breaking method and adapting the lengths of the rounds appropriately.
	Also, \Algo{Divide-and-Battle} is modified by means of its arm elimination strategy after a round: If there is an arm that is empirically preferred over each arm (in a direct pairwise comparison) in its corresponding subset by an $\epsilon$-dependent margin, then this arm proceeds to the next round and all others are eliminated, while if there is no such arm, then a randomly chosen arm proceeds.
	%
%	Thanks to the appealing properties of the Plackett-Luce model, rank-breaking does not introduce any bias in the empirical estimation of the pairwise preference probabilities, so that nearly matching upper bounds (up to logarithmic factors) on the sample complexity of the methods can be derived.
	
	Finally, if the action space is $\bA_{\leq l}$ and the learner again observes feedback in the form of the most preferred arm, the authors propose \Algo{Halving-Battle}, which is a tournament-based algorithm sharing similarities with the beat-the-mean algorithm (cf.\ Section \ref{subsec_btm_algorithm}).
	The set of arms is divided into groups of size $l$ and each group is compared for a specific number of times. 
	Subsequently, all arms that have won less times than the empirical median of the number of wins of arms inside the group are discarded, while the other ones are maintained. 
	Once again this procedure is repeated until only a single arm is left, which is likely to be an $\varepsilon$-best arm, due to its high chance of having a higher win count than the empirical group median.
	It is shown that \Algo{Halving-Battle} has a sample complexity of order $ \bigO \left( \frac{K}{\epsilon^2} \log \frac{1}{\delta} \right)$, and therefore improves upon \Algo{Trace-the-best} and \Algo{Divide-and-Battle} for the same feedback, but also has a more flexible action space.
	
	In a subsequent work, \cite{saha2020pac} propose an enhanced variant of \Algo{Divide-and-Battle}, which differs in the way how the lengths of each round are defined: While \Algo{Divide-and-Battle} ignores the overall utility of a subset and simply uses each subset a fixed number of times for its action (up to adjustments of $\delta$ and $\epsilon$), the enhanced version first estimates the overall utility of a subset via a subroutine and then adapts the length of a round accordingly to the estimated overall utility of the action subset of that round. 
	This results in a refined instance-dependent bound on the sample complexity of order  
	\begin{align}\label{sample_complex_divide_and_battle}
			 \bigO \left( \frac{K \max_{A \in \bA_{l}}\sum_{a_i \in A} \theta_i  }{l} \max\left\{1,\frac{1}{l' \epsilon^2}\right\} \log \frac{l}{\delta} \right). 
	\end{align}
	
	This enhanced variant is then combined with the sampling procedure of the algorithms discussed below in Section \ref{subsec_ranking_PL_multiwise} to an epoch-based tournament algorithm called \Algo{PAC-Wrapper}.
	In each epoch, first the enhanced \Algo{Divide-and-Battle} algorithm is used to find a good anchor arm.
	Then, the set of all non-eliminated arms except the anchor arm is divided into subsets of size $l-1,$ merged with the anchor arm, and thereafter each of these subsets is successively used as the action subset for a particular number of times based on the epoch.
	At the end of an epoch, i.e., after all subsets have been used the epoch-specific number of times, all arms within one subset, which are empirically not preferred over the anchor arm  up to some $\epsilon$-dependent margin, are eliminated, and the next epoch is started until only one arm remains.
	
	By adapting the epoch-wise approximation qualities and confidence levels appropriately, as well as using the estimation of the overall utility of an action subset, it is shown that  \Algo{PAC-Wrapper} is an $(\epsilon,\delta)$-PAC learner with sample complexity of order 
	\begin{align} \label{sample_complex_pac_wrapper}
			 \bigO \left( \frac{\max_{A \in \bA_{l}}\sum_{a_i \in A} \theta_i  }{l} \sum_{i\neq i^*} \max\left\{1,\frac{1}{ l' \max\{  \Delta_i^2(\theta) , \epsilon^2 \} } \right\} \log \frac{l}{\delta}\left( \log \frac{1}{ \max\{  \Delta_i(\theta) , \epsilon \}}	\right) \right), 
	\end{align}
	where $\Delta_i(\theta)$ is as in Section \ref{subsec_maxminUCB}.
	Moreover, it is shown that \Algo{PAC-Wrapper} returns the arm with largest latent utility with probability at least $1-\delta$ with a sample complexity as in \eqref{sample_complex_pac_wrapper} for $\epsilon$ set to $0.$
	Further, a lower bound of order 
	$$	\Omega \left(	\max\left\{	\frac{1}{l'} \sum_{i\neq i^*} \frac{\theta_i \theta_{i^*}}{\Delta_i^2(\theta) }		\, , \, \frac{K}{l} \log(1/\delta)		\right\}		\right)		$$
	for this learning scenario is derived, showing that \Algo{PAC-Wrapper}  has a nearly optimal sample complexity.
	
	Finally, the authors also consider the learning scenario in which the learner has only a limited budget available for the total number of its actions.
	Such learning scenarios have been considered in the MAB problem \citep{audibert2010best}, but have not received much attention in the PB-MAB problem yet.
	For this problem scenario, they propose the \Algo{Uniform-Allocation} algorithm, which is conceptionally similar to \Algo{Halving-Battle}, but adapts the lengths of each round with respect to the available budget, and also incorporates that the action space is $\bA_{l}$ as well as that feedback is provided in the form of the $l'$ most preferred arms.
	The following bound on its probability of error for returning the best arm is shown:
	$$	\bigO \left( \log n \exp \left(	 \frac{l' B \Delta_{min}(\theta) }{K+l\log(l)}	\right)	\right),	$$
	where $\Delta_{min}(\theta)=\min_{i\neq i^*} \Delta_i(\theta)$ and $B$ is the available budget.
	However, the authors also derive a lower bound of $\Omega(  \exp(-2 l'B H(\theta) ) ),$ where $H(\theta)= \left( \sum_{i\neq i^*} \frac{\theta_i^2}{\Delta_i^2(\theta)} \right)^{-1}$ is a complexity parameter of the underlying PL problem instance, showing that there is a gap between the error bound of \Algo{Uniform-Allocation} and the derived lower bound.

	\subsubsection{Nearly Best Arm of RUMs} \label{subsec_toparm_RUMS_multiwise}
	Considering the same learning scenario as in \citet{saha2019pac}, but relaxing the PL model assumption\footnote{The discussion below \eqref{eq:marginals} justifies to speak of a relaxation.} to a more general RUM assumption with identically distributed noise variables (IID-RUM), \cite{saha20a} analyze the \Algo{Divide-and-Battle} algorithm under the name \Algo{Sequential-Pairwise-Battle}.
	To this end, they introduce the notion of \emph{minimum advantage ratio} (Min-AR) of an arm $a_i$ over an arm $a_j$, which is defined as 
	$$	\min_{ A \in \bA_{l} : \, a_i,a_j\in A }  \frac{q_{i|I(A)}}{q_{j|I(A)}} \, .	$$
	By showing a lower bound on the Min-AR for any IID-RUM, the authors derive a condition under which the use of the rank-breaking method, for updating the pairwise preference estimates after receiving feedback in the form of the $l'$ most preferred arms, can still lead to appropriate estimates in the sense that it is still possible to infer which arm has the larger utility parameter.
	More specifically, if for any pair of arms $a_i,a_j$, the Min-AR of $a_i$ over $a_j$ is larger than $1+\frac{4 C_{\text{RUM}}(\theta_i-\theta_j)}{C_{\text{RUM}}},$ where $C_{\text{RUM}}>0$ is some constant depending on the distribution of the noise variables in the underlying IID-RUM, then \Algo{Sequential-Pairwise-Battle} returns the $\epsilon$-best arm with probability at least $1-\delta$ and has a sample complexity of order $\bigO \left( \frac{K}{l' \, C_{\text{RUM}}^2 \, \epsilon^2} \log \frac{l}{\delta} \right)$ if the feedback comes in the form of the $l'$ most preferred arms.
	Here, the round-wise approximation qualities of \Algo{Divide-and-Battle} have to be modified by incorporating the constant $C_{\text{RUM}}$, giving rise to the \Algo{Sequential-Pairwise-Battle} algorithm. 
	The authors also provide a lower bound for any $(\epsilon,\delta)$-PAC learning algorithm for an IID-RUM (fulfilling a qualitatively similar lower bound on its minimum advantage ratios as above) of order $\bigO \left( \frac{K}{l' \, C_{\text{RUM}}^2 \, \epsilon^2} \log \frac{1}{\delta} \right),$ showing that \Algo{Sequential-Pairwise-Battle} is optimal up to logarithmic terms.
	
	\subsubsection{Nearly Best Ranking of Plackett-Luce Model} \label{subsec_ranking_PL_multiwise}

	\cite{saha19a} investigate the PAC setting of finding an $\epsilon$-best ranking, that is, a ranking $\pi$ such that $\theta_{\pi^{-1}(k)} \geq \theta_{\pi^{-1}(j)} - \epsilon $ holds for any $1\leq j < k \leq K$, under the same scenario as in \citep{saha2019pac} (cf.\ Section \ref{subsec_toparm_PL_multiwise}).	
	For this setting, the authors suggest the \Algo{Beat-the-Pivot} and the \Algo{Score-and-Rank} algorithm, which are both making use of a subroutine \Algo{Find-the-Pivot} in order to obtain a ``good'' initial anchor arm for eventually inferring the final ranking.
	The \Algo{Find-the-Pivot} subroutine is an $(\epsilon,\delta)$-PAC algorithm for finding the best arm (the same problem as considered in Section \ref{subsec_toparm_PL_multiwise}) and is just the \Algo{Trace-the-best} algorithm.
	Both \Algo{Beat-the-Pivot} as well as \Algo{Score-and-Rank} use the same sampling procedure, but differ in the way how the final ranking is returned. 
	More specifically,  \Algo{Beat-the-Pivot} uses the estimated pairwise preference probabilities with respect to the anchor arm, say $A$, while \Algo{Score-and-Rank} uses the estimated relative utility parameters $\theta_i/\theta_A$ to infer the ranking.
	The sampling procedure of both algorithms is simply to divide all arms except the anchor arm into subsets of size $l-1,$ merge each subset with the anchor arm and thereafter use each of these subsets successively as the action subset for a particular number of times based on $\delta$ and $\epsilon$.

	Both algorithms have a sample complexity of order $ \bigO \left( \frac{K}{\epsilon^2} \log \frac{K}{\delta} \right),$ which is shown to be optimal by providing a corresponding lower bound. 
	Once again, there is no improvement of the sample complexity compared to the dueling bandits problem (see Table \ref{tab:regaxiom_second}).
	However, assuming feedback in the form of the $l'$ most preferred arms leads to a lower bound of order   $ \Omega \left( \frac{K}{l' \, \epsilon^2} \log \frac{K}{\delta} \right),$ which \Algo{Beat-the-Pivot} again achieves by exploiting the appealing properties of the PL model via the rank-breaking method.

	\subsubsection{Exact Top-$k$ Arms of Plackett-Luce Model} \label{subsec_topk_PL_multiwise}

	Allowing the learner to choose subsets with a size of up to $l$ as its actions, where $l\leq K$ is fixed, and assuming that the feedback in the form of the most preferred arm among the chosen subset is generated by a PL model (or MNL model), \cite{chen2018nearly} study the problem of identifying the top-$k$ arms.
	To this end, they consider two regimes for $l,$ one in which $l=\bigO(\log K)$ and another one for  $l=\Omega(\log K)$.

	For the logarithmic regime, the authors suggest the $\Algo{AlgPairwise}$ algorithm, which uses only pairs of arms as its action subset, and argue that this restriction to pairs results in a logarithmic term of $K$ in the resulting theoretical guarantee.
	$\Algo{AlgPairwise}$ maintains a set of active arms and builds the set of top-$k$ arms successively by proceeding in rounds, each of which eliminates active arms by checking whether an arm belongs to the set of top-$k$ arms or an arm belongs to the worst $K-k$ arms with a certain confidence.
	In each round, the algorithm samples uniformly at random $\tilde{\mathcal{O}}(n)$ many pairs of active arms for its action and computes confidence intervals for the pairwise preference probabilities to decide about the order of their underlying latent scores.
	Because the order for a specific pair, say $a_i$ and $a_j,$ can not necessarily be decided solely on the basis of the comparisons made between $a_i$ and $a_j$, the algorithm exploits some sort of transitivity to incorporate order relations between other active arms: If there exist active arms $a_{i_1},\ldots,a_{i_d}$ such that, based on the comparison made, $a_i$ has a higher (lower) utility than $a_{i_1},$ $a_j$ has a lower (higher) utility than $a_{i_d},$ and $a_{i_s}$ has a higher (lower) utility than $a_{i_{s-1}}$ for $s=1,\ldots,d-1,$ then it is inferred that $a_i$ has a higher (lower) utility than $a_{j}.$
	In light of this, an active arm is eliminated and added to the set of top-$k$ arms in case it has a higher utility than $k-k'$ many active arms, while if an active arm has a lower utility than $m-k'$ many active arms, it is simply removed from the set of active arms.
%	\\
	Here, $k'$ is the number of arms that are already added to the set of top-$k$ arms, and $m$ is the number of active arms.

	It is shown that the set of top-$k$ arms is returned with high probability, while the number of pairwise comparisons is of order
	\begin{align}\label{sample_complex_alg_pairwise}
	\begin{split}
				 \tilde{\mathcal{O}} \Bigg(	\frac{K}{l} + k + \frac{\sum_{i=k+1}^K  \theta_{(i)}}{ \theta_{(k)}} &+ \sum_{i\geq k+1: \, \theta_{(i)}>\theta_{(k)}/2} \frac{\theta_{(k)}^2}{(\theta_{(k)} - \theta_{(i)}  )^2}	\\ &+ \sum_{i \leq k: \, \theta_{(i)}\leq 2\theta_{(k+1)}} \frac{\theta_{(k+1)}^2}{(\theta_{(k+1)} - \theta_{(i)}  )^2}		\Bigg) \, ,
	\end{split}
	\end{align}
	where polylogarithmic terms of $K$ and $(\theta_{(k)}-\theta_{(k+1)})^{-1}$ are hidden in the latter $\tilde{\mathcal{O}}$ term, respectively.
	For the superlogaritmic regime, the authors suggest the $\Algo{AlgMulti-wise}$ algorithm, which has a sample complexity (i.e., total number of actions) of the same order as in \eqref{sample_complex_alg_pairwise}.
	Moreover, the authors derive a lower bound of the same order (without the polylogarithmic terms), showing that the algorithms are nearly optimal.

	\subsection{Related Frameworks}
	
	As mentioned at the beginning of this section, there are some other frameworks that share similarities with the multi-dueling bandits regarding the qualitative nature of the feedback observed by the learner.
	In the following, we give a brief overview of such frameworks, explaining similarities but also highlighting differences. 
	
	\subsubsection{Stochastic Click Bandits}
	The online learning-to-rank problem as described in Section \ref{subsec_multileaving} covers web search as one of its most prominent fields of application.
	However, the modeling approach by means of the multi-dueling bandits disregards in some way how the user usually interacts with a displayed list of objects such as web pages or documents: The user scans the list from top to bottom, so that the order in which the objects are presented may play a central role. Moreover, the user might not choose any of the presented objects at all.

	In stochastic click bandits, these specialties are taken into account by means of a click model, which is a stochastic model of the choice (or click) behavior over an ordered set.
	A commendable introduction to this setting (and beyond that) can be found in Chapter 32 of \cite{lattimore2020bandit}, with a thorough overview of the literature on this topic.
	Nevertheless, it is important to note that multi-dueling bandits and stochastic click models are quite different, mainly due to their different views on the importance on the order of the arms in the subset,  and neither one can be considered a special case of the other.
	
	\subsubsection{Dynamic assortment optimization}
	The dynamic assortment optimization problem \citep{caro2007dynamic} is motivated by the area of retailing, where a retailer seeks to find an optimal assortment (subset) of her/his available products (arms) by repeatedly offering different assortments to a customer, and observe a/no purchase of one product within the offered assortment.
	Quite naturally, the goal is to maximize the expected revenue over time, which can be expressed as a regret minimization task.
	This problem framework differs from the multi-dueling bandits in two important points. 
	First, the arms in the former are assumed to be equipped with a priori known revenues, while in the latter, such revenues are simply not present. 
	As a consequence, the optimal action subset might be different in both settings.
	Second, the customer in the dynamic assortment optimization problem may refuse the purchase.
	Technically, in the setting  of multi-dueling bandits, this can be  modeled by extending the set of arms by means of a dummy arm representing this no-choice option and adding this dummy arm to all possible action subsets.
	While this seems to be an interesting direction for future work, it would require a rethinking of many of the learning tasks and related performance measures.
	
	\subsubsection{Best-of-$k$ bandits} 
	In the setting of best-of-$k$ bandits, the learner is allowed to use a $k$-sized subset of arms as an action (similarly as in multi-dueling bandits with action space $\bA_{k}$) and obtains specific forms of feedback.
	In contrast to multi-dueling bandits, feedback is usually assumed to be numerical.
	However, \citet{simchowitz2016best} consider another type of feedback, called \emph{marked-bandit feedback}, in which the learner observes the index of the arm with the largest (latent) reward if it is among the action subset, and otherwise a ``void" information.
	As the rewards are assumed to be binary, the index is chosen uniformly at random from the index set of all arms with positive rewards.
	Although this type of feedback is once again of a qualitative nature, the ``void'' information is conceptionally similar to a no-choice option in the dynamic assortment optimization problem.
	It is an interesting question whether the best-of-$k$ bandits with marked-bandit feedback can be formulated as a specific learning scenario within the multi-dueling bandits by incorporating a dummy arm.
%	 and consequently differing from the several types of feedback in Section \ref{subsec_feedback_multi}.
%	, which do not admit such feedback without information content.

	%###################################################
	\section{Applications} \label{sec_applications}
	%In preference-based reinforcement learning, \citet{BuSzWeChHu14} present a preference-based extension of the evolutionary direct policy search (EDPS) framework for policy learning based on evolutionary optimization, in which the value-based racing algorithm is replaced with a preference-based one that only requires qualitative comparisons between sample histories as training information. The authors provide theoretical guarantees for the introduced preference-based version of the Hoeffding race algorithm and show empirically that their approach performs well in practice.
	
	Multi-armed bandits have been used in various fields of application, and the more recent setting of dueling bandits is receiving increasing attention from a practical perspective, too. In the following, we provide a short overview of some recent applications of dueling bandits algorithms.

	%##########
	\citet{ChMa12} consider the problem of learning a model of gesture generation to automatically generate animations for dialogues. To this end, they make use of subjective human judgment of naturalness of gestures. In this regard, pairwise comparisons (one gesture is considered more natural than another one) appear to be much easier than absolute  judgments, which are often very noisy. This is why the authors tackle the task as a dueling bandits problem. Concretely, they use the \Algo{DBGD} algorithm (cf.\ Section \ref{subsec_DBGD}) and show empirically that the framework can effectively improve the generated gestures based on the simulated naturalness criterion.
	
	%##########
	In the context of information retrieval, \citet{HoShWhDe13} investigate to what extent historical data in the form of interaction data can lead to an improvement for learning in online learning-to-rank problems. They introduce an approach based on the \Algo{DBGD} algorithm using the Probabilistic Interleave (PI) method for interleaving (cf.\ Section \ref{subsec_multileaving}). 
	They evaluate the performance of their approach in terms of the discounted cumulative reward on several learning-to-rank data sets and find that historical data can indeed be useful to enhance the performance of a learner in online learning-to-rank problems.
%	whether and how previously collected historical interaction data can be used to speed up online learning to rank. 
%	. The latter is based on a probabilistic interpretation of interleaved comparisons that allows one to infer comparison outcomes using data from arbitrary result lists. 
%	They evaluate the performance of their approach in terms of discounted cumulative reward on several learning-to-rank data sets and find that historical data can indeed be used to increase the effectiveness of online learning-to-rank for information retrieval.
	
	%##########
	Supporting clinical research that aims at recovering motor function after severe spinal cord injury, \citet{SuBu14} set up a dueling bandits instance to help paralyzed patients regain the ability to stand on their own feet. The feedback consists of a stochastically ranked subset of $ K $ tests, each of which corresponds physically to an electrical stimulation period applied to the spinal cord with a specific stimulus. The goal is to identify the optimal stimulus for a patient, and the ranking is based on a combined scoring of certain standing criteria by the observing clinicians (under noisy conditions). The authors introduce the Rank-Comparison algorithm, a modified version of \Algo{BTM} (cf.\ Section \ref{subsec_btm_algorithm}).
%	 which has  an optimal expected total regret.
	
	\citet{SuYuBu17} address the same application. To overcome the issue of very large action spaces, which is due to the huge number of different stimulating configurations and hinders a fast convergence of algorithms attempting to solve this problem, they consider correlation structures on the action space and exploit dependencies among the arms. This allows them to update a set of active arms instead of only a pair in each iteration of the algorithm. The authors propose $ \Algo{CorrDuel} $, an algorithm based on BTM. This algorithm is applied in a synthetic experimental setup and shown to perform better than algorithms that do not exploit correlation information. In a live clinical trial of therapeutic spinal cord stimulation, $ \Algo{CorrDuel} $ performs as good as specialized physicians.
	
	%##########
	\citet{SuZhBuYu18} consider the safe Bayesian optimization problem, where the goal is to optimize an unknown utility function with absolute or preference feedback in a sequential manner, subject to some unknown safety constraints, in the case where a small region of the safe action space is known a priori and needs to be expanded in an iterative manner. They propose the \Algo{StageOpt} algorithm, which models both the safety and utility functions via Gaussian processes (GPs), and makes use of confidence bounds around the mean function for the sake of (safe) exploration and optimization. Moreover, \Algo{StageOpt} proceeds in two stages for the underlying optimization problem; in the first stage, the safe region is gradually expanded by means of the confidence bounds, while in the second stage, this expanded safe region is used as the domain for Bayesian optimization.
%	Moreover, \Algo{StageOpt} separates the optimization problem into two stages; one stage where it iteratively expands the safe region, followed by the second stage in which it performs Bayesian optimization within the safe region.
	For the case of preference feedback, in which the algorithm receives feedback in the form of a Bernoulli distribution with probability as in \eqref{eq_utility_based_pairwise} for the current and the previous sample points (infinite action space),
%	Bernoulli feedback according to some link function between the current and the previous sample points, 
	they use the multi-dueling bandit algorithm \Algo{KernelSelfSparring} (cf.\ Section \ref{subsec_selfsparring}).  Motivated by the clinical application setting described in \citet{SuBu14}, they consider the goal of finding effective stimulation therapies for patients with severe spinal cord injuries without introducing undesirable side effects, and use the preference-based version of their algorithm to safely optimize clinical spinal cord stimulation in order to help the patients regain physical mobility. They show that the therapies it suggests outperforms the ones proposed by experienced physicians. 
	
	%##########
	\citet{SoRiUr16} use the Structured Dueling Bandits algorithm, an extension of \Algo{DBGD}, for response-based structured prediction in Statistical Machine Translation (SMT). In a repeatable generate-and-test procedure, the learner is given partial feedback that consists of assessments of the quality of the predicted translation, which is used by the learner to update specific model parameters. In a simulation experiment, the authors show that learning from such type of feedback can indeed facilitate the supervision problem and allows a direct optimization of SMT for different tasks.
	
	%##########
	\citet{ChZhKi16} consider the problem of the allocation of assessment tasks among peers when grading open-ended assignments in Massive Open Online Courses (MOOCs), and formalize it as a sequential noisy ranking aggregation problem. More specifically, each student ranks a subset of his peers' assignments, and the goal is to aggregate all the partial rankings into a complete ranking of all assignments. The authors assume the existence of a ground-truth ranking, and that the underlying distribution is a Mallows model. Based on these assumptions, they propose \Algo{TustAwareRankingBasedMAB}, an algorithm based on merge sort and MallowsMPR (cf.\ Section \ref{subsec_mallows}).
	
	%##########
%	\citet{ChRaHo16} present a bandit-based approach to address the cold-start recommendation problem, i.e. the elicitation of preferences of new users, in an online learning setting, with a focus on the task of restaurant recommendation. They develop, i.a., a preference elicitation framework to identify which questions to ask new users to most quickly learn their preferences, and demonstrate its benefits.
	
	%##########
	\citet{ScKu17} investigate the problem of learning a user's task preferences in Human-Robot Interaction for socially assistive tasks. Concretely, they consider the goal of learning a user's exercise category. They formulate a dueling bandits problem, where arms represent exercises. Preference feedback is given by a user who, given two exercises that are presented to him as text on a display, selects the more preferred one. \Algo{DTS} (cf.\ Section \ref{subsec_dts}) is used as a dueling bandits learning algorithm. The simulation experiments show that the users were satisfied with the suggested preference rankings. Moreover, the results of a comparison of the preference learning approach against a simulated strategy that randomly selects preference rankings show that the preference learning approach leads to a significant reduction of ranking errors.
	
	Inspired by the application of MAB algorithms for implementing tree search policies in Monte Carlo tree search \citep{browne2012survey}, such as UCT (Upper Confidence bound for Trees),  \citet{joppen2018preference} introduce the Preference-Based Monte Carlo Tree Search (PB-MCTS) framework, where the feedback is of a qualitative nature (e.g., ordinal rewards).
	In contrast to classical tree policies, where a single successor is chosen at each node, resulting in an observation over a path, PB-MCTS triggers two roll-outs at each node, resulting in an observation along a binary tree. 
	Apparently, this variant induces an exponential growth in the number of explored paths.
	The authors propose a tree policy with a choice mechanism for the successor guided by the RUCB algorithm (cf.\ Section \ref{subsec_RUCB}) and investigate its performance for the 8-puzzle problem.

	For the task of automatically recommending suitable parameter settings for an algorithm or solver for sequentially arriving problem	instances,  \cite{el2020pool} propose the Contextual Preselection with Plackett-Luce (\Algo{CPPL}) algorithm. 
	Here, each algorithm parametrization corresponds to an arm having a latent utility parameter function, which depends on the features of the algorithm parametrization and varies with the features of a problem instance.
	It is assumed that for an incoming problem instance multiple parametrizations of a solver can be run in parallel for solving the instance, and the parallel solving process is stopped as soon as one of the parametrizations has found a solution.
	\Algo{CPPL} maintains a pool of candidate parametrizations  generated by a genetic crossover procedure using estimates of the utility parameter function in order to decide which algorithm parametrizations are suitable candidates.
	The candidates with the largest upper confidence on the utility parameter function evaluated for the current problem instance are used for the parallel solving process, and candidates within the pool are pruned by adopting a racing strategy \citep{MaMo94,MaMo97}, whereupon the pool is replenished with new candidates by the genetic crossover procedure.
	On a series of data sets corresponding to mixed-integer programming and satisfiability problems, the suggested algorithm reveals satisfactory empirical performance in choosing the configurations of corresponding solvers in a sequential manner and even outperforms state-of-the-art methods. 
%	

	%#################################
	\section{Summary and Perspectives} \label{sec_summary}
	
	In this paper, we surveyed the state of the art in preference-based online learning with bandit algorithms, a relatively recent research field that we referred to as preference-based multi-armed bandits (PB-MAB), and which is also known under the notion of ``dueling bandits''. In contrast to standard MAB problems, where bandit information is understood as (stochastic) real-valued rewards produced by the arms the learner decided to explore (or exploit), feedback is assumed to be of a more indirect and qualitative nature in the PB-MAB setting. This includes preference information in the form of comparisons between pairs of arms, which has been the focus of most approaches so far, but which has been extended toward more general feedback scenarios such as partial rankings in the recent past. We have given an overview of instances of the PB-MAB problem that have been studied in the literature, algorithms for tackling them, and criteria for evaluating such algorithms. Besides, we have given an overview of existing applications. 
	
	In spite of a growing body of literature, the field is still developing and certainly much less mature than research on standard multi-armed bandits. Obviously, this is due to its recency but also because the setting itself is more involved (for example, inconsistencies such as preferential cycles may occur, the definition of regret is less obvious than in the standard value-based setting, etc.). Various algorithms have been proposed so far, and different theoretical results have been produced. However, comparing these results and relating them to each other is far from obvious, mainly because different authors start from different assumptions and formalizations of the problem; in fact, there is no complete agreement on assumptions, targets, and performance measures so far, and a complete and coherent theoretical framework is still to be developed. Moreover, even for the standard setting, there is still a number of ``gaps'' in the PB-MAB landscape, i.e., open theoretical questions and algorithmic problems that have not yet been addressed.

	With this survey, we hope to contribute to the further popularization, development, and shaping of the field. We conclude the paper with a short (and certainly non-exhaustive) list of open problems that we consider particularly interesting for future work.

	%\item %(\emph{ii}) 
%	Similar questions can be asked for the regret. The RUCB algorithm achieves a high probability regret bound of order $K\log T$ by merely assuming the existence of a Condorcet winner. Yet, this assumption is arguable and certainly not always valid.
	
	%\item % (\emph{iii}) 
	For some of the settings and related learning tasks discussed in the paper, the lower bounds are still not known. For instance, the lower bound on the regret for general tournament solutions in Section \ref{subsec_gen_tournaments} has been left as an open question. Also for the task of weak regret minimization considered by \cite{ChFr17} (cf.\ Section \ref{sec_winner_stays_algo}), there seems to be still no proven lower bound. Thus, it is difficult to say whether the suggested algorithms are optimal with regard to the problem-dependent complexity terms or not.
	%For most of the settings discussed in the paper, such as those based on statistical models, a lower bound on the sample complexity is not known. Thus, it is difficult to say whether an algorithm is optimal or not. There are a few exceptions, however. For the regret optimization setup with the assumption of the existence of total order over arms, it is known that, for any algorithm $A$, there is a bandit problem such that the regret of  $A$ is $\Omega (K \log T)$ (see Theorem 2 by \cite{YuBrKlJo12}). Moreover, in the case of the utility-based bandit setup, the reduction technique of \cite{AiKaJo14} implies that the lower bound of the standard bandit setup~\citep{LaRo85} also applies for the utility-based setup. Obviously, these lower bounds are also lower bounds for all settings starting from weaker assumptions.	
	In the context of statistical approaches (such as Mallows and Plackett-Luce), the analysis of lower bounds would also be interesting for the problem of approximating the entire distribution $\prob$. \cite{BuSzHu14} address this problem for the Mallows model, using the Kullback-Leibler (KL) divergence as a measure of distance between $\prob$ and the learner's prediction $\widehat{\prob}$. The problem turned out to be hard, and the sample complexity of the authors' algorithm scales quite poorly with the number of arms. Thus, one may conjecture that a truly efficient algorithm does not exist, which in turn calls for the proof of a lower complexity bound. 
	
		%\begin{itemize}
	%\itemsep0em
	%\item %(\emph{i}) 
	Next, as we have seen, the difficulty in the basic PB-MAB learning scenario strongly depends on the assumptions on the preference relation $\bQ$: The more restrictive these assumptions are, the easier the learning task becomes. 
	Take the $(\epsilon,\delta)$-PAC learning scenario for finding a suitable ranking of the arms for instance, for which the results of \cite{falahatgar2018limits} show that strong stochastic transitivity and the stochastic triangle inequality, and consequently the exploitation of both properties is the key to achieve a sample complexity which is sub-quadratic with respect to the number of arms. However, in the multi-dueling bandit problem, such transitivity properties are not readily available. It would be an interesting question whether one could define properties giving a full-fledged substitute without imposing too strict requirements such as latent utilities of the arms or suchlike. 
	
	%	An interesting question in this regard concerns the ``weakest'' assumptions one could make while still guaranteeing the existence of an algorithm that scales linearly in the number of arms. \citet{RaRaAg16} partially addressed this question by introducing various properties on the preference relation that still allow efficient learning. Nevertheless, their assumptions, such as a utility-based preferences relation with linear link, appear to be still too restrictive. For example, the transitivity relation might be relaxed along the line of the work of \citet{YuJo09}. 
	%	More generally, regularity assumptions on $\bQ$ essentially correspond to assumptions about
	%	transitivity, and exploitation of transitivity of preferences is the key to achieving, \editvik[for instance,] sub-quadratic complexity \editvik[for finding a nearly optimal ranking \citep{falahatgar2018limits}]. Therefore, one may suspect that the difficulty of PB-MAB learning is mainly determined by the degree of transitivity of $\bQ$; indeed, note that PB-MAB learning reduces to simple sorting in the case of standard deterministic transitivity. 
	
	%\item %(\emph{iv}) 
	Related to the  previous issue, another important problem concerns the development of (statistical) tests for verifying the assumptions made by the different approaches in a real application. In the case of the statistical approaches based on the Mallows and PL distributions, for example, the problem would be to decide, based on data in the form of pairwise comparisons, whether the underlying distribution could indeed be Mallows or PL versus a reasonable alternative hypothesis class.
	In other words, suitable (online) hypothesis tests are needed that allow for deciding whether or not the data-generating process obeys a certain distribution, such as Mallows or Plackett-Luce. Surprisingly little work has been done on this problem in the statistical literature so far \citep{AlvoLu14,FlVe93}. 
	There is some recent work on testing properties of discrete distributions. For example, given the possibility to sample from a multinomial distribution, the goal is to decide whether the distribution belongs to a family of distributions that is given in advance to the learner, e.g., a family of monotone distributions \citep{AcDaKa15}. Similar questions can be addressed in the preference-based setup. For instance, given the possibility to sample pairwise preferences determined by $\bQ$, decide whether $\bQ$ belongs to some subset of relations with a certain property, such as relaxed or strong stochastic transitivity, satisfying the stochastic triangle inequality, or exhibiting a Condorcet winner as required by many methods. 
	Existing works seem to be restricted to testing the weak stochastic transitivity assumption in an offline setting \citep{IvFa1985,cavagnaro2014transitive}.

	%\item %(\emph{vi}) 
	The role of adaptivity\footnote{In the sense of learning in a ``non-batch'' setup, such as active or online learning, which provides the learner with an opportunity to influence and partly control the training data.} is not yet fully clear either. In preference-based online learning, we assume that the learner can sample from the underlying pairwise data-generating process, and hence partly control the data to learn from. To what extent does this additional freedom help the learner, facilitate the problem, and possibly improve performance? This question has not been investigated in detail so far. \citet{Agarwal0AK17} address it in a specific (utility-based) setup, however, it still remains open in most of the cases we discussed in this paper. In particular, the question seems to be hard to answer in the case of the ranking-distribution-based setup. As an important prerequisite, the sample complexity of the optimal learning should be characterized when the learner has sample access to full rankings. For the case of the Mallows model, the sample complexity of optimal learning has been obtained in a recent work by~\cite{Busa-FeketeFSZ19}. Nevertheless, to the best of our knowledge, there is no known sample-optimal learning algorithm for other parametric ranking distributions, such as Plackett-Luce or log-linear ranking models~\citep{Mesaoudi-PaulHB18}.

	%\item 
	The Kemeny consensus or minimum feedback arc set (MFAS) ranking can be considered as the holy grail among the ranking methods, because it corresponds to the ranking with the smallest violation of pairwise relations. It is known that if $\bQ$ is given, finding the MFAS ranking is NP-hard. There is a constant approximation algorithm~\citep{AiChNe05}, and what is more, there is a PTAS for this problem~\citep{KeSc07}. One can naturally address the question of learning the MFAS ranking with $\bigO(K\log K )$ sample complexity in the online preference learning setup. In other words, do we need to reveal the whole matrix $\bQ$, or might partial information of $\bQ$ be enough?

	Another interesting variant of the bandit problems themselves is a combination of the preference-based and real-valued MAB problems, such that the learner is allowed to either pull an arm in order to obtain a real-valued reward, or to choose a pair of arms and obtain preference feedback.
	Such a scenario has recently been considered  by \cite{xu20threshold} with the goal of finding all arms with a mean reward above some specific threshold, while keeping both the number of duels and pulls as small as possible.
	The motivation behind such a fusion of both settings is that in some practical applications it might be more difficult to obtain real-valued rewards of arms than a preference between arms.
	These considerations open up a number of interesting questions, while the main question would be certainly how much the possibility of dueling arms can support the pulling of arms in order to achieve a certain target or vice versa.
	Even though \cite{xu20threshold} already provide some insights and first answers to this question, there are definitely more open questions to be addressed in this regard.
%	, especially for the task of regret minimization

	Last but not least, there are of course various practically motivated extensions of the basic PB-MAB setting one may think of, along the lines of those summarized in Section \ref{sec:extensions} and even beyond. In addition to working on such extensions and generalizations, it would be important to test methods and algorithms in real applications, such as crowd-sourcing platforms.
	 However, up to now it seems that there is no available code repository including all (or at least a substantial share) of the established algorithms of the PB-MAB setting. 
	An attempt to address this issue for the programming language \Algo{Python} is made by the \emph{duelpy} package\footnote{\url{https://gitlab.com/duelpy/duelpy}}.

	%For example, in the basic setting considered by most approaches so far, two arms are chosen and compared in every time step. Despite the great practical relevance of pairwise comparisons, there are many applications in which preference information is revealed in a different way. Such applications motivate the extension of the PB-MAB framework so as to make it amenable to preference information that is more general than pairwise comparisons. For example, feedback is often provided in the form of a choice decision among more than two alternatives. Consider online advertisement as an example, where the system recommends ads to a user. Typically, a list of (more than two) ads is put up on the website opened, and the user clicks on one of them, presumably the most preferred one. Another example is the observation of partial rankings in movie recommendation, where each user provides ratings about a small subset of movies. More formally, these examples correspond to a 1-of-$k$ choice and a ranking of a subset of alternatives; for the latter, one can further distinguish between the case where the subset is chosen at random and where the subset is supposed to be the top-$k$ (i.e., observation of the top of the full ranking).

	%	\item Last but not least, it would be important to test the algorithms in real applications---crowd-sourcing platforms appear to provide an interesting test\-bed in this regard.
	%\end{itemize}
	
	\acks{Eyke H\"ullermeier, Adil El Mesaoudi-Paul and Viktor Bengs gratefully acknowledge financial support by the German Research Foundation (DFG). We would also like to thank two anonymous referees for their valuable comments and suggestions, which helped to significantly improve this survey.}
	
	\vskip 0.2in
	\bibliography{bandit}
	
\end{document}